\pdfoutput=1
\documentclass{these-dbl}
\usepackage{amsmath}
\usepackage{dsfont}
\usepackage{amssymb}
\usepackage{nicefrac}

\usepackage{booktabs}
\usepackage{multirow}
\usepackage{cellspace}

\usepackage{algorithm}
\usepackage{algpseudocode}
\usepackage{ifthen}

\usepackage{subcaption}
\usepackage{ragged2e}
\usepackage{pdflscape}
\usepackage{rotating}

\usepackage{enumitem}

\usepackage{hyperref}
\usepackage{xurl}

\usepackage{titlecaps}
\usepackage{sectsty}
\usepackage[all]{nowidow}

\usepackage{tikz}
\usepackage{xcolor}
\usepackage{fdsymbol}

\usepackage{ebgaramond}
\usepackage{epigraph}
\usepackage{etoolbox}

\usepackage{minitoc,titletoc}

\makeatletter
\newlength\epitextskip
\pretocmd{\@epitext}{\em}{}{}
\apptocmd{\@epitext}{\em}{}{}
\patchcmd{\epigraph}{\@epitext{#1}\\}{\@epitext{#1}\\[\epitextskip]}{}{}
\makeatother
\newcommand\cincludegraphics[2][]{\raisebox{-0.3\height}{\includegraphics[#1]{#2}}}

\newcommand{\quotes}[1]{``#1''}

\RequirePackage{caption}
\newcommand{\titlecaption}[3][]{\caption[#2]{\textbf{#2}\ifthenelse{\equal{#1}{}}{. }{ }#3}}

\definecolor{yellowdark}{HTML}{BC8C00}
\definecolor{bluedark}{HTML}{2F528F}
\definecolor{greendark}{HTML}{507E32}
\definecolor{bluefig}{HTML}{5B9BD5}

\definecolor{lightblueborder}{HTML}{41719C}
\definecolor{lightbluefill}{HTML}{5E9CD3}

\mtcsetoffset{minitoc}{-0.80em}
\mtcsetfeature{minitoc}{after}{\vspace{.5cm}}

\hypersetup{
}

    {\small\begin{center}%
    \bfseries{Abstract} \end{center}
    
    \begin{quote}
      \singlespacing\small
      \rule{14cm}{1pt}
      #1
      \vskip-4mm
      \rule{14cm}{1pt}
\end{quote}
}

\begin{document}

\selectlanguage{french}


\makeatletter
\newcommand\addcase[3]{\expandafter\def\csname\string#1@case@#2\endcsname{#3}}
\newcommand\makeswitch[2][]{%
    \newcommand#2[1]{%
        \ifcsname\string#2@case@##1\endcsname\csname\string#2@case@##1\endcsname\else#1\fi%
  }%
}
\makeatother

\newcommand\hauteurlogos[3]{
    \hauteurlogoecole{#1}
    \hauteurlogoetablissementA{#2}
    \hauteurlogoetablissementB{#3}
}


\newcommand\addecoledoctorale[6]{
    \direcole{#1}
    \numeroecole{#2}
    \definecolor{couleur-ecole-recto}{RGB}{#3}
    \definecolor{couleur-ecole-verso}{RGB}{#4}
    \nomecoleA{#5}
    \nomecoleB{#6}
}

\makeswitch[default]\ecoledoctorale{}

\addcase\ecoledoctorale{ALL}{\addecoledoctorale
    {ALL}
    {595}
    {255,165,139}
    {232,86,18}
    {Arts, Lettres, Langues}
    {}
}
\addcase\ecoledoctorale{DSP}{\addecoledoctorale
    {DSP}
    {599}
    {255,241,170}
    {255,214,12}
    {Droit et Science politiques}
    {}
}
\addcase\ecoledoctorale{EDGE}{\addecoledoctorale
    {EDGE}
    {597}
    {255,254,101}
    {255,237,0}
    {Sciences \'{e}conomiques et sciences de gestion - Bretagne}
    {}
}
\addcase\ecoledoctorale{EGAAL}{\addecoledoctorale
    {EGAAL}
    {600}
    {0,118,0}
    {0,93,49}
    {\'{E}cologie, G\'{e}osciences, Agronomie, Alimentation}
    {}
    \couleurpolice{white}
}
\addcase\ecoledoctorale{ELICCE}{\addecoledoctorale
    {ELICCE}
    {646}
    {255,207,114}
    {252,199,82}
    {\'{E}ducation, Langages, Interactions, Cognition, Clinique, Expertise}
    {}
}
\addcase\ecoledoctorale{ESC}{\addecoledoctorale
    {ESC}
    {645}
    {255,164,85}
    {240,138,0}
    {Espaces, Soci\'{e}\'{e}s, Civilisations}
    {}
}
\addcase\ecoledoctorale{MathSTICBO}{\addecoledoctorale
    {MathSTICBO}
    {644}
    {190,212,233}
    {139,181,221}
    {Math\'{e}matiques et Sciences et Technologies}
    {de l'Information et de la Communication en Bretagne Oc\'{e}ane}
}
\addcase\ecoledoctorale{MATISSE}{\addecoledoctorale
    {MATISSE}
    {601}
    {0,112,237}
    {0,84,160}
    {Math\'{e}matiques, T\'{e}l\'{e}communications, Informatique, Signal, Syst\`{e}mes,}
    {\'{E}lectronique}
    \hauteurlogos{1.8cm}{1.8cm}{1.8cm}
    \couleurpolice{white}
}
\addcase\ecoledoctorale{S3M}{\addecoledoctorale
    {S3M}
    {638}
    {159,19,90}
    {156,42,100}
    {Sciences de la Mati\`{e}re, des Mol\'{e}cules et Mat\'{e}riaux}
    {}
    \couleurpolice{white}
}
\addcase\ecoledoctorale{SML}{\addecoledoctorale
    {SML}
    {598}
    {19,139,112}
    {0,93,102}
    {Sciences de la Mer et du Littoral}
    {}
    \couleurpolice{white}
}
\addcase\ecoledoctorale{SPI}{\addecoledoctorale
    {SPI}
    {647}
    {136,191,255}
    {63,133,193}
    {Sciences pour l'Ing\'{e}nieur}
    {}
}
\addcase\ecoledoctorale{SPIN}{\addecoledoctorale
    {SPIN}
    {648}
    {161,173,255}
    {80,92,162}
    {Sciences pour l'Ing\'{e}nieur et le Num\'{e}rique}
    {}
}
\addcase\ecoledoctorale{SVS}{\addecoledoctorale
    {SVS}
    {637}
    {228,255,122}
    {200,210,0}
    {Sciences de la Vie et de la Sant\'{e}}
    {}
}


\newcommand\addetablissement[4]{
    \logoetablissementB{#1}
    \nometablissementC{#2}
    \nometablissementD{#3}
    \nometablissementE{#4}
}

\makeswitch[default]\etablissement{}

\addcase\etablissement{CS}{\addetablissement
    {CS}
    {}
    {}
    {CentraleSup\'{e}lec}
}
\addcase\etablissement{EHESP}{\addetablissement
    {EHESP}
    {}
    {l'\'{E}cole des Hautes \'{E}tudes}
    {en Sant\'{e} Publique}
    \hauteurlogos{2cm}{}{2.5cm}
}
\addcase\etablissement{ENIB}{\addetablissement
    {ENIB}
    {}
    {}
    {l'\'{E}cole Nationale d'Ing\'{e}nieurs de Brest}
}
\addcase\etablissement{ENS}{\addetablissement
    {ENS}
    {}
    {}
    {l'\'{E}cole Normale Sup\'{e}rieure de Rennes}
}
\addcase\etablissement{ENSAI}{\addetablissement
    {ENSAI}
    {}
    {l'\'{E}cole Nationale de la Statistique}
    {et de l'Analyse de l'Information}
}
\addcase\etablissement{ENSCR}{\addetablissement
    {ENSCR}
    {}
    {l'\'{E}cole Nationale Sup\'{e}rieure}
    {de Chimie Rennes}
}
\addcase\etablissement{ENSTA}{\addetablissement
    {ENSTA}
    {}
    {l'\'{E}cole Nationale Sup\'{e}rieure}
    {de Techniques Avanc\'{e}es Bretagne}
}
\addcase\etablissement{IMTA}{\addetablissement
    {IMTA}
    {l'\'{E}cole Nationale Sup\'{e}rieure}
    {Mines-T\'{e}l\'{e}com Atlantique Bretagne}
    {Pays de la Loire -- IMT Atlantique}
}
\addcase\etablissement{INSA}{\addetablissement
    {INSA}
    {}
    {l'Institut National des}
    {Sciences Appliqu\'{e}es de Rennes}
    \hauteurlogos{1.8cm}{}{2cm}
}
\addcase\etablissement{InstitutAgro}{\addetablissement
    {InstitutAgro}
    {}
    {}
    {l'Institut Agro Rennes Angers}
}
\addcase\etablissement{UBO}{\addetablissement
    {UBO}
    {}
    {}
    {l'Universit\'{e} de Bretagne Occidentale}
}
\addcase\etablissement{UBS}{\addetablissement
    {UBS}
    {}
    {}
    {l'Universit\'{e} Bretagne Sud}
}
\addcase\etablissement{UR}{\addetablissement
    {UR}
    {}
    {}
    {l'Universit\'{e} de Rennes}
}
\addcase\etablissement{UR2}{\addetablissement
    {UR2}
    {}
    {}
    {l'Universit\'{e} Rennes 2}
}

\newcommand\addpairetablissements[7]{
    \logoetablissementA{#1}
    \logoetablissementB{#2}
    \nometablissementA{#3}
    \nometablissementB{#4}
    \nometablissementC{#5}
    \nometablissementD{#6}
    \nometablissementE{#7}
}

\addcase\etablissement{ENSAB-UR2}{\addpairetablissements
    {ENSAB}
    {UR2}
    {}
    {l'\'{E}cole Nationale Sup\'{e}rieure}
    {d'Architecture de Bretagne}
    {d\'{e}livr\'{e}e conjointement avec}
    {l'Universit\'{e} Rennes 2}
    \hauteurlogos{2cm}{1.2cm}{2cm}
}
\addcase\etablissement{UR2-UR}{\addpairetablissements
    {UR2}
    {UR}
    {}
    {}
    {l'Universit\'{e} Rennes 2}
    {d\'{e}livr\'{e}e conjointement avec}
    {l'Universit\'{e} de Rennes}
    \hauteurlogos{1.8cm}{1.8cm}{1.5cm}
}
\addcase\etablissement{EHESP-UR}{\addpairetablissements
    {EHESP}
    {UR}
    {}
    {l'\'{E}cole des Hautes \'{E}tudes}
    {en Sant\'{e} Publique}
    {d\'{e}livr\'{e}e conjointement avec}
    {l'Universit\'{e} de Rennes}
    \hauteurlogos{2cm}{2cm}{1.5cm}
}
\addcase\etablissement{InstitutAgro-UR}{\addpairetablissements
    {InstitutAgro}
    {UR}
    {}
    {l'Institut Agro}
    {Rennes Angers}
    {d\'{e}livr\'{e}e conjointement avec}
    {l'Universit\'{e} de Rennes}
    \hauteurlogos{1.8cm}{1.2cm}{1.2cm}
}
\addcase\etablissement{ENIB-UBO}{\addpairetablissements
    {ENIB}
    {UBO}
    {}
    {l'\'{E}cole Nationale}
    {d'Ing\'{e}nieurs de Brest}
    {d\'{e}livr\'{e}e conjointement avec}
    {l'Universit\'{e} de Bretagne Occidentale}
    \hauteurlogos{2cm}{1.6cm}{1.6cm}
}

\ecoledoctorale{MATISSE}

\etablissement{UR}

\spec{INFO}

\author{Julien DELAUNAY}

\title{Explainability for Machine Learning Models: From Data Adaptability to User Perception}
\lesoustitre{Strategies for Faithful and Understandable Explanations}

\date{20 decembre 2023}
\lieu{Rennes}

\uniterecherche{IRISA}


\jury{
{\normalTwelve \textbf{Rapporteurs avant soutenance :}}\\ \newline
\footnotesizeTwelve
\begin{tabular}{@{}ll}
Marie-Jeanne LESOT & Professor, Univ. Sorbonne LIP6, France  \\
Andrea PASSERINI & Associate Professor, Trento University, Italia \\
\end{tabular}

\vspace{\baselineskip}
{\normalTwelve \textbf{Composition du Jury :}}\\
\footnotesizeTwelve
\begin{tabular}{@{}lll}

Pr\'{e}sidente :        & Elisa FROMONT &  Professor, Univ. Rennes, France \\
Examinateurs :         & Pierre MARQUIS & Professor, Univ. Artois, France \\
                       & Katrien VERBERT & Professor, KU Leuven, Belgium \\
                       & Niels VAN BERKEL & Associate Professor, Aalborg University, Danemark \\
Dir. de th\`{e}se :    & Christine LARGOUËT & Associate Professor, Institut Agro, Rennes, France \\
Co-dir. de th\`{e}se : & Luis GALARRAGA & Researcher INRIA/IRISA, Rennes, France \\
\end{tabular}

}

\maketitle

\selectlanguage{english}
\clearemptydoublepage

\frontmatter
\clearemptydoublepage
\dominitoc
\renewcommand{\contentsname}{Table of Contents}
\tableofcontents 

\clearemptydoublepage
\chapter*{Introduction}
\addcontentsline{toc}{chapter}{Introduction}
\chaptermark{Introduction}
\adjustmtc
\minitoc

In recent decades, the rapid advancement of artificial intelligence (AI), and particularly of machine learning (ML) models, has significantly impacted our daily lives. This remarkable progress can be attributed to the exponential growth in the availability of data and the enhanced accuracy of these models. As a result, AI and ML models have become capable of remarkable achievements such as providing medical diagnoses, generating coherent texts, and efficiently identifying environmental issues. These advancements have transformed numerous industries and have the potential to further revolutionize our society.

However, this progress has also led to an increase in complexity, which has turned ML models into black boxes. Their opaque nature makes it challenging to inspect their reasoning, conduct audits, or gain insights from them. The question then arises: Can we rely on these models in critical situations, even when we are unaware of their limitations and potential failures? In scenarios like predicting personal preferences for entertainment such as Spotify or Netflix, the consequences of model inaccuracies may be minor. But in cases like predicting natural disasters or making crucial decisions in areas such as medicine, job offers or justice, understanding the model's reliability and reasoning becomes paramount. Indeed, a lack of trust or misunderstanding in a model may lead to erroneous decisions. Moreover, these models have demonstrated vulnerabilities in the form of biases against minorities and adversarial attacks invisible to human eyes. 

To address these issues, public discussions have brought these biases and drawbacks to light. The deployment of ML models has resulted in reported problems, as observed in media coverage of incidents. For instance, Amazon initiated a project aimed at automating their company's hiring process using an ML algorithm. However, this algorithm was found to exhibit gender bias, leading to discrimination against women and raising concerns about fairness and equity\footnote{\url{https://aclu.org/news/womens-rights/why-amazons-automated-hiring-tool-discriminated-against}}. Similarly, in the case of Compas, the system used for assessing American prisoner recidivism, it has been observed that black prisoners are more likely to be classified as at risk of recidivism\footnote{\url{https://www.propublica.org/article/machine-bias-risk-assessments-in-criminal-sentencing}}. A more recent controversy concerns the Dutch government's allegations of welfare fraud against numerous families, many of whom have dual nationalities or immigrant backgrounds~\footnote{\url{https://www.politico.eu/article/dutch-scandal-serves-as-a-warning-for-europe-over-risks-of-using-algorithms/}}. These accusations have resulted in unjust penalties and severe financial difficulties for these families. Consequently, questions arise about whether we, as a society, can fully trust ML models based solely on metrics such as accuracy or precision. To tackle these concerns, laws like the GDPR and the AI Act have been introduced to regulate and guide the usability of ML models. These regulations emphasize aspects such as model robustness, accuracy, and transparency~\cite{gdpr, European}.

In response to these challenges, the research community has recognized the critical importance of explaining AI models' decision-making processes. The field of eXplaining Artificial Intelligence (XAI) has experienced a significant surge as a means to provide explanations for model predictions~\cite{trends_in_xai}. In this thesis, we focus on generating explanations to identify the primary factors influencing an ML model's predictions. By clarifying the decision-making process, these explanations aim to enhance user trust and improve the reliability and accountability of ML models.

\section*{Context and Motivation}
\addcontentsline{toc}{section}{Context and Motivation}
Complex machine learning or black box models often lack interpretability, making it difficult to understand their decision-making processes and potential biases. Moreover, the lack of transparency in these models has elicited criticism from various legal structures, which recognize the importance of accountability and trust in automated decision-making systems~\cite{gdpr, European}. To address these issues, XAI has emerged as a key area of research. By providing interpretable explanations for model predictions, explainability techniques aim to enhance transparency and facilitate a better understanding of the underlying mechanisms of complex models. These efforts not only address legal concerns but also promote ethical and responsible AI practices. In the following, we will delve deeper into the specific challenges posed by complex machine learning models, explore the implications surrounding their lack of transparency, and provide an overview of how explainability techniques can help overcome these challenges.

\subsection*{Lack of Transparency}
\addcontentsline{toc}{subsection}{Lack of Transparency}
One key issue in complex machine learning models is their lack of interpretability, which stems from their intricate structure. Deep neural networks, for instance, often consist of numerous hidden layers, each transforming the input data in a non-linear manner, resulting in an enormous number of parameters\footnote{BERT base: 110 million parameters; GPT4: 1.76 trillion parameters}. Understanding the specific reasons behind the classification decisions made by these models can be elusive. For example, comprehending why GPT4~\cite{gpt4} may translate ``a nurse'' with the female word ``une infirmière'' in French and ``a doctor'' with the male word ``un médecin'' may require exposing the complex transformations and learned features within the network~\cite{chatGPT_gender}. 

The black-box nature of these models further intensifies the interpretability challenge. In ensemble models, multiple algorithms are combined to make predictions, adding a supplementary layer of complexity to understand their decision-making process. Each algorithm within the ensemble contributes with its own logic, making it challenging to trace the specific factors that influenced the final prediction. Consequently, the inner workings of ensemble models remain obscure, making it difficult to explain their outputs transparently and intuitively. The question may even arise for decision trees or linear models that are considered simple and transparent models. For example, when the depth of the tree or the number of coefficients of the linear model is too large~\cite{Lipton}.

From the lack of interpretability and the black-box nature of these complex models significant challenges in various domains arise~\cite{Rudin}. In fields where explainability is crucial, such as healthcare, finance, or legal settings, it becomes essential to bridge the gap between model predictions and human understanding. Efforts to unravel the black-box nature of these models not only improve transparency and accountability but also cultivate trust and adoption in critical applications~\cite{BeenKim_thesis}. By making complex models more interpretable, users can gain a deeper understanding and confidence in the model's predictions and ultimately drive wider acceptance and responsible use of these powerful machine learning techniques~\cite{ribeiro_thesis}.

\subsection*{Explainable Artificial Intelligence}
\addcontentsline{toc}{subsection}{Explainable Artificial Intelligence}
\begin{figure}
    \includegraphics[width=\textwidth]{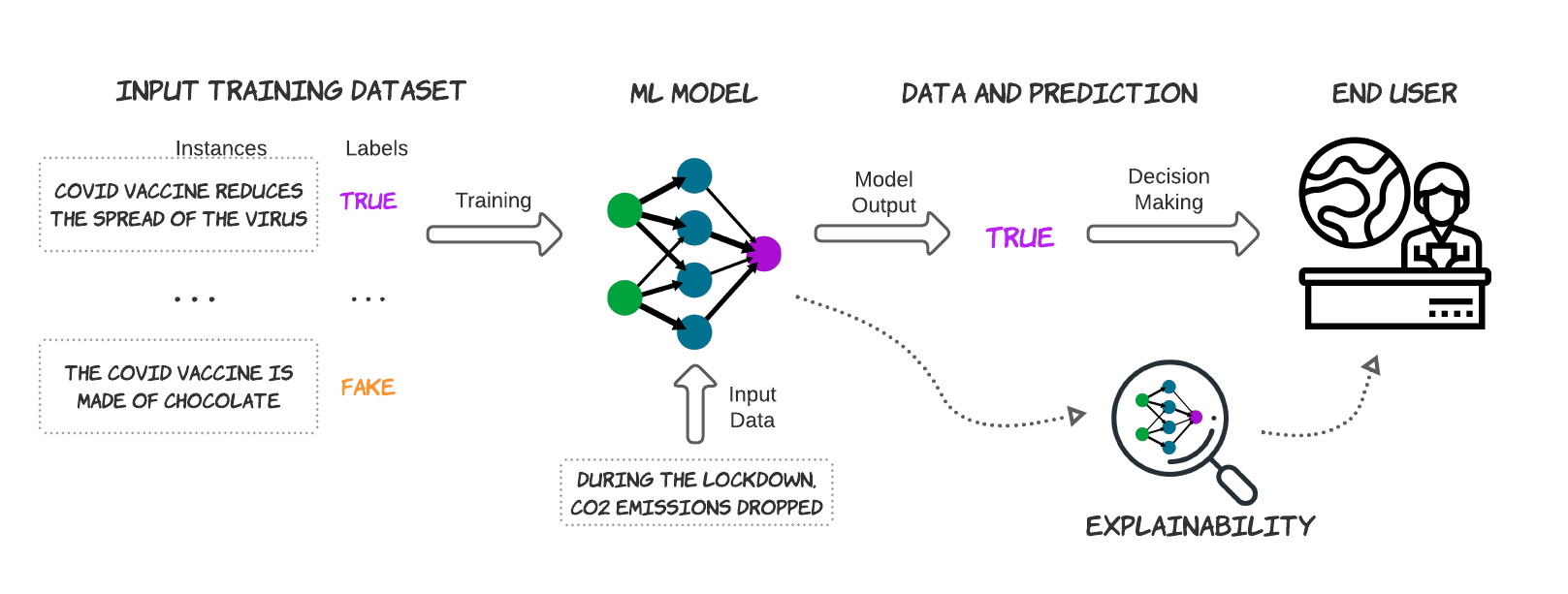}
    \caption[Illustration of eXplainable AI]{Visual representation of a machine learning model designed to detect fake news within newspaper titles. This model is trained on diverse labeled examples to accurately classify news articles as either ``fake'' or ``true''. The aspect of explainability is added to the system to provide comprehensive and transparent insights into the model's predictions.}
    \label{fig: xai_illustration}
\end{figure}
The problem of generating transparent models has been a topic of research for many years. However, recent advancements in the field, in particular Deep Learning (DL) methods, have brought about a renewed focus on this area. This renewed interest arises from recognizing the limitations of complex models that have been widely applied without a comprehensive understanding. Researchers are actively exploring techniques and methodologies to enhance the interpretability of complex models~\cite{survey_pisa, trends_in_xai}. These approaches aim to shed light on the decision-making process of complex models, enabling users to gain insights into how specific inputs are transformed and influence the final prediction. 

These interpretability methods can be applied to many domains, tasks, and data types~\cite{survey_pisa}. To illustrate, consider the toy example in Figure~\ref{fig: xai_illustration}, which portrays a model designed to detect fake news based on newspaper titles. This model is trained on existing newspaper headlines, each labeled as either fake or true. This training step enables the model to predict the label of a novel instance, such as a newspaper title in our example. The resulting prediction is then given to the end user, in this case, a journalist, who must make a decision based on this prediction. The explainability step of this process allows the journalist to investigate which elements from the input article contribute to the model's decision. For example, an explainability method can indicate the specific words in the newspaper title that led the model to classify it as true. Alternatively, another explanation technique might identify sensitive words that if removed or replaced in a certain way, would have resulted in predicting ``fake''. These methods exhibit considerable diversity and can explain, for instance, why an image recognition model classified an object as a bird~\cite{ThisLooksLikeThat} revealing how specific parts of the object resemble features from known bird images. 

However, relying solely on a single explanation method to investigate the significance of these elements comes with limitations. Various techniques may excel in explaining specific model types, may be adapted to different data formats, or may address specific instances, making it clear that no single method is universally effective across all scenarios. Consequently, it is mandatory to consider the data context, such as the model to explain or the type of instances before producing meaningful explanations. Moreover, providing identical explanations to diverse audiences and stakeholders can pose challenges and potential risks. For example, in recommendation systems, the expectations of the company developing the AI model may differ from those of consumers receiving recommendations. For example, an AI developer may be interested in debugging the model, whereas the consumer may require information to build trust in the system. Likewise, in healthcare systems, the questions and concerns of the company deploying the AI model may differ from those of doctors or hospitals utilizing the technology. Therefore, another important aspect of generating a good explanation for a model is to consider the person to whom the explanation is tailored.

\section*{Research Questions}
\addcontentsline{toc}{section}{Research Questions}
At the time of the explainability boom, most of the research on generating explanations for AI models was conducted by machine learning researchers. Therefore, the first aspect tackled by the XAI community was to generate precise and reliable explanations for model prediction. In this context, global explainability's central goal is the understanding of the overall functioning of the model. However, in this thesis, our focus is the study of local explanations, which aim to explain the prediction of a model for a \emph{target instance}. This instance may take various forms such as information about an individual or textual data as in Figure~\ref{fig: xai_illustration}. The XAI community has developed numerous and various methods to explain the prediction of a model for a given instance~\cite{survey_pisa, trends_in_xai}. These methods span from simple decision rules~\cite{decision_sets, Anchor}, linear models~\cite{iBreakDown, LIME}, and showing similar inputs that convey different outputs from the model~\cite{gs, wachter}. A notable observation emerges: existing research has mostly focused on generating the best explanation techniques that should work on every instance, model, and user. However, as demonstrated in this thesis, the quest for such universal solutions may resemble a pursuit of the mythical El Dorado. No single technique can adapt to all data and user contexts.

Since there exists already a plethora of explanation techniques, in this thesis we aim to identify some limits of the existing explanation methods. As the quality of the explanation may be impacted by different aspects such as the kind of model to explain or the users' profile, this thesis is divided into two parts. Firstly, we focus on generating local explanations adapted to the data. Secondly, we study the impact of the chosen explanation techniques and representations on users. Therefore, our first research question is: \textbf{How to generate the best explanation from a data perspective?}. Conversely, while it is largely accepted that explanations should be tailored to the users receiving them, the users-centric aspect has been underrepresented in the literature~\cite{survey_pro_hcxai2, survey_pro_hcxai}. As such, the second part of our research seeks to answer: \textbf{How to generate the best explanation from a user perspective?}. 

The research presented in this thesis focused on local explanations for supervised machine learning classification models trained on tabular and textual data. The manuscript is composed of seven chapters, including a preliminary of explainable AI and a conclusive summary. As the research included in this thesis addresses both of these research questions, we have chosen to structure the thesis into two parts, each focusing on one of these questions. The first part, devoted to the data perspective, initiates in Chapter~\ref{chap: anchors} by studying the impact of an appropriate parametrization on the quality of explanations, specifically on rule-based explanations. Chapter~\ref{chap: ape} follows and proposes to adapt the explanation technique to the target instance. Finally, Chapter~\ref{chap: emnlp} studies the influence of the conversion space utilized for embedding input text before generating an explanation. In the second part, which concentrates on the user perspective, Chapter~\ref{chap: context} introduces a methodological framework for the design of user studies that assess the impact of explanation techniques. Subsequently, Chapter~\ref{chap: chi} applied this framework to investigate the impact of different explanation techniques and their representations on users' trust and understanding. This two-part structure allows us to comprehensively explore the diverse facets of the explainability landscape and contribute with valuable insights to the field. These insights are supported by the publications produced during the Ph.D. and listed in the following section.

\section*{Publications}
\addcontentsline{toc}{section}{Publications}
We conclude this section with a list of the articles published during my Ph.D. as well as those still under review:
\boitemagique{Mentioned in this Thesis}{
    \begin{itemize}
        \item Published:
        \begin{itemize}
            \item Improving Anchor-based Explanations. Julien Delaunay, Luis Gal{\'{a}}rraga, and Christine Largouët, in: Proc. International Conference on Information and Knowledge Management \textit{{CIKM}}, 2020.
            \item When Should We Use Linear Explanations? Julien Delaunay, Luis Gal{\'{a}}rraga, and Christine Largouët, in: Proc. International Conference on Information and Knowledge Management \textit{{CIKM},} 2022.
            \item Adaptation of AI Explanations to Users' Roles. Julien Delaunay, Luis Gal{\'{a}}rraga, Christine Largouët, and Niels van Berkel, in: \textit{Conference on Human Factors in Computing Systems {CHI}, workshop on Human-Centered Explainable Artificial Intelligence {HCXAI}}, 2023.
        \end{itemize}
        \item Under review:
        \begin{itemize}
            \item Impact of the Explanations Techniques and Representations on Users' Trust and Understanding. Julien Delaunay, Luis Gal{\'{a}}rraga, Christine Largouët, and Niels van Berkel, \textit{under review in: Conference on Human Factors in Computing Systems {CHI}}, 2024.
            \item Explaining a Black Box without a Black Box. Julien Delaunay, Luis Gal{\'{a}}rraga, and Christine Largouët, \textit{under review in: Conference of the North American Chapter of the Association for Computational Linguistics {NAACL},} 2024.
        \end{itemize}
    \end{itemize}
}

The following publications, have been conducted through collaborations during my PhD, but are not mentioned in this manuscript.
\boitemagique{Other Works (joint collaboration)}{
\begin{itemize}
    \item s-LIME: Reconciling Locality and Fidelity in Linear Explanations. Romaric Gaudel, Luis Gal{\'{a}}rraga, Julien Delaunay, Laurence Rozé, and Vaishnavi Bhargava, in: \textit{Proc. {IDA},} 2022.
    \item On Moral Manifestations in Large Language Models. Joël Wester, Julien Delaunay, Sander De Jong, Niels van Berkel, in: \textit{Proc. {CHI} Workshop on Moral Agents,} 2023
    \item ``Honey, Tell Me What's Wrong'', Explicabilité Globale des Modèles de TAL par la Génération Coopérative. Antoine Chaffin and Julien Delaunay, in: \textit{Proc. Le Traitement Automatique des Languages Naturelles,} 2023.
\end{itemize}
}

\clearemptydoublepage
\mainmatter
\clearemptydoublepage
\chapter{Foundations of Explainability}
\label{chap: preliminaries}
\chaptermark{Foundations of Explainability}
\minitoc

The precise definitions of many fundamental terms used in the explainable AI literature continue to vary among different authors, indicating a lack of consensus~\cite{Lipton}. For instance, the terms interpretability and explainability are often used interchangeably since they have been introduced in a close time gap. As the field of eXplainable AI (XAI) has grown, researchers have tried to unify the definitions of the terms employed. While some researchers have proposed or preferred to use interchangeably the terms interpretability, explainability, and transparency to name just a few of them, Lipton~\cite{Lipton} proposed a definition of interpretability and how it differs from transparency. This definition has been well-accepted in the research community, amassing over 4000 citations in 2023. It suggests that a model should be considered transparent if a person can read it in its entirety -- without necessarily understanding it. On the other hand, Lipton defined a model as interpretable if a human can understand it or can take input data together with the model parameters and calculate the model's prediction in a reasonable amount of time. Finally, in the realm of XAI, the concept of explainability is centered on the process of extracting information from the model and translating this information to the user. 
  
In this thesis, the term ``interpretability'' refers to the capacity of a user to understand the inner workings of a machine learning model. In contrast, ``explainability'' is used to describe the methods that are put on top of a machine learning model to elucidate its functioning to users. Lastly, ``transparency'' characterizes methods through which the model's mechanisms are directly observable. Transparency indicates how exactly the model works by presenting details about the model's inner workings, parameters, etc. 

In this chapter, we begin in Section~\ref{sec: exploring_explanations}, by categorizing machine learning explanation techniques and defining the taxonomy used in this thesis. Subsequently, in Section~\ref{sec: surrogate_techniques}, we define formally some notations and terms to enhance the clarity of this manuscript. Additionally, we present the three families of explanation methods. Finally, Section~\ref{sec: evaluation} discusses methods for evaluating the performance of explanation techniques.

\section{Explainable AI}
\label{sec: exploring_explanations}
This section delves into the realm of explanation methods for AI systems, specifically focusing on three fundamental dimensions defined by the community~\cite{survey_XAI, survey_pisa, survey1, survey0_pisa}: self-explainable vs. post-hoc explanations, global vs. local explanations, and model-dependent vs. model-agnostic explanation methods. Each dimension plays a crucial role in shaping the interpretability landscape, and understanding their distinctions is crucial to comprehend the stakes and the scope of this thesis.

\begin{figure*}[!ht]
    \begin{subfigure}{.47\linewidth}
        \centering
        \advance\leftskip-2cm
        \includegraphics[width=1.2\linewidth]{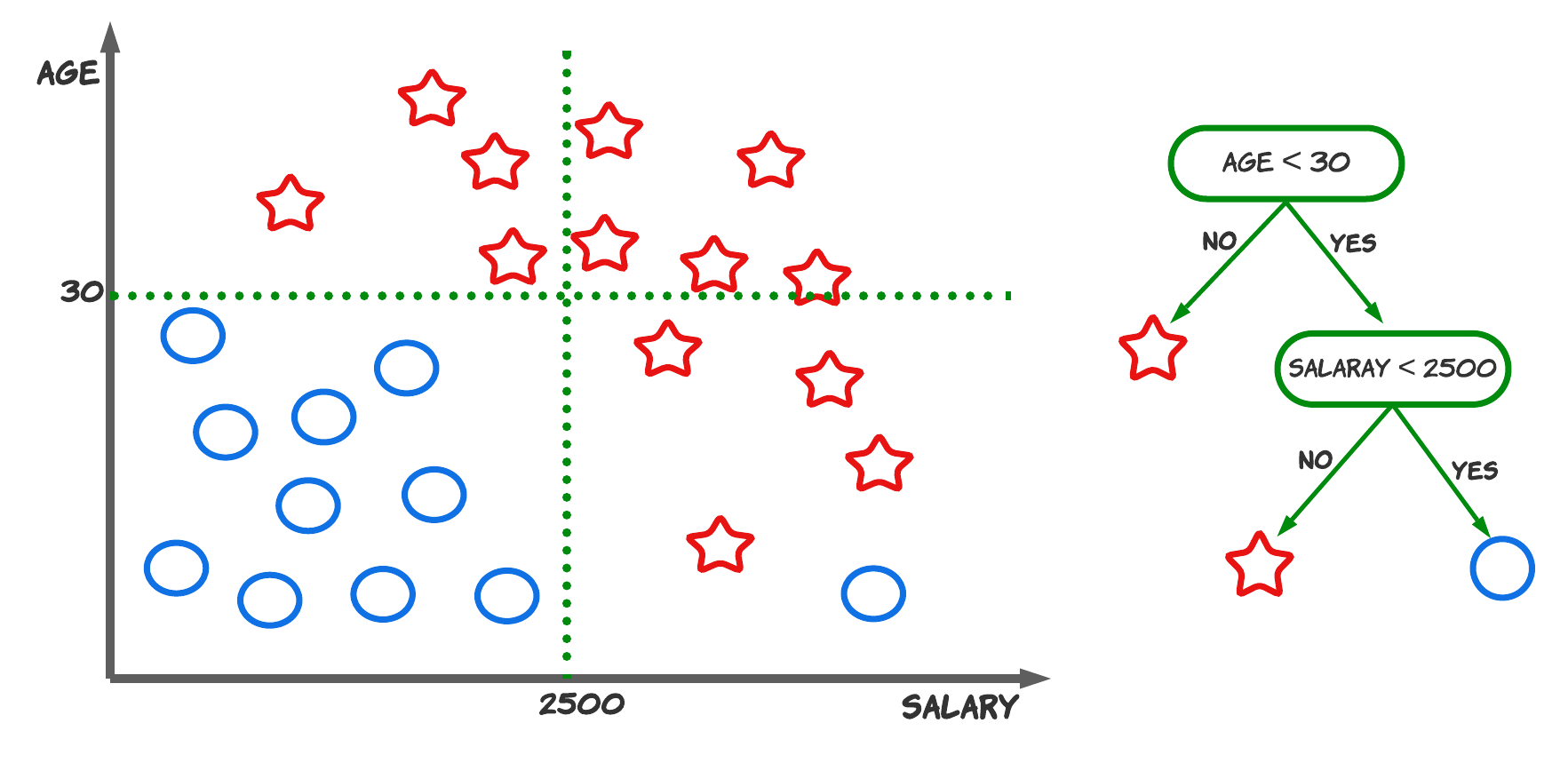}
        \caption{Decision tree}
        \label{fig: decision_tree}
    \end{subfigure}
    \centering
    \begin{subfigure}{.47\linewidth}
        \centering
        \advance\rightskip-2cm
        \includegraphics[width=1.2\linewidth]{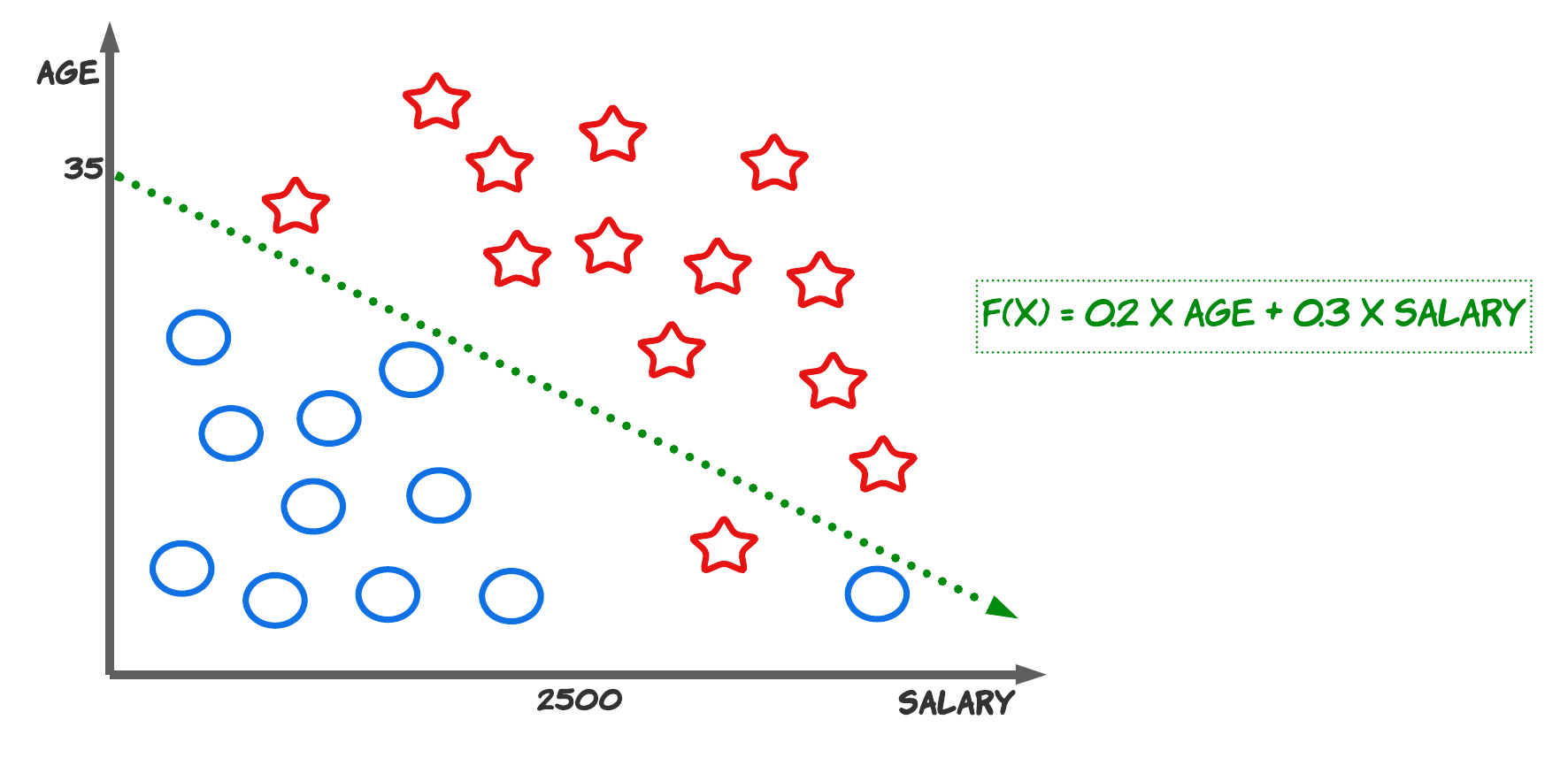}
        \caption{Linear model}
        \label{fig: linear_model}
    \end{subfigure}
    \caption[Illustration of two models, a linear model and a decision tree, employed to predict loan approval likelihood based on applicants' age and salary.]{Illustration of two models, a linear model and a decision tree, employed to predict loan approval likelihood based on applicants' age and salary. Blue circles represent loan rejections, while red stars represent loan approvals.}
    \label{fig: transparent_models}
\end{figure*}

\subsection{Self-Explainable vs Post-Hoc Explanations}
The distinction between self-explainable and post-hoc explanations sets the foundation for the taxonomy. Self-explainable models are constructed in a way that inherently incorporates transparency and intelligibility as integral features. These models possess inherent interpretability and can provide easily understandable insights into their decision-making processes without relying on external explanation techniques. These approaches resort to interpretable-by-design models such as decision trees~\cite{decision_sets} and simple linear models~\cite{pkdd_global}. To illustrate these concepts, we provide an illustrative example using a dataset that considers loan approval decisions based on individuals' age and salary in Figure~\ref{fig: transparent_models}. In this context, Figure~\ref{fig: decision_tree}, introduces a decision tree model, while Figure~\ref{fig: linear_model} shows a simple linear model. Decision trees are naturally interpretable as they allow humans to follow the conditions outlined within the tree structure until they reach a terminal leaf, which represents a model prediction or decision. This step-by-step navigation simplifies the comprehension of decision logic. Specifically, in the decision tree shown in Figure~\ref{fig: decision_tree}, the explanation reveals that if an individual is younger than 30 years old, their loan application is declined; conversely, if their monthly salary exceeds \$2500, the loan is accepted. On the other hand, when it comes to linear models, users can gain insights by inspecting the numerical coefficients associated with each feature. As a clear example, we observe that, as depicted in Figure~\ref{fig: linear_model} by the linear function: $f(x) = 0.2 \cdot \text{age} + 0.3 \cdot \text{salary}$, salary is the most important factor influencing the loan decision. This function indicates that increasing both age and salary is important for approval. 

Post-hoc explanation methods, in contrast, come into play after a complex model, commonly a black-box algorithm, has been trained to generate predictions. This is useful when we do not have access to the model's internal mechanisms. This inaccessibility can happen due to privacy constraints, for example. Consequently, these techniques employ various approaches to analyze the model's behavior and provide explanations, as reported in various surveys~\cite{survey0_pisa,survey_pisa}. The most common approach involves utilizing the complex model to label training data and subsequently training a simplified model or surrogate on this new dataset~\cite{LIME, LORE, Anchor}. It is worth noting that this approach has faced criticism for potentially oversimplifying complex models, which can result in the omission of crucial information~\cite{Rudin}. Nonetheless, post-hoc explanations offer a key advantage by enabling the interpretation of models without requiring retraining, which proves valuable in terms of time and energy consumption. This can prove particularly valuable for critical production models that are integral to a business and cannot be replaced from one day to another. Therefore, this thesis focuses on post-hoc explanation techniques that enable explainability for already deployed black-box models.


\subsection{Global vs Local Explanations}
Global explanations provide a holistic understanding of a model's behavior across its entire decision space (see e.g.,~\cite{decision_sets, pkdd_global, ribeiro_global}). These explanations are valuable for gaining high-level insights into the model's overall behavior and for identifying patterns or biases. Some common global explanation techniques include decision rules, feature importance scores, partial dependence plots, and decision boundary visualization~\cite{molnar}.

In contrast, local explanations zoom in on individual predictions, offering precise insights into why a specific instance received a particular output from the model. In other words, it seeks to answer the question, ``Why did the model predict this outcome for this specific input?''. Local explanation methods are especially valuable when dealing with complex models that lack inherent interpretability, such as deep neural networks or ensemble models. They aim to shed light on the black-box nature of these models by highlighting the specific features or input characteristics that had the most significant impact on the model's decision for that particular instance. Techniques like Local Interpretable Model-agnostic Explanations (LIME)~\cite{LIME} and SHapley Additive exPlanations (SHAP)~\cite{SHAP} stand as two popular and widely-used post-hoc local explanation methods~\cite{trends_in_xai}. LIME for instance is a method that explains a complex model by learning a linear surrogate on the outputs of the original model. 

Recently, hybrid approaches have combined various local explanations and synthesized them into a global explanation~\cite{LIME} to obtain more comprehensive insights. Thus, researchers have proposed methods such as Black box model Explanations by Local Linear Approximations (BELLA)~\cite{BELLA} or Natively Interpretable t-SNE~\cite{t-sne}. BELLA is a method that combines linear models on specific neighborhoods to explain a regression model. Similarly, Natively Interpretable t-SNE generates the best set of linear explanations and their associated coverage for dimensionality reduction. These approaches aim to improve coverage and accuracy in global explanations.

A critical challenge in designing explanation techniques lies in finding the right balance between fidelity to the complex model and simplicity or understandability. Global explanations may lack fidelity to the complex model, as simpler methods like decision trees or logistic regression may not fully approximate and explain every output variation of a model with a vast number of parameters. On the other hand, explaining the decision boundary locally for complex models allows explanation techniques to retain high fidelity to the model in a local context while remaining highly readable. 

Throughout this thesis, our focus centers on generating post-hoc and local explanations, to reveal the underlying mechanisms of intricate models and enhancing their interpretability, transparency, and applicability in real-world scenarios.

\subsection{Model Dependent vs Model Agnostic}
Lastly, we discuss the distinction between model-dependent and model-agnostic explanation techniques. Model-dependent methods are specifically tailored to explain the outputs of a particular model or family of models. These techniques leverage the internal characteristics and structure of the model to provide insights into its decision-making process. The explanations derived from model-dependent methods tend to be more faithful, as they may leverage control over the model's training~\cite{tree_regulation} or exploit the unique architecture of the model, such as in neural networks~\cite{deeplift} or tree ensembles~\cite{steve_RF_explanation}. 

On the other hand, model-agnostic methods are not tied to any specific model architecture and can be applied universally to various algorithms. This flexibility enables post-hoc explanations to be applied across a wide range of models, domains, and scenarios. These techniques prioritize transparency and generalizability, enabling a more versatile and inclusive approach to model interpretability. Furthermore, model-agnostic explanations provide a consistent and standardized approach to understanding model behaviors, making them invaluable in scenarios where the deployment of different models is commonplace. However, it is essential to note that while model-agnostic methods provide this flexibility, they may introduce some trade-offs, such as potential loss of explanation fidelity or increased computational complexity. Despite these challenges, model-agnostic techniques empower researchers and practitioners to gain insights into the decision-making processes of various models without the need for specialized adaptations or modifications of the explanation methodology. During this thesis, I co-proposed Therapy~\cite{therapy}, a global explanation method for textual data\footnote{not presented in this thesis}, which leverages constraint generation to produce representative text instances representing the most different classes. Therapy stands out as the first method capable of generating accurate explanations while being both model and data-agnostic.

\begin{figure}
    \centering
    \includegraphics[width=\textwidth]{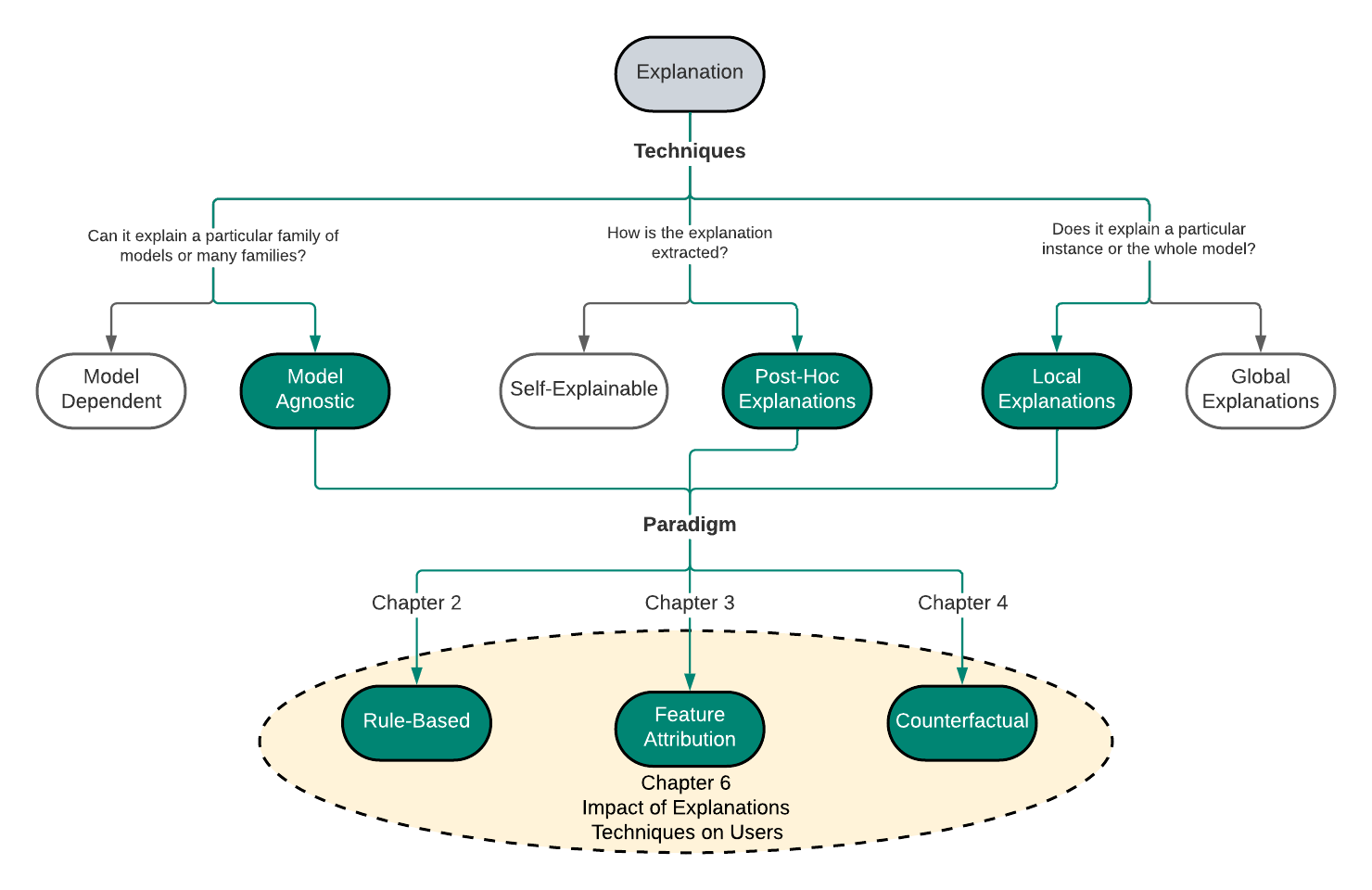}
    \caption[Taxonomy of the explanation techniques.]{Taxonomy of the explanation techniques. Paths in green represent the explanation techniques studied in this thesis.}
    \label{fig: taxonomy}
 \end{figure}

Figure~\ref{fig: taxonomy} illustrates the taxonomy and highlights the specific categories that are the primary focus of this thesis. Our emphasis is placed on model-agnostic methods due to their inherent complexity compared to model-dependent explanations. The latter explanations are tailored to specific model architectures and may not be easily transferable to other models or domains. In the subsequent section, we introduce the distinct explanation paradigms employed to interpret a model. Each of these paradigms assumes a central position in each of the following chapters, ultimately contributing to our user study presented in the second part.

\section{Explanation Paradigms}
\label{sec: surrogate_techniques}
In this section, we provide a review of three major types of explanation techniques, commonly employed in the domain of tabular and textual data~\cite{survey_pisa, survey0_pisa, survey_nn_xai}. Before delving into these different families of paradigms, we introduce some notation used throughout the thesis to enhance readability and comprehension. This notation will serve as a helpful tool for readers to navigate within the content effectively.

\subsection{Notation}
\label{subsec: notation}
\paragraph*{Classifier and Instances.}
In the context of this thesis, we employ standardized notation to describe our problem domain. Our objective centers around explaining the rationale behind the predictions made by a black-box classifier $f:X \rightarrow Y$, for a target instance represented as $x=(x_1,\cdots,x_d) \in X$. This instance is defined differently depending on the data type. For example, it may take the form of a set of features or attributes in the case of tabular data or a sequence of words in textual data. Depending on the domain, these features are represented as either numerical or categorical values (in tabular data) or words/tokens (in textual data). Additionally, $Y$ denotes a set of classes, which can represent categories such as low-risk versus high-risk or newspaper topics.

\paragraph*{Surrogate and Neighborhood.}
To explain the reasoning behind $f(x) = y$, we may leverage surrogate explanation methods. These methods train a white-box surrogate model $g$, which can be a linear model or decision tree, among others. The surrogate model $g$ approximates the behavior of the classifier $f$ within a local vicinity of the instance $x$. This locality is determined by a function $\nu_x : X \rightarrow \{0, 1\}$ such that $\nu_x(x')=1$ if an instance $x'$ is considered a neighbor of $x$ and 0 otherwise. The complete set of all possible neighbors of $x$ is then denoted as $\Phi_x = \nu_x(X) =\{x' \in X \mid \nu_x(x') = 1 \}$. It is important to note that the specific implementation of $\nu_x$ may vary depending on the chosen explanation method. 

Most surrogate methods adopt a common approach to generate local explanations. They train a surrogate model $g$ using a sample of instances created through a generative process. This process produces what is referred to as {\it artificial instances} denoted as $z \in Z \subset \Phi_x$ within the neighborhood of the instance $x$~\cite{LORE, LIME, Anchor}. Additionally, if available, these methods also consider the presence of \emph{real instances} that belong to the same neighborhood. These real instances are represented as $t \in T \cap \Phi_x$. 

\paragraph*{Counterfactuals and Friends.}
Within this context, we introduce the terms {\it counterfactual} or {\it enemy}~\cite{gs} to describe any instance denoted as $e \in E \subset X$ where $f(e) \ne f(x)$. Conversely, when the classifier assigns the same label to an instance $x'$ as it does to the target instance $x$ (\textit{i.e.,} $f(x') = f(x)$), we refer to $x'$ as a \textit{friend} of $x$. Counterfactual instances that are close to the target instance $x$ can serve as informative contrastive explanations for $f(x)$.

\begin{table}[ht!]
   \centering
   \footnotesize
   \begin{tabular}{||c|c|c|c||}
       \hline
       Symbol & Definition & Symbol & Definition \\ 
       \hline\hline
       $f(\cdot)$ & Black-box classifier & $g(\cdot)$ & Surrogate\\
       $X$, $x$ & Input domain, target instance & $Y$ & Output domain\\
       $T$, $t$ & Input dataset, instance &  $F$ & Target's friend instances\\ 
       $E$, $e$ & Target's enemies, enemy & $\Phi$, $\nu_x(\cdot)$ & Locality, Locality function \\
       $Z$, $z$ & Artificial instances, instance & $R$ & Feature-attribution ranking  \\
       $m(\cdot)$ & Adherence metric & $\tau$ & Adherence threshold \\
       \hline
   \end{tabular}
   \caption{Notation used in the thesis.}
   \label{tab: notation}
\end{table}

In the following, we will further describe the three categories to generate local explanations. Examples of these explanation methods are available for tabular data in Table~\ref{tab: explanation_example}. 

\begin{table}[!ht]
   \centering
   \begin{tabular}{@{}ll@{}} 
       \toprule
       \multirow{2}{*}{Instance $x$} & family=False, age=18, monitor=True,\\
       & meals=`Low', high-c=`No', {\color{blue}(y=30)} \\ \midrule
       \textbf{Expl. Technique} & \textbf{Explanation} \\ \midrule
       \multirow{2}{*}{Feature attribution} & (family=False) $\rightarrow-6$,\\
       & (meals $\geq$ `Low') $\rightarrow-5$ \\\midrule
       Rule & If age $\leq 20 \land$ monitor=True $\Rightarrow$ {\color{blue}non-obese} \\\midrule
       \multirow{2}{*}{Counterfactual} & meals=`Sometimes', \\
       & high-c =`Yes', ${\color{red}(y=70)}$\\ \bottomrule
   \end{tabular}
   \caption[Example of counterfactual, feature attribution, and rule-based explanations on tabular data.] {Explanations for a classifier $f$ computing the risk of obesity $y \in[0,100]$ with the outcome of `non-obese' if $y \leq 50$. The attributes consist of the patient's family's obesity antecedents (family), age (age), monitoring calorie consumption (monitor), consumption of food between meals (meals), and high-caloric food (high-c).}
   \label{tab: explanation_example}
\end{table}  

\subsection{Rule-based Explanations}
Logic rules are widely recognized for their interpretability and have a rich history of research. Consequently, rule extraction stands as an attractive approach for interpreting complex models. The rule-based methods determine the necessary conditions on the target instance's features that make the AI predict a particular outcome. These conditions take the form of one or multiple decision rules applied to the input features and are commonly represented as:

$$ \text{If }P \text{, then } Q \text{.}$$ 
Here, $P$ is referred to as the antecedent, while $Q$ serves as the consequent, which, in our context, signifies the prediction of a classifier, such as a class label. Generally, $P$ is a combination of conditions related to various input features. Additionally, it is worth noting that explanation rules take on various forms, including propositional rules, first-order logic rules or fuzzy rules~\cite{stepin}.
 
\begin{figure}[!h]
   \centering
   \includegraphics[width=\textwidth]{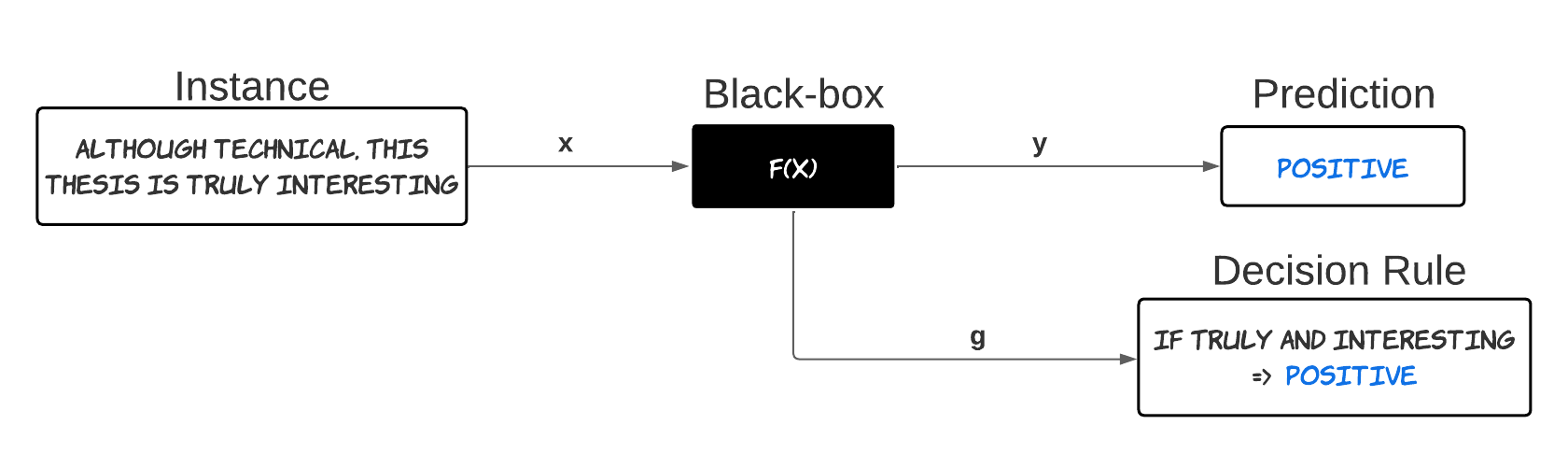}
   \caption[Rule-based explanation for a sentiment classification model.]{Rule-based explanation for a sentiment classification model. This rule specifies that the model predicted positively due to the presence of the words `truly' and `interesting'.}
   \label{fig: rule-explanation}
\end{figure}
As an example, in Table~\ref{tab: explanation_example} on the obesity dataset, a rule-based explanation specifies that ``if an individual who is under the age of 20, monitors their calorie consumption, then our example random forest classifier predicts no risk of obesity''. Another example of a rule-based explanation, shown in Figure~\ref{fig: rule-explanation}, indicates that a classifier assigns a positive classification to a text document if it contains the words ``truly'' and ``interesting''. These rule-based explanations offer a structured and interpretable way to understand the AI model's decision-making process based on specific feature conditions~\cite{Rudin, decision_sets}. It is important to note that while there may be differences between rules, decision trees, and their respective extraction techniques, we do not make a clear distinction in this context. This is because both rules and decision trees provide similar types of explanations since a decision tree can be seen as a collection of decision rules. 

Anchors~\cite{Anchor} generates random artificial instances to learn a rule-based explanation. It resorts to a multi-armed bandit exploration that generates those instances gradually. Anchors then computes a single general and accurate decision rule that mimics the black box's behavior on the target instance. The instances are used by a breadth-first-search rule mining procedure that favors shortest rules. Anchors mines for the shortest decision rule as such a rule will cover more instances~\cite{garreau_anchors}. The antecedent of the rule consists of conditions on the input features, e.g., $\mathit{salary} > 50k\; \land\;\mathit{status}=`\mathit{single}$', which are used to predict the class assigned to the target instance by the black box. This assignment to a class is made with a high level of confidence, over a given threshold, often set at 95\%.

Other methods such as LORE (LOcal Rule-based Explanations)~\cite{LORE}, xSPELLS (e\textit{x}plaining Sentiment Prediction generating Exemplars in the Latent Space)~\cite{xspells}, and ABELE (Adversarial Black box Explainer generating Latent Exemplars)~\cite{ABELE} rely on decision trees. These techniques have been proposed by Guidotti et al.~\cite{LORE, xspells, ABELE} and extract rules from a decision tree trained on artificial instances that resemble the target instance. They benefit from the tree structure to propose rule-based explanations. Indeed, starting from the leaf in which the instance falls, these methods generate rules by going up until the root of the tree. Alternatively, they can also search for paths in the tree that lead to a leaf node associated with a different black-box prediction, providing this contrastive path as a counterfactual explanation. These methods work differently depending on the data type. For instance, LORE uses a genetic algorithm to create similar instances to the target, whereas xSPELLS and ABELE employ a variational autoencoder to encode the target instance into a latent space and make slight perturbations, specifically tailored to textual and image data, respectively.

In their work, Dhurandhar et al.~\cite{pertinents_negatifs} introduce a novel concept known as ``pertinent negative and positive explanations''. These explanations are designed to be rule-based explanations and provide insights into why a certain input $x$ is classified as class $y$ based on the presence of certain features $f_i, \cdots, f_k$, and the absence of some other features $f_m, \cdots, f_p$. The researchers achieve this by identifying small, sparse perturbations that either preserve the same prediction when applied to the original input or change the prediction when applied to a target input.

\subsection{Feature-Attribution}
\begin{figure}[!h]
   \centering
   \includegraphics[width=\textwidth]{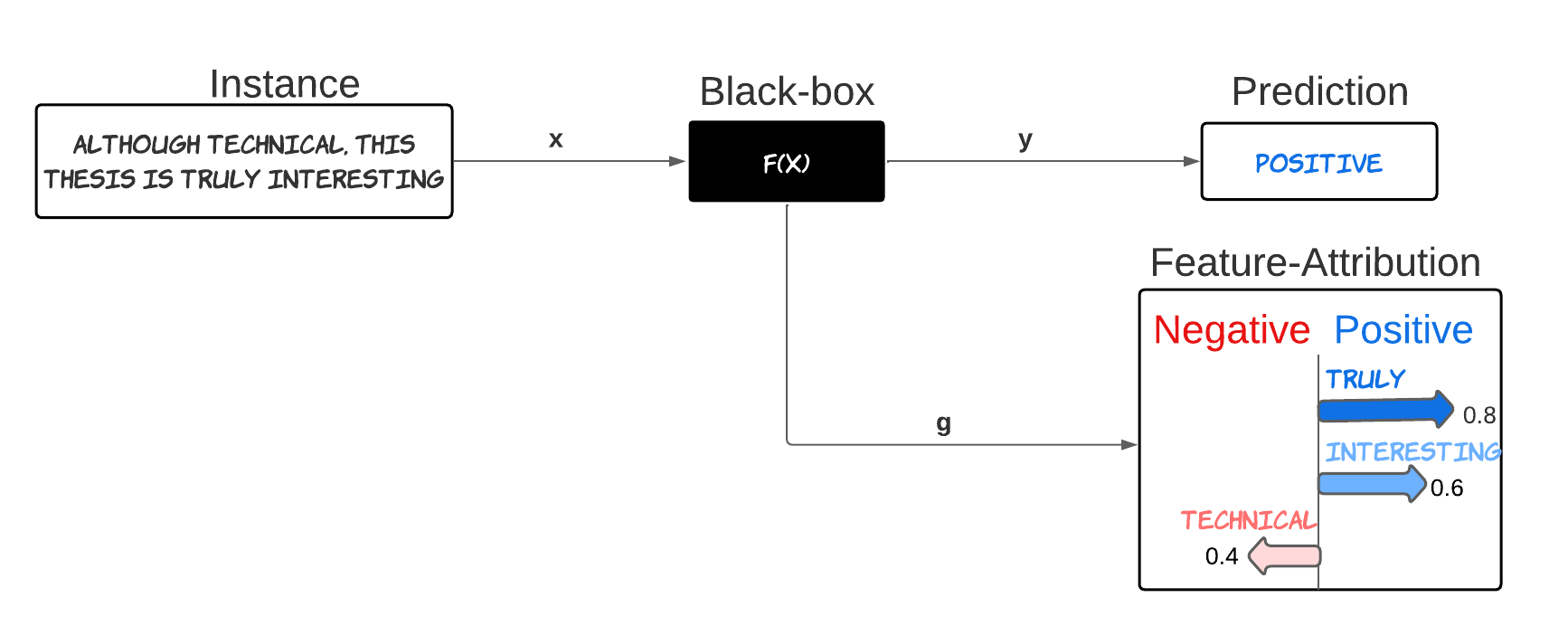}
   \caption[Feature-attribution explanation for a sentiment classification model.]{Feature-attribution explanation for a sentiment classification model. The length of each bar represents the extent to which the presence of a specific word in the sentence influences the model's prediction toward the corresponding class (positive or negative).}
   \label{fig: linear-explanation}
\end{figure}

Feature-attribution techniques compute the contribution of a black box's input features to the classification of a target instance. The magnitude of the contribution tells us the importance of the feature for a particular prediction outcome, which can correlate positively or negatively with the answer provided by the black box. For instance, consider the information in Table~\ref{tab: explanation_example}, which suggests that in the case of a random forest classifier predicting the risk of obesity using factors like family history and daily routines, a decreased risk of obesity is linked with habits such as low food consumption between meals and having non-obese parents. In another case, for a sentiment prediction model as the one in Figure~\ref{fig: linear-explanation}, the word `truly' pushes the model towards positive predictions with a positive weight of 0.8. Conversely, the word `technical' makes a negative prediction more likely with a weight of 0.4. 

Several well-known feature-attribution methods, such as LIME~\cite{LIME}, SHAP~\cite{SHAP}, Integrated Gradient (IG)~\cite{ig}, among others, are the most frequently employed in the field of XAI~\cite{nauta, trends_in_xai}. Among these methods, LIME (Local Interpretable Model-Agnostic Explanations)~\cite{LIME} is the most prominent approach to compute surrogate explanations~\cite{trends_in_xai}. Its core concept involves the training of a linear model to approximate the complex model's behavior around a specific instance. 

LIME initiates the explanation process by creating an artificial neighborhood in the locality of the target instance. The specifics of this process depend on the data type under consideration. For tabular data, LIME perturbs the numerical attributes of the target according to a $\mu$-centered and $\sigma$-scaled normal distribution, where $\mu$ and $\sigma$ are the attribute's mean and standard deviation in the training set. For categorical attributes, LIME uses the empirical distribution of the attribute values. For the textual data, LIME randomly `hides' part of the input instance (words) to generate an artificial neighborhood.

Subsequently, LIME assigns weights to these artificial neighbors based on their proximity to the target. The weighting is done using an exponential kernel that considers the $l_2$-distance between neighbors and the target instance. This weighting scheme ensures that closer neighbors are given more importance. LIME then trains a linear model on this weighted neighborhood. The coefficients associated with each input feature in this model form the basis of the explanation. Additionally, LIME resorts to a regularization term that limits the number of features used by the linear model. This regularization helps simplify the complexity of the explanation.

The effectiveness of LIME has led to a significant body of research exploring the impact of the different components and parameters of LIME on the quality of the resulting explanations. To provide a comprehensive overview of the landscape of LIME extensions, we survey some of these extensions.

While SHAP and LIME produce the same type of explanation, the semantics of their explanations are different~\cite{leaf}. Indeed, if properly parametrized, LIME approximates the instantaneous gradient of the black box w.r.t. the input features~\cite{explaining-lime}. Conversely, SHAP takes root from game theory and computes -- or rather approximates -- the Shapley values~\cite{shapley_values}. These values quantify the feature contributions to the difference between the model's answer on a baseline instance and the target. The baseline depends on the use case, e.g., a single-color image, or an empty sentence. This contribution is measured by iteratively replacing some parts of the target instance with values from the baseline and observing its impact on the classifier prediction. The model-agnostic version of SHAP (KernelSHAP) computes the Shapley values by (i) generating artificial instances as in LIME but concentrating the training weights on the closest and farthest instances in the neighborhood, (ii) dropping the regularization term used in LIME to reduce the complexity of the linear surrogate. SHAP's explanations offer interesting theoretical guarantees such as local accuracy, i.e., the surrogate is always accurate on the target instance~\cite{SHAP}. 

Among the local explanation techniques, it has been demonstrated that applying LIME within a neighborhood defined by the classifier's decision boundary can lead to more locally faithful explanations~\cite{Laugel_Defining_locality}. In that vibe, the Local Surrogate approach centers the generative process not on the target instance but on its closest enemy, which by itself provides a complementary explanation for the model prediction. Local Surrogate then constructs a linear surrogate within a hyper-sphere centered at the closest counterfactual, further enhancing the fidelity and stability of explanations. This method will be further developed in Chapter~\ref{chap: ape}.

Additionally, various extensions of LIME have emerged to address the inherent instability of the original LIME algorithm~\cite{ALIME, OptiLIME, DLIME}. In the original LIME algorithm, two executions with the same input may yield different explanations due to randomness in different algorithmic steps. These extensions offer diverse approaches to tackle this instability issue. For example, the authors of Optimized Local Interpretable Model Explanation (OptiLIME)~\cite{OptiLIME} delve into the relationship between the bandwidth parameter, the adherence, and the instability in LIME. OptiLIME underscores the critical role of selecting an appropriate bandwidth value for each instance and uncovers an inverse correlation between bandwidth and explanation instability. Building upon this insight, OptiLIME introduces a method to select the optimal bandwidth value that strikes the ideal balance between adherence and stability. Throughout this thesis, I collaborated on Smoothed LIME (s-LIME)~\cite{slime}, an extension of LIME that specifically examines the impact of this bandwidth parameter on the quality of generated explanations. s-LIME achieves this by generating neighbor instances in a continuous space. The size of the neighborhood defined by those instances and the magnitude of the perturbation depend on the distance defined by the bandwidth parameter. Thus, s-LIME generates a neighborhood that is more nuanced and allows for higher faithful linear explanations.

\subsection{Example-based Explanations}
\label{subsec: counterfactual}
Example-based explanations provide users with similar instances or examples that are either classified as the target instance (prototype) or differently (counterfactual). Prototypes are instances representative of a given class~\cite{VCNET}. Conversely, counterfactuals illustrate the minimum changes in the target instance necessary to modify the AI's prediction. They, therefore, identify the most \emph{sensitive} features in the AI agent's decision process. Counterfactual explanations have gained prominence for two main reasons. Firstly, they align more closely with the way humans naturally explain concepts, drawing from cognitive processes and human reasoning~\cite{Counterfactual_Thought, miller, cf_review}. Secondly, they fulfill legal requirements for explaining prediction, particularly in accordance with regulations such as the GDPR~\cite{wachter}.

\begin{figure}[!h]
   \centering
   \includegraphics[width=\textwidth]{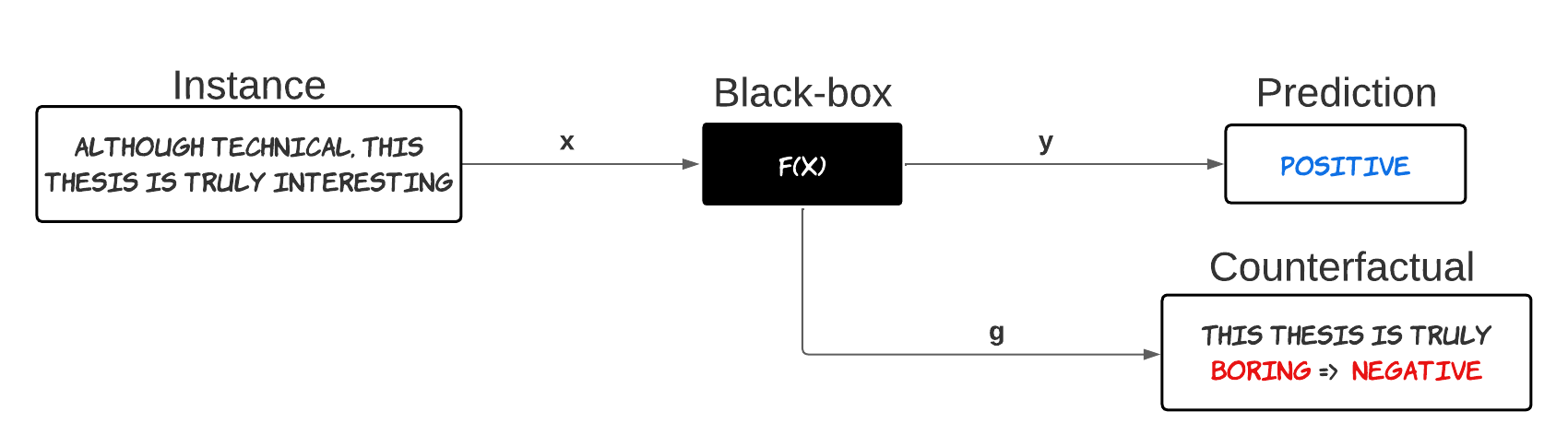}
   \caption[Counterfactual explanations for a sentiment classification model that predicts a text as positive.]{Counterfactual explanations for a sentiment classification model that predicts a text as positive. This explanation illustrates that by substituting the word `interesting' with `boring', the text would have been classified as negative. Hence, the word `interesting' is indeed important for the positive class.}
   \label{fig: counterfactual-explanation}
\end{figure}

We provide in Figure~\ref{fig: counterfactual-explanation}, an example of a counterfactual explanation for a model that initially predicts a text as positive. This explanation highlights how replacing the word `interesting' with `boring' alters the model's prediction. Table~\ref{tab: explanation_example} offers another instance of a counterfactual explanation, this time for a random forest classifier predicting an individual as having no risk of obesity. This explanation shows that changing an individual's frequency of food consumption between meals from `low' to `sometimes', along with consuming high-caloric food, would result in the model predicting a high risk of obesity for that individual. 

In recent times, a multitude of innovative counterfactual explanation techniques have emerged, leading to an expansion of the field. This rapid growth is confirmed by the fact that, within just one year, more than 50 new methods were proposed between the surveys conducted by Bodria et al.~\cite{survey_pisa} and the one from Guidotti~\cite{survey_cf_Guidotti}. Therefore, among the example-based explanations, we first differentiate between the ``non-synthetic'' methods which return training samples~\cite{natural_cf_survey} and the methods that generate the instances~\cite{cf_review}. We find among the non-synthetic category, methods such as Nearest Unlike Neighbor (NUN)~\cite{NUN}, which is derivative from the nearest neighbors method and looks for the nearest element that belongs to a different class. Additionally, we have methods like MMD-critic~\cite{mmd_critic} and Nearest-CT~\cite{nearest_ct} which select prototypes and counterfactuals from the original data points. 

However, in this thesis, we focus on methods that generate artificial instances to produce counterfactual explanations because this generating process may be utilized to construct linear or rule-based explanations, as demonstrated in Chapter~\ref{chap: ape}. One of these methods is Growing Spheres~\cite{gs}, which produces the closest valid counterfactual based solely on the target instance and the model to explain. This method, detailed further in Section~\ref{sec: counterfactual_methods}, iteratively perturbs the target instance until it identifies an artificial instance that elicits a different prediction from the complex model. Growing Spheres generates instances within a hypersphere centered on the target instance. This hypersphere radius is extended until the closest counterfactual is found. 
Another approach, Reinforcement Learning Agent eXplainer (ReLAX)~\cite{Relax} leverages reinforcement learning (RL) techniques to generate counterfactual instances. The primary goal of ReLAX is to find a counterfactual instance that is at the same time close to the original instance and classified differently from it. ReLAX does this by formulating the task as a sequence of actions that an RL agent must take to maximize its expected reward. In this context, the optimal policy represents the objective of finding an instance that is both differently classified and close to the original instance. The RL agent in ReLAX operates iteratively, making decisions about which features to modify and by what magnitude. Until the closest counterfactual is found, the agent, at each iteration, selects from the set of features that have not yet been modified and determines the extent to which a chosen feature should be increased or decreased. 

Other methods, such as Diverse Counterfactual Explanations (DiverseCF)~\cite{DICE} aim to obtain the highest feasibility and diversity. Thus, DiverseCF achieves diversity by leveraging a mathematical concept known as determinantal point processes. This concept has previously been used in solving problems involving the selection of diverse subsets. In this context, determinantal point processes help ensure that the counterfactuals generated cover a wide range of possibilities. In addition to diversity, DiverseCF incorporates various constraints to make the counterfactual explanations more meaningful and applicable. These constraints include proximity to the instance being explained, sparsity in the generated counterfactual compared to the target, and user-defined constraints to enhance trustworthiness. 

Furthermore, it is worth noting that some of the methods we explored in rule-based explanations can also be considered as counterfactual explanations. For example, Pertinent Negative~\cite{pertinents_negatifs} identifies elements that must be absent to maintain the prediction for the target. On the other hand, xSPELLS~\cite{xspells}, LORE~\cite{LORE}, and ABELE~\cite{ABELE} use decision trees to generate explanatory rules, with the paths within the decision tree that lead to different classifications being utilized as contrastive explanations.

\section{Evaluating Explanations Techniques}
\label{sec: evaluation}
The evaluation of explanation techniques is a crucial aspect of assessing their effectiveness and utility. While the fundamental goal of generating explanations is to make them comprehensible to humans, the evaluation of most existing methods has predominantly relied on performance criteria~\cite{finale, ribeira_user_centered}. This focus on performance evaluation can be attributed, in part, to the significant presence of machine learning researchers within the eXplainable AI (XAI) community. Leveraging metrics commonly employed in their respective domains, these researchers quantify the quality and performance of explanation techniques.

As surprising as it may sound, certain researchers have presented arguments against using human validation to assess the accuracy of explanation methods~\cite{nauta}. Indeed, it is pertinent to note that while an explanation may appear plausible to a person, it does not guarantee that it accurately reflects the underlying reasoning of the model~\cite{evaluation_nlp_xai}. Conversely, when confronted with models that learn nonsensical correlations to derive predictions, explanations that appear implausible should not be penalized, as their purpose is to reflect the model's intrinsic logic. Furthermore, the evaluation of explanations can be significantly influenced by the user and the specific context. The provision of examples that seem reasonable can be a way to evaluate explanations, but there exists a risk of cherry-picking examples to pass a face-validity test, potentially undermining the evaluation process~\cite{face-validity}.

In this section, we explore a series of criteria that illustrate the effectiveness of explanation methods. We differentiate between two fundamental approaches: surrogate evaluations and instance-based assessments due to their distinct nature and the specific focus of their evaluation. In the case of surrogate-based explanations, the evaluation process tends to provide more general insights, offering a comprehensive summary of the model's behavior across multiple instances. These evaluation standards often encompass comparisons with a ground truth or assessments of the stability of explanations across various instances. In contrast, example-based explanations are renowned for their specificity, offering insights into individual instances and how they can be altered to achieve different predictions. When it comes to assessing example-based techniques, the process primarily involves evaluating the instances themselves rather than the explanation models.

\subsection{Surrogate-Based Evaluation Criteria}
This section examines a variety of criteria that provide insights into the performance of surrogate explanation methods. These metrics establish performance criteria capable of objectively gauging the effectiveness of these methods in providing comprehensible explanations. Some prominent criteria include fidelity, adherence, uncertainty, stability, and complexity. This focus on performance-based criteria not only aligns with the research community's background but also acknowledges the need for objective assessment in a field where the subjectivity of human interpretation can often be intricate.

\subsubsection{Adherence and Fidelity}
The \textbf{adherence} of an explanation refers to the degree to which the explanation accurately represents the decision-making process of the underlying model. This property is crucial and often the first aim of explanation techniques. It is often measured by comparing the prediction of the explanation surrogate and those of the classifier to explain~\cite{LORE, LIME}. This comparison may take the form of measuring the accuracy or precision between the two predictions on a set of artificial or real instances. A similar criterion, sometimes confounded with adherence is the \textbf{fidelity}. This standard also named agreement, measures whether the features involved in the explanation are effectively important for the model to explain. One way to measure the fidelity of an explanation is to resort to a transparent or glass-box model. By doing so, we can control the features involved in the prediction and compare them with those indicated as important in the explanation~\cite{LIME}. Another approach involves inserting or deleting elements in the target indicated as important in the explanation. This helps assess the impact of these elements on the classifier's prediction for this target. The intuition behind deletion is that removing the ``cause'' will force the model to change its decision~\cite{insertion_deletion}. Similarly, adding an element indicated by the explanation as important for another class should lower the confidence of the model in the original class.

\subsubsection{Stability and Uncertainty}
When generating an explanation for a model, it is important to assess the extent to which it can be relied upon. Therefore, significant efforts have been made to measure the \textbf{stability} of explanation techniques~\cite{alvarez_robustness, visani_vsi_csi}. Robustness serves as a metric to quantify the stability of explanation techniques. Highly robust explanations should exhibit minimal changes in response to slight perturbations for the instance to explain. This indicates that the explanation may serve as a reliable substitute for the complex model within the vicinity of the explanation. Alvarez and Jaakkola~\cite{alvarez_robustness} proposed a novel metric that relies on the concept of local Lipschitz continuity. This metric perturbs artificial instances located within a ball centered on the target instance and then measures the ratio between the variance observed in the original feature space and the explanation space. In essence, it quantifies the ratio between the distance of (a) the perturbed instances from the target instance, and (b) the original explanation and the modified one. Similarly, Visani et al.~\cite{visani_vsi_csi} introduced two metrics known as the Variables Stability Index (VSI) and Coefficients Stability Index (CSI) to assess the stability of linear explanations. Since linear explanations resort to random perturbations to generate artificial instances, the explanations they produce may vary with each execution. VSI tracks the top features indicated by successive execution of the explanation module, evaluating whether these variables or features consistently appear, while CSI assesses whether the associated attribution scores remain similar across repeated runs.

Another aspect that contributes to users' trust in explanations is the measurement of \textbf{uncertainty}. In the work presented in~\cite{iBreakDown}, the authors proposed the use of bootstrapping to generate a sample of different explanations and measure the stability of contribution values. This approach involves first selecting a model along with an explanation and generating random samples with varying values. Subsequently, explanations are generated for each of these new samples, and uncertainty is measured as the variation in contribution values between these explanations.
Moreover, noteworthy findings from research~\cite{ DBLP:conf/eacl/KavumbaBHI23, DBLP:conf/aaai/LiHCXTZ22} have highlighted the positive relationship between improving a model's explainability and the robustness of explanation methods. This implies that enhancing the transparency and interpretability of a model can lead to more stable and reliable explanation techniques.

\subsubsection{Simplicity and Conciseness}
One facet of explanation quality involves assessing the \textbf{simplicity} and \textbf{conciseness} of the explanations. Simple, concise explanations are more accessible to users and facilitate their understanding of the model's decisions. Various metrics and qualitative assessments can be employed to gauge this aspect of complexity effectively. One quantitative approach involves examining the length of explanations, which can be measured by counting the number of terms or conditions contained within the explanation. For instance, in rule-based explanations, the complexity could be quantified through the number of predicates in the rule~\cite{Anchor}. Similarly, for feature-attribution explanations, the number of coefficients associated with input features can serve as a measure of complexity. In the case of decision trees, the depth is a significant factor in assessing complexity. The shorter the path within a tree, the simpler and more concise the explanation it provides~\cite{LORE}. 

Furthermore, an alternative approach to evaluating conciseness is to employ ground-truth models tailored to different explanation types (e.g., linear or rule-based). These transparent models serve as benchmarks~\cite{guidotti_measuring_xai}, enabling comparisons between the identified important feature coefficients in feature-attribution explanations and their true values. In the case of rule-based explanations, the features included in the paths of decision trees can be assessed for conciseness by comparing them to the ground truth. This approach helps quantify the level of alignment between the explanation and the actual importance of features, shedding light on the simplicity and conciseness of the explanations provided.


\subsection{Instance-Based Criteria}
The instance-based explanation approaches are designed to find optimal instances based on one or multiple desired criteria. In the context of counterfactual explanation techniques, Guidotti~\cite{survey_cf_Guidotti} has outlined eight criteria, including validity, similarity, sparsity, diversity, actionability, causality, plausibility or realisticness, and discriminative power.

\subsubsection{Validity and Similarity}
A counterfactual is deemed \textbf{valid} if its classification by the classifier differs from the prediction on the target instance. This notion of validity is inherent to every counterfactual. Moreover, we consider a counterfactual to be \textbf{similar} to the target when the distance between them, determined by a distance function, is low~\cite{miller, wachter}. This criterion is also referred to as ``minimality''.

\subsubsection{Sparsity and Diversity}
The \textbf{sparsity} condition indicates that the counterfactual should possess the fewest distinct features when compared to the target~\cite{gs}. This principle of sparsity ensures that no other valid counterfactual exhibits fewer different attribute value pairs~\cite{wachter}. Conversely, when generating multiple counterfactuals, we expect \textbf{diversity} among them, meaning they should be close to the target but distant from one another. For instance, consider two counterfactuals proposing that an individual should be younger to be classified as having no risk of obesity. In this scenario, the insights gained might be limited compared to two explanations that highlight different attributes such as monitoring calorie consumption and increasing vegetables intake. Indeed, the former explanation fails to provide a nuanced understanding of the model's decision~\cite{DICE}.

\subsubsection{Actionability and Causality}
\textbf{Actionability} aims to increase users' trust in the explanation. These counterfactuals ignore unchangeable features such as age or gender. Thus, an actionable counterfactual only differs from the target in other attributes~\cite{face}. \textbf{Causality} is connected with these two properties, as a causally generated counterfactual validates actionability and plausibility by maintaining any causal relationship between features.

\subsubsection{Realisticness and Discriminative Power}
A counterfactual is considered \textbf{realistic} if its attribute values align with those of instances from a broader population. In other words, a realistic counterfactual does not deviate significantly from typical instances in the original dataset. The \textbf{realistic} metrics serve as a means to assess whether generated examples lie within the data manifold and remain coherent with the underlying data distribution. Such metrics prevent artificial instances from being considered outliers~\cite{cf_review}. One practical approach to measuring realisticness involves calculating the average distance of the generated counterfactual to the k-closest instances from a dataset~\cite{cf_review}. This metric gauges whether a generated instance aligns with the original data distribution, helping identify potential outliers. Moreover, Laugel et al.~\cite{laugel_plausible_cf,laugel_unjustified} have contributed to this area with their work on the Local Risk Assessment (LRA) metric. LRA provides a means to evaluate whether a counterfactual explanation is justified by verifying the existence of an $\epsilon$-chain between the counterfactual instance and an instance from the training dataset. An $\epsilon$-chain is established by generating intermediate instances between the counterfactual and the real instance. This chain effectively serves as a path connecting two instances, where the model's prediction remains consistent for every artificial instance along this path. 

Finally, the \textbf{discriminative power} principle stipulates that changes should be discernible and comprehensible to humans. It is important to note that even small alterations, such as modifying a few pixels in an image, may not be noticeable by humans but can significantly impact the model's prediction~\cite{adversarial_into_cf}. This aspect is vital for distinguishing between adversarial attacks which aim to fool a model by slightly perturbing the input and counterfactuals which explain why the model made a prediction~\cite{adversarial_into_cf}.

\section{Outline of this Thesis}
After having introduced the key concepts of explainability and defined how they are used in the rest of this thesis, this chapter presented an overview of three kinds of explanation techniques: rule-based, feature-attribution, and example-based. Finally, it presented the evaluation process for these techniques, differentiating between surrogate and instance-based methods. 

\begin{sidewaysfigure}[ht]
    \includegraphics[width=\linewidth]{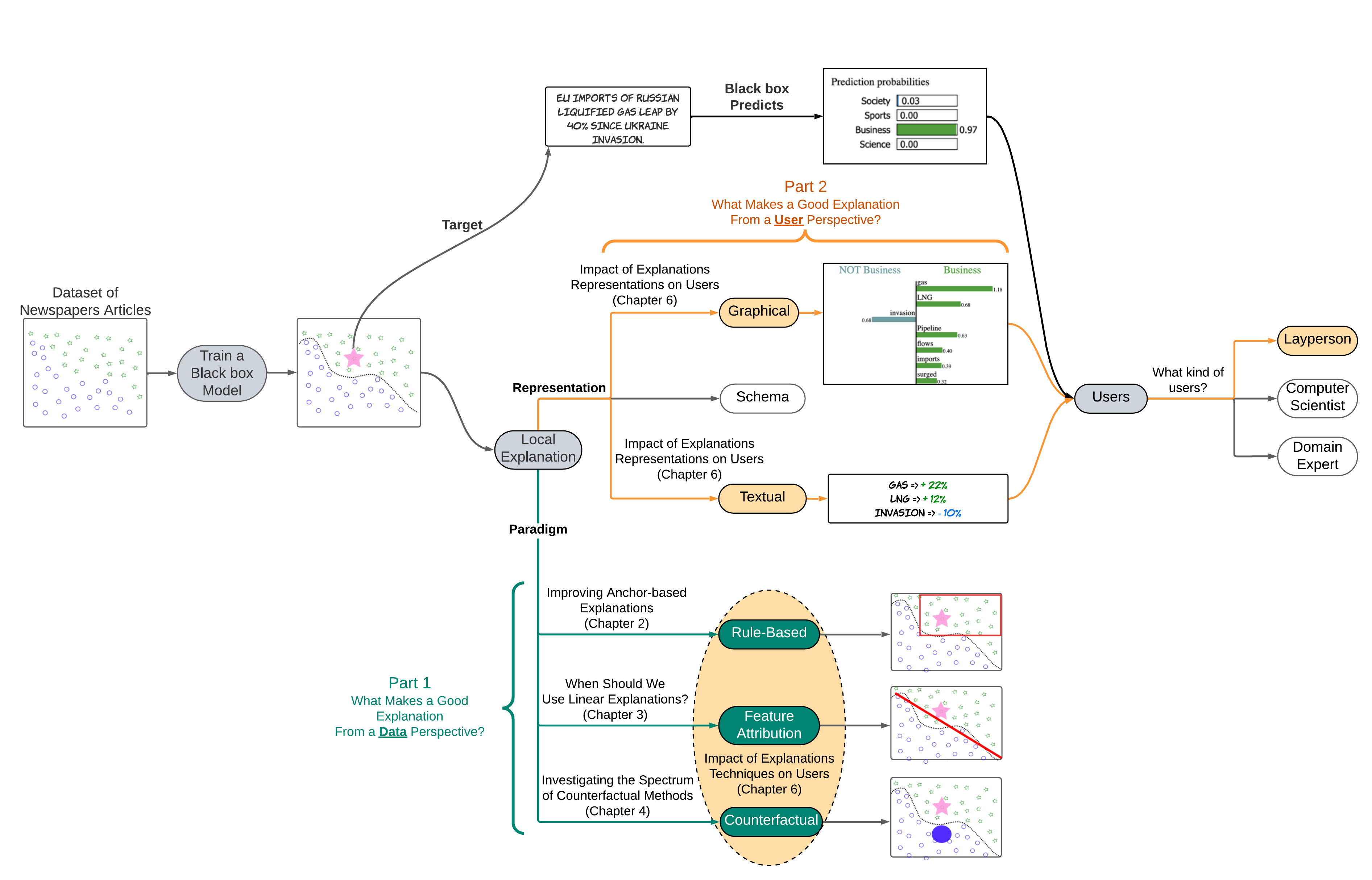}
    \caption[Schema of the research plan followed in the thesis.]{Diagram depicting the aspects of explainability studied in this thesis. The first part depicted by the green boxes, mainly focuses on the different explanation techniques. The focus of the second part is more on the users' aspects and is represented by the orange ovals.}
    \label{fig: thesis_plan}
\end{sidewaysfigure}

As we have seen through this overview, numerous aspects are important to generate an explanation adapted to the situation. Such aspects encompass the type of explanation (rules, feature-attribution or counterfactual), the criteria they aim to maximize (fidelity, stability, realisticness) and the model to explain (model agnostic or model dependent). Consequently, this thesis tackles two essential dimensions depicted in Figure~\ref{fig: thesis_plan} to generate adapted explanations. First, it delves into the data-centric perspective, with each chapter focusing on one explanation paradigm:
\begin{itemize}
    \item In the first chapter of this section (Chapter~\ref{chap: anchors}), we introduce two improvements to the widely used rule-based explanation method, Anchors~\cite{Anchor}. These improvements address the inherent limitations on tabular data by relying on better discretization techniques and broadening the search space of rules for textual data.  
    \item Chapter~\ref{chap: ape} proposes an approach to determine when it is appropriate to employ linear explanations. This decision is driven by a study of the decision boundary's shape in the locality of the instance to explain. Within this chapter, Section~\ref{sec: ape} introduces a framework for identifying scenarios in which a linear explanation is most suitable and proposes a rule-based alternative in other cases.
    \item  More and more counterfactual explanation techniques are developed with the goal of generating more and more accurate, faithful, and plausible explanations. However, similarly to the original model that explainability aims to open, these methods are becoming more and more complex. Therefore, in Chapter~\ref{chap: emnlp}, we study the benefits of using complex black boxes to explain other black boxes.
\end{itemize}

Moving beyond the data perspective, a crucial aspect to consider when generating an explanation is the person who receives it. As a consequence, the second part of this thesis studies how to generate the best explanation from a user perspective. We thus conducted user studies on laypersons with diverse explanation representations and techniques as illustrated in Figure~\ref{fig: thesis_plan}.
\begin{itemize}
    \item In the first chapter of this second part (Chapter~\ref{chap: context}), we argue for the need for user studies. We thus provide an overview of how existing researchers have evaluated explanation techniques with users. Then, we introduce a methodology to conduct user studies and delve into the various scales and metrics employed to assess the impact of explanations on user perception and behavior.
    \item In Chapter~\ref{chap: chi}, a comparative analysis evaluates the effectiveness of three distinct explanation techniques: rule-based, feature attribution, and counterfactual. These techniques are evaluated across two distinct representations: graphical and textual. The chapter examines the influence of these techniques on users' trust and comprehension, presenting the merits and limitations of the methods and representations.
\end{itemize}
This thesis concludes with Chapter~\ref{chap: conclusion} where all the works of the past three years are summarized. Furthermore, this chapter introduces some open challenges and highlights opportunities for future work.

\clearemptydoublepage
\ifenvsetTF{COMPILE_ALL}{
	\part[Good Explanation From Data Perspective]{What Makes a Good Explanation From a Data Perspective?}

\chapter{Improving Anchor-based Explanations}
\label{chap: anchors}

\minitoc

\section{Context}
Explanations based on logical rules are a popular strategy to explain the logic of complex black-box machine learning (ML) classifiers~\cite{miller, wachter}. However, approximating a complex model with human-readable rules incurs an inevitable trade-off: 
Fidelity can only be achieved at the expense of complexity, and complex explanations miss the whole point of explainable ML.
For this reason, recent approaches, such as Anchors~\cite{Anchor}, focus on explanations of local scope. These are if-then rules -- also called {\it anchors} -- that mimic the black box in the vicinity of a target instance. This strategy relies on the assumption that the black-box classifier is simpler to approximate when we focus on a particular region of the space. 

While local rule-based explanations yield simple and locally faithful explanations, their quality can still be very sensitive to some design factors. One of such factors is the discretization of the numerical attributes for tabular data. Figure~\ref{fig: discretization_example} illustrates the anchors obtained for the same dataset with two discretization methods. When running Anchors with a suitable discretization method on the left-hand side of the figure, we obtain the anchor $x > -5.78 \Rightarrow \mathit{Red}$ that matches the black box's behavior more faithfully than the anchor obtained by the discretization method on the right-hand side. 

\begin{figure}[!h]
    \centering
    \begin{subfigure}{.49\textwidth}
        \addtolength{\leftskip} {-1cm}
        \centering
        \includegraphics[width=0.9\linewidth]{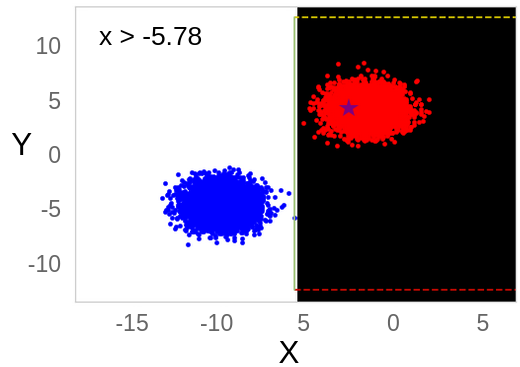}
        \caption{A good discretization}
        \label{fig: mdlp_discretization}
    \end{subfigure}
    \begin{subfigure}{.49\textwidth}
        \addtolength{\rightskip}{-1cm}
        \centering
        \includegraphics[width=0.9\linewidth]{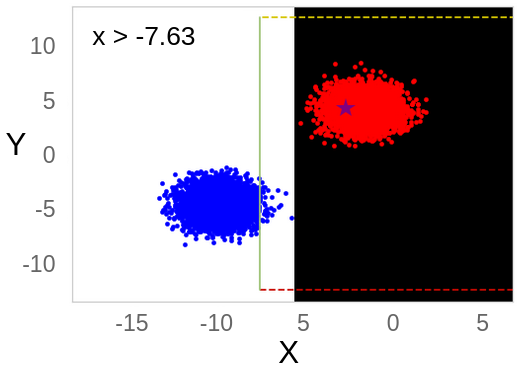}  
        \caption{A suboptimal discretization}
        \label{fig: Quartile_discretization}
    \end{subfigure}
    \caption{Two anchors (depicted as green lines) learned with different discretizations of the numerical features. The target instance is marked as a violet star}
    \label{fig: discretization_example}
\end{figure}

Another factor that can impact the quality of an anchor is the training set used to learn the explanation. Anchors~\cite{Anchor} generates training samples by perturbing the instance of interest according to a neighborhood generation strategy. Figure~\ref{motivate_pertinent} shows the average anchor length (number of conditions on the rule's antecedent) and precision across 10 instances of three explanations learned with different neighborhood generation methods. The strategy in dark blue (\emph{pertinent negatives} explained later) provides the explanation with the best trade-off between rule length and precision.
\begin{figure}[!h]
    \centering
    \includegraphics[width=0.9\linewidth]{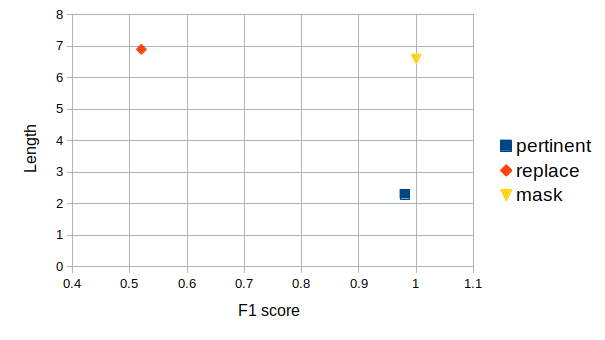}
    \caption{Trade-off between F1 score and anchor length for three neighborhood generation methods}
    \label{motivate_pertinent}
\end{figure}

In this chapter, we study the impact of discretization and neighborhood generation on different metrics that define the quality of anchor-based explanations. Our contributions focus on the tabular and text variants of Anchors and include (i) the application of Minimum Description Length Binning (MDLP)~\cite{MDLP} to discretize the numerical attributes on tabular data, and (ii) the definition of ~\emph{pertinent negatives} on text classifiers. Before elaborating on our contributions, we provide a proper introduction to Anchors in the next section.

The research discussed in this chapter formed the basis of the paper entitled: Improving Anchor-based Explanations, which was published at the CIKM 2020 conference~\cite{Delaunay}.

\section{Anchors}
\label{sec: precision}
Ribeiro et al.~\cite{Anchor} define an \emph{anchor} as a logical rule $R$ that explains a black box $f$ around a target instance $x$. This instance $x$ is a vector of $d$ attributes that can be either categorical or numerical, and $x[j]$ denotes the value of $x$ for the $j$-th attribute. The anchor operates on a surrogate interpretable space defined via a conversion function $\eta : \mathcal{X}^d \rightarrow \{0, 1\}^{d'}$. The generated rule has the form:
\[R: \mathbf{B} \Rightarrow f(\eta^{-1}(z)) = f(x) \;\;\;\;\;\;\;\text{with} \;\; \mathbf{B} = \bigwedge_{j \in F \subseteq \{1, \dots, d' \}}{z[j]} \] 
\noindent The left-hand side (or antecedent) of the rule is a conjunction of conditions that predicts $f(x)$, i.e., the class of the target instance $x$ according to the black box. An example is the rule $\mathit{age} \in [28, 37] \land$ \textit{workclass}=``\textit{private}'' $\Rightarrow$ \quotes{\textit{well-paid}} (for simplicity we write only the predicted value on the right-hand side). For tabular data, the interpretable space can be obtained by discretizing the numerical variables -- to turn them categorical --  and then binarizing the resulting conditions. For text classifiers, the surrogate space is usually defined as the presence or absence of words.

The method proposed by Ribeiro et al.~\cite{Anchor} learns rule-based explanations from the answers of the black box $f$ on a randomly generated neighborhood $\mathcal{Z} \subseteq \{0, 1\}^{d'} $ constructed around $z=\eta(x) \in \mathcal{Z}$. Anchors applies principles of depth-first search and multi-armed bandit theory to output the shortest anchor with the largest coverage that satisfies the precision guarantee $\mathit{prec}(R) = P(f(\eta^{-1}(z)) = f(x)\;|\;\textbf{B} \land z \in \mathcal{Z}) \ge \tau$ for a user-defined precision threshold $\tau$. The coverage of an anchor is the ratio of instances in $\mathcal{Z}$ that match the anchor's antecedent, i.e., $\mathit{cov}(R) = P(\textbf{B}\;|\;z \in \mathcal{Z})$.

The next two sections present our contributions that highlight some of the limitations of Anchors and propose improvements. 

\section{Impact of discretization on Tabular Data}
The variant of Anchors for tabular data assumes we have access to the black box's training dataset 
$\mathcal{D} \subseteq \mathcal{X}^{d}$. Tabular Anchors uses $\mathcal{D}$ to discretize the numerical attributes properly according to the data distribution. 
It supports three discretization methods. Two of them, \emph{decile} and \emph{quartile}, are based on classical quantile discretization. In contrast, the \emph{entropy} discretization method splits the domain of an attribute $j$ in a dataset $\mathcal{D}$ (denoted by $\mathcal{D}[j]$) so that the information entropy of $f(x)$ -- for $x\in \mathcal{D}$ and black box $f$ -- is minimized. Anchors' entropy-based discretization outperforms quantile-based discretization in terms of coverage, precision, and anchor length. However, as we show later in this section, it can still lead to relatively long anchors. On these grounds, we investigate the performance of two new discretization methods.

\subsection{New Discretization Methods}
\subsubsection{K-means} 
We propose a baseline discretization method based on the k-means clustering algorithm~\cite{k-means}. This method splits the domain $\mathcal{D}[j]$ of an attribute into $k$ clusters that minimize the intra-cluster distance while maximizing the inter-cluster distance. Distance is based on the absolute difference of the values in $\mathcal{D}[j]$. Therefore, and unlike the entropy-based method, our adaptation of k-means does not make use of the labels provided by the black box $f$. The parameter $k$ is chosen using the Elbow method~\cite{elbow_method}.

\subsubsection{MDLP Discretization} 
Fayyad and Irani~\cite{MDLP} proposed a method for discretization of continuous-valued attributes into multiple intervals based on the Minimum Description Length Principle (MDLP). Continuous-valued attributes are discretized using the information entropy minimization heuristic. Intuitively, MDLP returns the minimal number of \quotes{pure} intervals needed to separate instances from distinct classes. Compared to a traditional entropy-based method, MDLP focuses on compression minimality, hence it outputs as few intervals as possible. Its key heuristic lies in the selection of the \quotes{best} cut points.

\subsection{Experimental Evaluation}
We evaluate the quality of Anchors when used with five discretization methods. This includes the three methods already supported, i.e., quartile (Q), decile (D), entropy (E), and the methods proposed in this work, i.e., MDLP (M) and k-means (K). The precision threshold $\tau$ (Section~\ref{sec: precision}) is set to the default value, that is, $\tau=0.95$. 

\subsubsection{Metrics} 
The quality of an anchor is defined by its \emph{coverage}, \emph{precision}, and \emph{length}. Precision, as previously discussed in Section~\ref{sec: evaluation} refers to the degree of adherence, while the length of the rule gauges the succinctness of the explanation. Thus, shorter anchors with high coverage and precision are preferred. We highlight that the coverage of an anchor is almost analogous to the \emph{recall}: It co-exists in trade-off with the precision. However, unlike recall, coverage can never reach 1, as this would mean that every instance is classified as the target instance. To account for this issue, we define the normalized coverage $\mathit{ncov(R)}$ of an anchor $R$ as: 
\[\mathit{ncov}(R) = \frac{\mathit{cov}(R)}{P(f(\eta^{-1}(z)) = f(x) \;|\; z \in \mathcal{Z})}.\]

\noindent That is, the standard coverage is now normalized by the maximal attainable coverage of an anchor that explains the class given by $f(x)$. With this formulation, we can define the F1 score of an anchor as the harmonic mean of the precision and the normalized coverage. This score provides a trade-off between coverage and precision. 

\[\mathit{F1}(R) = \frac{2}{\mathit{ncov}(R)^{-1} + \mathit{prec}(R)^{-1}}\]


\subsubsection{Datasets} 
We use three synthetic and two real datasets for our evaluation. The synthetic datasets were generated by randomly drawing 10k instances with the functions \texttt{make\_blobs}, \texttt{make\_moons}, and \texttt{make\_circles} available in scikit-learn\footnote{\url{https://scikit-learn.org}}. The real datasets comprise (i) \emph{Titanic}\footnote{\url{https://www.kaggle.com/c/titanic/data}}, where the goal is to predict if a passenger of the Titanic survived based on her age, sex, class, etc., and (ii) \emph{Adult}\footnote{\url{https://archive.ics.uci.edu/ml/datasets/adult}} where we aim at predicting if a person earns more than 50k USD also based on personal characteristics.

\subsubsection{Black-box models} 
\label{subsubsec: bb-tabular} 
We tested our contributions on a variety of black-box classifiers, namely logistic regression, support vector machines, multi-layer perceptron, and random forests. 

\begin{table*}[!ht]
    \centering
    \addtolength{\leftskip} {-1cm}
    \addtolength{\rightskip}{-1cm}
    \footnotesize 
    \begin{tabular}{|c|c|c|c|c|c||c|c|c|c|c||}
    \hline
        & \multicolumn{5}{c||}{Support Vector Machines} & \multicolumn{5}{c||}{Logistic Regression} \\ 
              \cline{2-11} & K & M & D & Q & E & K & M & D & Q & E \\ \hline
        \textbf{Blobs} & 0.89 & \textbf{1} & 0.84 & 0.85 & \textbf{1} & 0.89 & \textbf{1} & 0.84 & 0.85 & \textbf{1} \\ \hline
        \textbf{Circles} & 0.45 & \textbf{0.87} & 0.43 & 0.46 & 0.77 & 0.45 & \textbf{0.87} & 0.43 & 0.46 & 0.77 \\ \hline
        \textbf{Moons} & 0.6 & 0.7 & 0.66 & \textbf{0.72} & 0.68 & 0.6 & 0.7 & 0.66 & \textbf{0.72} & 0.68 \\ \hline
        \textbf{Adult} & 0.66 & 0.66 & 0.66 & \textbf{0.98} & 0.66 & 0.66 & 0.96 & 0.66 & \textbf{0.98} & 0.66 \\ \hline
        \textbf{Titanic} & 0.42 & \textbf{0.93} & 0.33 & 0.42 & 0.42 & 0.42 & \textbf{0.93} & 0.33 & 0.42 & 0.42  \\ \hline \hline
        \textbf{MR} & 3 & \textbf{1.4} & 3.6 & 2 & 2 & 3.2 & \textbf{1.4} & 3.8 & 2 & 2.2 \\ \hline
    \end{tabular}
    \hspace{1cm}
    \begin{tabular}{|c|c|c|c|c|c||c|c|c|c|c||}
        \hline
        & \multicolumn{5}{c||}{Multilayer Perceptron} & \multicolumn{5}{c||}{Random Forest}  \\
        \cline{2-11} & K & M & D & Q & E & K & M & D & Q & E \\ \hline
        \textbf{Blobs} & 0.89 & \textbf{1} & 0.84 & 0.85 & \textbf{1} & 0.89 & \textbf{1} & 0.84 & 0.85 & \textbf{1} \\ \hline
        \textbf{Circles} & 0.45 & \textbf{0.87} & 0.43 & 0.46 & 0.77 & 0.45 & \textbf{0.87} & 0.43 & 0.46 & 0.77 \\ \hline
        \textbf{Moons} & 0.6 & 0.7 & 0.66 & \textbf{0.72} & 0.68 & 0.6 & 0.7 & 0.66 & \textbf{0.72} & 0.68 \\ \hline
        \textbf{Adult} & 0.66 & 0.96 & 0.66 & \textbf{0.98} & 0.66 & 0.66 & 0.65 & 0.66 & \textbf{0.98} & 0.66 \\ \hline
        \textbf{Titanic} & 0.42 & \textbf{0.93} & 0.33 & 0.42 & 0.42 & 0.42 & \textbf{0.93} & 0.33 & 0.42 & 0.42 \\ \hline \hline
        \textbf{MR} & 3.2 & \textbf{1.4} & 3.8 & 2 & 2.2 & 3 & \textbf{1.6} & 3.6 & 2 & 2 \\ \hline 
    \end{tabular}
    \caption{F1 score for Anchors using different discretization methods. MR denotes the mean rank of the method.}
    \label{F1_tab}
\end{table*}

\begin{table*}
    \centering
    \addtolength{\leftskip} {-1cm}
    \addtolength{\rightskip}{-1cm}
    \footnotesize    
    \begin{tabular}{|c|c|c|c|c|c||c|c|c|c|c||}
        \hline
        & \multicolumn{5}{c||}{Support Vector Machines} & \multicolumn{5}{c||}{Logistic Regression} \\ 
        \cline{2-11} & K & M & D & Q & E & K & M & D & Q & E \\ \hline
        \textbf{Blobs} & \textbf{1} & \textbf{1} & \textbf{1} & \textbf{1} & \textbf{1} & \textbf{1} & \textbf{1} & \textbf{1} & \textbf{1} & \textbf{1}  \\ \hline
        \textbf{Circles} & 8.14 & \textbf{2.39} & 4.79 & 3.69 & 3.69 & 8.14 & \textbf{2.39} & 4.79 & 3.69 & 3.69  \\ \hline
        \textbf{Moons} & 2.47 & 2.46 & \textbf{1.95} & 2.49 & 2.72 & 2.47 & 2.46 & \textbf{1.95} & 2.49 & 2.72 \\ \hline
        \textbf{Adult} & 9.37 & 7.43 & 7.6 & \textbf{6.48} & 8.54 & 9.16 & 8.03 & 7.89 & \textbf{6.41} & 8.23 \\ \hline
        \textbf{Titanic} & 4.18 & \textbf{2.58} & 4.72 & 3.52 & 3.52 & 4.18 & \textbf{2.58} & 4.72 & 3.52 & 3.52 \\ \hline \hline
        \textbf{MR} & 3.2 & \textbf{1.4} & 2.4 & 2 & 2.8 & 3.2 & \textbf{1.6} & 2.2 & 2 & 2.4 \\ \hline
    \end{tabular}

    \centering
    \addtolength{\leftskip} {-1cm}
    \addtolength{\rightskip}{-1cm}
    \footnotesize    
    \begin{tabular}{|c|c|c|c|c|c||c|c|c|c|c||}
        \hline
        & \multicolumn{5}{c||}{Multilayer Perceptron} & \multicolumn{5}{c||}{Random Forest} \\
        \cline{2-11} & K & M & D & Q & E & K & M & D & Q & E \\ \hline
        \textbf{Blobs} & \textbf{1} & \textbf{1} & \textbf{1} & \textbf{1} & \textbf{1} & \textbf{1} & \textbf{1} & \textbf{1} & \textbf{1} & \textbf{1} \\ \hline
        \textbf{Circles} & 8.14 &\textbf{ 2.39 }& 4.79 & 3.69 & 3.69 & 8.14 & \textbf{2.39} & 4.79 & 3.69 & 3.69 \\ \hline
        \textbf{Moons} & 2.47 & 2.46 & \textbf{1.95} & 2.49 & 2.72 & 2.47 & 2.46 & \textbf{1.95} & 2.49 & 2.72  \\ \hline
        \textbf{Adult} & 8.99 & 7.26 & 8.34 & \textbf{6.67} & 8.49 & 9.11 & 7.21 & 8.26 & \textbf{6.77} & 8.84  \\ \hline
        \textbf{Titanic} & 4.18 & \textbf{2.58} & 4.72 & 3.52 & 3.52 & 4.18 & \textbf{2.58} & 4.72 & 3.52 & 3.52 \\ \hline \hline
        \textbf{MR} & 3.2 & \textbf{1.4} & 2.4 & 2 & 2.8 & 3.2 & \textbf{1.4} & 2.4 & 2 & 2.8 \\ \hline
    \end{tabular}
    \caption{Anchor length using different discretization methods. MR denotes the mean rank of the method.}
    \label{size_tab}
\end{table*} 

\subsection{Results}
Table~\ref{F1_tab} summarizes the F1 performance of Anchors for a set of instances\footnote{10k for the synthetic datasets, 100 for Titanic and Adult.} of each dataset for all the studied black-box models and discretization methods. The labels K, M, D, Q, and E denote k-means, MDLP, decile, quartile, and entropy respectively. The different discretization methods exhibit similar performance on the artificial datasets because all black boxes behave equivalently on those datasets (e.g., they all achieve 95\% accuracy). Thus, the difference between each model is very slight. All discretization methods obtain good performance on the highly structured dataset \emph{Blobs} (depicted in Figure~\ref{fig: discretization_example}). 
We observe that overall, MDLP achieves the best F1 followed by quartile and entropy. In some cases, however, MDLP is worse than quartile (e.g., for the \emph{Adult} dataset). In particular, MDLP and quartile split the domain of attributes into fewer intervals, leading to less specific conditions with potentially higher coverage. The use of black-box labels for binning usually gives MDLP a significant advantage over a simple quartile discretization. Besides, the focus on compression minimality makes MDLP output fewer intervals than the \emph{entropy} strategy. Table~\ref{size_tab} confirms our intuitions as we observe that MDLP yields on average the shortest anchors. 
An example of an anchor using MDLP in the \emph{Adult} dataset is $\mathit{age} \le 22 \land \mathit{relationship} =$ \emph{\quotes{own-child}} $ \Rightarrow \mathit{<\,50k USD}$.

\section{Improving Anchors on Text}
In some of our experiments with Anchors on text data, it was impossible to attain the default precision threshold $\tau=0.95$. This phenomenon makes Anchors output rules with a precision smaller than $\tau$. We argue that the maximal attainable precision of Anchors depends on (i) the distribution of the training neighborhood and (ii) the expressiveness of the rule language. In this section, we study the performance of Anchors for two different neighborhood generation strategies and propose an extension of the rule language by considering negated conditions, known as pertinent negatives in the explainable AI literature. 

\subsection{Neighborhood Generation Strategies}
The variant of Anchors for text classification converts a textual instance into a surrogate binary vector where each entry defines the absence or presence of a word of the target phrase. Consider, for example, a black-box classifier $f$ for sentiment analysis and the target instance ``This is a good book''. Anchors will convert this instance into a five-component vector, i.e., 11111, and generate neighbors by randomly toggling off bits of this binary representation. Examples are the instances 10101 or 11101. An anchor is induced from that set of neighbors and their class labels according to the black box $f$. However, $f$ operates in a different space than Anchors. Hence, the inverse conversion function $\eta^{-1}$ must map the generated neighbors to actual text instances. The strategy called \emph{mask words} (MW) does so by replacing the words of each zero component with a neutral wildcard unseen before by the black box. In our example the neighbor 11101 becomes \quotes{This is a $\mathbb{W}$ book} for wildcard $\mathbb{W}$. The strategy called \emph{replace words} (RW), on the other hand, replaces toggled-off words with random words that have the same syntactic role, i.e., they would be assigned the same part-of-speech tag. For instance, the neighbor 11101 could become \quotes{This is a \textbf{great} book}. 

\subsection{Pertinent Negatives}
We highlight that anchors are defined on conjunctions of non-negated conditions. For text data, this entails conditions on the presence of words in phrases. This design decision guarantees simpler rules while keeping the search space under control. On the downside, it imposes limits on the expressiveness of explanations. Inspired by the work presented by~\cite{pertinents_negatifs}, we propose to change the language of Anchors and provide explanations on the absence of words. Those words are known as ~\emph{pertinent negatives} (PN) and can be seen as counterfactual explanations, i.e., words whose presence would change the answer of the black box. 

Considering the absence of all possible words in the corpus makes the search space for anchors prohibitively large. Hence, we apply two mechanisms to alleviate this fact. First, we focus on a limited set of words. This set consists of the top $k$ most frequent words that co-occur next to the words of the target instance, stopwords excluded. For our example ``This is a good book'', our algorithm would consider words such as \emph{scientific}, \emph{interesting}, or \emph{very} as they may often appear with ``book'' and  ``good''. Second, we set an upper bound $p$ in the number of pertinent negatives allowed in explanations. 

It follows that a neighborhood generation method purely based on pertinent negatives represents a phrase as a vector of $m + p$ components where $m$ is the number of words in the target phrase and $p$ is the number of pertinent negatives. The target instance is mapped to a vector where the first $m$ elements are set to 1 and the remaining are set to 0. Neighbors are then generated by randomly toggling on the zero entries of the pertinent negatives, which instructs Anchors to add the word to the phrase. Our goal is to show the potential and viability of pertinent negatives in Anchors, thus we leave as future work the implementation of a hybrid approach that combines pertinent negatives with one of the classical strategies for neighborhood generation based on present words. 

\subsection{Experimental Evaluation}
We evaluate the discussed neighborhood generation strategies using the F1 measure and the anchor length as quality criteria. For pertinent negatives, we use $p=20$.

\subsubsection{Datasets.} Our experimental datasets comprise (i) \emph{Polarity}\footnote{\url{http://www.cs.cornell.edu/people/pabo/movie-review-data/}}, a set of movie reviews for sentiment analysis, and (ii) \emph{Tweets}\footnote{\url{https://competitions.codalab.org/competitions/17344\#learn_the_details-data}}, a set of tweets used for a multi-class classification task focused on predicting the occurrence of emojis. 

\subsubsection{Black-box models.} We use the same black-box models as in Section~\ref{subsubsec: bb-tabular}. Those models were trained on a vector representation of the phrases based on word counts and provided by the class \texttt{CountVectorizer} of scikit-learn. In this representation, a phrase is converted into a sparse vector such that each component is associated with a word seen in the corpus -- the set of phrases used for training -- and stores the frequency of the word in the phrase.

\subsection{Results} 
\begin{table*}
    \centering
    \addtolength{\leftskip} {-1cm}
    \addtolength{\rightskip}{-1cm}
    \begin{tabular}{|c|c|c|c||c|c|c||c|c|c||c|c|c|}
    \hline
         & \multicolumn{3}{c||}{\footnotesize{Support Vector Machines}} & \multicolumn{3}{c||}{\footnotesize{Logistic Regression}} & \multicolumn{3}{c||}{\footnotesize{Multilayer Perceptron}} &  \multicolumn{3}{c|}{\footnotesize{Random Forest}} \\
        \cline{2-13} & RW & MW & PN  & RW & MW & PN & RW & MW & PN & RW & MW & PN \\ \hline
        \textbf{Tweets} & 5.1 & \textbf{4.3} & 8.1 & \textbf{4.6} & 6.4 & 9.1 & \textbf{2.1} & 4.6 & 7.2 & \textbf{3.7} & 5.4 & 4.2 \\ \hline
        \textbf{Polarity} & 8.3 & 7.7 & \textbf{4.9} & 6.7 & \textbf{3.6} & 4.5 & 6.9 & 6.6 & \textbf{2.3} & \textbf{2} & \textbf{2} & \textbf{2} \\ \hline
    \end{tabular}
    \caption{Length of textual Anchors for different neighborhood generation strategies.}
    \label{size_text}
\end{table*}

\begin{table*}
    \centering
    \addtolength{\leftskip} {-1cm}
    \addtolength{\rightskip}{-1cm}
    \begin{tabular}{|c|c|c|c||c|c|c||c|c|c||c|c|c|}
    \hline
         & \multicolumn{3}{c||}{\footnotesize{Support Vector Machines}} & \multicolumn{3}{c||}{\footnotesize{Logistic Regression}} & \multicolumn{3}{c||}{\footnotesize{Multilayer Perceptron}} &  \multicolumn{3}{c|}{\footnotesize{Random Forest}} \\
        \cline{2-13} & RW & MW & PN  & RW & MW & PN & RW & MW & PN & RW & MW & PN \\ \hline
        \textbf{Tweets} & 0.63 & 0.41 & \textbf{0.82} & 0.56 & 0.31 & \textbf{0.91} & 0.85 & 0.44 & \textbf{0.95} & 0.57 & 0.27 & \textbf{0.95} \\ \hline
        \textbf{Polarity} & 0.35 & \textbf{1} & 0.87 & 0.35 & \textbf{0.98} & 0.86 & 0.52 & \textbf{1} & 0.98 & 0.47 & \textbf{1} & 0.79 \\ \hline
    \end{tabular}
    \caption{F1 score of textual Anchors for different neighborhood generation strategies.}
    \label{F1_text}
\end{table*}

We summarize the aggregated results for the F1 measure and the anchor size among 10 randomly selected instances in Tables~\ref{size_text} and~\ref{F1_text}. We first observe that the \emph{replace words} strategy (RW) lies far behind the pertinent negative (PN) and mask words (MW) for the \emph{Polarity} dataset. While it usually produces anchors of high precision, the coverage of those anchors is very low, in other words, it generates overly specific explanations. This intuition is confirmed by Table~\ref{size_text}, where we can observe that \emph{replace words} yields, on average, longer anchors than the other strategies. 
These are a consequence of the neighborhood generation strategy. By replacing toggled-off words with other words of the same syntactic role, the neighbor instances become very unstable: the addition of a single word can change the meaning of the phrase as well as the black box's answer. We observe this phenomenon to a lesser extent when using the strategy \emph{mask words} in the \emph{Tweets} dataset. This happens because replacing a word with the wildcard forces the black box to decide based on fewer words. We observe that pertinent negatives lead to short and still fairly accurate anchors on \emph{Polarity}, and achieve the best F1 score on \emph{Tweets} -- at the expense of length -- while it returns more complex rules on Tweets and a slightly less F1 score on \emph{Polarity}. 
These long anchors are due to the fact that the training set for PN consists of many more features. This is also aggravated by the large number of classes as \emph{Tweets} is a multiclassification problem with 20 different emojis to predict. 
An example anchor with PN is $\neg \textit{caring} \land \textit{downpur} \Rightarrow $ \includegraphics[height=2.5ex]{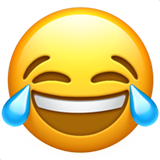} for the tweet 
``Totally worth getting caught in this evening's downpour. \#jacquelineonassisreservoir''. The addition of the word ``caring'' may have resulted in the classifier predicting a different emoji.

\section{Conclusion}
In this chapter, we delved into the core aspects that influence the effectiveness of anchor-based explanations for machine learning models. Specifically, our investigation centered on two critical components: the discretization of numerical features in tabular data and the strategy employed for generating neighborhoods in text data.

Our findings reveal that a meticulous adjustment of these components yields substantial enhancements in the precision, coverage, and length of anchor-based explanations. This illustrates the potential of finding more accurate and comprehensive explanations for other interpretability methods in machine learning models.

We identify two promising directions for future research. Firstly, we envision assessing the impact of discretization on other explanation methods, such as LIME, thereby broadening the scope of our insights. Secondly, we aim to explore post-hoc discretization techniques, aiming to expand the coverage of Anchors. A possible approach is to expand the boundaries of each variable within an anchor until precision begins to decline. This approach has the potential to increase the coverage of the explanations.

Furthermore, we plan to investigate the synergy between pertinent negatives and mask word strategies, aiming to create more expressive and accurate anchors that encompass both the presence and absence of specific words. This holistic approach holds the promise of further enriching the explanatory power of anchor-based explanations. The code, data, and experimental results are available at \url{https://github.com/juliendelaunay35000/anchors}.

In this chapter, we assumed that a user would apply Anchors without considering its suitability for a given use case. This is because Anchors was proposed as a general, model-agnostic approach. However, as we saw, Anchors is not always optimal on all black-box classifiers and datasets. In the next chapter, we put to test the question of whether a one-size-fits-all approach for surrogates is ideal. In this chapter, we emphasized refining the components that underpin the effectiveness of rule-based explanations. Our exploration now revolves around linear surrogates, which, while widely used for their simplicity and fidelity, may not always be the most suited for delivering unambiguous, faithful explanations. This leads us to introduce a novel method that characterizes black-box classifier decision boundaries and identifies the ideal scenarios for deploying linear models as reliable explanations. This shift in focus represents our quest to make interpretability more nuanced, adaptable, and data-driven.

\clearemptydoublepage

\chapter{When Should We Use Linear Explanations?}
\label{chap: ape}
\minitoc

\section{Context}
\label{sec: ape_introduction}
One way to explain a black-box model in a post-hoc manner is to learn a surrogate white-box model that mimics the black box. Even though there may be multiple surrogates or paradigms to explain the verdict of a black box for a given scenario, this decision is rarely based on the particularities of the use case. In the previous chapter, we focused on how to improve the quality of rule-based explanations, which are fairly popular. In this chapter, we shift the focus to address the nuanced question of selecting the most suitable explanation method in a given context. Linear surrogates, appreciated for their simplicity and faithfulness, are another popular choice for locally approximating black-box models~\cite{trends_in_xai}. Despite the popularity of local linear explanations, they may not always be the most adapted method to explain a black-box outcome. 

To illustrate this, consider the two scenarios depicted in Figure~\ref{fig: multiple_linear}. In Figure~\ref{fig: multiple_linear}a, the instance of interest lies in a zone where there is clearly a single local linear approximation for the black-box classifier. In contrast, the target instance in Figure~\ref{fig: multiple_linear}b depicts a scenario where three possible linear explanations are possible. Since these approximations exhibit different slopes, the attribution scores assigned to the input features are obviously contradictory -- a situation that would harden interpretation. While we could provide one of the explanations for Figure~\ref{fig: multiple_linear}b, that would tell an incomplete story.

\begin{figure}[!h]
    \centering
    \includegraphics[width=0.95\linewidth]{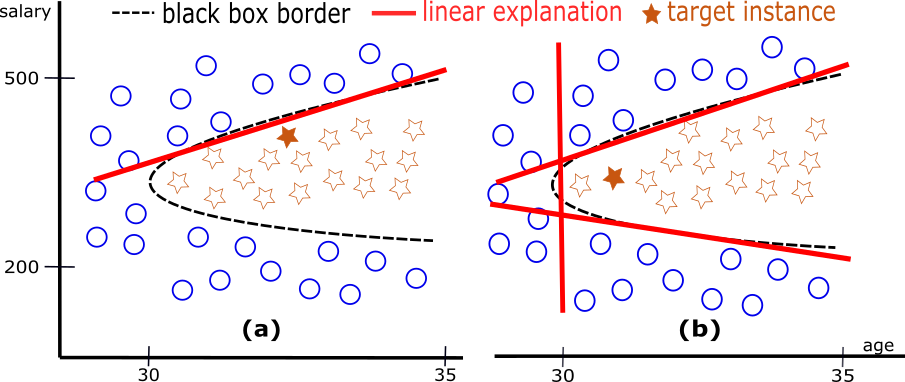}
    \caption{Two explanation scenarios for a classifier and a target instance (the filled star): (a) a suitable single linear explanation; (b) three contradictory linear explanations.}
    \label{fig: multiple_linear}
\end{figure}

Based on the aforementioned arguments, this chapter proposes APE, which stands for Adapted Post-hoc Explanations, a novel method to determine \emph{a priori} whether a black-box classifier and a target instance admit a faithful and unambiguous local linear explanation. When this is not the case, APE does not stop but recommends a different explanation paradigm. Building upon the knowledge acquired in the preceding chapter, we investigate the use of rule-based explanations in our experimental work. Through our empirical evaluation, we demonstrate that these explanations serve as a valuable complement to linear explanations. APE operates by characterizing the classifier's decision boundary, which is achieved by identifying the target's closest counterfactual instance. We recall that counterfactual instances, also called enemies, are instances that are close to the target instance but are classified differently by the black box. Such instances can be used as explanations that highlight the minimal changes required on the target instance to change the classifier's outcome. All in all, our contributions are:

\begin{itemize}[leftmargin=*]
    \item A definition of \emph{suitability} for explanations based on local linear surrogates. This definition builds upon existing notions such as {\em adherence} and {\em locality}, which we also define formally.   
    \item Three novel algorithms for counterfactual exploration are introduced: \emph{Growing Fields} (GF), \emph{Growing Language} (GL), and \emph{Growing Net} (GN). These techniques build upon the Growing Spheres (GS) algorithm~\cite{gs}. Notably, Growing Fields is tailored for tabular data and handles categorical attributes. Growing Fields also accounts for the distribution of input features, employing the standardized Euclidean distance as a robust metric. Growing Language and Growing Net are specifically designed to generate counterfactual explanations for textual data. They rely on external knowledge to explore the search space and restrict it to pertinent words.
    \item The APE Oracle is a linear suitability test that tells users whether a black-box classifier can be locally approximated by a single and faithful linear surrogate. To do so, APE characterizes the distribution of the instances around the decision boundary.
    \item The APE algorithm that returns a linear explanation if suitable. Otherwise APE proposes a rule-based explanation. In all cases, APE computes complementary counterfactual explanations.
\end{itemize}

\noindent The chapter is structured as follows. After formulating the problem and introducing preliminary concepts in Section~\ref{sec: problem_statement}, Section~\ref{sec: counterfactual_methods} introduces our novel counterfactual explanation techniques. Section~\ref{sec: ape} elaborates on the APE approach before we evaluate it on a handful of datasets and classifiers in Section~\ref{sec: ape_experiments}. This is followed by a discussion of our insights.

Most of the work presented in this chapter was the subject of the paper: When Should We Use Linear Explanations?, published at the CIKM 2022 conference~\cite{Delaunay_cikm}. The results on textual data are novel works.

\section{Preliminaries}
\label{sec: problem_statement}
\subsection{Problem Statement}
Given a black-box classifier $f:X \rightarrow Y$ trained on a dataset $T \subset X$, and a target instance $x=(x_1,\cdots,x_d) \in X$, our goal is to construct an Oracle that tells us whether a linear surrogate $g$ learned on a locality $\Phi_x \subset X$ (defined below) is suitable to explain $f(x)$. By ``suitable'' we mean that two \emph{contradictory} linear explanations $g$, and $g'$ may not have the highest adherence in $\Phi_x$ -- the adherence being the outcome agreement between $f$ and $g$. In this formulation, $\Phi_x$ is a region of the space that (i) covers $x$, (ii) is traversed by $f$'s decision boundary, and (iii) is maximal, otherwise stated, the surrogate $g$ cannot attain the quality guarantee $m(g) \ge \tau$ in any locality $\Phi_x' \supset \Phi_x$ for some adherence metric $m$.

Requirement (i) guarantees that the target instance $x$ is included in the surrogate's training set. Moreover, requirement (ii) ensures that this training set contains both instances inside and outside the class $f(x)$. 
Consequently, the minimal locality satisfying these two requirements should be centered on the decision boundary -- more precisely on $x$'s closest counterfactual. This alignment places the target instance $x$ directly on the boundary, as illustrated by the inner dotted circle in Figure~\ref{fig: locality}. Incorporating requirement (iii) implies that the scope $\Phi_x$ might be expanded if the surrogate $g$ maintains a strong adherence. Thus, Figure~\ref{fig: locality} illustrates two localities represented by distinct dashed circles. The blue circle denotes the initial locality, with a radius equivalent to the distance between the target instance and its associated closest counterfactual. The orange circle represents the largest possible locality to approximate a black box model with a linear surrogate while preserving adherence between the linear model and the complex model above a given $\tau$ threshold. In such a scenario, the explanation generalizes to broader regions of the data space, as depicted by the larger dashed circle.

\begin{figure}[htbp]
    \centering
    \includegraphics[width=0.9\linewidth]{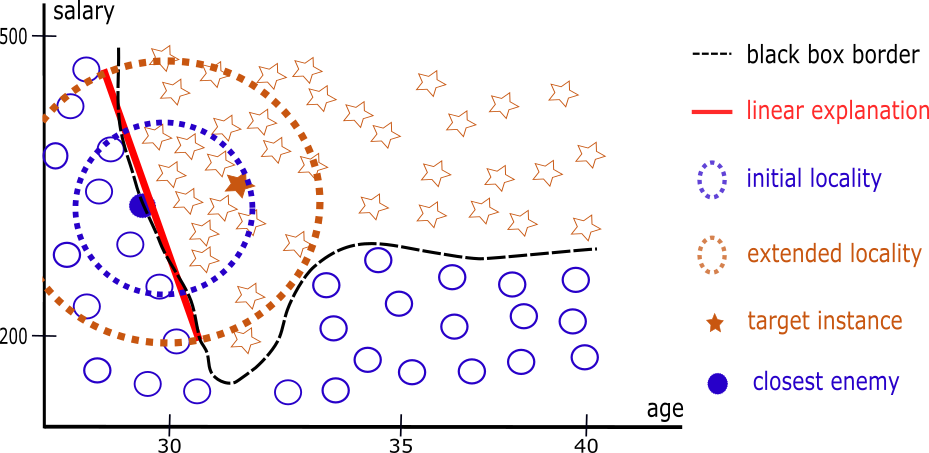}
    \caption[Illustration of a linear explanation for a classifier and a target instance $x$.]{A linear explanation for a classifier and a target instance $x$. The inner circle (dotted in blue) is the minimal locality $\Phi_x$ that covers $x$ and is traversed by the decision boundary. Locality can be extended  (orange circle) and still provide an equally good linear approximation for the black box. Friends $F$ of $x$ are represented by yellow stars, and enemies $E$ by blue circles.}
    \label{fig: locality}
\end{figure}

\subsection{Linear Explanations}
In order to explain the outcome $f(x)$ of a classifier $f$ on a target instance $x$, methods such as LIME~\cite{LIME} or Local Surrogate~\cite{Laugel_Defining_locality} provide a signed feature-attribution ranking $R(g)$ that consists of ordered sets of features $R^+(g)$ and $R^-(g)$. The features in $R^+(g)$ contribute positively to predicting the class $f(x)$, whereas the features in $R^-(g)$ push towards predicting a different class. The ranking is based on the coefficients of a linear surrogate $g$ that approximates $f$ in a locality or neighborhood around $x$. We say that two linear explanations $g$ and $g'$ for $f(x)$ are contradictory if they induce different attribute rankings, more formally, if $R(g) \neq R(g')$. 

\subsection{Adherence and Fidelity} 
The quality of a surrogate model $g$ for a black-box classifier $f$ is evaluated through the notions of adherence and fidelity. The \emph{adherence} of a surrogate model $g$ for a black-box model $f$ is the degree of agreement between $f$'s and $g$'s outcomes. The \emph{fidelity}, on the other hand, assesses the surrogate's ability to identify the features truly employed by the black-box model. When $f$ is a true black box, users can only rely on adherence to estimate the quality of explanations.

\subsection{Existing Methods}

LIME~\cite{LIME} stands out as the predominant method for computing local linear explanations. 
It has been shown by Laugel et al.~\cite{Laugel_Defining_locality} that we can learn more locally faithful explanations if we apply LIME on a neighborhood traversed by $f$'s decision boundary. In that vibe, the Local Surrogate (LS) approach~\cite{Laugel_Defining_locality} centers the generative process not on the target instance $x$ but on its closest enemy $e$ -- which by itself constitutes a complementary explanation for $f(x)$. LS then learns a linear surrogate on a neighborhood defined by a hyper-sphere centered at $e$, as depicted by the inner circle in Figure~\ref{fig: locality}.  

\section{Counterfactual Explanation Methods}
\label{sec: counterfactual_methods}
To inspect the decision boundary of a classifier $f$ in the vicinity of a target instance $x$, the first step is to find its closest enemies. Growing Spheres (GS)~\cite{gs} is a method that searches for enemies of $x$ by drawing instances uniformly within the volume of a $l_2$-sphere of radius $r$ centered at $x$. The value of $r$ is adjusted so that the resulting sphere traverses $f$'s decision boundary and encompasses enemies of $x$ lying close to the border. While Growing Spheres is known for its simplicity and versatility, it comes with several inherent limitations. Notably, it does not handle categorical attributes and does not consider the variance within feature distributions. Additionally, Growing Spheres is primarily designed for generating counterfactuals in tabular data settings, making it unsuitable for data types like text. Therefore, we developed three enhancements of the Growing Spheres algorithm that we call Growing Fields (GF), Growing Language (GL), and Growing Net (GN). We first develop in this section on Growing Fields, the tabular extension which proceeds likewise, but tackles some of the limitations of Growing Spheres as explained next. We then present Growing Language and Growing Net, our two extensions for textual data.

\subsection{Counterfactuals for Tabular Data}
\label{subsec: growing_fields}
\begin{algorithm}[t]
    \caption{The $\mathcal{F}$ instance generation process}
    \label{pseudo-code_distribution}
    \begin{algorithmic}[1]
    \Require $\text{a dataset}\;T \subset X$, $\text{a radius}\;r \in (0, 1]$, 
    an instance $x = ( x_1, \dots, x_d ) \in X$ 
    \Ensure An artificial instance $z = ( z_1, \dots, z_d )$
    \For{$i \in 1 \dots d$}
        \If{$x_i$ is numerical} \Comment{\footnotesize $A_i = \mathit{max}_i - \mathit{min}_i$ }
            \State {$a = \textit{min}(0, r \times A_i(T) - \sigma_i(T))$}  
            \State {$b= a + \sigma_i(T)$}
            \State {$z_i \gets x_i + \rho_k$ with $\rho_k \thicksim  \mathcal{U}(a, b)$}
        \Else
            \State {$z_i \gets (x_i \ \text{with prob.} \ 1-\rho_k )$ with $\rho_k \thicksim  \mathcal{U}(0, r)$}
        \EndIf
    \EndFor \\
    \Return{$z$}
    \end{algorithmic}
\end{algorithm}

Growing Fields generates instances by slightly perturbing the target instance until it identifies an instance classified differently by the black box, thus providing a counterfactual explanation. The instances generating process of Growing Fields is outlined in Algorithm~\ref{pseudo-code_distribution}, while the main algorithm is detailed in Algorithm~\ref{pseudo-code_Growing_fields}. Growing Fields extends Growing Spheres through two key enhancements: firstly, it incorporates the distribution of the input features, and secondly, it deals with categorical attributes.

\subsubsection{Attribute-dependant perturbations}
By drawing instances uniformly in a $l_2$-sphere, Growing Spheres assumes that all numerical attributes should be perturbed at the same rate. In reality, the attributes may have different amplitudes, variances, and distributions. Consequently, in Growing Fields the perturbation added to a numerical attribute $x_i$ follows a uniform distribution that depends on both the radius $r$ and the attribute's domain amplitude $A_i(T)$, and at the same time preserves the attribute's std. deviation in the input dataset $T$ -- denoted by $\sigma_i(T)$. Therefore, Growing Fields generates uniformly perturbed instances depending on the radius $r$ of the field and follows a uniform distribution whose variance is $\sigma_i(T) = \frac{(b-a)^2}{12}$ given lower and upper bound parameters $a$ and $b$. Growing Fields computes $a$ and $b$ by solving:
\begin{alignat*}{4}
    \label{eq: uniform}
    v & {}={} & \sqrt{12\sigma_i(T)} \\
    a & {}={} &  min(0, r - v)  \\
    b & {}={} & a + v.
\end{alignat*}
The product $\rho \times v$ where $\rho \thicksim \mathcal{U}(0, r')$, $v \thicksim \mathcal{N}(0, \sigma_i(X))$, and $\sigma_i(X)$ is the standard deviation of the $i$-th attribute of $x$ in the input dataset. This implies that the vicinity generated by Growing Fields around an instance $x$ is not anymore a sphere, but rather a volume or, as we call it, a \emph{field}. The actual shape of this field depends on the distance function. We highlight that taking into account the data distribution guarantees a data-aware exploration of the space, which results in a speed-up of up to 2 orders of magnitude w.r.t. Growing Spheres as shown in the results section.

Another limitation of Growing Spheres is that all attributes have the same impact when computing the distance between two instances. That said, a salary ``distance'' of 30 EUR is insignificant compared to an age ``distance'' of 30 years. On those grounds, Growing Fields normalizes the contribution of attribute $i$ using the mean $\mu_i$ and standard deviation $\sigma_i$ in the training set, which boils down to the standardized Euclidean distance\footnote{This is a special case of the Mahalanobis distance when the covariance matrix is diagonal.}:

\begin{equation} \label{eq: distance}
    \mathit{dist}(x, x') = \sqrt{\sum_{i = 1}^{d} (\frac{(x_i - \mu_i) - (x'_i - \mu_i) }{\sigma_i})^2}
\end{equation}
\noindent Equation~\ref{eq: distance} assumes that the categorical attributes have been one-hot encoded. We normalized the distance by dividing it by the maximum distance between the target instance and the instances in the dataset. This normalization guarantees that the generated instances do not exceed the distance of real instances when their distance is below 1. In contrast, Growing Spheres lacks a predefined upper limit on the radius value $r$, allowing for unconstrained variations. This enables us to establish a meaningful distance between the counterfactual and the target instance, one that can be compared to the rest of the dataset. 

\subsubsection{Support for categorical features}
The original Growing Spheres algorithm does not support categorical attributes, such as the gender or the marital status of a person. We can now handle those attributes by treating them as random continuous variables distributed uniformly within the range of $[0, r]$. Consider a field with radius $r=0.5$ and a target instance with the attribute $\mathit{sex}=F$. If by drawing a random value in $[0, 0.5]$ we obtain, for example, a value of 0.2, we interpret it as throwing a biased coin that keeps the sex of the target instance with probability $1 - 0.2 = 0.8$. If the attribute defines more than two categories, e.g., \{single, married, divorced, widowed\} and we have to change the category, we employ the re-adjusted empirical probabilities of the other categories in the input dataset $T$ to randomly choose the new category. Note that our way of handling categorical attributes requires a parameter $r$, the radius of the sphere that lies in $(0,1]$.

\begin{algorithm}[t]
    \small
    \caption{\textsc{Growing Fields (GF)}}
    \label{pseudo-code_Growing_fields}
    \begin{algorithmic}[1]
    \Require a dataset $T \subset X$, $\text{a target instance}\; x = ( x_1, \dots, x_d ) \in X$, $\text{a classifier}\;f:X \rightarrow Y$, 

    \Statex Hyper-parameters: $r_0=0.1$, $\theta=1.8,\ n=2000$ as defined by Growing Spheres~\cite{gs} 
    \Ensure Set $Z$ of instances; resulting field radius $r$
    \State $r \gets r_0$
    \State{$Z \thicksim \mathcal{F}(T, r, x)_{i \leq n}$}
    \While {$\exists \ e \in Z \mid f(e) \neq f(x)$} %
        \State $r \gets r / 2$
        \State{Update $Z \thicksim \mathcal{F}(T, r, x)_{i \leq n}$}
        \EndWhile
    \While {$\nexists \ e \in Z \mid f(e) \neq f(x)$}
        \State{$r \gets$ min(1,$\ \theta \times r)$} 
        \State{Update $Z \thicksim \mathcal{F}(T, r, x)_{i \leq n}$}
        \EndWhile\\
    \Return {$Z, \arg \min_e\{\textit{dist}(x, e) \; | \; e \in Z \ \textbf{and} \ f(e) \neq f(x) \}$}
    \end{algorithmic}
\end{algorithm}

Algorithm~\ref{pseudo-code_distribution} details the resulting generation process, called $\mathcal{F}$ (which stands for field), used to draw random artificial instances with both numerical and categorical attributes. The result of integrating $\mathcal{F}$ into Growing Spheres gives rise to the Growing Fields algorithm detailed in Algorithm~\ref{pseudo-code_Growing_fields}. Growing Fields starts with an initial field of radius $r_0$ and reduces it until no enemies are found (lines 3-6). In the second stage, the field is \emph{gradually expanded} until the decision boundary is crossed and close counterfactual instances can be reported (lines 7-10). The algorithm then returns $x$'s closest counterfactual.

\begin{algorithm}
    \caption{Generating Counterfactual Explanations}
    \label{alg: growing}
    \begin{algorithmic}[1]
    \Require $\text{a target instance}\;x = ( x_1, \dots, x_d ) \in X$,  $\text{a black-box classifier}\;f:X \rightarrow Y$; 
    \Statex $Store\_Similar\_Words() \rightarrow$ a function to store similar words for an input word
    \Statex Hyper-parameters: $n = 2000$; $\theta$ = 0.8 (similarity\_threshold)
    \Ensure one or multiple counterfactual instances
    \State Initialize $d$ sets $W = (W_1, \dots, W_d)$ of candidate replacement words
    \For{$i \in 1 \dots d$}
        \State $W_i \gets Store\_Similar\_Words(x_i, \theta$, part-of-speech($x_i$))
    \EndFor
    \State Initialize $Z = (z_1, \dots, z_n)$ as $n$ copy of $x$
    \State Initialize $C$ set of valid counterfactuals
    \While{number of words modified $< d$ \textbf{and} $C$ is empty}
        \For{$j \gets 1$ \textbf{to} $n$}
            \State $k = random(0, d)$ \Comment{Ensure that $z_{j,k}$ has not been already modified} 
            \State Replace $z_{j,k}$ with a word randomly taken from $W_k$
            \If{$f(x) \neq f(z_j)$}
                \State $C \gets z_j$
            \EndIf
        \EndFor
    \EndWhile
    \State \textbf{return} $C$ the set of valid counterfactual instances
\end{algorithmic}
\end{algorithm}
\begin{figure}[t]
    \centering
    \includegraphics[width=\textwidth]{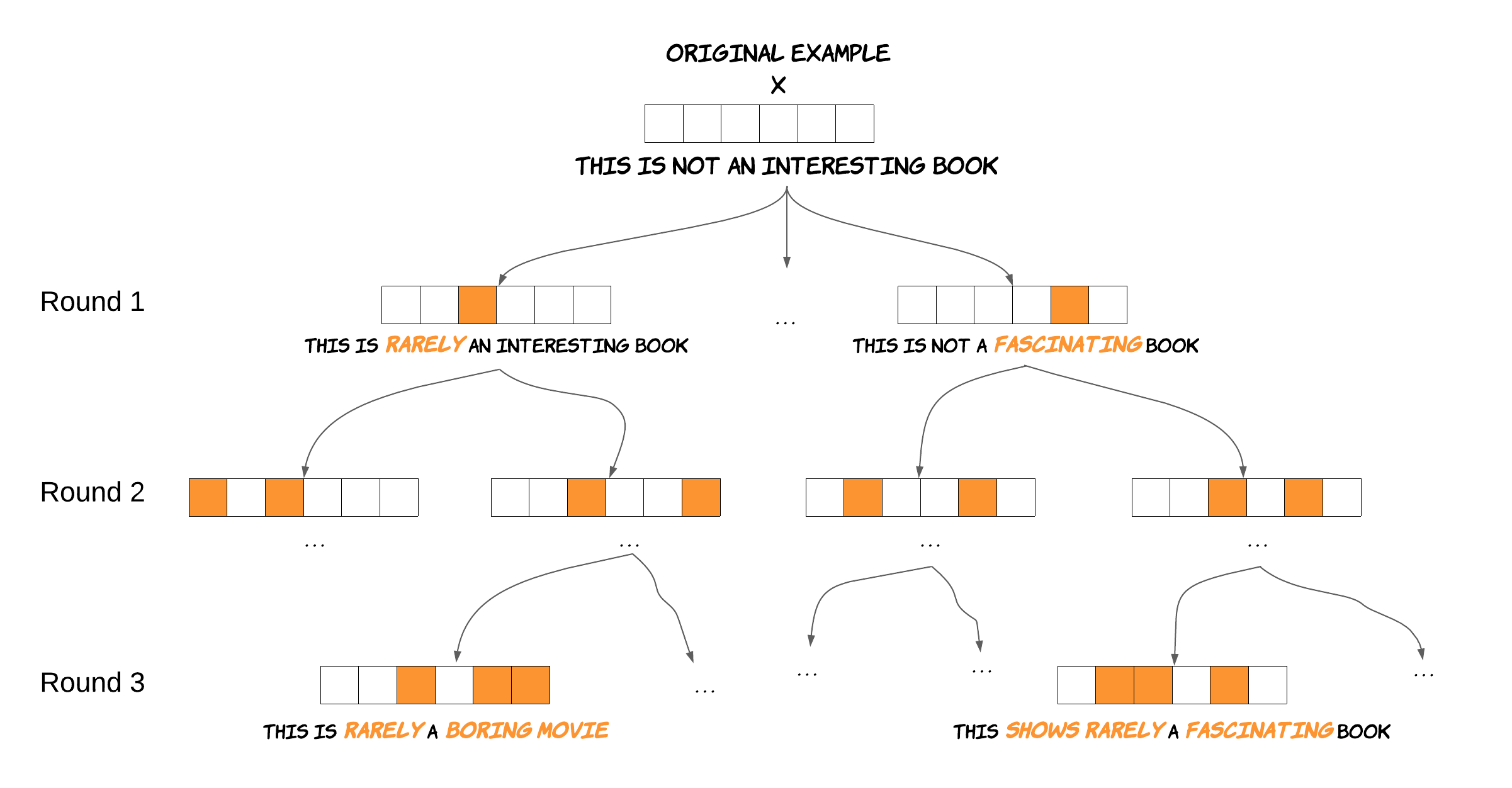}
    \caption[Illustration of the tree structure of the algorithm used to iteratively perturb the target document.]{The tree structure of the algorithm used to iteratively perturb the target document. At each round, a word from the target text is iteratively replaced by a word from its corresponding set of potential replacement words. Thus, with each successive round, the number of word replacements for generating artificial documents increases.}
    \label{fig: bfs_growing}
\end{figure}
\subsection{Counterfactual Techniques for Textual Data}
\label{subsec: counterfactual_text}
The Growing Fields algorithm (Algorithm~\ref{alg: growing}) works only for tabular data. In this section, we introduce a pair of innovative techniques referred to as Growing Language and Growing Net, for generating counterfactuals to explain the predictions of any text-based ML model. These techniques are adaptations of the Growing Spheres method~\cite{gs} to textual data. Growing Language and Growing Net both involve an iterative process in which words within the target text are successively replaced. The objective of this iterative process is to generate sparse counterfactual explanations, aiming to minimize the number of modified words. 

Algorithm~\ref{alg: growing} outlines the iterative process employed by Growing Language and Growing Net. In the first step (lines 1 to 4), both approaches generate $d$ sets of potential word replacements ($W = (W_1, \dots, W_d)$) for each word in the target document ($x = (x_1, \dots, x_d)$). The external module employed to generate the set of potential word replacements is the main distinction between these two methods. These modules are detailed in the following. Subsequently, Growing Language and Growing Net create artificial documents iteratively, detailed between lines 7 and 15 and exemplified as a tree structure in Figure~\ref{fig: bfs_growing}. These documents are generated until every word in the original document is replaced or a valid counterfactual is discovered. During each iteration, they initialize a set of $n$ artificial copies of the original document ($x$) and progressively replace individual words ($x_j$) with randomly selected words from their respective sets of potential replacements ($W_j$). 

For example, consider the target review, \textit{``This is not an interesting book''}, classified as negative by a sentiment analysis model (Figure~\ref{fig: bfs_growing}). In the first round, both Growing Language and Growing Net generate artificial documents with only one modified word. Subsequent rounds involve the replacement of two words and so on. In this process, counterfactuals are identified, and the closest one is returned as the explanation. These methods prioritize counterfactuals closely related to the original document to provide concise and meaningful explanations.

\subsubsection{Growing Net}
\label{subsec:growing_net}
\begin{figure*}[!ht]
    \begin{subfigure}{.47\linewidth}
        \centering
        \includegraphics[width=\linewidth]{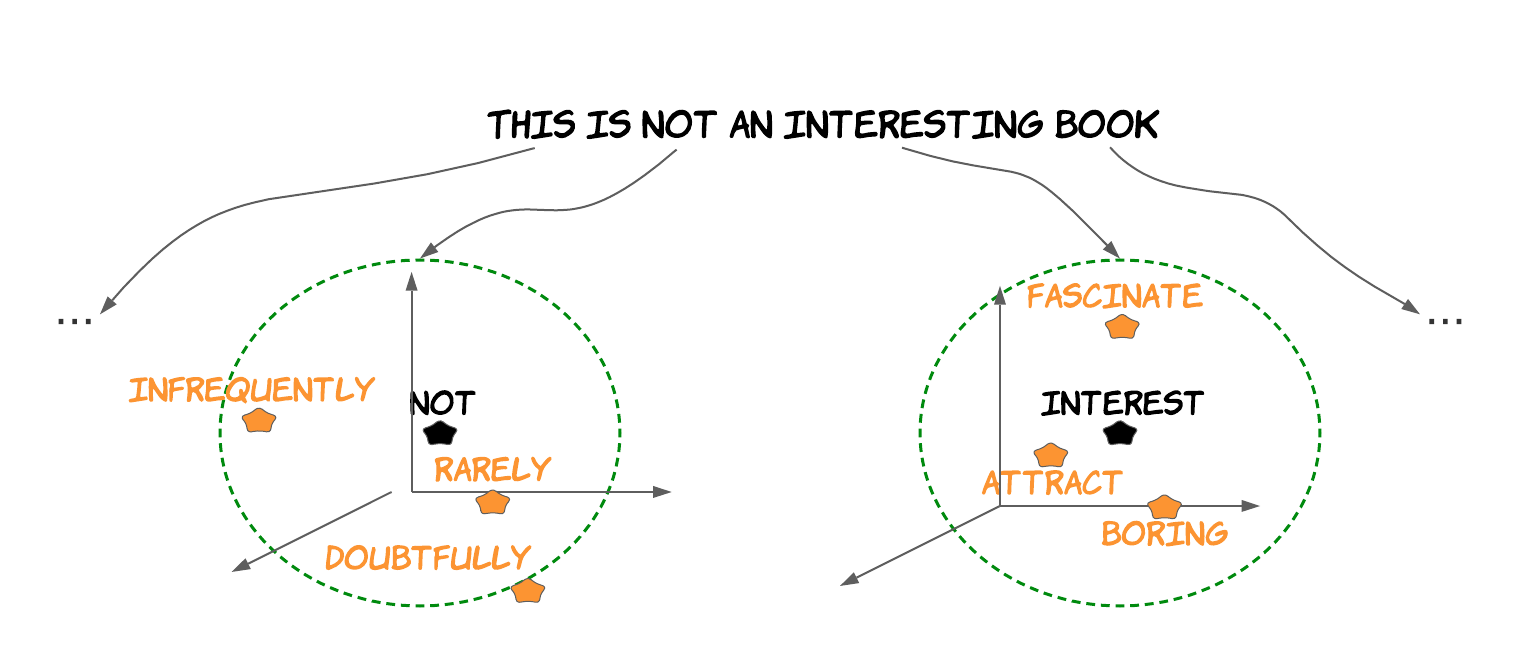}
        \caption{Growing Language}
        \label{fig: growing_language}
    \end{subfigure}
    \centering
    \begin{subfigure}{.47\linewidth}
        \centering
        \includegraphics[width=\linewidth]{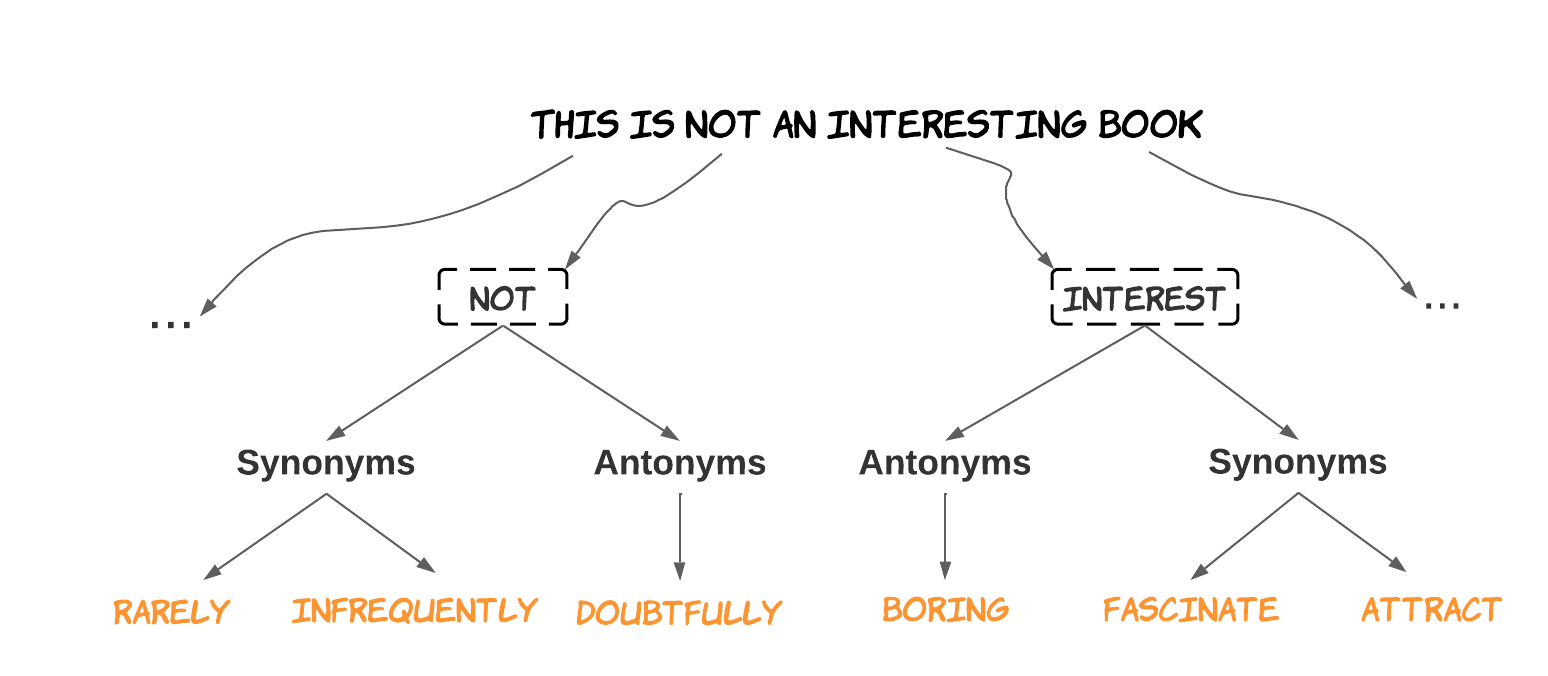}
        \caption{Growing Net}
        \label{fig: growing_net}
    \end{subfigure}
    \label{fig: growing_methods}
    \caption[Representation of the mechanism employed to generate sets of potential word replacements in Growing Language and Growing Net.]{The mechanism employed to generate sets of potential word replacements in Growing Language involves embedding words from the target text into a latent space and capturing nearby words in this space. Conversely, Growing Net leverages the tree structure of WordNet to create sets of potential word replacements.}
\end{figure*}
Growing Net capitalizes on the rich structure of WordNet~\cite{wordnet} to construct sets of closely related words. WordNet is a lexical database and thesaurus that organizes words and their meanings into a semantic tree of interrelated concepts. Therefore, Growing Net begins by creating sets of similar words for each term within the target document through WordNet as illustrated in Figure~\ref{fig: growing_net}. These sets are constructed based on the part-of-speech tags, such as verbs, nouns, determiners, and more, which ensures that the added words share the same grammatical roles as the original terms. Additionally, Growing Net identifies words that are close in the WordNet hierarchy, which includes synonyms, antonyms, hyponyms, and hypernyms. Hyponyms are words or phrases that are more specific than the target word ($x_i$), while hypernyms are more general.

Subsequently, Growing Net modifies the target document by randomly substituting words in the target sentence with words from their corresponding sets. Growing Net systematically increases the gap between the artificial instance and the original document with each successive round. Consequently, at each round, Growing Net expands the number of modified words ($k$) until it identifies counterfactuals. Ultimately, Growing Net returns the counterfactual with the smallest Wu-Palmer Similarity (Wu-P) distance~\cite{wup} as the final explanation. The Wu-Palmer Similarity is a similarity metric specifically designed for measuring the relatedness of concepts in WordNet. It considers the depth of terms in the WordNet hierarchy and the path length to their most common ancestor. In this context, the advantage of using the Wu-Palmer Similarity is that it captures not only surface-level similarity but also the depth of relatedness in the WordNet hierarchy. This is crucial because it ensures that the substituted words are not just superficially similar but also conceptually relevant to the original terms.

\subsubsection{Growing Language}
\label{subsec: growing_language}
Growing Language leverages the power of large language models to restrict the space of possible perturbations. Large language models are powerful natural language processing AI systems. They are employed in this context to embed words into numerical representations, often referred to as a latent space. In simpler terms, a latent representation consists of high-dimensional vectors that capture the underlying semantic and contextual information of the original text. This high-dimensional space facilitates the measurement of similarity between words. In the context of Growing Language, individual words from the target document are projected onto this latent space. Within this latent space, Growing Language assembles sets of candidate replacement words for each word in the original document, as illustrated in Figure~\ref{fig: growing_language}. To qualify for inclusion in these sets, words must fulfill two criteria. Firstly, they should share the same part-of-speech tags as the original words, ensuring grammatical consistency. Secondly, they must exhibit a similarity score that surpasses a predefined threshold ($\theta$). The similarity score is computed by comparing word vectors in a high-dimensional space, where each dimension represents semantic and contextual information of the words, and the score is based on the similarity between these vectors. In our experiments, we set this threshold to 0.8 on a scale from 0 to 1. This allows Growing Language to maintain computational efficiency, as starting with a low threshold could result in longer processing times. Once these sets are established, Growing Language initiates the generation of artificial documents.

The process of generating artificial documents is iterative, with Growing Language progressively substituting more words in each successive round. This iterative process continues until an artificial document is classified differently by the black-box model, effectively finding a counterfactual. Growing Language also incorporates an adaptive mechanism to handle scenarios where replacing all the words in the original document does not produce a valid counterfactual. In such instances, Growing Language proceeds to extend each set of similar words by decreasing the similarity threshold. This threshold reduction occurs iteratively and is set at two times the initial step size, which we initialize at 0.01. Consequently, the sets of replacement words become larger, containing more words that are less similar, allowing for an expanded search for counterfactuals. Should multiple counterfactuals be found, Growing Language selects the one with the fewest modifications compared to the original document. Importantly, any language model capable of converting words and measuring word similarities could be used in this process.


\section{Adapted Post-hoc Explanations}
\label{sec: ape}
\begin{figure}[t]
    \centering
    \includegraphics[width=\textwidth]{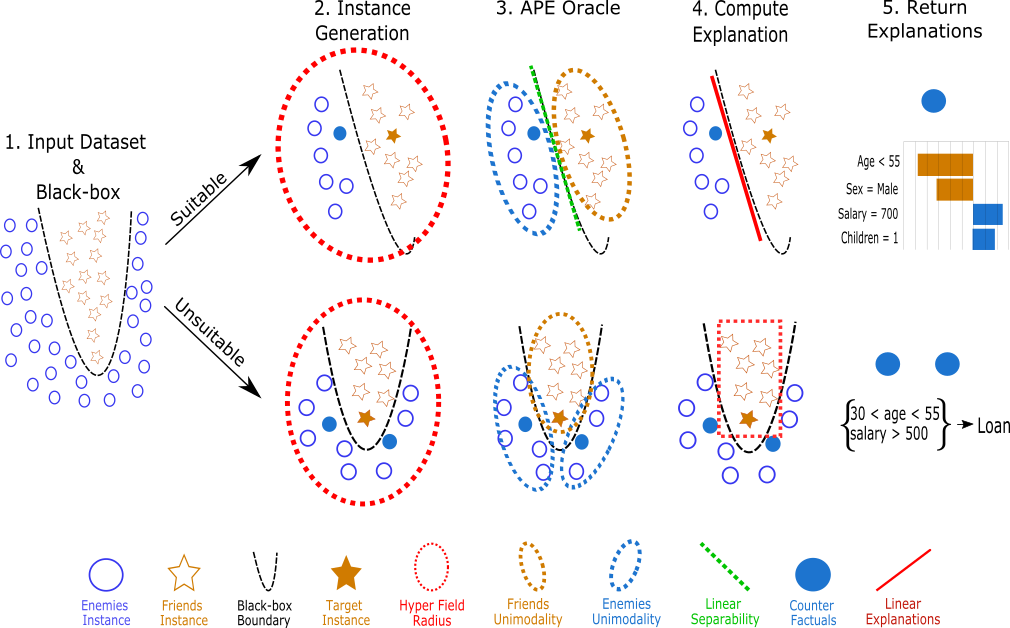}
    \caption[Illustration of the Adapted Post-hoc Explanation framework.]{The Adapted Post-hoc Explanation framework. In the first row, the Oracle's test indicates that a linear explanation is suitable, resulting in the return of a counterfactual and a linear explanation. Conversely, in the second row, the Oracle determines that a linear explanation is unsuitable, leading APE to generate a counterfactual for each cluster center along with a rule-based explanation.}
    \label{fig: framework_ape}
\end{figure}

We now elaborate on APE, our approach to compute adapted post-hoc explanations for a target instance $x$ and a black-box classifier $f$. When the decision frontier of $f$ admits a single local linear surrogate according to our problem statement in Section~\ref{sec: problem_statement}, APE returns a linear-based explanation complemented with a counterfactual explanation. Otherwise, APE recommends a different explanation paradigm such as a rule-based surrogate.

\begin{algorithm}[t]
    \small
    \caption{\textsc{APE}}\label{pseudo-code}
    \begin{algorithmic}[1]
    \Require $\text{a training dataset}\;T \subset X$, $\text{a target instance}\;x = ( x_1, \dots, x_d ) \in X$,  
    \Statex $\text{a black-box classifier}\;f:X \rightarrow Y$; number of samples $n$

    \Ensure one or multiple counterfactual instances, a surrogate classifier $g$
    \State $e \gets \;$\textsc{Growing Fields}$(T, x, f) $ 
    \State $r \gets \nicefrac{1}{\delta} \times \mathit{dist}(x, e)$ \Comment{\footnotesize $\delta$ is the largest distance in $T$ } 
    \State $Z \thicksim \mathcal{F}(T, r, e)_{i \le n}$
    \If{\textsc{APE Oracle}$(Z, x, f)$} \\
        \hspace{2ex}\Return $e$, $LS_{\mathit{APE}}(Z,f,x,e)$ trained on $e$-centered field of radius $r' \ge r$
    \Else \\
    \hspace{2ex}\Return $\{e_1, \dots, e_k\} \subset Z$, \textsc{Rule-based Surr.}$(f)$ 
    \EndIf
    \end{algorithmic}
\end{algorithm}

APE is detailed in Algorithm~\ref{pseudo-code} and Figure~\ref*{fig: framework_ape}. In 
the first stage (line 1), APE invokes an algorithm to find the black-box decision boundary such as {\it Growing Fields} for tabular data or \textit{Growing Net} and \textit{Growing Language} for textual data. This is achieved by identifying $x$'s closest enemy -- denoted by $e$. Then, APE generates a set of random instances $Z$ uniformly distributed in a locality around $e$ (line 3). This locality constitutes a \emph{field}, which APE samples using the $\mathcal{F}$ generation process already explained in Section~\ref{subsec: growing_fields}. The size of the field depends on a radius parameter that is proportional to $\mathit{dist}(x, e)$, i.e., the distance between $x$ and its closest enemy.  
More precisely, we set $r=\nicefrac{1}{\delta} \times \mathit{dist}(x, e)$, where $\delta$ is the farthest distance from $x$ to a real instance in $T$, i.e., $f$'s training set. By normalizing the radius, we (a) provide users with a clear notion of distance, and (b) reduce the risk of sampling instances beyond the limits of the attribute domains. By centering the generative process at $e$ with radius $r$, APE makes sure that $Z$ covers $x$ and contains diverse subsets $E$ and $F$ of friends and enemies of $x$ -- in concordance with the requirements (i) and (ii) in the problem statement in Section~\ref{sec: problem_statement}. The $\mathcal{F}$ generation procedure as well as the Growing Fields algorithm are detailed in Section~\ref{subsec: growing_fields}.

In the next step (line 4), APE characterizes the decision boundary of $f$. To this end, the algorithm invokes the APE Oracle (Section~\ref{subsec: ape_Oracle}), which runs efficient unimodality and linear separability tests~\cite{Alban_unimodality,thornton2002truth} on $E$ and $F$ to determine whether a linear surrogate is suitable or not. The Oracle recommends a linear explanation if both sets $E$ and $F$ exhibit an unimodal distribution, that is, if there is only one cluster per class and we can separate those clusters with a single linear surrogate. In that case, APE returns a linear explanation and the closest enemy of $x$ as a counterfactual explanation for $f(x)$. The linear explanation is learned via an extension of Local Surrogate~\cite{Laugel_Defining_locality}, called $\mathit{LS}_{\mathit{APE}}$, applied on a superset of $Z$, consisting of real and artificial instances. Those instances constitute a field with a radius of at least $r$. We elaborate on those details in Section~\ref{subsec: linear_explanations}. 

When the APE Oracle deems linear explanations unsuitable, APE proposes a rule-based surrogate. This happens when the instances in $E$ or $F$ form multiple clusters, or because $Z$ is not linearly separable. Rule-based alternatives are Anchors~\cite{Anchor} or shallow decision trees. In the first case, the user obtains a single rule of the form:
\[R: \mathbf{B} \Rightarrow f(\eta^{-1}(z)) = f(x) \;\;\;\;\;\;\;\text{with} \;\; \mathbf{B} = \bigwedge_{j \in F \subseteq \{1, \dots, d' \}}{z[j]} \] 
\noindent where $\eta : \mathcal{X}^d \rightarrow \{0, 1\}^{d'}$ is a conversion function into a surrogate interpretable space and the left-hand side (or antecedent) of the rule is a conjunction of conditions that predicts $f(x)$. This rule $R$ guarantees a precision $\mathit{prec}(R) = P(f(\eta^{-1}(z)) = f(x)\;|\;\textbf{B} \land z \in \mathcal{Z}) \ge \tau$ for a user-defined precision threshold $\tau$~\cite{Anchor}. In the second case, the user gets a decision tree trained on a superset of $Z$, consisting of real and artificial instances. Since the decision boundary may consist of several disconnected instance clusters, APE completes its explanation with a counterfactual instance per cluster in $E$ (see Section~\ref{subsec: rule_explanations}). That way users can have a comprehensive view of the different ways to change the black box's outcome $f(x)$.

Since we have already covered two of APE's foundational components, the instance generation process and the counterfactual methods, we will now elaborate on the APE Oracle and the procedures to compute the linear and rule-based surrogates.

\begin{algorithm}[!h]
    \small
    \caption{\textsc{APE Oracle}}\label{pseudo-code_ape_Oracle}
    \begin{algorithmic}[1]
    \Require instances $Z \subset X$, $\text{target instance}\;x = \{ x_1, \dots, x_d \} \in X$, $\text{ classifier}\;f:X \rightarrow Y$ 
    \Ensure Is $f$ linearly separable in $Z$ w.r.t. the class $f(x)$?
    \If{$E \subset Z$ and $F \subset Z$ are unimodal}
        \If{$Z$ is linearly separable w.r.t. $f$} \\
            \hspace{6ex}\Return {True}
        \EndIf
    \EndIf \\
    \Return {False}
    \end{algorithmic}
\end{algorithm}
\subsection{APE Oracle}
\label{subsec: ape_Oracle}
\begin{figure}
    \centering
    \includegraphics[width=\textwidth]{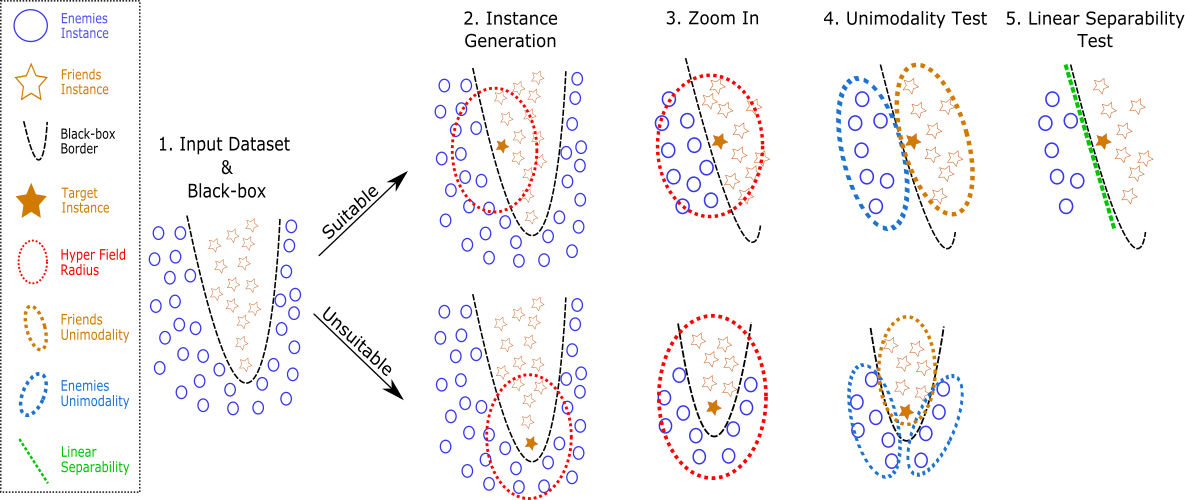}
    \caption[Description of APE Oracle assessing the suitability of a linear surrogate for approximating a classifier locally.]{A description of the APE Oracle that assesses the suitability of a linear surrogate for approximating a classifier locally. In the first row, a linear explanation is deemed suitable as friends and enemies of the target instance are grouped into a single cluster. In contrast, the second row shows that a linear surrogate is inadequate due to the enemies being distributed across two clusters.}
    \label{fig: ape-Oracle}
\end{figure}
The core of the APE algorithm is the APE Oracle described in Algorithm~\ref{pseudo-code_ape_Oracle} and Figure~\ref*{fig: ape-Oracle}. This Oracle determines whether the black-box decision boundary is separable by a single linear approximation. To achieve this, the Oracle applies the Libfolding unimodality test~\cite{Alban_unimodality} separately on the sets of friends $F$ and enemies $E$ of the target instance $x$ in $Z$. If the test is passed, it means that $F$ and $E$ form each a single cluster in $Z$. 

Various methods exist for converting textual data into numerical representations. One approach involves using vectorized representations like the `bag of words'~\cite{bag_of_word} or term-weighting~\cite{term-weighting} techniques. Another approach transforms text into latent representations, making use of methods such as AutoEncoder~\cite{autoencoder} and word2vec~\cite{word2vec}. In our case, we adapt the Libfolding test to accommodate these two text representation strategies. The first variant transforms text into a numerical vector using a bag of words vectorizer. The second variant, on the other hand, employs a sophisticated language model~\cite{spacy} to embed text into a latent space. These transformed text representations are then used as inputs for the Libfolding test, enabling us to evaluate the unimodality around the target text.

This, however, does not suffice for linear separability; ergo the Oracle carries out a quick linear separability test to determine whether these clusters of friends and enemies can be told apart with a linear approximation. The test is carried out on a balanced sample $Z_b \subseteq Z$. We enforce $Z_b$ to contain an equal number of friends and enemies of $x$ because $Z$ can be highly imbalanced towards the enemies of $x$ for very small localities.  

There are multiple methods to determine whether there exists a linear function that separates a two-class dataset. Such methods range from linear and quadratic programming to approaches based on computational geometry and neural networks~\cite{linear_separability_problem}. Nevertheless, all these strategies are at least as expensive as running a linear regression on the input dataset. 
On those grounds, APE tabular resorts to a simple test based on the Thornton's \emph{separability index} $\mathit{si}$~\cite{thornton2002truth}. If $\Gamma_{X'}(x)$ returns the closest neighbor $x'$ of $x$ in a set $X' \subseteq X$, the separability index measures the ratio of instances for which that closest neighbor is a friend of $x$. In our setting, this can be computed according to the following formula:
$$\mathit{si}(X') = \frac{\sum_{x' \in X'}{\mathds{1}_{f(\Gamma_{X'}(x')) = f(x')}}}{|X'|}.$$
We remark that $\mathit{si}$ lies between 0 and 1 and that higher values denote higher separability. Line 2 in Algorithm~\ref{pseudo-code_ape_Oracle} checks if $\mathit{si}((T \cap \Phi_x) \cup Z_b) = 1$. That is, the test also considers real instances that fall within the field from which $Z$ was drawn. If the test is passed, the decision boundary is considered linearly separable enough and the Oracle returns true.

\subsection{Linear Explanations}
\label{subsec: linear_explanations}
If the APE Oracle estimates that $f$'s decision boundary is linearly separable around the target instance $x$, APE resorts to the routine $\mathit{LS}_{\mathit{APE}}$ (described in Algorithm~\ref{pseudo-code_lsape}) to learn a linear surrogate $g$ on $Z$ and to explain $f(x)$. We could center the generative process to learn $g$ on the target instance $x$ as in standard LIME, or around the decision boundary as in LS. We opt for the latter alternative since LS has been shown to identify more accurately the features that influence the black box locally~\cite{Laugel_Defining_locality}.

We recall that $Z$ is a sample drawn from a field centered on $e$ with radius $r= \nicefrac{1}{\delta} \times \textit{dist}(x, e)$ where $e$ is $x$'s closest enemy. We could therefore learn $g$ from the instances used for the linear separability test because these are exactly what LS needs for training. We highlight, however, that nothing prevents our linear surrogate from attaining a good adherence in larger scopes. In concordance with our maximality requirement (Section~\ref{sec: problem_statement}), $\mathit{LS}_{\mathit{APE}}$ carries out a posteriori expansion of the training field before reporting the linear explanation to the user. While the adherence does not decrease, that is while $m(g) \ge \tau$, $\mathit{LS}_{\mathit{APE}}$ extends the field radius and trains a new linear explanation (line 5-10 in Algorithm~\ref{pseudo-code_lsape}). The threshold $\tau$ is set to the adherence of $g$ in the initial field. We recall that the function $m$ serves as a metric for quantifying the adherence of a surrogate relative to a model within a specific locality and can encompass metrics such as accuracy, precision, or recall. The radius is increased using the same expansion strategy of Growing Fields (lines 8-9 in Algorithm~\ref{pseudo-code_Growing_fields}). Besides, we precise that we do not decrease the radius of the field because we want to guarantee the presence of the target instance within the field.

\begin{algorithm}
    \small
    \caption{\textsc{Extended Local Surrogate ($\mathit{LS}_{\mathit{APE}}$)}}
    \label{pseudo-code_lsape}
    \begin{algorithmic}[1]
    \Require $\text{instances}\;Z \subset X \text{ drawn from a field}$, $\text{a classifier}\;f:X \rightarrow Y$,
    \Statex $\text{target and counterfactual instance}\,x,e  \in X$, an adherence metric $m$;
    \Statex 
    Hyper-parameters: $\theta=0.05$

    \Ensure a linear surrogate classifier $g$
    \State $r \gets \nicefrac{1}{\delta} \times \mathit{dist}(x, e)$
    \State Split $Z$ into $Z_{\textit{train}}, Z_{\textit{test}}$ 
    \State $g \gets$ \textsc{Linear Regression}$(Z_{\textit{train}}, f(Z_{\textit{train}}))$
    \State $a \gets \tau \gets  m(g) \text{ on } Z_{\textit{test}}$
    \While {$a \ge \tau \land r < 1$}
        \State $r \gets \theta \times r$
        \State $Z \thicksim \mathcal{F}(T, r, e)_{i \leq n}$
        \State Split $Z$ into $Z_{\textit{train}}, Z_{\textit{test}}$ 
        \State $g \gets$ \textsc{Linear Regression}$(Z_{\textit{train}}, f(Z_{\textit{train}}))$
        \State $a \gets  m(g) \text{ on } Z_{\textit{test}}$    \EndWhile \\
    \Return $g$
    \end{algorithmic}
\end{algorithm}

\subsection{Rule-based Explanations}
\label{subsec: rule_explanations}
If the decision frontier in the vicinity of our target instance is too complex to be approximated with a single linear surrogate, users may apply clustering techniques on the neighborhood $Z$ and provide different linear explanations for each of the instance clusters at the decision boundary. This would provide a complete picture of the black box behavior around the target. However, such an explanation is potentially difficult to grasp for users, because it might consist of potentially contradicting feature-attribution rankings. On those grounds, APE proposes by default a rule-based explanation when linear surrogates are considered unsuitable. Alternatives are anchors or shallow decision trees. Anchors~\cite{Anchor} learns a single explanation rule of the form $p \Rightarrow f(x)$ such that $p$ is a conjunction of conditions of maximal coverage and the rule has a precision of at least $\tau$. The decision tree is learned on the set $Z$ containing both friends $F$ and enemies $E$ of $x$ in the field centered on $e$, the closest enemy of $x$. We remark, nevertheless, that our framework could be coupled with other explanation approaches~\cite{SHAP, LORE, Delaunay}. This is an interesting avenue for future research.

Finally, APE complements the rule-based explanation with a set of counterfactual instances $\{e_1, \dots, e_k\} \subset E^*$. These are the centroids of the clusters defined by an extended set of enemies $E^* \supseteq E$ (generated using the $\mathcal{F}$ generation process from Algorithm~\ref{pseudo-code_distribution}). This set can be obtained by increasing the field ratio $r$ while the precision of the explanation is above $\tau$. The clusters are computed using K-means~\cite{k-means} and the number of clusters $k$ is determined using the Elbow method~\cite{elbow_method}. 

\subsection{Illustrative Example}
\begin{figure}[t]
    \centering
    \includegraphics[width=0.9\textwidth]{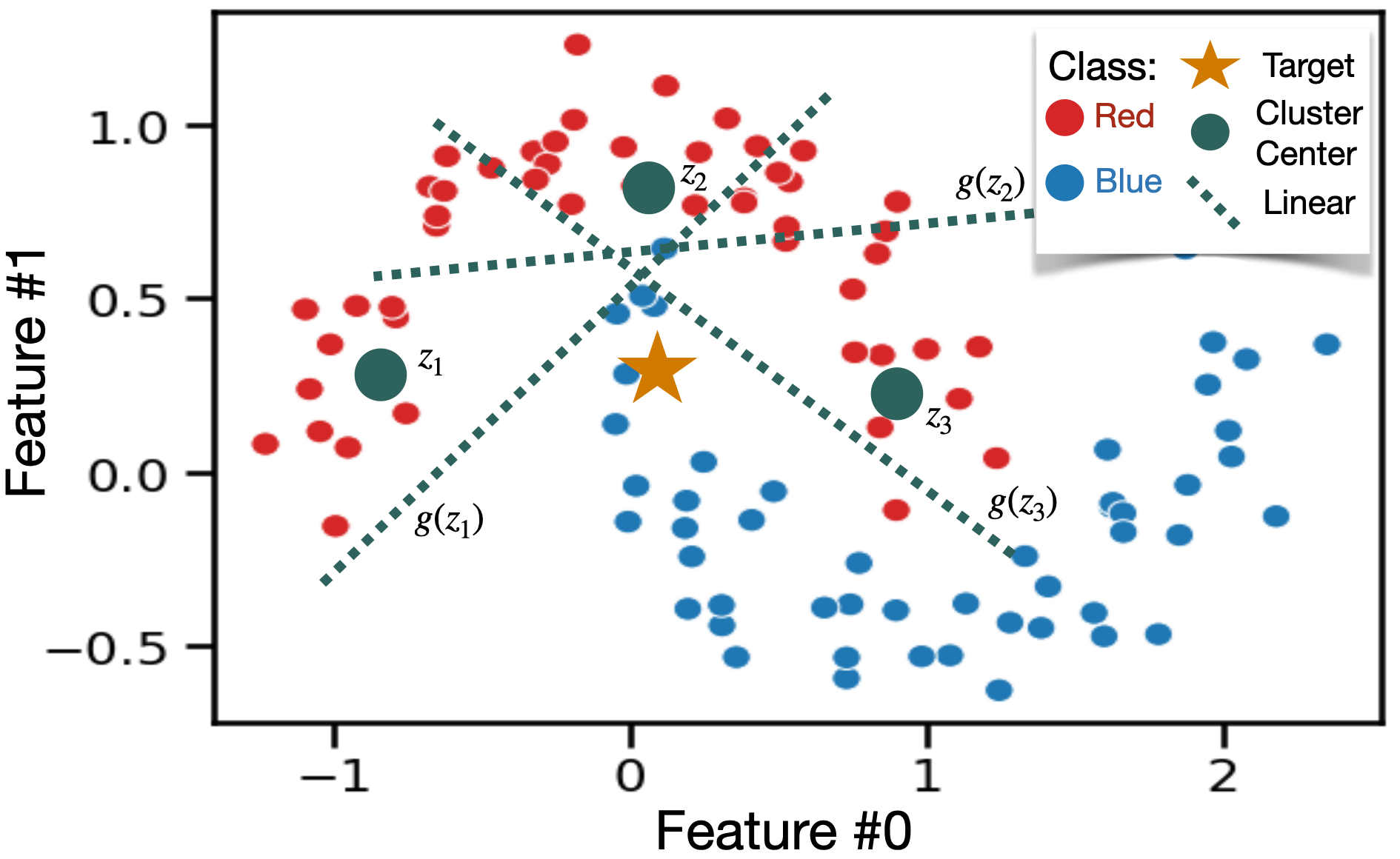}
    \caption[An illustrative example of the moon dataset.]{An illustrative example of the moon dataset. The yellow star depicts the target instance while the green circles $z_1$, $z_2$, and $z_3$ represent the cluster centers of the counterfactuals. Corresponding linear explanations are indicated by the dashed lines of the same color.}
    \label{fig: illustrative_example}
\end{figure}

We further motivate the utility of APE through an example drawn from the \emph{Moons} dataset consisting of 2 features as shown in Figure~\ref{fig: illustrative_example}. The application of the Libfolding unimodality test on the set of closest enemies $E$, represented by red circles surrounding the target instance $x = [0.05, 0.29]$ reveals a multimodal distribution. The k-elbow method identifies three enemy clusters whose centers are: $z_1 = [-0.91, 0.25]; z_2 = [0.01, 0.76]$ and $z_3 = [0.96, 0.19]$. Subsequently, we apply LS$_{\mathit{APE}}$ on those counterfactual instances, serving as centers for the generative process to learn linear explanations. Our observations reveal contradictory explanations since the attribution of the first feature for $z_1$ is $0.79$ whereas for $z_3$ it is $-0.53$.

We also demonstrate examples of the explanations provided by APE using the \emph{Titanic} dataset. In this dataset, the task is to predict whether a passenger of the Titanic survived or not. We consider two instances, $x_1$ and $x_2$ with the following attributes values: 
\begin{align*}
    x_1 = \{ \mathit{age}=42, \mathit{class}=1, \mathit{sex}=F, \mathit{siblings}=0, \mathit{children}=3 \},\\
    x_2 = \{ \mathit{age}=25, \mathit{class}=2, \mathit{sex}=M, \mathit{siblings}=1, \mathit{children}=0 \}. 
\end{align*}
A multi-layer perceptron classifier predicts $f(x_1)=\mathit{Survived}$ and $f(x_2)=\mathit{Died}$. APE Oracle determines that a linear explanation is suitable to approximate the model in the locality of $x_1$. Thus, it provides the following explanations $g_1$:
\begin{align*}
    g_1 = \{ \{ e \}, \ g(x)_{\mathit{survived}} = 0.5\times \mathit{sex} + 0.2 \times \mathit{class} + 0.2 \times \mathit{children}  \},
\end{align*}
with the counterfactual explanation (only perturbed attributes shown) $e=\{ \dots, sex=M, \dots \}$. This explanation suggests that changing the individual's gender is sufficient to predict survival, highlighting gender as the most important factor. The linear explanation also indicates that the number of brothers and sisters does not impact the model prediction.

Conversely, APE Oracle determines that the region around $x_2$ is not linearly separable, and a rule-based along with counterfactual explanations would be more appropriate. Therefore, APE produces the explanation $g_2$:
\begin{align*}
    g_2 = \{ \{ e_1, e_2, e_3 \}, \ \mathit{sex}=M \land \mathit{children}=0 \Rightarrow \mathit{Died} \}, \\
\end{align*}
with counterfactuals $e_1=\{ \dots, \mathit{sex=F}, \dots \}$, $e_2= \{ \dots, \mathit{children}=2, \dots \}$, and $e_3=\{ \dots \mathit{sex}=F, \ \mathit{children}=1, \dots \}$. The rule-based explanation specifies the conditions under which the model confidently predicts that an individual died on the Titanic. In contrast, the counterfactual explanations reveal that, for instance, if the individual were a female, as depicted by $e_1$, or had two children, as suggested by $e_2$, the model would have predicted survival.

\section{Experiments}
\label{sec: ape_experiments}
We executed three series of experiments to evaluate the performance of APE on tabular and textual data:
\begin{itemize}[leftmargin=*]
    \item In the first set of experiments (Section~\ref{sec: unimodal_test_evaluation}), we focused on the APE's Oracle, specifically on its ability to identify when a linear explanation provides an accurate approximation for a given black-box classifier and target instance.
    \item In the second set of experiments, we compared the explanations generated by APE with those produced by LIME~\cite{LIME} and LS~\cite{Laugel_Defining_locality} in terms of adherence (Section~\ref{subsec: fidelity_coverage_evaluation}).
    \item The third set of experiments (Section~\ref{sec: ablation_study}) involved an ablation study of the APE Oracle's two components to assess their impact on the adherence of APE.
\end{itemize}
Then, we compare in Section~\ref{subsec: counterfactuals_eval}, the quality of the counterfactual generated by Growing Fields with those output by Growing Spheres~\cite{gs}. The source code of Growing Fields, APE, as well as the experimental datasets and additional results, are available on Github\footnote{\url{https://github.com/j2launay/APE}}.

\subsection{Experimental Setup}
\subsubsection{Datasets}
Table~\ref{tab: datasets} describes our tabular experimental datasets. The list comprises 6 real and 6 synthetic datasets, the latter generated with scikit-learn\footnote{\url{http://scikit-learn.org}}. Five of those synthetic datasets contain only numerical features. The real datasets were chosen to provide a mix of numerical and categorical features. For the experiments on textual data, we used four datasets described in Table~\ref*{tab: datasets_text}. This includes three real-world and one dataset created by combining two distinct sets of articles, one with real articles~\footnote{\url{https://www.kaggle.com/datasets/rmisra/news-category-dataset}}, and one with fake articles~\footnote{\url{https://www.kaggle.com/competitions/fake-news/overview}}. Each of these datasets was designed for binary classification tasks, except for the Ag News dataset, which features four target classes. Moreover, it is worth noting that a multi-class classification problem can always be redefined as a collection of binary classification problems, with one binary problem per class.

\begin{table}[ht]
    \small
        \centering
        \begin{tabular}{lcccccccc}
            \multirow{2}{*}{Name} &  \multicolumn{2}{c}{Features} & \multirow{2}{*}{Instances} & \multicolumn{5}{c}{Models} \\ 
            \cmidrule(r){2-3} \cmidrule(r){5-9}
            & Numerical & Categorical & & GB & MLP & RF & VC & SVM \\ \toprule
            Adult  & 2 & 10 & 48842 & 86\% & 84\% & 84\% & 79\% & 71\% \\
            Blob $\dagger$  & 2 & 0 & 1000 & 94\% & 93\% & 94\% & 94\% & 94\% \\
            Blobs $\dagger$  & 12 & 0 & 5000 & 97\% & 98\% & 97\% & 98\% & 99\% \\
            Blood  & 4 & 0 & 748 & 74\% & 62\% & 75\% & 50\% & 44\% \\
            Cat Blobs $\dagger$ & 4 & 4 & 5000 & 97\% & 98\% & 98\% & 99\% & 97\% \\
            Cancer & 10 & 20 & 569 &  97\% & 94\% & 98\% & 98\% & 92\% \\
            Circles $\dagger$  & 2 & 0 & 1000 & 94\% & 95\% & 94\% & 96\% & 97\% \\
            Diabetes  & 8 & 0 & 768  & 70\% & 66\% & 74\% & 74\% & 71\% \\
            M Blobs $\dagger$ & 20 & 0 & 7500 & 99\% & 99\% & 99\% & 99\% & 100\%  \\
            Moons $\dagger$ & 2 & 0 & 1000 & 97\% & 88\% & 97\% & 94\% & 97\% \\
            Mortality & 15 & 52 &  1614 & 64\% & 69\% & 67\% & 66\% & 69\% \\
            Titanic  & 1 & 5 & 1046 & 83\% & 87\% & 84\% & 87\% & 61\% \\ \bottomrule
        \end{tabular}
        \caption[Number of instances, numerical and categorical features for each dataset.]{Number of instances, numerical and categorical features for each dataset. The $\dagger$ indicates that datasets are synthetic. The last five columns represent the black-box model's test set accuracy.}
        \label{tab: datasets}
\end{table}

\begin{table}[ht]
    \centering
    \normalsize
    \begin{tabular}{lcccccccc}
        \multirow{2}{*}{Name} &  \multicolumn{3}{c}{Nb Words} & \multirow{2}{*}{Instances} & \multicolumn{4}{c}{Model} \\ 
        \cmidrule(r){2-4} \cmidrule(r){6-9}
        & Total & Average & STD & & NN & RF & NB & BERT\\ \toprule
        Polarity  &  11646 & 20.8 & 9.3 & 10660 &  72\% & 67\% & 72\% & 83\% \\
        Fake $\dagger$  &  19419 & 11.8 & 3.2 & 4025  &  84\% & 84\% & 87\% & 91\% \\ 
        Ag News & 92806 & 38.8 & 11 & 12000 & 86\% & 87\% & 90\% & 97\%\\
        \bottomrule 
    \end{tabular}
    \caption[Information about the experimental datasets with textual data.]{Information about the experimental datasets with textual data. $\dagger$ indicates generated datasets. The three columns under ``Nb Words'' represent respectively (a) the total number of distinct words in the whole dataset, (b) the average number of words per sentence, and (c) the standard deviation. The fourth column indicates the number of text documents per dataset. The final columns show the average accuracy of the different complex models.}
    \label{tab: datasets_text}
\end{table}

\subsubsection{Black-box Classifiers}
We assessed APE's performance on a variety of classifiers with different architectures -- i.e., ensemble methods, piecewise-constant functions, smooth functions -- implemented in scikit-learn~\cite{scikit_learn} with default values for the hyperparameters unless stated otherwise. For tabular data, the classifiers included: (i) Gradient Boosting (GB) with 20 tree estimators, (ii) Multi-layer Perceptron (MLP) with a logistic activation function, (iii) Random Forest (RF) with 20 tree estimators, (iv) Gaussian Naive Bayes (NB), (v) Support Vector Machine (SVM) with a balanced class weight, (vi) Decision Tree (DT), (vii) Logistic Regression (LR) and (viii) a Voting ensemble (VC) composed of LR, SVM, and NB classifiers. For textual data, the classifiers comprised: (i) Multi-Layer Perceptron (MLP) with 100 neurons and four layers, (ii) Random Forest with 500 trees, (iii) Gaussian Naive Bayes (NB), and (iv) a BERT base model (BERT) implemented in the python package ``transformers''~\cite{DistilBERT}, fine-tuned on the dataset. In addition to the class of an instance, the classifiers can provide class probabilities. The classifiers were trained on 70\% of the data points and their accuracy was tested on the remaining 30\%. Table~\ref{tab: datasets} presents the accuracy of the tabular models while Table~\ref{tab: datasets_text} exhibits the accuracy of the textual models. 

\subsubsection{Explanation modules} 
APE and the competitors were tested on a random sample of 100 target instances drawn from the test datasets. All the explanation modules had access to the training set used for training the classifiers (referred to as $T$ in Algorithm~\ref{pseudo-code}). APE's evaluation considered two variants: one using Anchors and the other employing shallow decision trees (with a maximum depth of 3) as explanation solutions when linear explanations were considered unsuitable. These variants are denoted as $\textit{APE}_a$ and $\textit{APE}_t$, respectively. Anchors requires a precision goal parameter $\tau$ for rules, which we set to 0.95. Nevertheless, the semantics of $\tau$ are purely indicative, as the algorithm will always generate an explanation even if the specified precision goal is unattainable in the surrogate's training set. In line with LIME and LS, the training instances for learning the linear surrogate were labeled with the class probabilities of the target class $f(x)$ as provided by the black-box classifiers.

\subsubsection{Metrics} 
To assess the quality of explanations, we employed various metrics. Adherence, which measures how well explanation surrogates match the behavior of the original model, was evaluated by calculating the accuracy score of the surrogate models within the region (e.g., field) on which they were trained. This evaluation used 70\% of the generated artificial instances (lines 5 and 7 in Algorithm~\ref{pseudo-code}) for training the surrogates, while the remaining 30\% were reserved for accuracy evaluation. Explanation fidelity was assessed differently for tabular and textual data. In cases where the actual features used by the input classifier were known, fidelity was measured using precision and the Kendall rank correlation coefficient on the sets of features reported by the explanations. The precision score indicates the proportion of features in the explanation that were effectively used by the black-box classifier. The Kendall coefficient quantifies the agreement between the feature attribution rankings of the explanation and the actual contribution ranking in the black box. For textual methods, explanation fidelity was evaluated by removing words identified as important by the explanation and measuring the resulting change in the black box prediction. 

\begin{table*}[t]
    \centering
    \scriptsize
    \addtolength{\leftskip} {-3cm}
    \addtolength{\rightskip}{-2.5cm}
    \begin{tabular}{lccccccccccccccc}
    \toprule
    & \multicolumn{15}{ c }{Is a Linear Explanation Suitable?}\\
    \midrule
    &    \multicolumn{3}{ c }{GB} &  \multicolumn{3}{ c }{MLP} &  \multicolumn{3}{ c }{RF} &   \multicolumn{3}{ c }{VC} &  \multicolumn{3}{ c }{SVM}  \\
    \cmidrule(r){2-4}   \cmidrule(r){5-7} \cmidrule(r){8-10}   \cmidrule(r){11-13} \cmidrule(r){14-16}
     & Yes  & No & \textit{$Prop$}$_{no}$  & Yes  & No & \textit{$Prop$}$_{no}$ & Yes  & No & \textit{$Prop$}$_{no}$ & Yes  & No & \textit{$Prop$}$_{no}$ & Yes  & No & \textit{$Prop$}$_{no}$ \\
    \midrule
    Adult &  \textbf{0.555} &  0.486 & \textcolor{orange}{0.65} &  \textbf{0.507} &  0.397 & \textcolor{orange}{0.60} &  \textbf{0.659} &  0.483   & \textcolor{orange}{0.47} &  \textbf{0.334} &  0.304 & \textcolor{blue}{0.25} &  \textbf{0.679} &  0.643 & \textcolor{orange}{0.35} \\
    Blob &  \textbf{0.891} &  0.782 & \textcolor{orange}{0.57} &  \textbf{0.890} &  0.760 & \textcolor{orange}{0.49} &  \textbf{0.874} &  0.730 & \textcolor{orange}{0.56} &  \textbf{0.899} &  0.748 & \textcolor{orange}{0.46} &  \textbf{0.894} &  0.744 & \textcolor{orange}{0.43} \\
    Blobs &  \textbf{0.855} &  0.636 & \textcolor{red}{0.78} &  \textbf{0.723} &  0.606 & \textcolor{red}{0.86} &  \textbf{0.783} &  0.655 & \textcolor{red}{0.82} &  \textbf{0.745} &  0.610 & \textcolor{red}{0.68} &  \textbf{0.717} &  0.599 & \textcolor{red}{0.80}\\
    Blood &    \textbackslash &  0.437 & \textcolor{red}{0.99} &    \textbackslash &  0.497 & \textcolor{red}{1.00} &    \textbackslash &  0.283 & \textcolor{red}{1.00} &    \textbackslash &  0.223 & \textcolor{red}{1.00} &    \textbackslash &  0.622 & \textcolor{red}{1.00} \\
    Cancer & \textbf{0.502} & 0.381 &  \textcolor{blue}{0.20} & \textbf{0.501} & 0.499 & \textcolor{blue}{0.12} & 0.510 & \textbackslash &  \textcolor{blue}{0.00} & \textbf{0.411} & 0.382 & \textcolor{blue}{0.21} & 0.499 & \textbackslash &  \textcolor{blue}{0.02} \\
    Cat Blobs &  \textbf{0.910} &  0.898 & \textcolor{red}{0.70}  &  \textbf{0.958} &  0.900 & \textcolor{red}{0.86}  &  0.874 &  \textbf{0.958}  & \textcolor{orange}{0.50} &  \textbf{0.967} &  0.936 & \textcolor{red}{0.72} &  \textbf{0.883} &  0.794 & \textcolor{orange}{0.48} \\
    Circles &  \textbf{0.945} &  0.723 & \textcolor{blue}{0.09} &  0.958 &    \textbackslash & \textcolor{blue}{0.00}  &  \textbf{0.950} &  0.708 & \textcolor{blue}{0.04}  &  0.948 &    \textbackslash & \textcolor{blue}{0.00}  &  0.949 &    \textbackslash & \textcolor{blue}{0.00}\\
    Diabetes &  \textbf{0.630} &  0.399 & \textcolor{red}{0.92} &  \textbf{0.802} &  0.585 & \textcolor{red}{0.96} &    \textbackslash &  0.453 & \textcolor{red}{0.98} &  \textbf{0.673} &  0.258 & \textcolor{red}{0.96} &  \textbf{0.717} &  0.518 & \textcolor{red}{0.88} \\
    M Blobs &    \textbackslash &  0.833 & \textcolor{red}{0.97} &    \textbackslash &  0.967 & \textcolor{red}{1.00} &  \textbf{0.863} &  0.845 & \textcolor{red}{0.82} &    \textbackslash &  0.947 & \textcolor{red}{0.99} &  \textbf{0.944} &  0.942 & \textcolor{red}{0.71}\\
    Moons &  \textbf{0.923} &  0.708 & \textcolor{orange}{0.55} & \textbf{0.917} &  0.802 & \textcolor{orange}{0.59} &  \textbf{0.918} &  0.727 & \textcolor{orange}{0.42} &  \textbf{0.916} &  0.881 & \textcolor{red}{0.85} &  \textbf{0.920} &  0.750 & \textcolor{orange}{0.50} \\
    Mortality &    \textbackslash &  0.826 & \textcolor{red}{1.00} &    \textbackslash &  1.000 & \textcolor{red}{1.00} &    \textbackslash &  0.839 & \textcolor{red}{1.00} &    \textbackslash &  0.518 & \textcolor{red}{1.00} &    \textbackslash &  0.420 & \textcolor{red}{1.00}  \\
    Titanic &  \textbf{0.761} &  0.667  & \textcolor{blue}{0.06} &  0.919 &    \textbackslash & \textcolor{blue}{0.00} &  0.973 &  \textbf{1.000} & \textcolor{blue}{0.04} &  \textbf{0.999} &  0.997 & \textcolor{blue}{0.16} &  0.715 &    \textbackslash & \textcolor{blue}{0.00} \\
    \bottomrule
    \end{tabular}
    \caption[Average accuracy per black-box model and tabular dataset for $\mathit{LS}_{\mathit{APE}}$ in relation to both Oracle outcomes.]{
        Average accuracy calculated on 100 instances per black-box model and tabular dataset for $\mathit{LS}_{\mathit{APE}}$ in relation to both Oracle outcomes. The columns labeled ``Yes'' and ``No'' represent the average accuracy of $\mathit{LS}_{\mathit{APE}}$ when the Oracle indicates that a linear explanation is suitable or unsuitable. ``\textbackslash{}'' denotes a non-meaningful accuracy score, i.e., there were fewer than 3 instances in that case. The columns labeled $\textit{Prop}_{no}$ denote the proportion of cases where the Oracle does not predict linear suitability. The colors blue, orange, and red correspond to $\textit{Prop}_{no} \le 33\%$,  $33\% > \textit{Prop}_{no} \ge 66\%$, and $\textit{Prop}_{no} \ge 66\%$ respectively. Each row reports results for a specific dataset, such as ``Adult'' in the first row.
    } 
    \label{tab: score_lse_unimodal}
    \end{table*}

\subsection{APE Oracle Evaluation}
\label{sec: unimodal_test_evaluation}
In this section, we evaluate the APE Oracle's ability to determine whether a linear surrogate is appropriate for a given scenario through two key evaluations. Both evaluations involve comparing the adherence and fidelity of linear surrogates learned using $\mathit{LS}_{\mathit{APE}}$ when the Oracle predicts suitability and when it does not. We expect higher adherence and fidelity in the first case.

\subsubsection{Adherence Evaluation}
We computed the adherence (accuracy) of the linear surrogate for each black-box classifier across 100 test instances on our experimental datasets. The surrogates were computed using $\mathit{LS}_{\mathit{APE}}$. For each target instance, the APE Oracle determines whether or not the decision boundary admits a single accurate linear approximation (Yes or No).\\

\paragraph*{Tabular data.} Table~\ref{tab: score_lse_unimodal} presents the mean adherence (accuracy) of the linear surrogates. The results clearly confirm that (i) the modality of the instances at the decision boundary plays a pivotal role in the quality of a linear surrogate, and (ii) APE's linear suitability test is pertinent. Specifically, when the APE Oracle predicts a linearly separable decision boundary, the surrogate's accuracy is on average 0.124 points higher compared to cases where the Oracle predicts the opposite.\\

\begin{table*}[t]
    \centering
    \footnotesize
    \addtolength{\leftskip} {-1cm}
    \addtolength{\rightskip}{-1cm}
    \begin{tabular}{lcccccccccccc}
    \toprule
    \multicolumn{12}{ c }{Is a Linear Explanation Suitable?}\\
    \midrule
    &    \multicolumn{3}{ c }{RF} &  \multicolumn{3}{ c }{MLP} &  \multicolumn{3}{ c }{NB} &   \multicolumn{3}{ c }{BERT} \\
    \cmidrule(r){2-4}   \cmidrule(r){5-7} \cmidrule(r){8-10}   \cmidrule(r){11-13}
     & Yes  & No & \textit{$Prop$}$_{no}$  & Yes  & No & \textit{$Prop$}$_{no}$ & Yes  & No & \textit{$Prop$}$_{no}$ & Yes  & No & \textit{$Prop$}$_{no}$ \\
    \midrule
    Ag News & \textbf{0.994} & 0.988 & \textcolor{orange}{0.63} & \textbf{0.992} & 0.987 & \textcolor{orange}{0.58} & \textbf{0.992} & 0.988 & \textcolor{orange}{0.65} & 0.925 & \textbf{0.965} & \textcolor{orange}{0.41} \\
    Fake & \textbf{0.941} & 0.891 & \textcolor{blue}{0.14} & \textbf{0.822} & 0.819 & \textcolor{blue}{0.11} & 0.833 & \textbf{0.864} & \textcolor{blue}{0.11} & \textbf{0.952} & 0.938 & \textcolor{blue}{0.08} \\
    Polarity & \textbf{0.808} & 0.770 & \textcolor{orange}{0.36} & \textbf{0.87} & 0.851 & \textcolor{blue}{0.27} & \textbf{0.857} & 0.851 & \textcolor{blue}{0.27} & \textbf{0.965} & 0.956 & \textcolor{orange}{0.41} \\
    \bottomrule
    \end{tabular}
    \caption[Average accuracy per black-box model and textual dataset for $\mathit{LS}_{\mathit{APE}}$ with respect to both oracle outcomes.]{
        Average accuracy computed on 100 instances per black-box model and textual dataset for $\mathit{LS}_{\mathit{APE}}$ with respect to both oracle outcomes. The columns labeled ``Yes'' and ``No'' represent the average accuracy of $\mathit{LS}_{\mathit{APE}}$ when the oracle indicates the suitability or unsuitability of a linear explanation. ``\textbackslash{}'' denotes a non-meaningful accuracy score, i.e., there were less than 3 instances in that case. The columns labeled $\textit{Prop}_{no}$ indicate the ratio of cases where the oracle does not predict linear suitability. The colors blue, orange, and red correspond to $\textit{Prop}_{no} \le 33\%$,  $33\% > \textit{Prop}_{no} \ge 66\%$, and $\textit{Prop}_{no} \ge 66\%$ respectively.
    } 
    \label{tab: score_lse_unimodal_text}
    \end{table*}
\paragraph*{Textual data.} Table~\ref{tab: score_lse_unimodal_text} reports the average accuracy of $\mathit{LS}_{\mathit{APE}}$ for both possible outcomes of the APE Oracle. The results we obtained demonstrate that the APE Oracle's suitability test also works for textual data. Therefore, we observe that learning linear explanations in line with the recommendations of the APE Oracle test guarantees explanations with higher adherence. This underscores the importance of inspecting the decision boundary before generating an explanation. The success of the APE Oracle in improving explanation quality for both tabular and textual data, suggests its potential practical application in real-world scenarios.\\

\paragraph*{General Insights.} Our observations also reveal that the proportion of linearly separable cases is primarily influenced by the characteristics of the dataset. For example, datasets such as Blood and Mortality exhibit little suitability for linear explanations, while datasets such as Circles, Fake, and Titanic tend to have a higher proportion of cases that are adapted for linear surrogates. However, it is worth noting that the choice of the black-box classifier's architecture can also impact this proportion. For instance, in the Adult dataset, only 25\% of the target instances within the Voting Ensemble (VC) are considered unsuitable for a linear explanation, which contrasts with other datasets where this proportion is higher. Interestingly, this pattern differs when we consider the Gradient Boosting (GB) classifier, where 65\% of the target instances are deemed unsuitable for linear explanations according to the Oracle. 

Moreover, we observe that even in cases where the Oracle rejects linear suitability, the adherence of the linear surrogate can still be high. For instance, the Cat Blobs dataset with the GB black box exemplifies this phenomenon. This can be attributed to the fact that multimodal scenarios characterized by clustered data, may still exhibit a degree of linear separability if the individual clusters mostly consist of instances from the same class. In such cases, the APE framework tends to favor rule-based explanations accompanied by multiple counterfactual instances. This preference for rule-based explanations, along with the production of several counterfactuals, demonstrates various ways to alter the classifier's prediction. This highlights the complex nature of the decision boundary in the locality of the target instance. Consequently, APE adopts a two-step process, first testing for unimodality and then assessing linear separability, to effectively address these diverse scenarios.

\subsubsection{Fidelity Evaluation}
\label{subsec: simulated_users_evaluation}
This evaluation aims to show that when the APE Oracle indicates the suitability of a linear explanation, then this explanation faithfully reflects the inner workings of the black box. A good explanation is one that relies solely on features genuinely used by the complex model. To gauge the fidelity of linear surrogates under both potential outcomes of the APE Oracle, we employ two distinct metrics. For tabular models, we resort to a set of  ``glass'' black-box classifiers, essentially white-box classifiers treated as black boxes, where we have control over the features used for predicting the class of any instance. Conversely, since it is more difficult to control the words used by a prediction model on textual data, we assess explanation fidelity through the insertion and deletion score. This score measures the difference between the black box model's prediction of the original text and the same text where the words considered as important by the explanation are removed.\\

\begin{figure}[!t]
    \centering
    \includegraphics[width=0.9\linewidth]{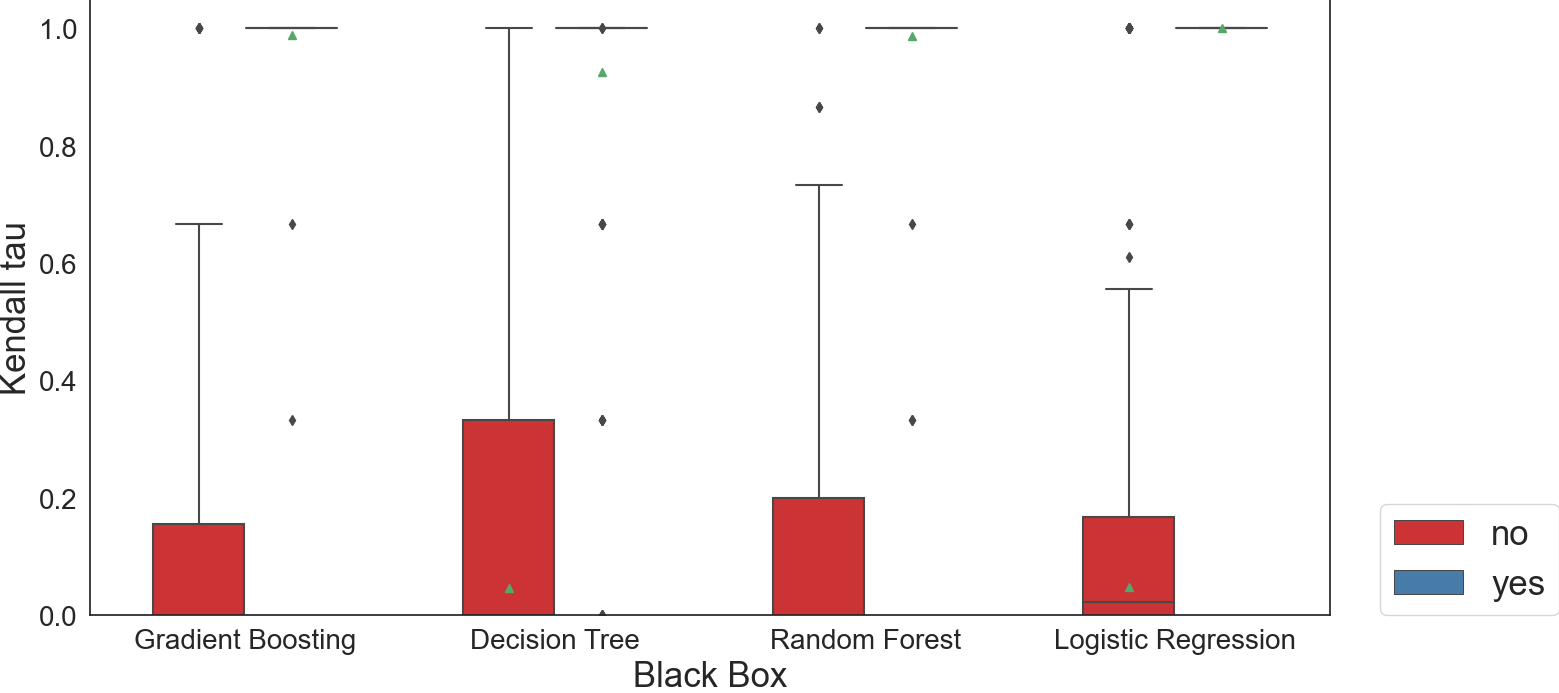}
    \caption{Average Kendall's rank correlation coefficient of the $\mathit{LS}_{\mathit{APE}}$'s explanations computed on 100 instances for 7 tabular datasets and 4 ``glass'' black-box models across the Oracle's outcomes.}
    \label{fig: kendall}
\end{figure}

\paragraph*{Tabular data.}
Our evaluation on tabular data focuses on datasets with a minimum of 8 features. We train classifiers on modified datasets where half of the features were set to 0 for all instances. Those features were randomly chosen. For example, in the context of the Blobs dataset with 12 features, our Decision Tree model considers only 6 of these features, as the others do not contribute to the classification. When evaluating a linear explanation, we consider it a satisfactory explanation if it employs the same features as the ``glass'' black box, and it is an unsatisfactory explanation if it relies mostly on features not used by our decision tree model. To perform this assessment, we use the upper half of the features according to the ranking provided by $\mathit{LS}_{\mathit{APE}}$, based on the absolute value of the attribution coefficient, from the linear surrogates.

Figure~\ref{fig: kendall} illustrates the Kendall rank correlation coefficient for various classifiers, including Gradient Boosting (GB), Decision Tree (DT), Random Forest (RF), and Logistic Regression (LR). For LR, we use the feature coefficients of the logistic function as the ground truth. Similarly, we extract the ground truth for DT by collecting the features encountered along the classification path of the instances. For the GB and RF classifiers, we construct feature rankings using the Gini importance score~\cite{gini} provided by scikit-learn.

Our observations indicate that when the APE Oracle indicates linear suitability, the rank correlation is consistently very close to 1. This suggests that $\mathit{LS}_{\mathit{APE}}$ accurately captures the actual feature importance ranking within the complex model. However, when the Oracle discourages linear explanations, $\mathit{LS}_{\mathit{APE}}$ faces challenges in identifying the features employed by the ``glass'' black-box classifier. These findings underscore the effectiveness of APE's linear suitability test as a reliable indicator of the expected quality of a linear surrogate, which translates into faithful explanations for black-box classifiers. While precision and Kendall's tau metrics yield similar results, we opt for Kendall's tau in Figure~\ref{fig: kendall} since it accounts for the ranking order of each feature.\\

\begin{table*}[t]
    \centering
    \footnotesize
    \addtolength{\leftskip} {-1cm}
    \addtolength{\rightskip}{-1cm}
    \begin{tabular}{lcccccccccccc}
    \toprule
    \multicolumn{12}{ c }{Is a Linear Explanation Suitable?}\\
    \midrule
    &    \multicolumn{3}{ c }{MLP} &  \multicolumn{3}{ c }{NB} &  \multicolumn{3}{ c }{RF} &   \multicolumn{3}{ c }{BERT} \\
    \cmidrule(r){2-4}   \cmidrule(r){5-7} \cmidrule(r){8-10}   \cmidrule(r){11-13}
     & Yes  & No & \textit{$Prop$}$_{no}$  & Yes  & No & \textit{$Prop$}$_{no}$ & Yes  & No & \textit{$Prop$}$_{no}$ & Yes  & No & \textit{$Prop$}$_{no}$ \\
    \midrule
    Ag News & \textbf{0.505} & 0.342 & \textcolor{orange}{0.63} & \textbf{0.3} & 0.286 & \textcolor{orange}{0.58} & 0.012 & \textbf{0.061} & \textcolor{orange}{0.65} & \textbf{0.563} & 0.206 & \textcolor{orange}{0.41} \\
    Fake & \textbf{0.398} & 0.361 & \textcolor{blue}{0.14} & \textbf{0.268} & 0.225 & \textcolor{blue}{0.11} & \textbf{0.125} & 0.123 & \textcolor{blue}{0.11} & \textbf{0.244} & 0.233 & \textcolor{blue}{0.08} \\
    Polarity & \textbf{0.323} & 0.283 & \textcolor{orange}{0.36} & \textbf{0.199} & 0.184 & \textcolor{blue}{0.27} & \textbf{0.076} & 0.071 & \textcolor{blue}{0.27} & \textbf{0.244} & 0.223 & \textcolor{orange}{0.41}\\
    \bottomrule
    \end{tabular}
    \caption[The average difference in the model prediction's probability between the original text and the same text without the words identified as important by $\mathit{LS}_{\mathit{APE}}$.]{
        The average difference in the model prediction's probability between the original text and the same text without the words identified as important by $\mathit{LS}_{\mathit{APE}}$. This score is computed based on 100 instances per black-box model and dataset, taking both Oracle outcomes. The columns labeled ``Yes'' and ``No'' represent the average change in the prediction probability of the classifier when the Oracle indicates the suitability or unsuitability of a linear explanation. The columns labeled $\textit{Prop}_{no}$ denote the ratio of cases where the Oracle does not predict linear suitability. The colors blue, orange, and red indicate  $\textit{Prop}_{no} \le 33\%$, $33\% > \textit{Prop}_{no} \ge 66\%$, and $\textit{Prop}_{no} \ge 66\%$ respectively.
    } 
    \label{tab: deletion_lse_unimodal_text}
    \end{table*}
\paragraph*{Textual data.}
We assess explanation fidelity by examining the average changes in the black box model's prediction when the words identified as important by the explanation are removed~\cite{insertion_deletion}. We report this mean probability difference in Table~\ref{tab: deletion_lse_unimodal_text}. We consider that an explanation is good if removing a word considered important from the original text decreases the classifier's confidence in the same class. Conversely, an explanation is bad if removing the top words does not decrease confidence or, worse, increases confidence in the classifier's prediction.

Our results consistently demonstrate that judiciously selecting when to employ a linear explanation ensures that the elements highlighted as important by the explanation genuinely influence the model's classification. Notably, for cases where the APE Oracle test indicates the suitability of a linear explanation (column 'Yes'), the average difference between the model's prediction before and after removing the most important words consistently increases. This implies that the words identified as important by the adapted linear explanations have a more significant impact on the model's prediction compared to the top words from non-adapted explanations. This observation holds true for most scenarios, except for the Random Forest classifier on the Ag News dataset. We note that the Ag News dataset exhibits a mean change exceeding 0.5 for both MLP and BERT models when linear explanations are suitable, which means that such explanations effectively impact the model's prediction. These findings underscore the benefit of studying the decision boundary before approximating it with a linear surrogate, as it serves as a reliable indicator of the potential fidelity of the computed explanations for black-box classifiers.

\begin{figure}[htbp]
    \centering
    \includegraphics[width=0.9\linewidth]{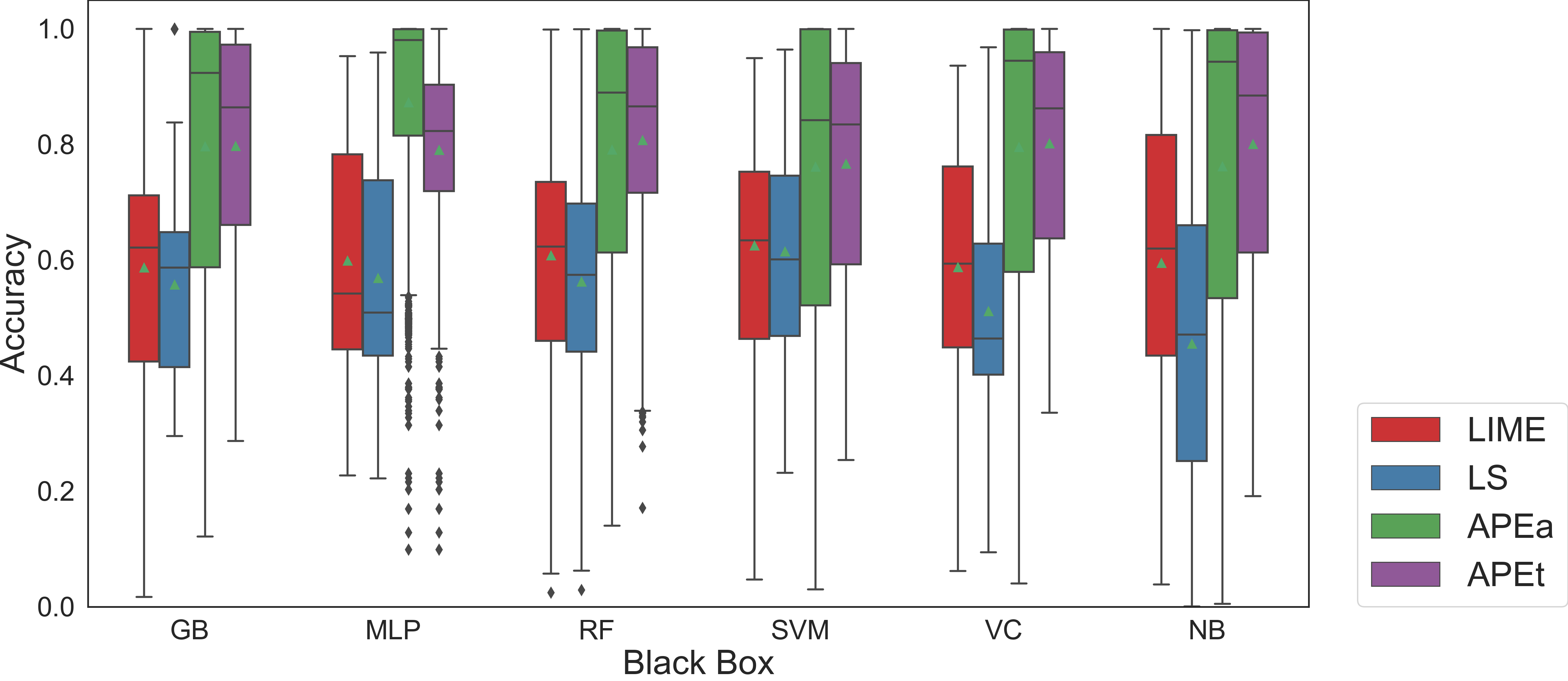}
    \caption{Average accuracy per black-box model on 100 instances of the experimental tabular datasets for $\textit{APE}_a$ and $\textit{APE}_t$.}
    \label{fig: accuracy_apea_apet}
\end{figure}
\subsection{Comparison with other Explanation Methods}
\label{subsec: fidelity_coverage_evaluation}
We compare the average accuracy of linear surrogates generated by LIME, LS, and the APE's variants $\textit{APE}_a$ and $\textit{APE}_t$ across 100 target instances. We exclude SHAP~\cite{SHAP} from this evaluation because it does not compute a linear approximation of the black box such as LIME and LS~\cite{explaining-lime}. Even though its variant Kernel SHAP uses linear regression, this method approximates the Shapley values. APE's variants produce either an anchor or a shallow decision tree when the APE Oracle does not predict linear suitability, otherwise, they both invoke $\mathit{LS}_{\mathit{APE}}$.\\

\paragraph*{Tabular data.}
The results are depicted in Figure~\ref{fig: accuracy_apea_apet}. For LS, we exclude the datasets with categorical attributes since they are not supported by this method. The findings suggest that regardless of the rule-based surrogate, APE achieves the highest accuracy. However, we observe that the performance of its two variants depends on the specific characteristics of the black-box model. On average, $\textit{APE}_a$ offers higher adherence but also exhibits greater variability. All in all, this evaluation demonstrates that the judicious selection between linear and rule-based explanations on a per-instance basis brings an average adherence gain of 0.21 points when compared to always choosing LIME or LS.\\

\paragraph*{Textual data.}
\begin{figure}[htbp]
    \centering
    \includegraphics[width=0.9\linewidth]{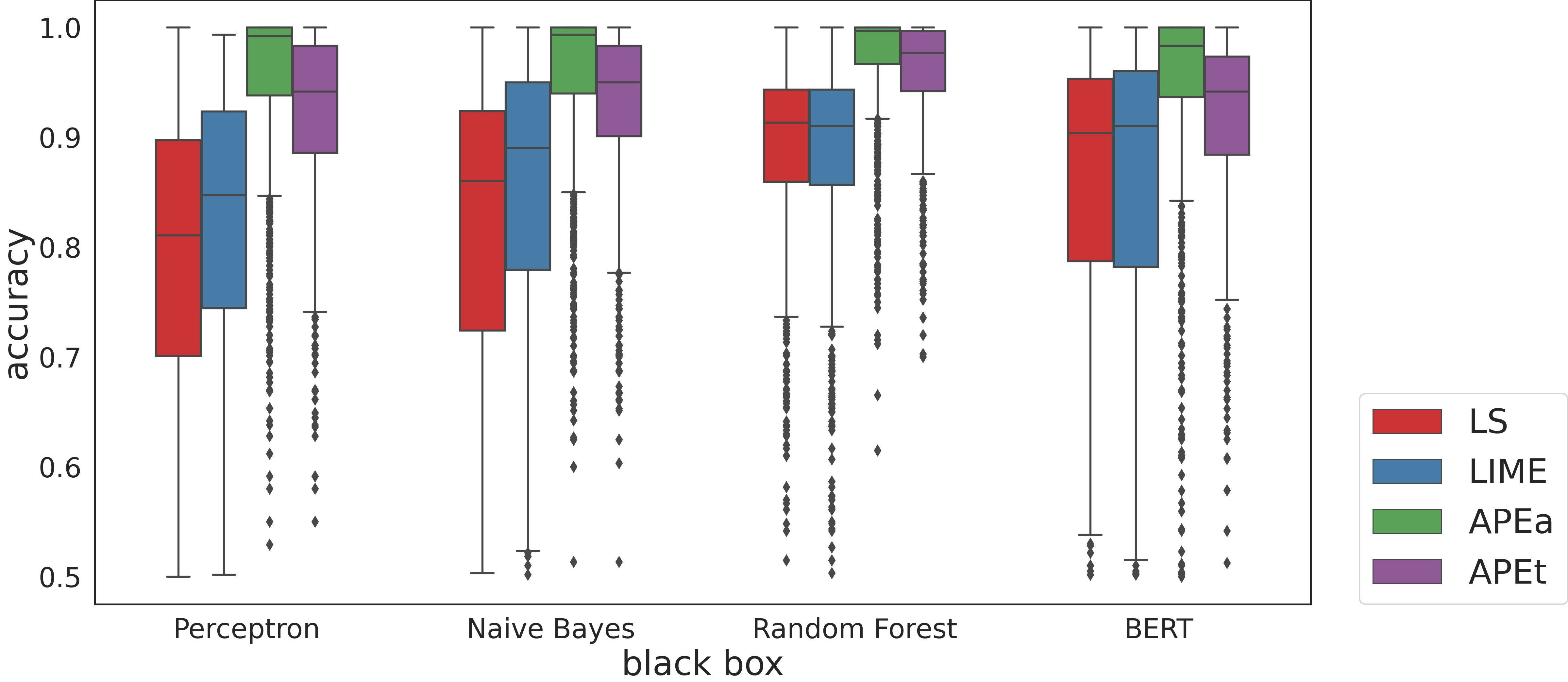}
    \caption{Average accuracy achieved by $\textit{APE}_a$ and $\textit{APE}_t$ on each black-box model based on 100 instances drawn from the textual datasets.}
    \label{fig: accuracy_apea_apet_text}
\end{figure}
Figure~\ref{fig: accuracy_apea_apet_text} displays the average accuracy of LIME, LS, $\textit{APE}_a$, and $\textit{APE}_t$. Notably, both methods employing the APE framework consistently outperform LIME and LS across all complex models. Additionally, we observe that the variance of the techniques restricted to linear surrogates is higher than those of $\textit{APE}_a$ and $\textit{APE}_t$.\\

\paragraph*{General Insights.} These findings provide valuable insights, suggesting that when APE chooses to report a linear explanation, the decision frontier is indeed linearly separable. This is supported by the superior performance of both $\textit{APE}_a$ and $\textit{APE}_t$ compared to using linear surrogates alone. This observation holds true even for black box models with a relatively high proportion of linearly suitable frontiers, e.g., SVM and tabular RF (see Table~\ref{tab: score_lse_unimodal}).

While these results may suggest that APE's performance primarily relies on Anchors or decision trees and that we should consistently pick a rule-based surrogate, there are nuances to consider. We provide, in Tables~\ref{tab: score_lse_unimodal} and~\ref{tab: score_lse_unimodal_text}, the proportion of times APE selects a rule-based explanation over a linear surrogate for tabular and textual data, respectively. We remind the reader that APE Oracle tests the suitability of a linear explanation. Therefore, it cannot predict the performance of an anchor or a rule-based explanation. Thus, in Figure~\ref{fig: accuracy_apea_apet} and~\ref{fig: accuracy_apea_apet_text}, we compare the accuracy of both $\textit{APE}_t$ and $\textit{APE}_a$, our two methods that differ only in the choice of the rule-based surrogate model. We observe that $\textit{APE}_a$ with Anchors enhances the fidelity of the surrogate while using a shallow decision tree reduces standard deviation error. Therefore, the advantages of one rule-based method over another warrant further investigation.

\begin{figure*}[!ht]
    \begin{subfigure}{.47\linewidth}
        \centering
        \advance\leftskip-3cm
        \includegraphics[width=1.1\linewidth]{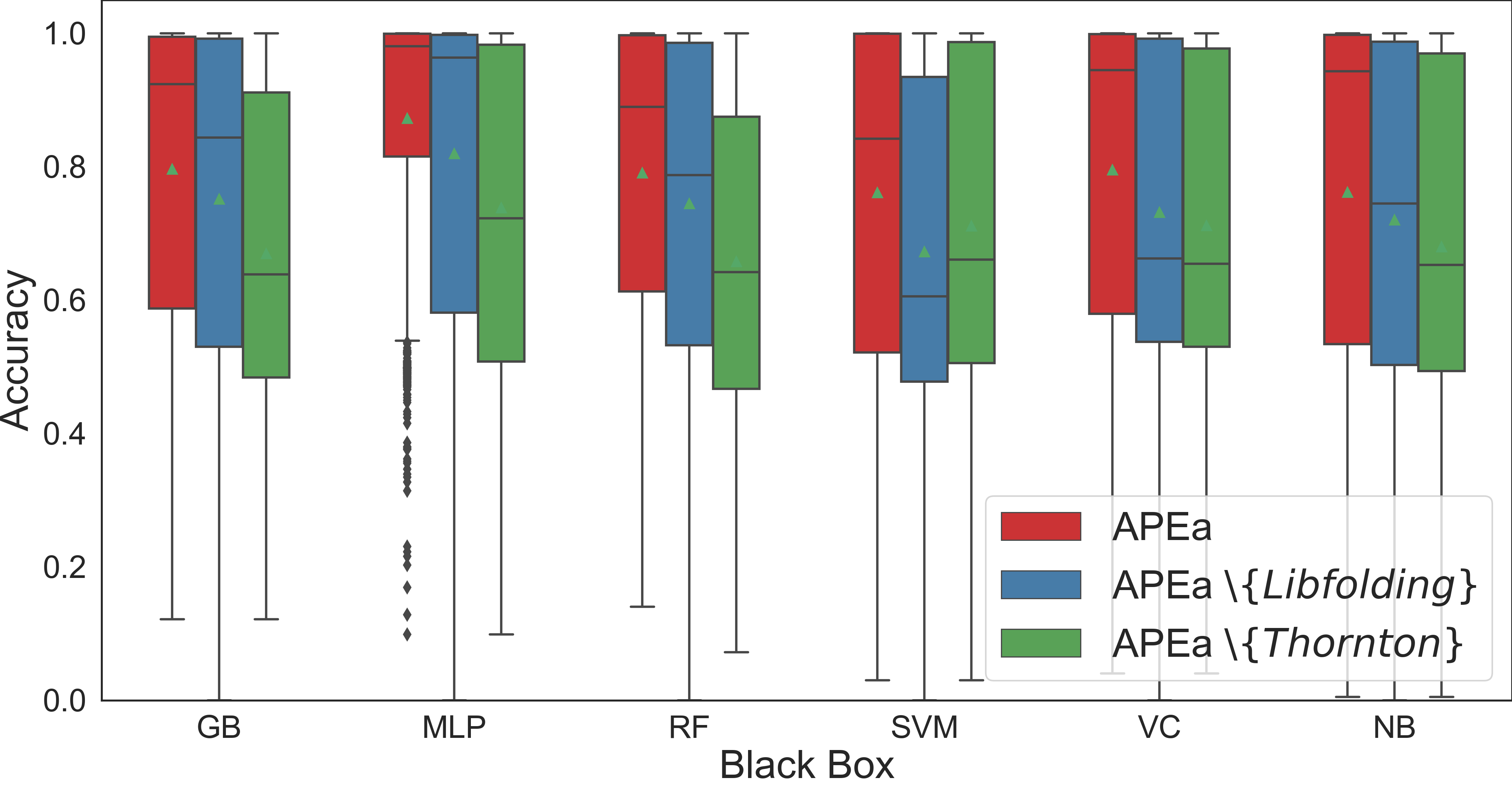}
        \caption{$\textit{APE}_a$}
        \label{fig: ablation_study_apea}
    \end{subfigure}
    \centering
    \begin{subfigure}{.47\linewidth}
        \centering
        \advance\rightskip-3cm
        \includegraphics[width=1.1\linewidth]{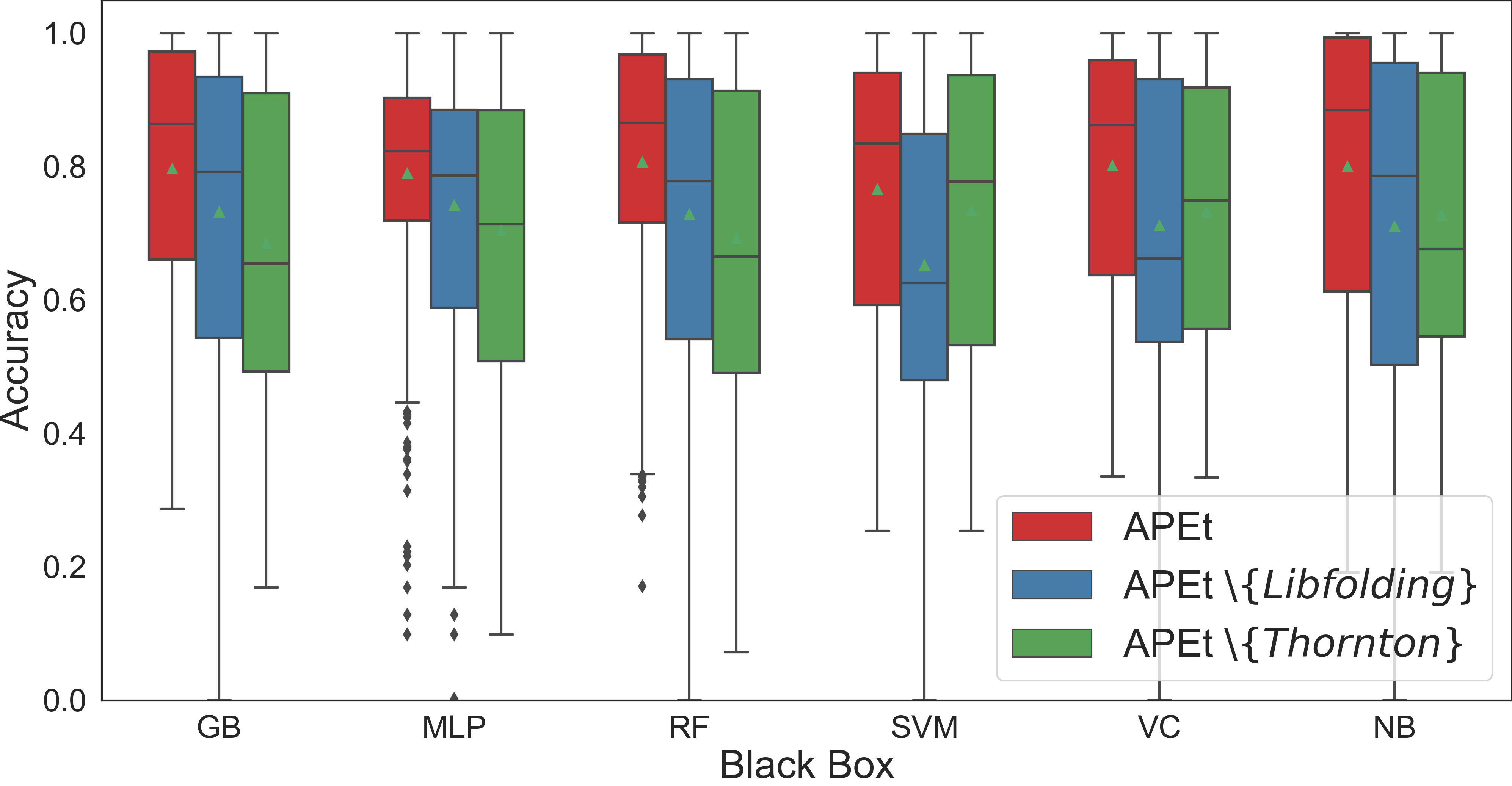}
        \caption{$\textit{APE}_t$}
        \label{fig: ablation_study_apet}
    \end{subfigure}
    \caption{Average accuracy per black box computed on 100 instances of the tabular datasets for $\textit{APE}_a$ in (a) and for $\textit{APE}_t$ in (b) when we remove the Libfolding unimodality and linear separability tests from the APE Oracle.}
    \label{fig: ablation_study}
\end{figure*}
\subsection{Ablation Study}
\label{sec: ablation_study}
We now carry out an ablation study to assess the individual contributions of the components of the APE Oracle, in particular the unimodality and separability tests. This study is conducted through the adherence metric.\\

\paragraph*{Tabular data.}
We report the average accuracy of $\textit{APE}_a$ and $\textit{APE}_t$ in Figures~\ref{fig: ablation_study_apea} and~\ref{fig: ablation_study_apet}, compared to the same variant of APE that excludes either the Libfolding unimodality test ($\textit{APE} \; {\backslash\{ \textit{Libfolding}\}}$) or Thornton's separability test ($\textit{APE} \; {\backslash\{ \textit{Thornton}\}}$). The results show that the unimodality and separability tests complement each other, and solely assessing linear separability around the decision boundary, as suggested by the accuracy of $\textit{APE} \; {\backslash\{ \textit{Libfolding}\}}$, is insufficient to predict linear suitability.\\

\begin{figure*}[!ht]
    \begin{subfigure}{.47\textwidth}
        \centering
        \advance\leftskip-3cm
        \includegraphics[width=1.1\linewidth]{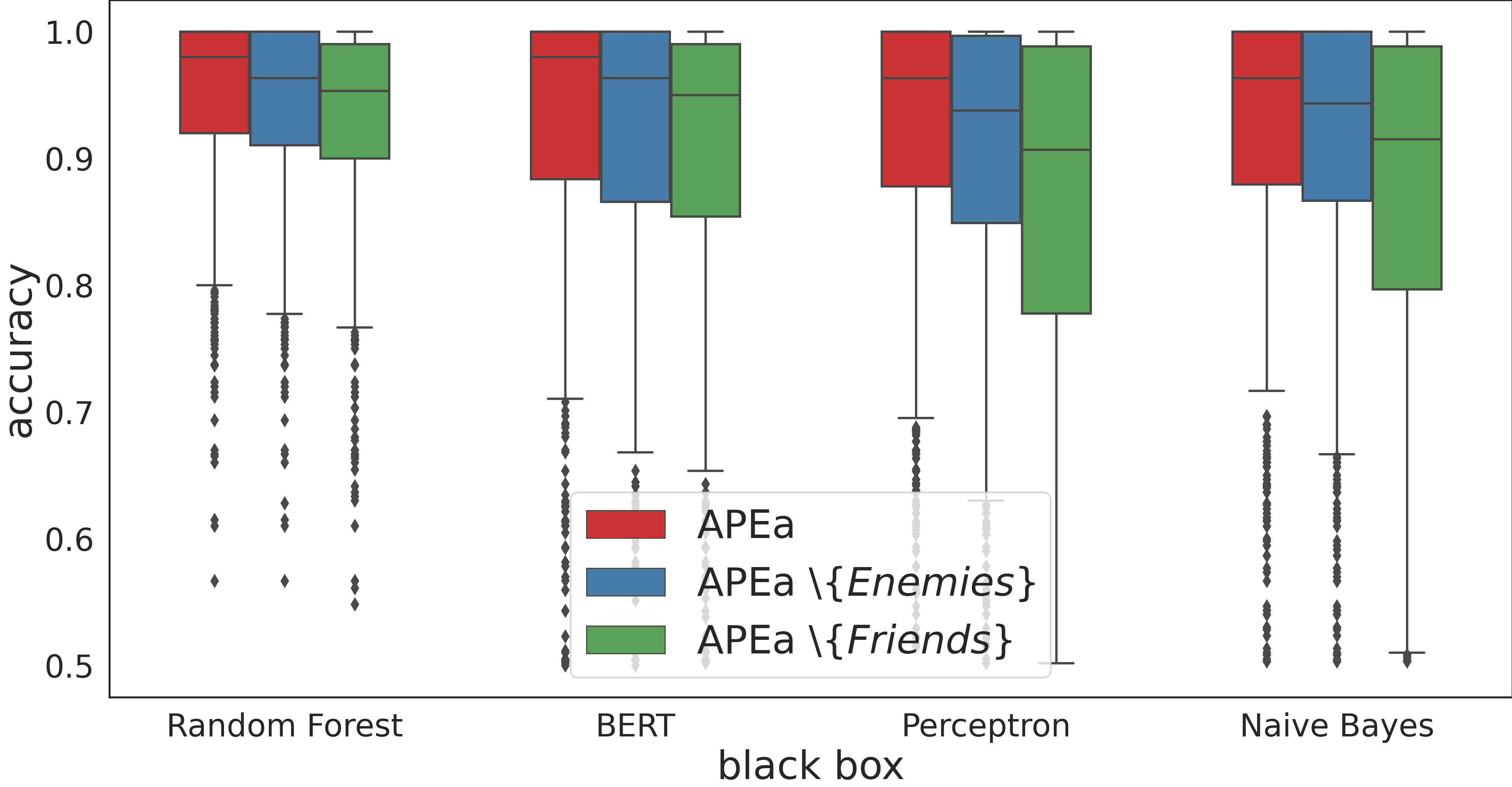}
        \caption{$\textit{APE}_a$}
        \label{fig: ablation_study_apea_text}
    \end{subfigure}
    \centering
    \begin{subfigure}{.47\textwidth}
        \centering
        \advance\rightskip-3cm
        \includegraphics[width=1.1\linewidth]{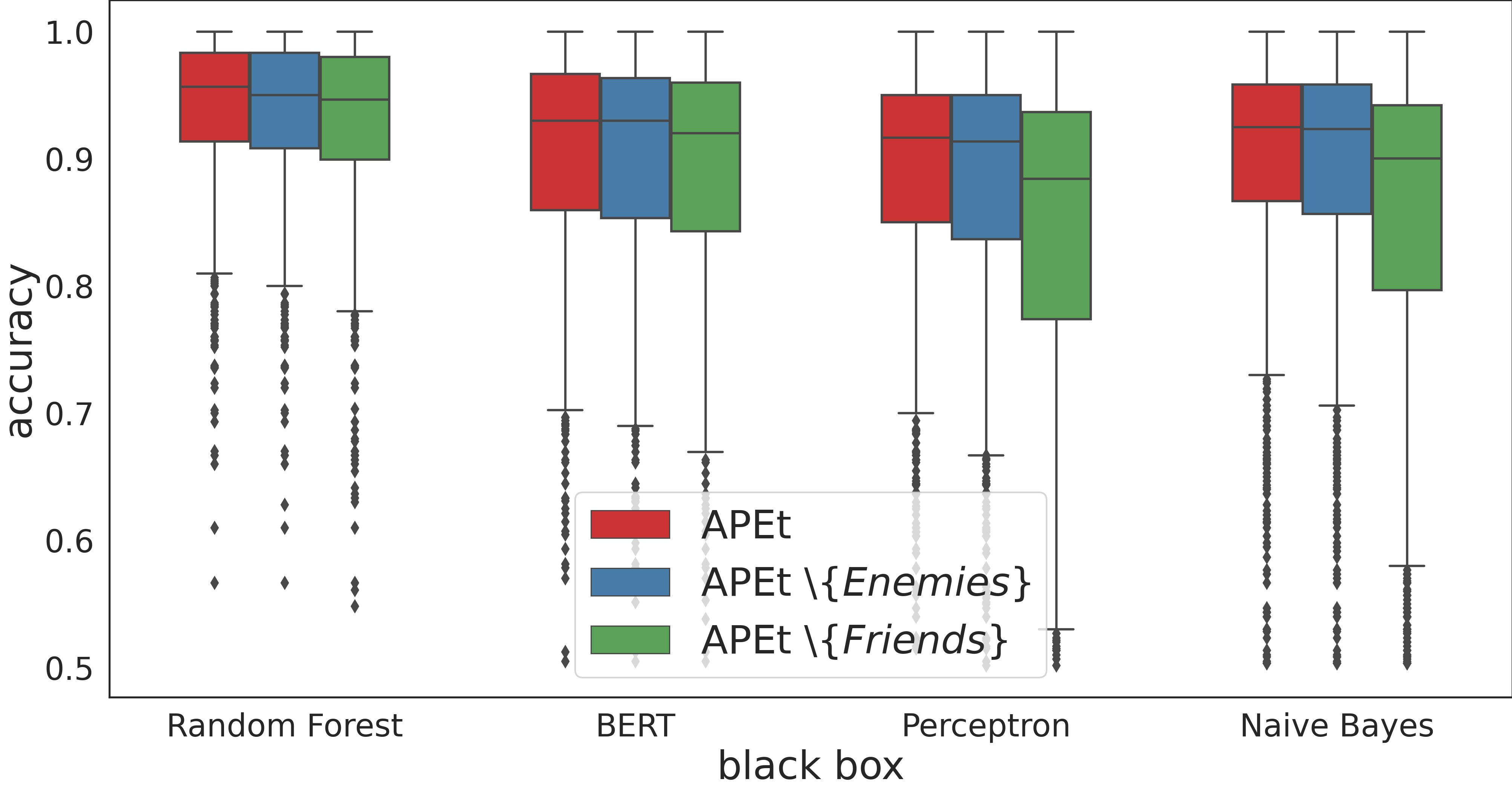}
        \caption{$\textit{APE}_t$}
        \label{fig: ablation_study_apet_text}
    \end{subfigure}
    \caption[Average accuracy attained by $\textit{APE}_a$ in (a) and for $\textit{APE}_t$ in (b) across different black-box models and textual datasets.]{Average accuracy attained by $\textit{APE}_a$ in (a) and for $\textit{APE}_t$ in (b) across different black-box models and textual datasets. The computation is based on 100 instances sampled from the experimental datasets. The results are displayed for both methods using the Libfolding unimodality test from the APE Oracle, on the friends or enemies instances.}
    \label{fig: ablation_study_text}
\end{figure*}

\paragraph*{Textual data.} In the context of textual data, we conduct two separate ablation studies. The first study is designed to examine the benefits resulting from the implementation of the unimodality test on both the friends and enemies of the target instance. The second study compared how the choice of the data space on which the unimodality test is applied,(either bag of words or latent space) influences the test's effectiveness.

Therefore, we report in Figure~\ref{fig: ablation_study_apea_text}, the average accuracy of $\textit{APE}_a$, compared to its variants that apply the Libfolding unimodality test to specific subsets of instances: either the friends (APEa$\backslash\{Enemies\}$) or the enemies (APEa$\backslash\{Friends\}$). Similarly, we examine the results for $\textit{APE}_t$ and its respective variants in Figure~\ref{fig: ablation_study_apet_text}. These findings highlight a critical insight: characterizing the distribution of either the friends or the enemies alone is not enough to detect multimodality. The limitation arises from the fact that analyzing the distribution of one group (friends or enemies) alone does not provide a comprehensive understanding of the decision boundary. As a result, it is crucial to examine the distribution of both friends and enemies before selecting an appropriate explanation technique.

\begin{figure}[htbp]
    \centering
    \includegraphics[width=0.9\linewidth]{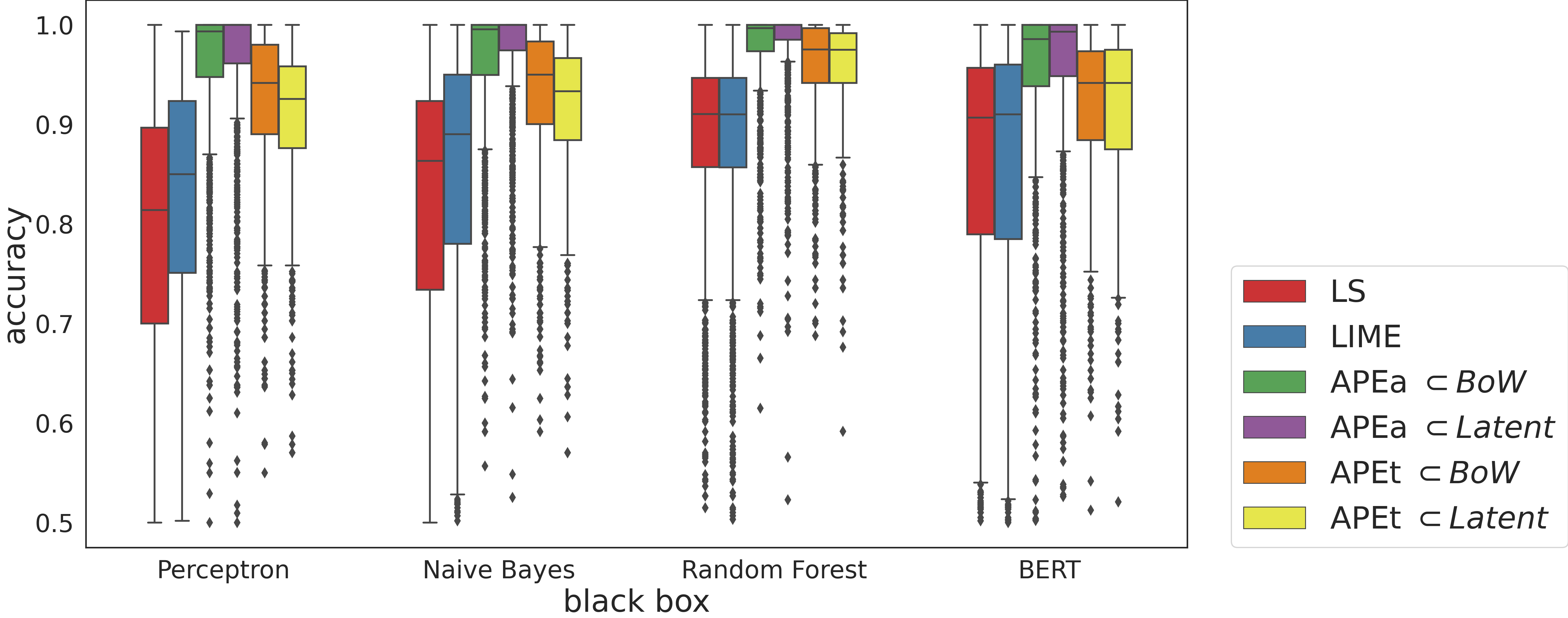}
    \caption[Average accuracy comparison between the outcomes of $\textit{APE}_a$ and $\textit{APE}_t$ based on the space used for the APE Oracle's test.]{Average accuracy calculated for 100 instances per textual dataset and complex model. We present a comparison between the outcomes of $\textit{APE}_a$ and $\textit{APE}_t$ based on the space used for the APE Oracle's test (`BoW' or `Latent').}
    \label{fig: accuracy_ape_latent}
\end{figure}
In Figure~\ref{fig: accuracy_ape_latent}, we examine the impact of employing different data spaces for the Libfolding test, specifically the bag of words representation (BoW) and the latent space (Latent). Our focus is to understand how this choice influences the performance of two APE variants, $\textit{APE}_a$ and $\textit{APE}_t$, in comparison to traditional linear explanation methods such as LIME and LS.
The findings from this analysis reveal that regardless of the space chosen for testing the instance distribution, both $\textit{APE}_a$ and $\textit{APE}_t$ consistently outperform linear explanation methods. 
Nevertheless, the choice of the data space does not have the same effect on $\textit{APE}_a$ and $\textit{APE}_t$, and this effect also depends on the target black box model. 
To illustrate this point, consider MLP and NB, where the bag of words representation (BoW) enhances the average accuracy of $\textit{APE}_t$, while the latent representation increases the average accuracy and reduces the variance of $\textit{APE}_a$. With these results, we gain valuable insights into the role of space selection in determining the effectiveness of the APE framework. They underscore the importance of employing adaptable strategies when interpreting textual data.

\subsection{Summary and Discussion of Linear Suitability Results}
The extensive evaluation presented in this section focuses on the concept of ``linear suitability'' within the APE framework. These evaluations aim to determine whether the APE framework accurately identifies scenarios where linear explanations are most appropriate. Several key findings emerge from the results.

Firstly, the APE Oracle's ability to assess linear suitability significantly impacts the quality of explanations. When the APE Oracle predicts that a linear explanation is suitable for a particular instance, the resulting explanation exhibits higher adherence and fidelity. In essence, these explanations not only have a good adherence but also accurately represent the inner workings of the complex black-box models, leading to more accurate and interpretable results.

Secondly, the evaluation underscores the nuanced nature of linear suitability. It reveals that the APE Oracle's test for linear separability in the decision boundary is a strong indicator of the potential success of linear explanations. In instances where the decision boundary is indeed linearly separable, APE's linear explanations perform exceptionally well, offering accurate and insightful results. These findings hold true even for complex black-box models, such as Support Vector Machines and Random Forests, which can often produce linearly suitable decision boundaries.

Furthermore, the results suggest that the proportion of linearly separable cases can be influenced by dataset characteristics. For instance, datasets like Circles, Fake, and Titanic exhibit a higher prevalence of linearly suitable boundaries, likely due to the nature of the data distribution. However, the choice of the black-box classifier architecture also plays a role, with different classifiers yielding varying proportions of linearly suitable cases.

In conclusion, the APE framework's linear suitability assessment offers a practical and effective method for distinguishing when linear explanations should be applied. This two-step process, involving unimodality and separability tests, allows APE to adapt to a wide range of scenarios, providing a flexible approach to generating interpretable explanations for complex black-box models. By leveraging this understanding of linear suitability, APE demonstrates its versatility and robustness in providing explanations that are both faithful to the model's behavior and accurate in representing the decision boundaries of complex classifiers.

\subsection{Counterfactuals Evaluation}
\label{subsec: counterfactuals_eval}
We now evaluate the quality of the counterfactuals generated by Growing Spheres and Growing Fields for our 100 test instances. Our evaluation considers two key aspects: (a) the resemblance of these counterfactual instances to real-world examples, and (b) the runtime performance of each method. Note that we detail the experimental results for Growing Language and Growing Net in Section~\ref{sec: resultscounterfactual}.

\subsubsection{Quality of the Counterfactuals}
In line with the literature in counterfactual explanations~\cite{cf_review}, we assess the quality of counterfactual instances generated by Growing Fields and Growing Spheres by measuring their resemblance to actual instances. We resort to the Mahalanobis and Euclidean distances, computed between the generated counterfactuals and the entire set of enemies in the test instances. The Mahalanobis distance quantifies to which extent our counterfactual explanations are outliers w.r.t. the distribution of non-synthetic enemies. For the sake of brevity, we report results based on the Mahalanobis distance since the results for Euclidean and Manhattan distances show the same behavior when comparing Growing Fields and Growing Spheres~\cite{gs}. 

Figure~\ref{fig: distance} shows the average distance between the counterfactual generated and the original instance. It is important to note that these distances are normalized with respect to the farthest instance from the input dataset $T$. As a result, we can consider a counterfactual to be realistic if the distance to the target instance is lower than one, which means it is lower than the farthest observed distance. However, the results show that this is not always the case. Specifically, our findings reveal that, on average, APE (Growing Fields) finds more realistic counterfactual instances than Growing Spheres. These observations hold especially true when examining the multi-layer perceptron classifier (MLP). On this classifier, Growing Spheres finds counterfactual instances that are very unrealistic, as the normalized distance to their closest real counterfactual is far higher than one. The counterfactuals generated by APE tend to be, on average, 0.356 points closer to actual instances. This improved performance of Growing Fields can be attributed to its ability to account for the variance and amplitude of the attributes when generating synthetic instances. Although Growing Spheres occasionally generates counterfactuals that are as similar to the target as Growing Fields, there are cases for which the counterfactuals produced by Growing Spheres are significantly distant from real-world examples, a shortcoming not observed in the case of Growing Fields. 

\begin{figure}[htbp]
    \centering
    \includegraphics[width=0.9\linewidth]{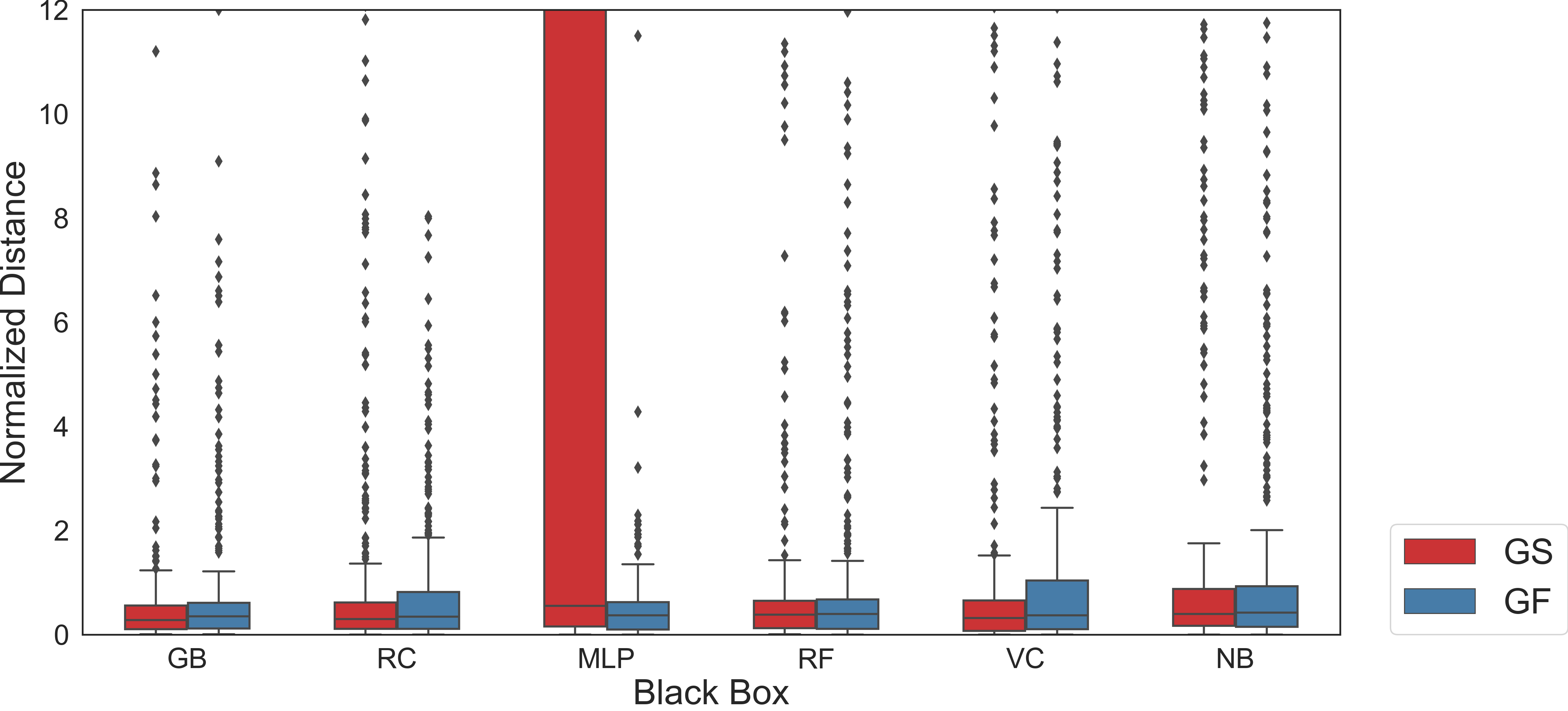}
    \caption{Average Mahalanobis distance between the Growing Spheres (GS) and Growing Fields (GF) counterfactual instances and their closest real enemy.}
    \label{fig: distance}
\end{figure}

\subsubsection{Runtime Performance}
We present in Table~\ref{tab: time} the runtime performances of both methods, Growing Spheres and Growing Fields. By leveraging the dataset distribution, Growing Fields efficiently samples artificial instances and generates the closest artificial counterfactual, outperforming Growing Spheres in terms of speed and realism. This significant improvement arises from Growing Fields' capacity to adjust the expansion speed of the field radius based on data distribution. Specifically, it ensures that features experience an expansion speed adapted to the magnitude of their values. We notice that on average over 100 instances, seven datasets and five black box models, Growing Fields achieves an average speed-up of two orders of magnitude compared to Growing Spheres to discover the closest counterfactual. 

Furthermore, the runtime of generating a counterfactual with Growing Fields ranges from 0.08 to 3.03 seconds. In contrast, Growing Sphere exhibits considerable runtime varying between 1.53 and 725 seconds. Consequently, Growing Fields offers a superior balance of speed and reliability for generating counterfactual explanations compared to Growing Spheres.

\begin{table*}[ht]
    \centering
    \footnotesize
    \begin{tabular}{p{0.1\textwidth}p{0.04\textwidth}p{0.04\textwidth}p{0.04\textwidth}p{0.04\textwidth}p{0.04\textwidth}p{0.04\textwidth}p{0.04\textwidth}p{0.04\textwidth}p{0.04\textwidth}p{0.04\textwidth}}
    \toprule
    &\multicolumn{2}{ c }{GB} &  \multicolumn{2}{ c }{MLP} &  \multicolumn{2}{ c }{RF} &  \multicolumn{2}{ c }{RC}  &  \multicolumn{2}{ c }{VC}\\
    \cmidrule(r){2-3}   \cmidrule(r){4-5} \cmidrule(rl){6-7}   \cmidrule(r){8-9} \cmidrule(r){10-11}
    & GS & GF & GS & GF & GS & GF & GS & GF & GS & GF \\
    \midrule
    Blood &  31.8 & \textbf{0.14} & 212 &  \textbf{0.26} &  96.4 & \textbf{0.70} & 371 & \textbf{0.11} & 725 &  \textbf{2.09} \\
    Blob &   6.71 & \textbf{0.12} &  15.6 &  \textbf{0.18} &  12.7 & \textbf{0.47} &  14.6 & \textbf{0.08} &  38.2 &  \textbf{1.50} \\
    Blobs &  82.9 & \textbf{0.42} & 158 &  \textbf{0.53} & 153 & \textbf{0.86} & 162 & \textbf{0.35} & 408 &  \textbf{2.47} \\
    Circles &   1.53 & \textbf{0.19} &   2.05 &  \textbf{0.30} &   2.21 & \textbf{0.65} &  10.2 & \textbf{0.12} &  26.8 &  \textbf{3.03} \\
    Diabetes &   4.36 & \textbf{0.21} & 101 &  \textbf{0.28} &  45.7 & \textbf{0.58} &  23.4 & \textbf{0.14} &  42.8 &  \textbf{2.14} \\
    M Blobs & 237 & \textbf{1.00} & 260 &  \textbf{0.84} & 315 & \textbf{1.69} & 241 & \textbf{0.62} & 573 &  \textbf{2.82} \\
    Moons &   1.69 & \textbf{0.15} &   4.24 &  \textbf{0.20} &   3.41 & \textbf{0.47} &   5.24 & \textbf{0.09} &  14.0 &  \textbf{1.56} \\
    \bottomrule
    \end{tabular}
    \caption{Average runtime (in seconds) of Growing Spheres (GS) and Growing Fields (GF) over 100 instances per black box and dataset.}
    \label{tab: time}
    \end{table*}


\section{Discussion and Conclusion}
\label{sec: ape_conclusion}
The fundamental question of what makes an explanation suitable for a particular use case lies at the junction of XAI and cognitive sciences. For this reason, this research question has not been addressed from a holistic perspective but rather from different, still complementary, angles.

On the one hand, the XAI community has put emphasis on the development of post-hoc explanation paradigms that identify the features that play a role in the predictions of an AI model. Among those, feature attribution rankings based on linear surrogates such as LIME~\cite{LIME} or LS~\cite{Laugel_Defining_locality}, enjoy notable popularity, because they can provide accurate per-instance explanations~\cite{trends_in_xai}. Besides, practitioners from most disciplines are familiar with linear models. These surrogate models are learned so that they optimize for user-agnostic criteria such as the \emph{fidelity}, i.e., the degree to which the surrogate mimics the black-box model it aims to uncover. Most of the literature in classical XAI has pushed the state of the art towards novel approaches -- or improvements of existing ones. As a result, none of these works tackles the question of when a linear surrogate is objectively a reliable explanation.

On the other side of the spectrum, cognitive and social sciences study the subjective and human aspects of explaining AI models. In that spirit, the suitability of an explanation is characterized by its comprehensibility and plausibility~\cite{Fuernkranz2020}. Comprehensibility captures the extent to which a user grasps an explanation and can use it to accomplish well-defined tasks~\cite{user_based_guidelines}, e.g., determine the features used by the black-box system, predict the black box's answer, etc. On the other hand, the plausibility dimension models the cognitive preferences and background of the users. As pointed out by several studies~\cite{Fuernkranz2020, too_much_too_little}, users can reject an explanation if it contradicts common sense, for instance, if the explanation is too simplistic given that the underlying problem is deemed complex. The consensus seems to indicate that showing plausible and sound explanations increases trust in AI systems~\cite{too_much_too_little,van_der_waa}, whereas the effects on comprehensibility and task efficiency are mixed.

While the XAI and cognitive science communities may appear somehow unreconciled, the relevance of the quality dimensions targeted by classical XAI methods has been justified by user studies. It has been suggested~\cite{too_much_too_little} that in the context of recommender systems, low fidelity harms trust in explanations. In this line of thought, we introduced in this chapter, a novel approach to determine \emph{a priori} the pertinence of local linear explanations for a given use case. Our decision is driven by standard user-agnostic desideratum, namely the fidelity of explanation. The results of our experiments provide evidence that characterizing the decision boundary of a black-box classifier around a target instance and making informed choices between linear and rule-based explanations are indeed feasible. In that spirit, the answers generated by APE, are promising for users of AI systems and linear surrogate explanations. When APE favors a rule-based explanation, it effectively communicates that the classification boundary is likely complex, and relying on a unique linear attribution explanation would result in incompleteness or inaccuracy. 

Furthermore, evidence also suggests that multi-paradigm explanations can have a positive impact on comprehensibility~\cite{miller, wachter}. In particular, counterfactual explanations can be a complement to attribution-based or rule-based explanations and enrich the user's experience. Indeed, when the classifier's decision boundary has a multimodal distribution, APE presents a diverse and representative set of scenarios. These scenarios effectively illustrate how changes in the input features impact the classifier's output. This perspective extends to the combination of linear attribution and rule-based explanations as well. Thus, our work does not discourage the exploration of alternative combinations of explanation paradigms; rather, it provides valuable insights into the nature of the classifier's decision boundary. Such insights can be particularly useful in scenarios where the objective is to replace the black-box model, such as reverse engineering, or when a single, comprehensive, and unambiguous explanation is required.

Looking ahead, we envision to explore several avenues for future works. We have shown that APE is adapted to diverse data types such as textual and tabular data, thus, one possible avenue may be to extend APE to accommodate various distance metrics, and a wider spectrum of machine learning tasks, such as regression. Furthermore, we envisage enhancing our framework to support other explanation paradigms. For instance, future versions of APE may determine when a rule-based explanation is suitable for a specific instance and which rule-based model should be employed, such as Anchors or a decision tree. Additionally, we see an opportunity for research that integrates concepts of coverage, complexity, and plausibility when deciding on the most suitable explanations for a given use case. Indeed, methods that produce global explanations by combining local explanations may prefer simple explanations of high coverage over faithful and very specific explanations.

In the present chapter, we explored the importance of selecting an appropriate explanation paradigm to enhance the quality of generated explanations. However, as we adapted the unimodality test to textual data, a new challenge emerged: select the correct conversion space used by the surrogate. When working with textual data, there are two primary approaches to convert text into numerical representations suitable for ML models. The first approach involves using an interpretable space that converts words into vectorized representations, such as a bag of words. The second approach employs more complex methods to embed text into a latent space. In the next chapter, our objective is to assess how the choice between these two conversion methods impacts the quality of the generated explanations. Specifically, we will focus on counterfactual explanation techniques and compare a handful of state-of-the-art counterfactual techniques, including Growing Language and Growing Net. The goal is to provide a detailed analysis of the spectrum of existing counterfactual methods. This analysis focuses on the techniques employed to perturb input text and offers valuable insights into the question of employing a black box (in the form of a latent space) to explain another black box.


\clearemptydoublepage

\chapter{Explaining a Black Box Without Another Black Box}
\label{chap: emnlp}
\minitoc

\section{Context}
\label{sec: introduction}
The latest advances in Artificial Intelligence, and more particularly in machine learning, have significantly transformed various natural language processing (NLP) tasks~\cite{bert, roberta, DistilBERT}, including text generation, fake news detection, sentiment analysis, and spam detection. These notable improvements can be attributed, in part, to the adoption of methods that encode and manipulate text data within latent representations. In this context, latent representations refer to abstract, high-dimensional vector spaces where text is transformed into numerical form, thus capturing the underlying semantics, structure, and patterns of language. These latent representations offer a bridge between raw text data and machine learning models, allowing algorithms to operate on a more structured and semantically rich representation of language. 

However, the impressive gains in accuracy achieved by modern algorithms, such as Transformers models~\cite{bert}, can be diminished by the lack of interpretability of those algorithms~\cite{interpret_gan}. Indeed, a model could make correct predictions for the wrong reasons~\cite{bias_nlp, bias_nlp2}. Unless the machine learning model is a white box, explaining the results of such an agent requires an explanation layer that interprets the internal workings of the black box in a post-hoc manner.

\begin{figure}[!h]
    \centering
    \includegraphics[width=0.85\textwidth]{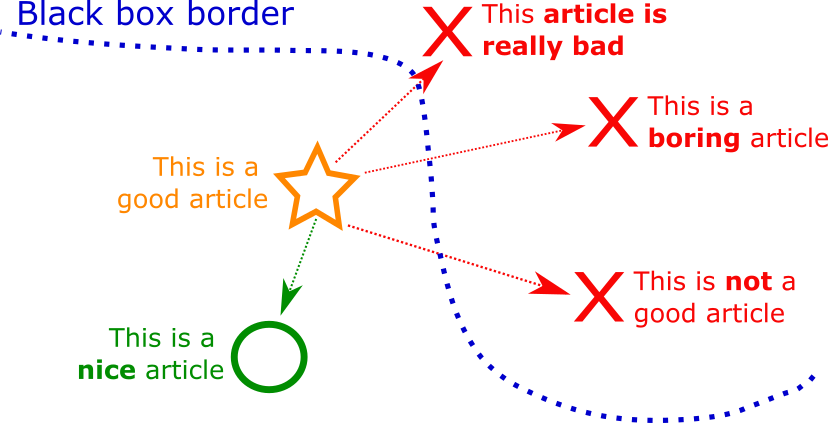}
    \caption[Illustration of an artificial decision boundary around a target text and its counterfactuals.]{Target instance is represented by the sentence ``This is a good article'' while other texts are artificial textual documents. The blue dashed line represents the classifier's decision boundary for predicting review polarity. Texts shown in red are classified as negative, and those in green are classified as positive.}
    \label{fig:cf_representation}
\end{figure}

While there are several ways to explain the outcomes of an ML model a posteriori, our previous chapter concentrated on feature attribution techniques. We now shift our attention to counterfactual explanations, a domain that experienced notable popularity over the last five years~\cite{survey_cf_Guidotti, miller}. A counterfactual explanation is a counter-example that is similar to the original text, but that elicits a different outcome in the black box~\cite{wachter}. Consider the classifier depicted in Figure~\ref{fig:cf_representation}, for sentiment analysis applied to the review ``This is a good article'' -- classified as positive. In this toy example, a counterfactual could be the phrase ``This is a boring article''. Through this explanation, the counterfactual technique conveys that the adjective ``good'' was a possible reason for this sentence to be classified as positive, and changing the polarity of that adjective may change the classifier's response. 

In the literature, counterfactual explanation methods operate by increasingly perturbing the target text until the classifier's answer changes. These methods lie in a spectrum spanning from fully transparent to fully opaque techniques. On one side of the spectrum, \emph{transparent} methods perturb the target text by adding, removing, or changing words and syntactic groups~\cite{martens, mice, plausible_counterfactual} in the original target text. To illustrate this concept, refer to Figure~\ref{fig: transparent_perturbation}, which provides a visual representation. Here, the yellow star represents the original target text, which is transformed into a binary vector. In this vector, a `1' indicates the preservation of the original word, while a `0' means the word will be replaced by another word. On the opposite side, more recent methods~\cite{decision_boundary, xspells, cfgan} embed the target text in a latent space on which perturbations are carried out subsequently. The embeddings are a compressed representation of the classifier's training data, which filters noise and retains the essential information for classification. An example of a latent perturbation is presented in Figure\ref{fig: opaque_perturbation}, where the input text is encoded into a high-dimensional numerical vector. Small perturbations are introduced to these vectors within the latent space before bringing them back to the original data space. These methods are classified as \emph{opaque} due to the inherent lack of interpretability of the latent space~\cite{gan_interpret_latent}. 

\begin{figure*}[!ht]
    \begin{subfigure}{.47\linewidth}
        \centering
        \includegraphics[width=\linewidth]{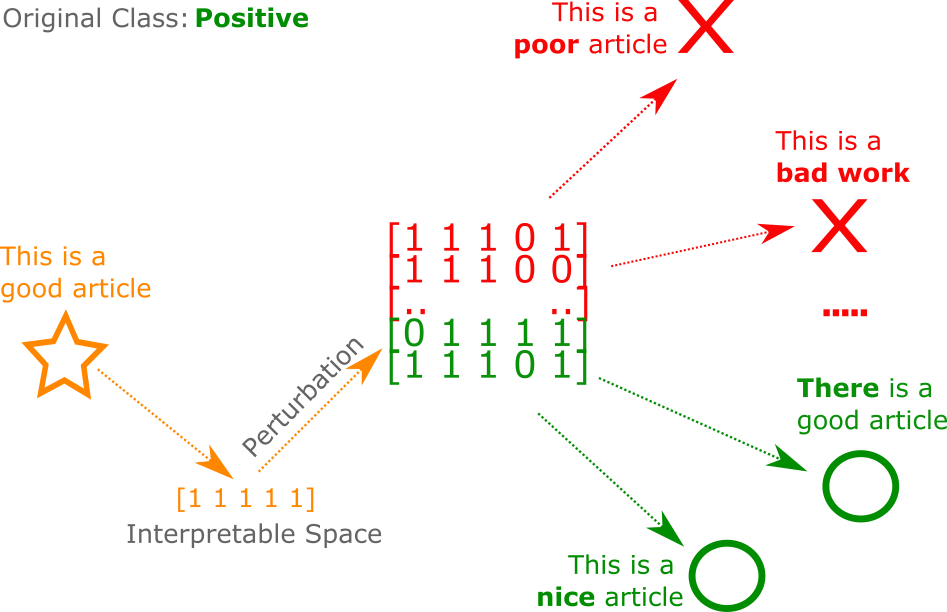}
        \caption{Transparent perturbation}
        \label{fig: transparent_perturbation}
    \end{subfigure}
    \centering
    \begin{subfigure}{.47\linewidth}
        \centering
        \includegraphics[width=\linewidth]{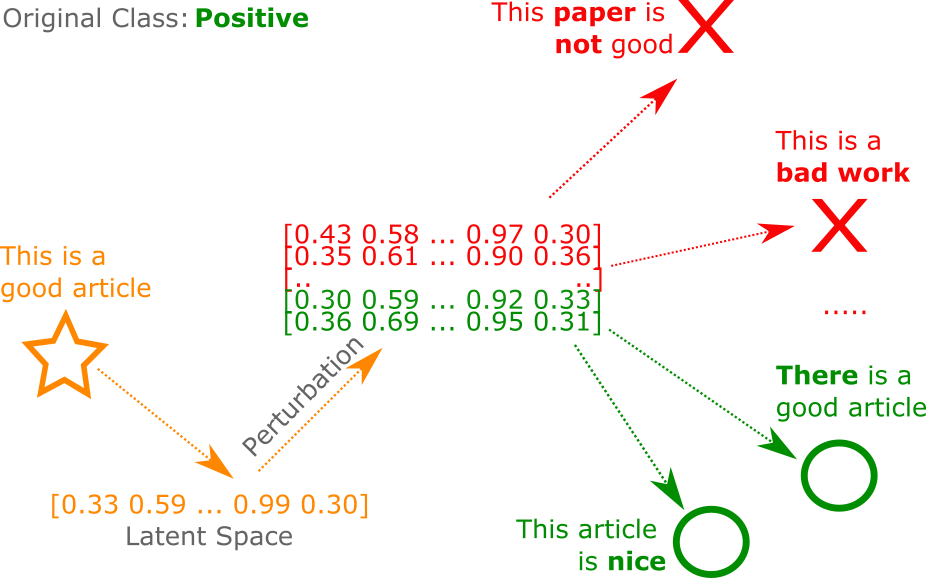}
        \caption{Opaque perturbation}
        \label{fig: opaque_perturbation}
    \end{subfigure}
    \label{fig: perturbation}
    \caption[Illustration of the mechanism employed to perturb the target documents by the transparent and opaque methods.]{The mechanism employed to perturb the target documents by the transparent and opaque methods. The transparent techniques convert the input text to a vector representation, where 1 indicates the presence of the input word and 0 denotes a replacement. The opaque methods embed words from the target text into a latent space and perturb the text in this high-dimensional space.}
\end{figure*}

In this chapter, our primary objective is to determine whether the use of complex latent spaces is an interesting approach for generating counterfactual explanations. To this end, we conduct a comparative examination of these two families of counterfactual approaches, in order to clarify the advantages of one over the other. Based on our empirical analysis, we discovered that for some downstream NLP tasks such as spam detection, fake news or sentiment analysis, learning a compressed representation can be overkill. To illustrate this, and as a proof of concept, we employed two transparent counterfactual explanation techniques, namely, Growing Language and Growing Net, introduced in Section~\ref{subsec: counterfactual_text}. Intriguingly, these transparent techniques outperform opaque methods, mainly because, opaque approaches often produce non-intuitive counterfactual explanations, i.e., counter-example texts that bear no resemblance to the target text. This not only contradicts the essence of counterfactual explanations but also raises questions about the actual level of transparency achieved when explaining a black box with another black box. 

In the previous chapter, we introduced a framework designed to enhance explanation fidelity. Our focus was primarily on enhancing fidelity from a data-oriented perspective, as prior research has established the profound impact of fidelity on user interaction with explanations~\cite{Fuernkranz2020, too_much_too_little}. On this basis, our current research aims to determine whether the complexity of opaque methods translates into significant performance improvements over transparent methods. Consequently, we evaluate the performance of transparent counterfactual explanation methods compared to opaque methods on three classic NLP tasks: fake news detection, sentiment analysis, and spam detection. Before delving into our experimental setup and results in Section~\ref{sec: experiments}, we will begin by reviewing the various counterfactual explanation methods found in the literature in Section~\ref{sec: rw}.

The work covered in this chapter served as the foundation for the paper titled: Explaining a Black Bow without a Black Box, which is planned for submission to the NAACL 2024 conference.

\section{Counterfactual Techniques for Textual Data}
\label{sec: rw}
Counterfactual explanation methods compute contrastive explanations for ML black-box algorithms by providing examples that resemble a target instance but that lead to a different answer in the black box. These counterfactual explanations 
convey the minimum changes in the input that would change a classifier's outcome. Social sciences~\cite{miller} have shown that human explanations are contrastive and Wachter et al.~\cite{wachter} have illustrated the utility of counterfactual instances in computational law. When it comes to NLP tasks, a good counterfactual explanation should be sparse~\cite{Pearl}, i.e., look like the target instance, and be fluent~\cite{text_attack}, i.e., read like something someone would say. 

Counterfactual approaches have gained popularity in the last few years. As illustrated by the surveys, first by Bodria et al.~\cite{survey_pisa} and later by Guidotti~\cite{survey_cf_Guidotti}, around 50 additional counterfactual methods appeared in a one-year time span. Despite this surge of interest in counterfactual explanations, their study for NLP applications remains underdeveloped~\cite{mice}. In the following, we elaborate on the existing counterfactual explanation methods for textual data along a spectrum that spans from transparent to opaque approaches.

\noindent \textbf{Transparent Approaches.} Given an ML classifier and a target text (also called a document), transparent techniques compute counterfactual explanations in a binary space. Each dimension represents the presence (1) or absence (0) of a word from a given vocabulary. Hence, to perturb a text, these methods toggle on and off 0s and 1s, where 0s are tantamount to adding, removing, or replacing words until the classifier yields a different answer. This was first proposed by Martens and Provost~\cite{martens} who introduced Search for Explanations for Document Classification (SEDC), a method that removes the words for which the classifier exhibits the highest \emph{sensitivity}. These are words that impact the classifier's prediction the most. Similarly, feature-attribution explanation methods such as LIME~\cite{LIME} and SHAP~\cite{SHAP}, mask words randomly from the target text. More recently, Ross et al.~\cite{mice} developed Minimal Contrastive Editing (MICE), a method that employs a Text-To-Text Transfer Transformer to fill masked sentences. Yang et al.~\cite{plausible_counterfactual} presented Plausible Counterfactual Instances Generation (PCIG), which generates grammatically plausible counterfactuals through edits of single words with lexicons manually selected from the economic domain. Since these methods are tailored for specific tasks or require manual selection, we removed these methods from our experiments.

\noindent \textbf{Opaque Methods.} We define opaque approaches as those perturbing the input text in a latent space in $\mathbb{R}^n$. Methods such as Decision Boundary~\cite{decision_boundary}, xSPELLS~\cite{xspells} or counterfactualGAN~\cite{cfgan} operate in three phases. First, they embed the target instance onto a latent space. This is accomplished by employing specific techniques such as Variational AutoEncoder (VAE) in the case of xSPELLS, or a pre-trained language model (LM) for counterfactualGAN. Second, while the classifier's decision boundary is not traversed, these methods perturb the latent representation of the target phrase. This is done by adding Gaussian noise in the case of xSPELLS, whereas counterfactualGAN resorts to a Conditional Generative Adversarial Network. Finally, a decoding stage generates sentences from the latent representation of the perturbed documents. There also exist methods such as Polyjuice~\cite{polyjuice}, Generate Your Counterfactuals (GYC)~\cite{gyc} and Tailor~\cite{tailor} that perturb text documents in a latent space, such as LM and Transformers, but can be instructed to change particular linguistic aspects of the target text, such as locality or grammar tense. Such methods are not particularly designed to compute counterfactual explanations but are rather conceived for other applications such as data augmentation.

Unlike pure word-based perturbation methods, latent representations are good at preserving \emph{semantic closeness} for small perturbations. That said, these methods are not free of pitfalls. First, methods such as xSPELLS and CounterfactualGAN are deemed opaque since a latent space is not human-understandable~\cite{interpret_gan}. Therefore, an intriguing paradox emerges between developing explanation techniques through latent space and methods explaining latent spaces~\cite{gan_interpret_latent}. It invites us to consider a fundamental question: Is it prudent to employ non-interpretable mechanisms to gain insights into complex classifiers? Moreover, existing latent-based approaches do not seem optimized for sparse counterfactual explanations. We prove this through experimental results that reveal a minor alteration in a latent space can cause a significant alteration in the original space.

\section{Comparative Study}
\label{sec: experiments}
Until now, we presented counterfactual explanation techniques as either opaque or transparent. However, the landscape is more nuanced. In this section, we provide an exploration of the complexity spectrum, which is depicted in Figure~\ref{fig: complexity_scale}. Subsequently, we delve into the details of the experiment setup in Section~\ref{subsec: experimentalinfo}, which is essential for replicating our experiments and understanding the methodology. 

\begin{figure}
    \centering
    \includegraphics[width=0.95\textwidth]{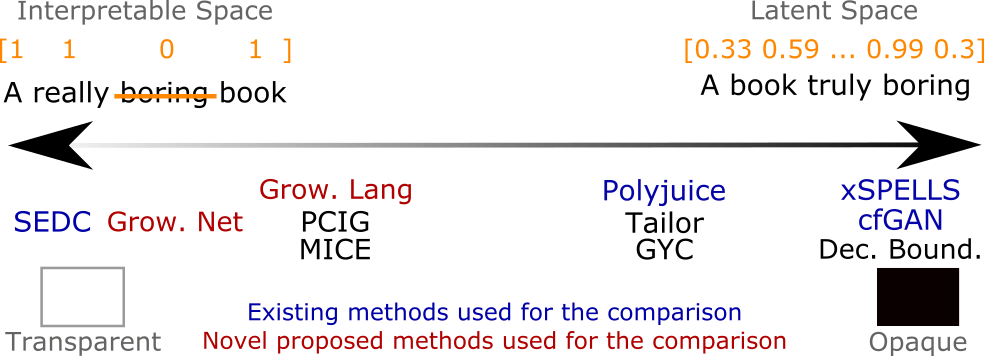}
    \caption[Spectrum for counterfactual explanation techniques that goes from the most transparent methods on the left to the most opaque methods, passing by our methods in red.]{Spectrum for counterfactual explanation techniques that goes from the most transparent methods on the left (e.g., SEDC) to the most opaque methods such as CounterfactualGAN, and xSPELLS, passing by our methods in red Growing Net and Growing Language. Transparent methods perturb documents in a binary space; opaque methods do it in a latent space.}
    \label{fig: complexity_scale}
\end{figure}

\subsection{Complexity Spectrum}
We introduce a complexity spectrum that serves as a visual guide to discern the transparency of various counterfactual explanation methods. The spectrum spans from the most transparent methods on the left to the most opaque ones on the right. 

\noindent\textbf{Fully Transparent.} At the leftmost end of the spectrum, we find the fully transparent approaches. This starts with the entirely transparent method SEDC~\cite{martens}, which perturbs text instances by hiding only highly sensitive words within the text. 

\noindent\textbf{Transparent Counterfactual Methods.} We positioned Growing Net as the second most transparent method as it goes beyond simple word masking by replacing words. Growing Net leverages the knowledge and tree structure of WordNet to select word substitutions more judiciously. This enables Growing Net to provide the user with information such as synonyms, antonyms, and hypernyms. Methods like PCIG, MICE, and Growing Language are classified as relatively more opaque among the transparent counterfactual techniques. They employ the latent space to identify semantically close word substitutions. However, despite their reliance on black-box techniques, these methods are considered transparent because their generated explanations preserve the document's structure while revealing which words should be replaced and by which other words.

\noindent\textbf{Partially Opaque.} Polyjuice, Tailor, and GYC fall in the category of partially opaque methods, as they leverage control codes to perturb the target document. Control codes are specific instructions that guide the model to perform a certain task, such as translating, summarizing, or changing the tense of a text. While these modifications occur in a latent space, the inclusion of control codes enhances the clarity regarding why a modification influences the model's prediction.

\noindent\textbf{Fully Opaque.} On the far right of the complexity spectrum, we encounter fully opaque approaches such as Decision Boundary, xSPELLS and counterfactualGAN. These methods operate in a completely opaque manner, making it challenging for users to discern the underlying process of counterfactual generation. 

This complexity spectrum provides valuable insights into the transparency and opacity of counterfactual explanation methods, enabling a more nuanced understanding of their capabilities.

\subsection{Experimental Information}
\label{subsec: experimentalinfo}
In this section, we provide a detailed account of the counterfactual generation process, the datasets employed, the classifiers to be explained, and the metrics applied in our experiments.

\subsubsection{Counterfactual Generation}
We start by outlining the six distinct counterfactual methods taken for generating counterfactuals. We excluded PCIG and MICE from our subsequent analysis for various reasons. Regarding PCIG, we highlight that this method relies on domain-specific rules from the field of economics, which limits its applicability to our diverse datasets. In the case of MICE, we observe this approach relies on transformer models to identify semantically relevant word replacements, which deviates from our focus on less complex methods. Furthermore, we have excluded methods such as Text Attack, introduced by Morris et al.~\cite{text_attack}, from our analysis. These adversarial methods are not designed for explanatory purposes but rather to fool the model. Similarly, we have removed Linguistically-Informed Transformation (LIT), introduced by Li et al.~\cite{LIT}, which is a method aimed at automatically generating sets of contrasts. These methods aim to generate documents that are outside the data distribution and are therefore unrealistic. This goes against the desired attributes for effective counterfactual explanations

We fill the middle ground with two methods, Growing Net and Growing Language, which implement a similar strategy to existing transparent methods. However, they do so with fewer methodological complexities. We adapted the code used to generate counterfactuals for the three transparent methods (SEDC, Growing Net, and Growing Language) and made it available on GitHub\footnote{\url{https://github.com/j2launay/ebbwbb}}. We employed the original code for the opaque methods, as described below.

\noindent\textbf{SEDC:} We modified the code used for word masking to ensure its compatibility with classification models that do not output class probabilities. This modified code version is accessible in Python on our GitHub as a variant of the counterfactual method class. This class proposes to choose from SEDC, Growing Net, or Growing Language, all specialized in generating transparent explanations.

\noindent\textbf{Polyjuice:} To generate counterfactuals, we utilized the code available in the repository \url{https://github.com/tongshuangwu/polyjuice}. Default hyperparameters were used to perturb texts from each test set until we found instances classified differently by the model.

\noindent\textbf{xSPELLS:} We employed the V2 version of xSPELLS, found in the repository \url{https://github.com/lstate/X-SPELLS-V2}, with default hyperparameters. 

\noindent\textbf{counterfactualGAN:} We used the code provided in the official release page of the paper, which can be accessed at \url{https://aclanthology.org/2021.findings-emnlp.306/}. We executed counterfactualGAN (cfGAN) with the default hyperparameters.

This comprehensive approach to counterfactual generation ensures a diverse set of methods to evaluate and compare in our experiments.

\subsubsection{Datasets} 
For our experiments, we used three datasets designed for three different applications: (a) spam detection in messages, (b) sentiment analysis, and (c) detection of fake news from newspaper headlines. Each of these datasets comprises two target classes and contains between 4000 and 10660 textual documents. The average number of words in each document is between 11.8 and 20.8 as reported in Table~\ref{tab: datasets_counterfactual}. 

Concerning the fake news detection dataset, we constructed it by taking real newspaper titles from a dataset~\footnote{\url{https://www.kaggle.com/datasets/rmisra/news-category-dataset}} and fabricated titles from a fake news dataset~\footnote{\url{https://www.kaggle.com/competitions/fake-news/overview}}. This combined dataset is publicly available on our GitHub\footnote{\url{https://github.com/j2launay/ebbwbb}}. As for the polarity~\footnote{\url{https://www.kaggle.com/datasets/nltkdata/sentence-polarity}} and spam\footnote{\url{https://www.kaggle.com/datasets/uciml/sms-spam-collection-dataset}} detection datasets, we took them from Kaggle. We divided each dataset into training and testing sets using the scikit-learn library's function:~\footnote{\url{https://scikit-learn.org/stable/modules/generated/sklearn.model_selection.train_test_split.html}}: $train\_test\_split$ with a test size of 30\% and a random seed of 1.

\begin{table}[ht]
    \centering
    \normalsize
    \begin{tabular}{lcccccc}
        \multirow{2}{*}{Name} &  \multicolumn{3}{c}{Nb Words} & \multirow{2}{*}{Instances} & \multicolumn{2}{c}{Models' Accuracy}  \\ \cmidrule(r){2-4} \cmidrule(r){6-7}
        & Total & Average & STD & & Neural Network & Random Forest  \\ \toprule
        Fake  &  19419 & 11.8 & 3.2 & 4025 &  84\% & 84\% \\
        Polarity  &  11646 & 20.8 & 9.3 & 10660 &  72\% & 67\% \\
        Spam  & 15587 & 18.5 & 10.6  & 8559 & 100\% & 100\%  \\
        \bottomrule \\
    \end{tabular}
    \caption[Information about the experimental datasets.]{Information about the experimental datasets. The three columns under ``Nb Words'' represent respectively (a) the total number of distinct words in the whole dataset, (b) the average number of words per sentence, and (c) the standard deviation. The ``Instances'' column indicates the number of text documents per dataset. The last columns show the average accuracy of the two classifiers for each dataset.}
    \label{tab: datasets_counterfactual}
\end{table}

\subsubsection{Black-box Classifiers}
\label{subsec: black-box_counterfactual}
Our evaluation uses two distinct black-box classifiers implemented using the scikit-learn library and already employed in~\cite{xspells}. These black boxes are (i) a Random Forest (RF) consisting of 500 tree estimators, and (ii) a straightforward neural network (DNN) with the same amount of neurons as there are words in the dataset. 

In addition to predicting the class associated with a textual document, these classifiers provide class probabilities. We trained both classifiers on 70\% of the dataset and their accuracy was tested on the remaining 30\%. We also selected the target instance to explain within this test set. Across all datasets, the average accuracy of these two classifiers ranges from 67\% to 100\%. Detailed results are presented in Table~\ref{tab: datasets_counterfactual}.

We employed a count vectorizer~\footnote{\url{https://scikit-learn.org/stable/modules/generated/sklearn.feature_extraction.text.CountVectorizer.html}} to convert the texts from the dataset into a matrix of tokens as input for the model. 

\section{Results}
\label{sec: resultscounterfactual}
In this section, we present the outcomes of our comprehensive evaluation of six counterfactual explanation techniques, each positioned along the transparency spectrum illustrated in Figure~\ref{fig: complexity_scale}. We conducted six rounds of experiments categorized into two main aspects. First, we assess the quality of the generated counterfactual explanations based on three essential criteria: (i) minimality, (ii) outlierness, and (iii) plausibility. Second, we evaluate the methods themselves in terms of (iv) flip change, (v) stability, and (vi) runtime. Our evaluation is based on three well-known NLP classification tasks: spam detection, polarity review, and fake news detection. For each task, we trained both a neural network, specifically a multi-layered perceptron, and a random forest classifier as described in Section~\ref{subsec: black-box_counterfactual}. In total, we generated counterfactual explanations for 100 target texts per dataset, black-box classifier, and counterfactual methods, serving as the inputs for our comprehensive evaluation. The explanation methods are ordered following their positions on the complexity spectrum from Figure~\ref{fig: complexity_scale}, ranging from the most transparent to the most opaque methods. Additionally, we have made our code and datasets available on GitHub\footnote{\url{https://anonymous.4open.science/r/ebbwbb-4B55/README.md}}.

\subsection{Counterfactual Quality} 
A high-quality textual counterfactual explanation adheres to several essential criteria~\cite{survey_cf_Guidotti}, including (i) minimal changes that make the counterfactual look closely to the target text, (ii) non-outlierness, meaning the counterfactual instance should resemble other phrases in the classifier's training/testing set, and (iii) linguistic plausibility, which entails that the counterfactual should sound like something a person would naturally write or say. The minimality criterion is quantified by measuring the distance between the counterfactual and the target sentence. Outlierness is operationalized as the distance of the counterfactual from the ``data manifold''. We thus measure outlierness through the distance between the counterfactual and the closest real instance in the dataset~\cite{Delaunay_cikm}. Linguistic plausibility, though typically evaluated through user studies~\cite{gyc, polyjuice}, is approximated here following Ross et al.~\cite{mice, tailor} and using perplexity scores based on a GPT language model~\cite{gpt3}, where lower scores indicate higher plausibility.

\begin{figure}[ht]
    \centering
    \includegraphics[width=0.95\textwidth]{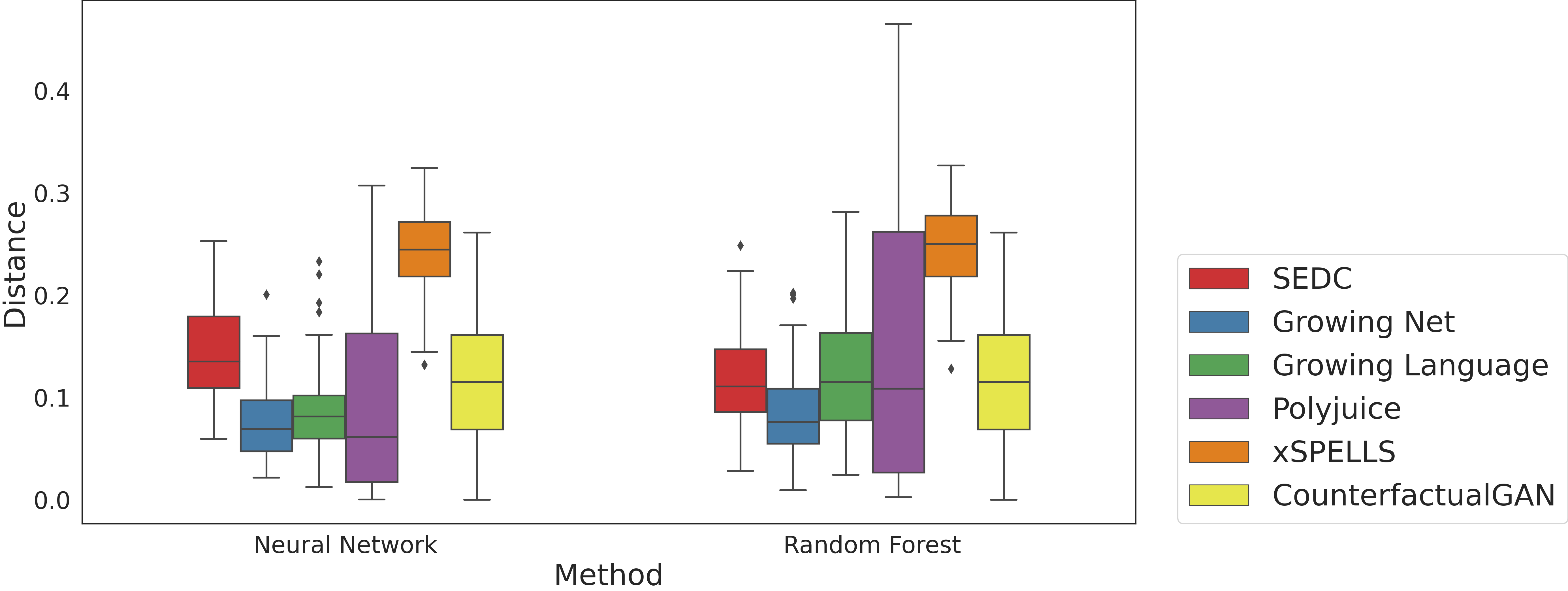}
    \caption{Minimality as the distance between the closest counterfactual and the target document (the lower the better).}
    \label{fig: minimality}
\end{figure}

Figure~\ref{fig: minimality} displays the results for minimality, which is the distance between the generated counterfactuals and their corresponding target texts. To assess minimality, we embedded both the counterfactual and the original text using a sentence model derived from GPT~\cite{distance-openai}. Subsequently, we computed the cosine distance between these embeddings. This ensures that our measurement considers a balance between lexical similarity and latent features, such as ``style''. Notably, our findings reveal that methods positioned in the middle-ground, particularly Growing Net, performed favorably compared to opaque approaches. It is worth noting that xSPELLS introduced the most significant changes to the original text. Similarly, we observe a high variance in the minimality of the counterfactuals generated by Polyjuice, indicating that some counterfactuals were notably distant from their corresponding target instances. While these methods introduced minor perturbations to the original text, these modifications occurred within a latent space. Nothing guarantees, however, that these minor adjustments translate into visually subtle modifications of the target phrase when the resulting phrase is brought back to the original space.

\begin{figure}[ht]
    \centering
    \includegraphics[width=0.95\textwidth]{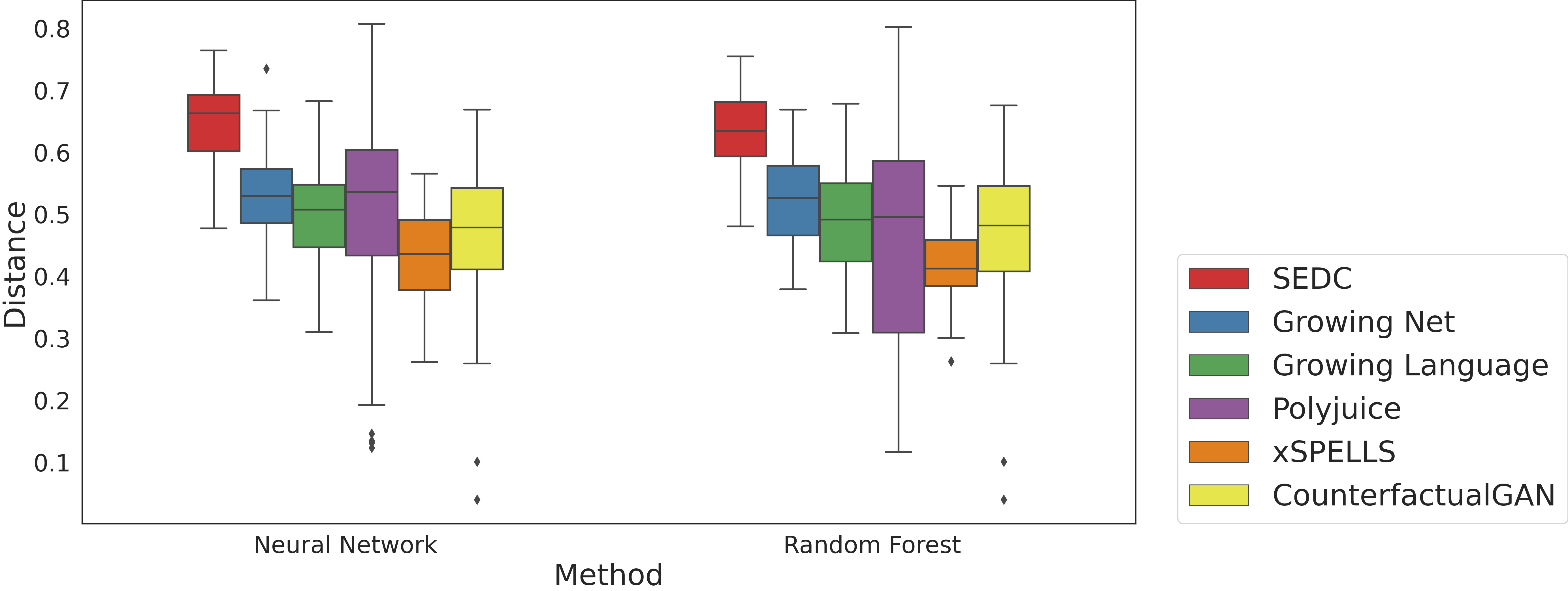}
    \caption{Outlierness, quantified as the measure of distance between the counterfactuals and the nearest instance within the test set.} 
    \label{fig: outlierness}
\end{figure}

Figure~\ref{fig: outlierness} illustrates the results for outlierness, which are computed as the distance between the generated counterfactuals and the closest test instances within our experimental datasets -- hence the lower the better. We observe that xSPELLS excelled in this criterion. This success can be attributed to its reliance on VAEs that are fine-tuned on the dataset and designed to create a compressed representation of it. At the other extreme, SEDC performed poorly, because removing words from the target text may lead to incomplete sentences that are identified as outliers. Notably, both Growing Net and Growing Language outperformed counterfactualGAN and Polyjuice in terms of outlierness, despite not relying extensively on heavy neural-network-based machinery.

\begin{figure}[ht]
    \centering
    \includegraphics[width=0.95\textwidth]{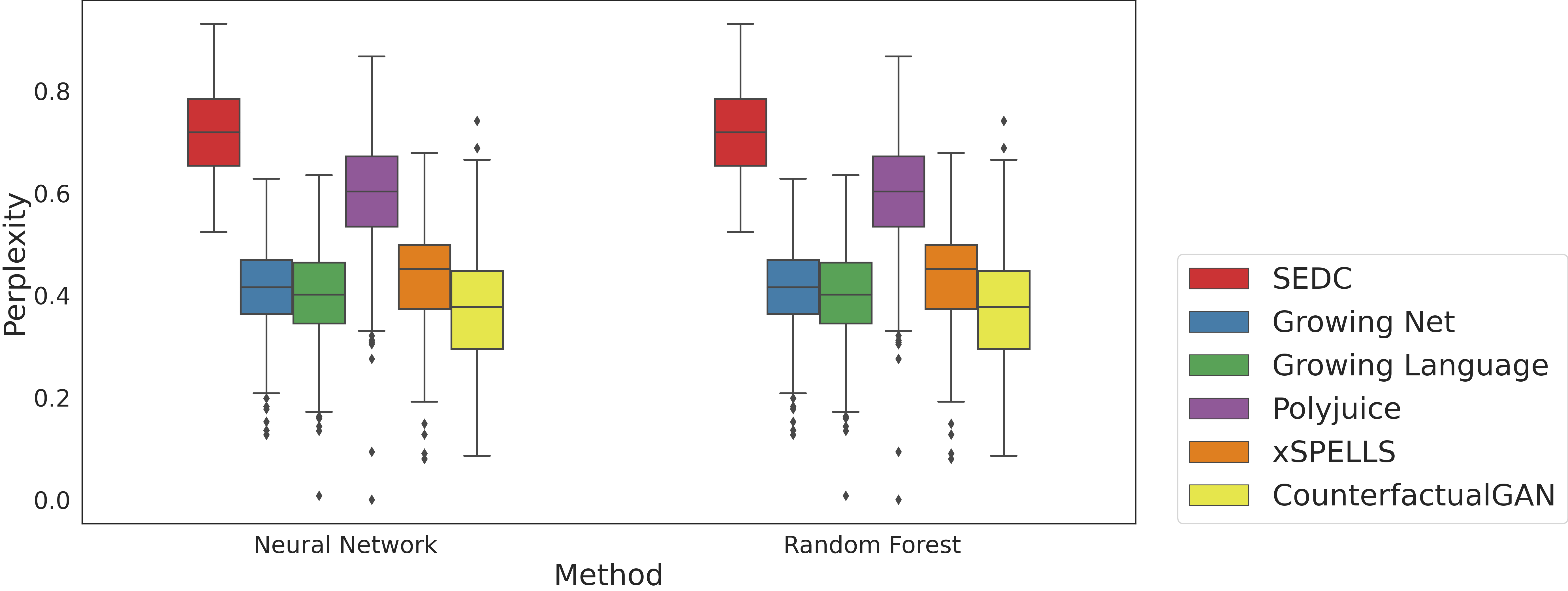}
    \caption{Perplexity as the MSE loss of a GPT model on the generated counterfactuals.} 
    \label{fig: perplexity}
\end{figure}

Figure~\ref{fig: perplexity} presents the plausibility of the counterfactuals, measured through the perplexity scores obtained from a language model. This score is computed by calculating the mean squared error (MSE) loss of a GPT model when predicting the next token in the counterfactual. We normalized the perplexity scores where lower scores indicate higher plausibility. SEDC and Polyjuice generated texts with the lowest plausibility, which is expected since SEDC masks words, leading to nonsensical sentences. In contrast, both Growing Net and Language achieved perplexity loss similar to those of xSPELLS and counterfactualGAN.

\begin{table*}[ht]
    \footnotesize
    \centering
    \begin{tabular}{lcccccccccccc}
    \toprule
    Method &    \multicolumn{2}{c}{SEDC} &    \multicolumn{2}{c}{Grow. Net}  &    \multicolumn{2}{c}{Grow. Lang.} &    \multicolumn{2}{c}{cfGAN} &    \multicolumn{2}{c}{xSPELLS} &    \multicolumn{2}{c}{Polyjuice}   \\ 
    \cmidrule(r){2-3} \cmidrule(r){4-5} \cmidrule(r){6-7}  \cmidrule(r){8-9} \cmidrule(r){10-11} \cmidrule(r){12-13}
    Black box & DNN & RF  & DNN & RF & DNN & RF & DNN & RF & DNN & RF & DNN & RF \\
    \midrule
        spam         & 0.57 &  0.55 &        0.36  &  0.35  & \textbf{0.65} & 0.49 & 0.57    &    0.57 &  0.34  &  \textbf{0.61} & 0.14 & 0.21\\
    \midrule
        fake &         0.90  &  0.98 &  0.79   & 0.59  &  0.86  &  0.86 &  \textbf{1}  &  \textbf{1} & \textbf{1}  & \textbf{1} & 0.38 & 0.28\\
    \midrule
    polarity & 0.95  & 0.92   &   \textbf{1} & 0.85    &  0.93  & 0.93 &  0.65  & 0.65  &  \textbf{1}  & \textbf{1} & 0.46 & 0.46 \\
    \bottomrule
    \end{tabular}
    \caption{Average label flip per dataset and black box of the six counterfactual methods.}
    \label{tab: recall}
\end{table*}
\subsection{Method Quality} 
In this section, we compare the quality of the counterfactual explanation methods based on three key criteria: (iv) label flip rate, which measures how frequently these methods successfully produce a counterfactual, that is, an instance classified differently by the model, (v) stability, the average similarity between five generated counterfactuals for the same document, and (vi) runtime, the time it takes for each method to generate a counterfactual explanation. 

Table~\ref{tab: recall} provides an overview of the label flip results, which indicates the methods' ability to find a counterfactual for a given text document. It is noteworthy that xSPELLS achieves the highest label flip rate, except for the spam detection dataset using neural networks. Additionally, our transparent methods consistently outperformed Polyjuice on every task and counterfactualGAN for polarity review. Notably, Growing Net exhibits strong performance for the polarity dataset. This highlights the effectiveness of replacing words with antonyms as a means to discover counterfactuals. We also emphasize that both Growing Net and Growing Language can be fine-tuned for a more exhaustive search by adjusting its parameters, for example by lowering the similarity threshold or incorporating additional terms from WordNet's tree structure. While this can enhance the label flip rate, it may result in longer runtimes. 

\begin{figure}[ht]
    \centering
    \includegraphics[width=0.95\textwidth]{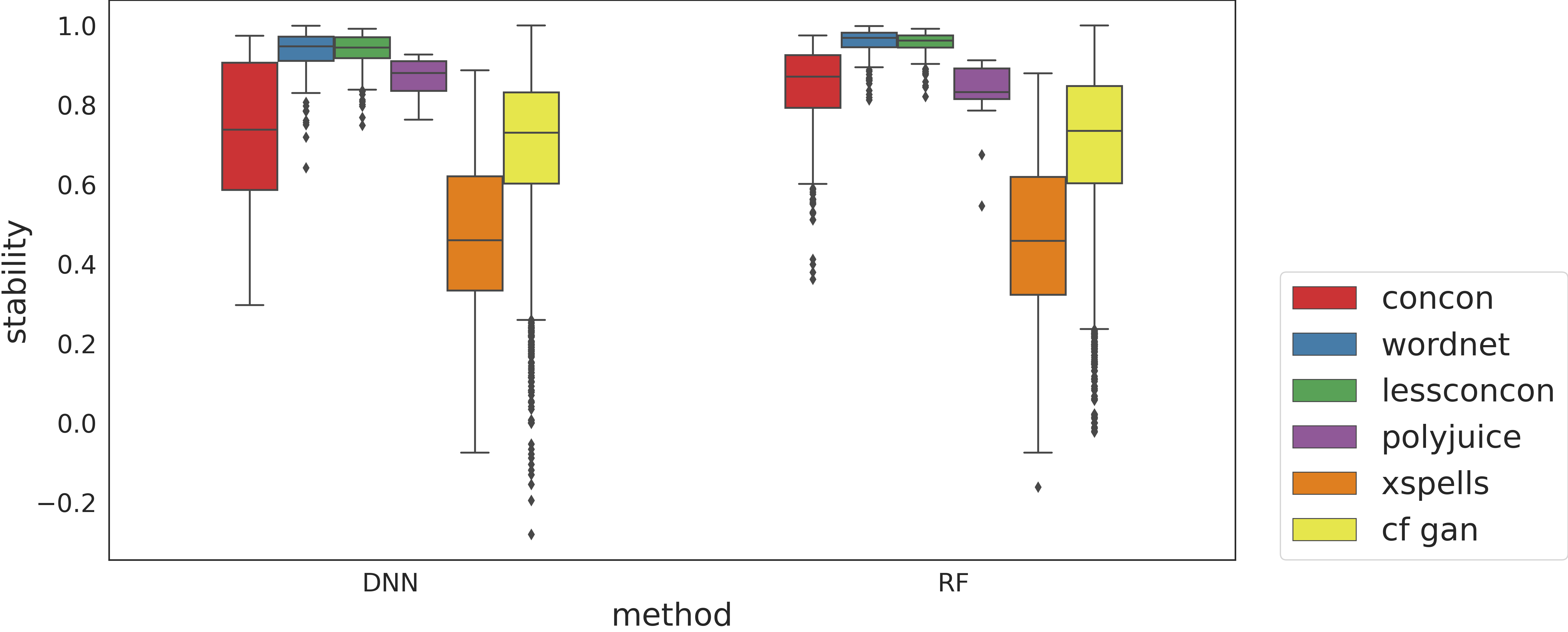}
    \caption[Stability as the average similarity between the generated counterfactuals for five successive runs on a single text document.]{Stability as the average similarity between the generated counterfactuals for five successive runs on a single text document. A similarity close to 1 indicates that the five counterfactuals generated are similar.} 
    \label{fig: stability}
\end{figure}

We present in Figure~\ref{fig: stability}, the stability results that measure the average similarity between five counterfactuals generated for the same target document. For each of the six approaches, we iterated five times over the same target document and measured the average similarity between each of the five generated counterfactuals. A method exhibits good stability if it systematically generates similar counterfactuals for the same document.

The results highlight that both Growing Language and Growing Net are the most stable. These methods obtain almost perfect stability (stability index close to one) while opaque methods and especially xSPELLS have a wide variability across different runs. We believe that these observations are due to variations in the set of similar words. Opaque methods, in particular, require re-training of the latent module, such as the VAE or the GAN, for each run. In contrast, the procedures employed by Growing Language and Growing Net are deterministic, and the sets of similar words remain consistent across successive generations. Our results also show similar stability for SEDC and Polyjuice, however, SEDC exhibits higher variance on the Neural Network classifier. We note that the high stability of Growing Language, Growing Net, and SEDC is because they have fewer steps that rely on randomness. Our results on stability suggest that transparent methods are less sensitive to random seeding than methods based on NN learning. 

\begin{table}[ht]
    \small
    \centering
    \begin{tabular}{llll}
    \toprule
    dataset & method & DNN & RF \\
    \midrule
        \multirow{5}{*}{spam} & SEDC & 21.45 (12.57) &   15.8 (9.36) \\
        & Growing Net &    0.97 (1.0) &    0.74 (0.8) \\
        & Growing Language & 60.21 (15.55) &  56.9 (13.84) \\
        & Polyjuice & 31.53 (7.07) & 62.23 (183.56)\\ 
        & counterfactualGAN &    1.4 (0.02) &   1.43 (0.03) \\
        & xSPELLS &   219.45 (16.78) &   197.63 (16.26) \\
        \midrule
        \multirow{5}{*}{fake} & SEDC &  30.7 (14.34) &  13.02 (6.15) \\
        & Growing Net &    1.54 (1.4) &   1.03 (0.76) \\
        & Growing Language & 54.81 (28.11) &     54,66 (12,16) \\
        & Polyjuice & 38.23 (7.75) & 70.04 (185.15) \\
        & counterfactualGAN &   1.03 (0.18) &    1.0 (0.13) \\
        & xSPELLS &    84.11 (6.47) &    85.63 (7.0) \\
        \midrule
        \multirow{5}{*}{polarity} & SEDC &  12.91 (9.77) &  12.39 (9.41) \\
        & Growing Net &   0.69 (0.91) &   0.63 (0.83) \\
        & Growing Language &   74.7 (33.3) & 73.56 (32.44) \\
        & Polyjuice & 81.07 (30.34) & 82.49 (47.66) \\
        & counterfactualGAN &   1.27 (0.23) &   1.29 (0.25) \\
        & xSPELLS &   135.69 (19.32) &   115.94 (10.9) \\
        \bottomrule
    \end{tabular}
    \caption[Average runtime in seconds for an instance (and standard deviation) per dataset and black box of the six counterfactual methods.]{Average runtime in seconds for an instance (and standard deviation) per dataset and black box of the six counterfactual methods. Note that the time for CounterfactualGAN (cfGAN) does not take into account the time to fine-tune the GAN on the target dataset. The training takes 6755, 4300, and 5770 seconds for the spam, fake, and polarity datasets, respectively.}
    \label{tab: runtime}
\end{table}

Finally, our runtime results are in Table~\ref{tab: runtime}. The table details the average and standard deviation of the runtime for each counterfactual explanation method across datasets and classifiers. Notably, CounterfactualGAN and Growing Net emerged as the fastest methods for generating counterfactuals. However, it is important to note that CounterfactualGAN requires the training of the Variational AutoEncoder (VAE) on each specific dataset, a process that incurs long training times. The time needed for fine-tuning varies, ranging from 4300 seconds for fake news title detection to 6755 seconds for spam detection.

Furthermore, we observe that xSPELLS and Growing Language exhibit the slowest runtime performance. Growing Language, for instance, requires approximately 60 seconds to generate a single counterfactual, while xSPELLS exhibits varying runtimes, ranging from 85 seconds for fake news detection to 219 seconds for spam detection. In summary, the two opaque methods necessitate a training step to adapt to specific datasets, whereas our two novel methods, along with SEDC, offer faster and less effortful development. 

To conclude, our runtime experiments reveal that methods like Growing Net are fast enough to be employed for the real-time generation of counterfactual explanations. This stands in sharp contrast to xSPELLS, which is two orders of magnitude slower due to its decoding phase from the VAE latent space to the original space. 

\section{Conclusion}
\label{sec: conclusion}
Our evaluation provides valuable insights into the landscape of generating counterfactual explanations for downstream NLP tasks. One of the most striking findings is that complexity, often associated with the use of neural networks and latent spaces, does not necessarily equate to superior performance in this context. Surprisingly, our results demonstrate that simpler approaches, characterized by a systematic and judicious strategy for word replacement within the target sentence, consistently yield satisfactory outcomes across a range of quality dimensions.


The results of our study prompt a deeper reflection on the optimal strategies for generating counterfactual explanations in the field of NLP. It invites readers to contemplate the broader implications of our findings and their implications for the development of transparent approaches versus improving opaque methods. The choice between these approaches should be made judiciously, considering the specific requirements and constraints of the application at hand.

Furthermore, our findings underscore the critical importance of transparency and interpretability in AI and machine learning. As we navigate in the complex landscape of increasingly sophisticated AI models, the need for transparency, accountability, and trust becomes paramount, especially in applications where human decisions are influenced by AI recommendations. The paradox of explaining one black box with another raises pertinent questions about the balance between model complexity, interpretability, and performance. It calls into question the unnecessary development of opaque approaches when transparent methods suffice, reinforcing the need for clarity and user-friendliness in AI systems.

In conclusion, our study contributes to the ongoing discourse on explainable AI by highlighting that effective counterfactual explanations can be achieved through simpler methods, challenging the prevailing assumption that complexity is always synonymous of better performance. We expect these findings will encourage the development of more transparent and interpretable AI systems, fostering trust and accountability in AI-driven decision-making processes.

In the second part of this thesis, our focus shifts to the user perspective. In this new part, we will delve into the impact of the three explanation methods used in each of the previous chapters. Before doing that, we reflect on the insights we collected throughout these chapters, primarily focusing on the data perspective.

First, our research journey through these chapters has revealed that there is ample room for improving existing explanation techniques. A noteworthy illustration of this is found in Chapter~\ref{chap: anchors}, where we demonstrate how pertinent and meticulous adjustments to the essential components of an explanation method can significantly improve the effectiveness of the resulting explanations. This suggests that further refinements and innovations in these methods hold the potential to enhance the quality and utility of explanations.

Secondly, it is imperative to assess judiciously when a particular explanation technique should be applied. Chapter~\ref{chap: ape} provides an example in this regard by demonstrating that linear explanations are not universally applicable and that there is a need for context-specific considerations. This insight is expected to extend to other explanation paradigms, highlighting the importance of choosing the right method for the dedicated task. This theme will be explored in the following part of this thesis.

Last but not least, the exploration of counterfactual explanation techniques conducted in this chapter offers an important lesson: the introduction of complexity does not inherently translate into improved performance. This lesson has brought implications not only for counterfactual explanations but extends to other explanation paradigms.

As we look ahead to the next part of this thesis, these insights serve as valuable guidance for our exploration of how to provide the best explanations from a user's perspective. We anticipate that these findings will continue to inform our research and contribute to the ongoing discourse on explainable AI and machine learning.

}{
	
}

\clearemptydoublepage
\ifenvsetTF{COMPILE_ALL}{
	\part[Explanations Tailored to the User]{What Makes a Good Explanation from the Perspective of the User?}

\chapter{User-Centered Evaluation of Explainability Methods}
\label{chap: context}
\minitoc
In the first part of this thesis, we have explored post-hoc explanation techniques, a widely popular family of methods that have gained significant attention within the XAI community. Despite the proliferation of post-hoc explainability methods, researchers have pertinently highlighted that XAI methods are not focusing sufficiently on the target users~\cite{finale,ribeira_user_centered}. In response to this observation, the second part of this thesis is dedicated to the evaluation of explanation techniques from a user-centric perspective. As highlighted by Doshi-Velez and Kim along with two other XAI surveys~\cite{survey_pro_hcxai2, survey_pro_hcxai}, a low number of XAI papers justify their novel methods through application-grounded evaluations or human-grounded metrics. Adadi et al.~\cite{survey_pro_hcxai2} found that in a sample of 381 XAI papers, only 5\% had an explicit focus on the evaluation of the proposed methods through human subjects. In other words, research in XAI is busy producing tons of local feature-attribution, rule-based, and example-based explanation methods for users without evaluating the resulting explanations with real users. This gap in research implies that we know little about the extent to which users understand explanations for AI systems. Likewise, it remains unclear whether the presence of explanations when using AI systems increases or not users' trust. In fact, some research findings have shown that explainable AI may offer limited benefits to users of AI recommendation systems~\cite{DBLP:journals/pacmhci/BucincaMG21, DBLP:conf/iui/GajosM22, DBLP:conf/fat/GreenC19, DBLP:journals/pacmhci/GreenC19, DBLP:conf/iui/NouraniRBHRRG21}. Similar results have been found for cognitive engagement~\cite{DBLP:conf/iui/ChromikEBKB21, placebic_explanations}.

To address these pressing issues, this chapter introduces a methodological framework for conducting user studies. The primary objective of this framework is to establish a robust methodology to investigate the impact of explanation techniques on users. Furthermore, we propose a comprehensive set of scales and metrics to gauge users' perceived and behavioral trust, understanding, and satisfaction. This methodological framework, along with the scales and metrics, will be subsequently applied in Chapter~\ref{chap: chi} to conduct a user study. This study aims to measure the benefits of different explanation techniques and their representations (whether graphical or text-based) on users' comprehension and trust. 

This chapter begins by providing in Section~\ref{sec: evaluating_xai_users} an overview of existing user studies that have been conducted to evaluate explanation techniques. Then, we introduce the methodological framework, along with the proposed scale and metrics, in Section~\ref{sec: method}. In Section~\ref{sec: context_conclusion}, we provide a discussion of how this framework may be applied.

\section{Evaluating Explanations with Users}
\label{sec: evaluating_xai_users}
This contribution lies at the intersection of eXplainable AI and Human-Computer Interaction (HCI) research. Therefore, we discuss the evaluation of XAI systems from a user's perspective. Then, we present existing guidelines and metrics to conduct user studies on XAI tools.

\subsection{Evaluating Explainable AI Systems with Users}
\label{subsec:evaluating_xai}
Miller argues that the development of effective explanation modules requires the joint effort of the XAI and HCI research communities~\cite{miller}. The HCI community has previously focused on the evaluation of  XAI models with the end-users~\cite{Niels_fairness,justice_bins,hcxai_liao_fairness,hcxai_counterfactual,van_der_waa}. We can group these evaluations into two categories: (a) assessment of novel explanation modules and representations, and (b) impact of the explanation's type on users' perception (e.g., trust, understanding). In contrast to these works, several surveys have highlighted the scarcity of XAI papers that evaluate novel explanation methods through user studies~\cite{survey_pro_hcxai2, survey_pro_hcxai, finale}. Consequently, most of the methods aimed at generating explanations for humans are evaluated without considering a human perspective. Among the studies that did consider the human perspective, most studies either assessed only the validity of their novel explanations method~\cite{decision_sets, kdd_21_user_studies, LIME, Anchor, cfgan, plausible_counterfactual} or the impact of the explanation's visual representation~\cite{cheng_explanation_style, tailored_visualisation, microsoft_manipulating, what-if}. A limitation of these works is that they are typically limited to the evaluation of one kind of explanation technique~\cite{decision_sets, tailored_visualisation, cfgan} and one application domain~\cite{microsoft_manipulating, plausible_counterfactual}. For instance, Lakkaraju et al. proposed a rule-based explanation method and measured the users' understanding in the healthcare domain~\cite{decision_sets}. While their results proved that their method increases the users' understanding, it is only valid for the healthcare domain and should not be extended to other domains. Moreover, some prominent explanation methods such as LIME and Anchors, evaluate the quality of the explanations with a small number of computer science students, who are already familiar with machine learning concepts~\cite{LIME, Anchor}. In our work, we set out to compare three different explanation techniques on two distinct datasets. 

To study the impact of explanations more broadly, prior works have evaluated users' trust and understanding in specific explanation conditions~\cite{edit_explanation, cheng_explanation_style, too_much_too_little, explanation_style_trust, van_der_waa}. For instance, Arora et al. studied the impact of interactive explanations on users' understanding~\cite{edit_explanation}. In their settings, an interactive explanation means that the user can edit the input text and directly visualize the effect on explanations. Their results confirmed that explanations might help users understand how the model works since they better identify key elements for the prediction. Cheng et al. compared the effect of interactive versus static explanations, as well as black box versus white box, on users' trust and understanding~\cite{cheng_explanation_style}. They observe that having access to the internal mechanism of the algorithm and interactive explanations both help to improve the users’ comprehension. Other researchers have studied the influence of explanation's representation on users' perceptions~\cite{Niels_fairness, cheng_explanation_style, explanation_style_trust}. As an example, Larasati et al.~\cite{explanation_style_trust} presented four explanation styles: contrastive, general, truthful, and thorough, and measured the impact of each style on human-AI trust. Interestingly, their results show that contrastive explanation styles were highly rated in terms of trust and personal attachment by users compared to general explanations, which offer simpler and broader insights. Van Berkel et al. compared textual and scatterplot representations and showed that the usage of a scatterplot visualization led to lower perceived fairness~\cite{Niels_fairness}. Other work has compared the effects of different explanation methods~\cite{edit_explanation, too_much_too_little, van_der_waa}, as for instance Van der Waa et al. who compared rule-based and example-based explanations in the domain of diabetes~\cite{van_der_waa}. Explanations were generated artificially by domain experts and the authors measured the impact of each explanation on users' trust and understanding. However, they were unable to identify significant factors in the explanation impacting the users' perception. Kulesza et al. proposed four different explanations based on various levels of soundness and completeness for a music recommender system~\cite{too_much_too_little}. The authors expected that too much completeness and soundness would confuse the users while their results showed that explanations with higher soundness and completeness increase the users' understanding of the recommender system. 

To summarise, a majority of research studies suggest that explanations influence users' perceived trust and understanding in human-AI interactions. However, certain studies have failed to reveal the impact of explanations on specific aspects of user perception, such as confidence or comprehension, as shown in previous works~\cite{edit_explanation, fail_HCXAI, van_der_waa}. Therefore, in line with recent guidelines for evaluating XAI applications~\cite{van_der_waa}, our proposed methodological framework focuses on various dimensions of user interaction with AI systems. Specifically, it investigates the effects of explanations on user trust, understanding, and satisfaction. Moreover, we combine metrics on both users' perceptual and behavioral to measure their trust and understanding when interacting with AI systems.

\subsection{Guidelines and Metrics to Conduct User Studies}
The evaluation of trust and understanding has been a prevalent topic in Psychology and Cognitive Sciences, which has led to numerous guidelines for conducting experiments. Hoffman et al.~\cite{hoffman_trust} surveyed several questionnaires from Social Sciences~\cite{trust_scale} and HCI~\cite{survey_trust} and combined them into a new satisfaction scale adapted for XAI. Cahour and Forzy~\cite{cahour-forzy} developed a trust scale based on three factors: reliability, predictability, and efficiency. This scale made of four questions directly asks users whether they are confident in the XAI system. Finally, Madsen and Gregor~\cite{Madsen-Gregor} proposed an eight-question scale measuring perceived technical competence and understandability.

Ribeira and Lapedriza~\cite{ribeira_user_centered}, as well as Doshi-Velez and Kim~\cite{finale}, proposed to group users into three categories: (a) machine learning practitioners, (b) domain experts, and (c) laypeople. Based on these three categories, Doshi-Velez and Kim propose to distinguish between application-grounded and human-grounded evaluations. While the former represents real tasks with computer scientists or domain experts, the latter are simplified tasks such as giving humans access to an input and an explanation and asking them to simulate the model's prediction. Doshi-Velez and Kim also indicated that running evaluations with laypeople offers the advantage of (a) evaluating the impact of the explanations more broadly, and (b) facilitating the conduct of the experiments since it is easier to control the factors of explanations in simplified tasks. In this part of the manuscript, we evaluate the impact of explanation techniques and their visualizations with laypersons. We also closely followed the advice from Van der Waa et al. by proposing an evaluation framework with several measurements, including trust, understanding, satisfaction, and completion time~\cite{van_der_waa}.

\section{Method}
\label{sec: method}
In response to the recognized need for user studies and guidelines to evaluate explanation techniques in XAI, this section presents a methodological framework. This framework is designed to comprehensively assess user interactions with XAI systems through surveys, wherein participants engage in specific tasks. This provides valuable insights into the participants' perceptions and behaviors when interacting with AI models. To ensure a structured approach and reproducibility, we organized the surveys into three distinct phases: the introduction, the task rounds, and the post-questionnaires. Moreover, we introduce a range of scales and metrics, to quantify the impact of different variables. These measurements provide a clear picture of how various factors influence user behavior, trust, and understanding when interacting with AI systems. In the next chapter, we apply this methodological framework, alongside the defined scales and metrics to conduct a user study.

\subsection{Methodological Framework}
\label{subsec: task}
The proposed methodological framework consists of online surveys where participants are confronted with tasks. These tasks encompass a range of activities, including making predictions based on provided information (for example, predicting the risk of obesity based on lifestyle habits or determining the topic of a newspaper article based on its title). Participants may also be tasked with replicating the prediction of an AI model based on an explanation for a similar instance. Figure~\ref{fig: user_study_standard} outlines the process followed by such a survey. Given a task, the only difference among the surveys is the ``Read Explanation'' phase. Each survey is composed of three phases: (i) introduction, (ii) task rounds, and (iii) post-questionnaires.

\noindent\textbf{Introduction.} The survey begins with an introductory explanation of the tasks assigned to the user and the information used by the AI model to make recommendations. Then the user is asked questions to verify whether they understood the task.

\begin{figure}
    \centering
    \includegraphics[width=0.95\textwidth]{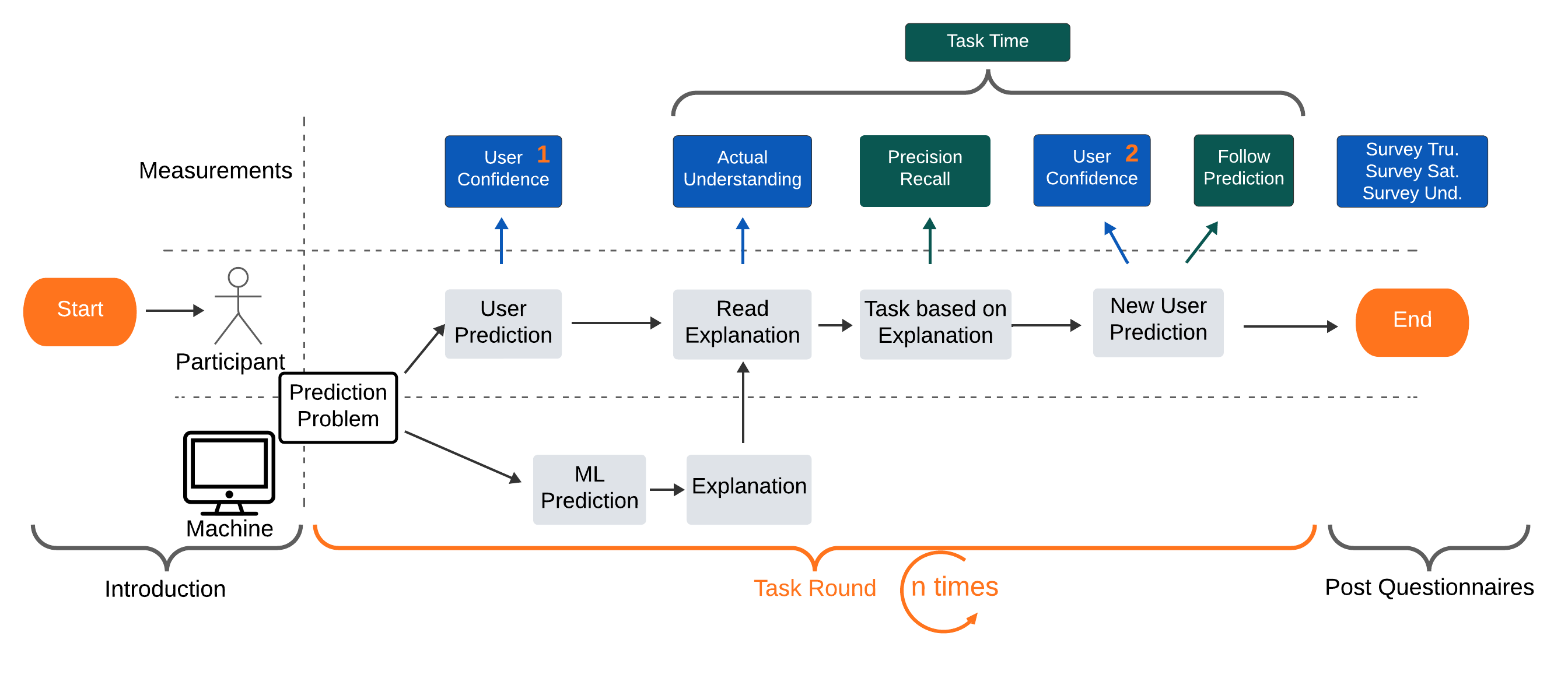}
    \caption{Diagram depicting the experimental workflow proposed in this thesis and used to assess user perception and behavior when interacting with a given explanation technique. Behavioral measurement steps are indicated in green, while self-reported measurement steps are in blue. The task rounds are repeated $n$ times.}
    \label{fig: user_study_standard}
\end{figure}

\noindent\textbf{Task Round.} After the introductory explanations, users are presented with the prediction tasks. Each task is divided into two steps. First, the users are asked to make an assessment based on the given information. After stating their assessment, the participants have access to the AI model's prediction along with its associated explanation. Based on this explanation, we then ask the users to complete a task. This task may be to select the features, among the list of all possible features, that were used by the AI model to make its recommendation. Another task could be to ask the users to provide counterarguments or point out potential drawbacks related to the AI model's recommendation. In the second stage, users can reconsider their prediction and answer two questions on a 5 Likert scale. A Likert scale offers a range of responses, from ``strongly disagree'' to ``strongly agree''. Those questions gauge the extent to which users believe they understand the provided explanation and their level of confidence in their prediction.

\noindent\textbf{Post-Questionnaires.} After the prediction tasks, our methodology involves the utilization of three standardized questionnaires, each comprising 20 questions. Through these questionnaires, users are encouraged to report their trust, understanding, and satisfaction concerning the AI model. The proposed framework ends with two open questions, designed to get a deeper understanding of the users' perception: (1) ``According to the scenarios you have seen and the corresponding explanation, how does the artificial intelligence tool predict?'', and (2) ``What was good in the explanation? What was bad in the explanation?''. Participants should not be restricted by word limits and are encouraged to provide as many points as they feel necessary. 

\subsection{Scales \& Metrics}
\label{subsec:metrics}
To assess the impact of the independent variables, which are the factors or conditions we control in our study, we propose to employ a range of scales and metrics. In the context of this thesis, the independent variables may include different explanation techniques, their representations, or a combination of paradigms. We propose to evaluate users' behavior and perception regarding trust and understanding of the AI system through various metrics. The proposed methodology seeks to distinguish between the subjective experiences of users and the objective outcomes. Additionally, we measured self-reported satisfaction with the system, which serves as a dependent variable, that reflects how users' experiences are influenced by the independent variables. Furthermore, we recorded the time taken to complete the task round, another dependent variable, which can provide insights into user efficiency and task engagement. Figure~\ref{fig: user_study_standard} illustrates the points in the study where these parameters, both independent and dependent, are measured.

This comprehensive approach enables the analysis of the relationships between the independent variables and the observed dependent variables, providing a deeper understanding of how various factors impact users' interactions with the AI system.

\textbf{Trust.} 
Jarvenpaa et al.~\cite{trust_collaboration} have shown that trust influences several factors and should be considered as an essential factor for humans to successfully collaborate with computers. Therefore, we build upon the methodology of Broon and Holmes~\cite{belief_trust} to measure users' behavioral trust and adapt scales for self-reported trust to align with the XAI domain. We distinguish two categories of trust metrics:

\begin{itemize}
    \item \textbf{Behavioral Trust (Follow Pred.)} This is the proportion of times users modify their initial prediction in favor of the AI's prediction. In this context, we consider only scenarios where the user's initial prediction diverged from the AI's prediction. This metric is particularly relevant when users are prompted to reevaluate their own predictions due to contrast with the AI's prediction, as it provides insights into the level of trust users place in the AI's recommendations.
    \item \textbf{Self-Reported Trust ($\Delta$ Confidence and Survey Tru.)} The assessment of perceived trust is composed of two distinct metrics. Firstly, it includes users' self-reported confidence in the AI model. Secondly, it resorts to a post-survey evaluation based on a questionnaire developed by Cahour and Forzy~\cite{cahour-forzy}, as exemplified in Section~\ref{subsec: metrics_illustration}.
    \begin{itemize}
        \item \textbf{$\Delta$ Confidence.} These are changes in self-reported trust before and after accessing AI predictions and explanations ('User Confidence 2' - 'User Confidence 1' in Figure~\ref{fig: user_study_standard}).
        \item \textbf{Survey Tru.} This is based on the answers from a four-question questionnaire, originally developed by Cahour and Forzy~\cite{cahour-forzy}, to assess users' perceptions of the AI system's reliability, predictability, and efficiency. The questionnaire has been adapted for application in XAI by Hoffman et al.~\cite{hoffman_trust}, and participants rate their responses on a seven-point Likert scale.
    \end{itemize}
\end{itemize}

\textbf{Understanding.} A widely accepted definition of a good explanation is its capacity to be understood by a human within a reasonable time frame, as defined by Lipton~\cite{Lipton}. We thus gauge the users' comprehension of the model through various aspects divided into behavioral and self-reported metrics.

\begin{itemize}
    \item \textbf{Behavioral Understanding (Prec. and Rec.)} Building upon the methodology proposed by Weld and Bansal~\cite{weld-bansal}, we assess users' behavioral understanding through a simple task. Therefore, it is important to note that the metrics may be adapted to the nature of the task. In the following chapter, we ask users to identify the features that have the highest impact on the classifier's prediction after viewing an explanation. Subsequently, we measure the behavioral understanding through two metrics:
    \begin{itemize}
        \item \textbf{Precision (Prec.)} Measure of alignment between features identified by users and top features reported in explanations.
        \item \textbf{Recall (Rec.)} Measure of users' ability to identify all the influential features indicated in explanations.
    \end{itemize}
    \item \textbf{Self-Reported Understanding (Immediate Und. and Survey Und.)} The measurement of perceived understanding combines self-reported comprehension during explanation review and post-survey assessment using Madsen and Gregor's questionnaire~\cite{Madsen-Gregor}.
    \begin{itemize}
    \item \textbf{Immediate Und.} Self-reported comprehension of the system's prediction on a five-point Likert scale while looking at the explanation.
    \item \textbf{Survey Und.} Adapted questionnaire by Madsen and Gregor~\cite{Madsen-Gregor} on perceived technical competence and understandability, using a five-point Likert scale for consistency.
    \end{itemize}
\end{itemize}

\textbf{Additional measures.} 
According to Lipton's definition of a good explanation~\cite{Lipton}, the time taken to understand the explanation is a key element for measuring its quality. Therefore, we consider the time taken to understand explanations and user self-reported satisfaction as additional measurements to provide a comprehensive evaluation. 
\begin{itemize}
    \item \textbf{Task Time.} The time required to complete the requested task based on the explanation.
    \item \textbf{Satisfaction (Survey Sat.)} Users' satisfaction with the AI model is assessed using the Explanation Satisfaction (ES) questionnaire by Hoffman et al.~\cite{hoffman_trust}, comprising eight items rated on a seven-point Likert scale.
\end{itemize}

\section{Conclusion}
\label{sec: context_conclusion}
In this chapter, we have delved into an important yet underexplored area within the field of explainable AI (XAI): user-centered evaluation. We have highlighted a pressing need to measure the real impact of explanation methods on users. To address this gap, we have introduced a set of metrics and a robust methodological framework that can be widely applied across various contexts. This user-centric methodology, tailored for conducting user studies, relies on specific indicators designed to assess understanding, trust, and other aspects related to the explainability of AI systems. It aims to provide a standardized framework for gauging the effects of explanation methods on users, enabling objective and reproducible result comparisons.

Our approach aims to contribute to the advancement of XAI by placing the user at the center of evaluation. With this powerful tool, we are well-prepared to measure the actual influence of explanation methods in AI on users. By understanding how users interact with explanations and how this interaction affects their perception of AI systems, we as a community are more prepared. To shape the future of these technologies we will therefore be able to make them more comprehensible and trustworthy to a wider audience. In the upcoming chapter, we apply this methodology through a user study. This study is designed to measure the impact of three different explanation methods and their representations on users. It aims to provide valuable insights for comparative analysis, enhancing the transparency and acceptance of AI systems.

\clearemptydoublepage

\chapter[Explanation Techniques and Representations on Users]{Impact of Explanation Techniques and Representations on Users}
\label{chap: chi}
\minitoc



As we highlighted in the previous chapter, there is a clear lack of comprehensive user studies that investigate how users perceive and interact with these explanations. As a result, the focus of the XAI community is starting to shift toward comprehensive user studies. In the first part of this thesis, our focus was primarily on optimizing explanations from a data perspective. In each chapter of the first part, we focused on a specific explanation paradigm, developing techniques to enhance the quality of explanations. In this chapter, we explore a distinct aspect, namely the impact of the explanation paradigm on users' trust and understanding. Our objective is to gauge how these factors shape users' perceptions and behavior. Consequently, we aim to provide valuable insights that can inform the design of AI interfaces and user-centered XAI systems. We shed light on these issues by addressing the following research question: 
\begin{enumerate}[label= \textbf{RQ\arabic*:}, leftmargin=1cm]
    \item Which local explanation technique, \textit{i.e.,} feature-attribution, rule-based, or counterfactuals, provides the best explanations in terms of users' trust and comprehension of the AI model?
\end{enumerate}

\noindent Moreover, we examined eight different XAI toolkits\footnote{AI360, Dalex, H2O, eli5, InterpretML, What-if-Tool, Alibi, Captum.} which employ at least one of the three aforementioned explanation types, and note that each explanation method is commonly represented under a certain form (graphical for feature-attribution explanations and textual for counterfactual and rule-based). This leads us to this research question:
\begin{enumerate}[label= \textbf{RQ2:}, leftmargin=1cm]
    \item Does the explanation's visual representation impact the users' trust and understanding?
\end{enumerate}

To answer these research questions, we employed the methodological framework proposed in the previous chapter to conduct two user studies. These studies evaluate the impact of the three aforementioned explanation techniques and two visual representations (graphical vs. text) on users' trust and understanding. By applying the set of scales and metrics introduced earlier, we obtain multiple findings. Notably, the choice of explanation technique significantly impacts the users' comprehension. Moreover, the selection between graphical and textual representation has a greater impact on users' trust. In summary, we discover that a graphical representation for explanations is perceived as more trustworthy, whereas rule-based explanations are the most effective at conveying the most important features of the AI decision process. Conversely, we find that users confronted with counterfactual explanations exhibit an understanding of the AI model comparable to our control group, that is, users who receive an AI-assisted prediction but with no explanation.

Most of the research presented in this chapter was the subject of the paper: Impact of Explanation Techniques and Representations on Users Comprehension and Trust in Explainable AI, submitted at the CHI 2024 conference.

\section{Explanation Techniques and Representations}
\label{subsec:system}
We next elaborate on the two datasets used for the studies and the implementation details of the explanation methods. Further, we discuss and present the chosen representation approaches. 

\subsection{Datasets \& AI models}
\label{subsec:data_and_ai}
\begin{table}[ht]
    \centering
    \begin{tabular}{@{}lccc@{}}
        \toprule
        \multirow{2}{*}{\textbf{Dataset}} &  \multicolumn{2}{c}{\textbf{Features}} & \multirow{2}{*}{\textbf{Instances}} \\ \cmidrule(r){2-3}
        & \textbf{Numerical} & \textbf{Categorical} & \\ \midrule
        Compas & 1 & 7 & 5364 \\ 
        Obesity & 2 & 13 & 2111 \\ \bottomrule
    \end{tabular}
    \caption{Datasets composition.}
    \label{tab:datasets}
\end{table}
Our evaluation is conducted on two widely used datasets among the XAI community, namely the COMPAS~\cite{compas_original} and Obesity datasets~\cite{obesity}. These datasets represent two domains where explainability is often deemed crucial: law and healthcare. COMPAS is a tabular dataset used to train a model that predicts a criminal defendant's likelihood of re-offending. The data comprises American defendants. 

The Obesity dataset~\cite{obesity} is used to predict the risk of developing obesity based on an individual's Body Mass Index (BMI) and answers to various questions. The data originates from Colombia, Peru, and Mexico. We chose those datasets in line with the recommendations from Van Berkel et al.~\cite{Niels_fairness} and Van der Waa et al~\cite{van_der_waa}, who suggest that a meaningful application-agnostic XAI evaluation should preferably include more than one domain, and strike a balance between simplicity -- users should understand the domain of the AI --, and plausibility -- the task should be difficult enough to justify the need for AI assistance. In the spirit of these principles, we conducted some basic feature selection on both datasets to reduce the number of features. We, for instance, removed the BMI index from the obesity dataset, which otherwise would have oversimplified the prediction task. Table~\ref{tab:datasets} contains the final number of features and instances for both datasets as used in our experiments. As both these datasets are public, our results could be used as a baseline for future researchers aiming to evaluate the effectiveness of their explanation method.

\textbf{AI Model and Explanations.} We used our experimental datasets to train an AI model based on a Multi-Layer Perceptron (MLP) classifier\footnote{\url{https://scikit-learn.org/stable/modules/generated/sklearn.neural_network.MLPClassifier.html}}. On Compas, the AI model was trained to predict the risk of recidivism among four classes: `very low risk', `low risk', `high risk', and `very high risk'.  The original Obesity dataset considers seven weight categories which we simplified into four ordinal classes (to stay consistent with Compas): `underweight', `healthy', `overweight', and `obese'. In both cases, we trained the MLP on 70\% of the instances and evaluated it on the remaining 30\%. We obtained an accuracy of  67\% and 78\% on Compas and Obesity, respectively. Then, for each instance in the test set, we generated three different types of explanations: a feature-attribution explanation based on LIME~\cite{LIME}, a rule-based explanation based on Anchors~\cite{Anchor}, and a counter-factual explanation using the Growing Fields algorithm~\cite{Delaunay_cikm}. We set the default parameters for each of these methods except that (a) we changed the discretization routine in Anchors to take into account the improvements proposed in Chapter~\ref{chap: anchors}, and (b) we reported the importance of all features in the LIME explanation -- contrary to the default configuration that only picks the top 6. 
Moreover, while nothing guarantees that the sum of the coefficients of the linear model trained by LIME is consistent with the final probability prediction of the MLP classifier, we normalized the coefficients with the probability prediction to be more user-friendly.

For each dataset, we selected five target individuals in the test set to be presented to the user --- one for each of the four predicted classes plus an additional individual used as an example. Figure~\ref{fig:example_users} depicts how information about an individual is shown to the user for both datasets. The grey column represents the various features while the corresponding prisoner or patient data are in the second column. Code, datasets, and results are available on GitHub\footnote{\url{https://anonymous.4open.science/r/user_eval-1776}}.

\begin{figure}
    \centering
    \begin{subfigure}{.45\textwidth}
      \centering
      \includegraphics[width=0.9\textwidth]{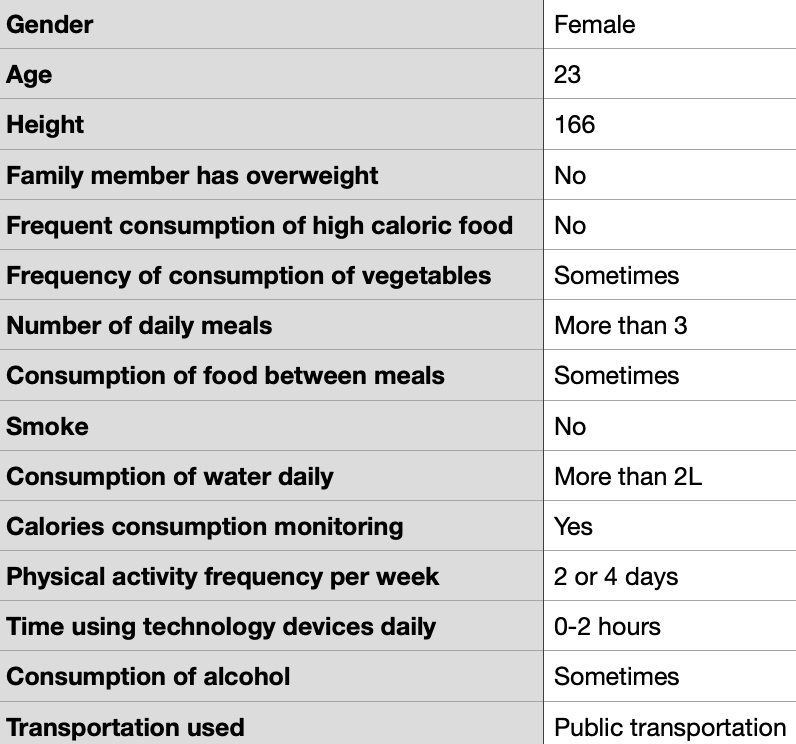}
      \label{fig:example_obesity}
    \end{subfigure}
    \begin{subfigure}{0.1\textwidth}
        
    \end{subfigure}
    \begin{subfigure}{.45\textwidth}
      \centering
      \includegraphics[width=0.9\textwidth]{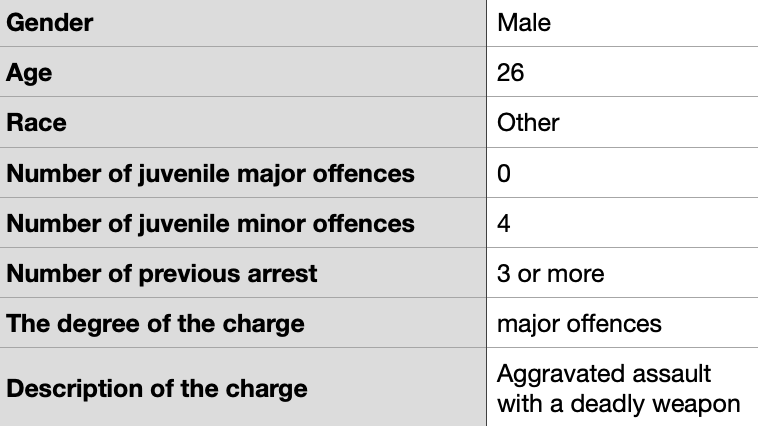}
      \label{fig:example_compas}
    \end{subfigure}
    \caption[Example of two individuals presented to the users for the Obesity and COMPAS datasets.]{Example of two individuals presented to the users for the Obesity (left) and COMPAS (right) datasets. The first column (in grey) represents the features description and the second column (in white) is the patient's or defendant's information.}
    \label{fig:example_users}
\end{figure}

\subsection{A Common Representation for Explanations}
\begin{figure}[ht]
    \centering
    \addtolength{\leftskip} {-1cm}
    \addtolength{\rightskip}{-1cm}
    \begin{tabular}{@{}lll@{}} 
        \toprule
        \textbf{Explanation technique} & \textbf{Graphical representation} & \textbf{Textal representation} \\ \midrule
        Feature-attribution & \cincludegraphics[width=5.9cm]{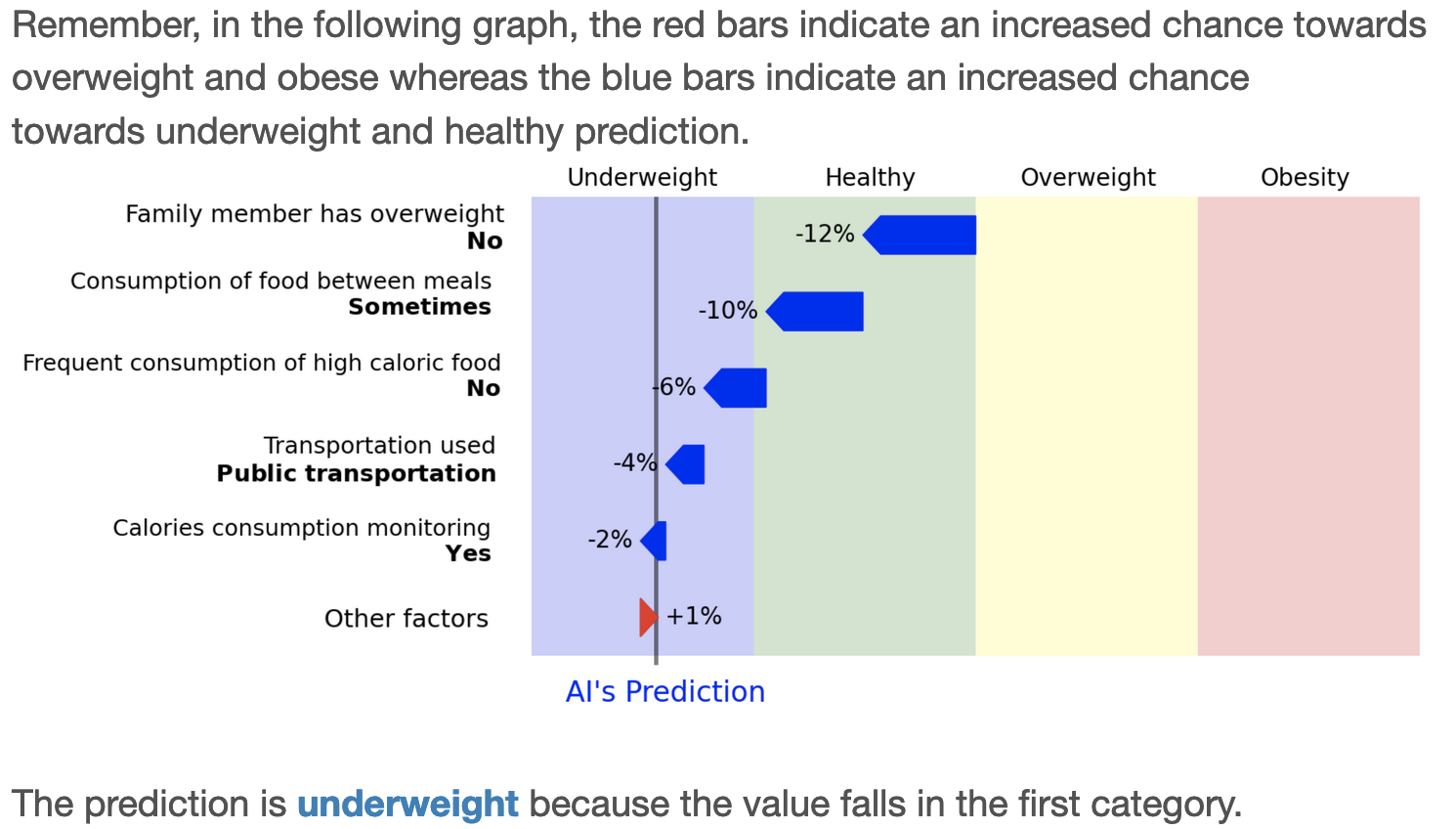} & \cincludegraphics[width=5.9cm]{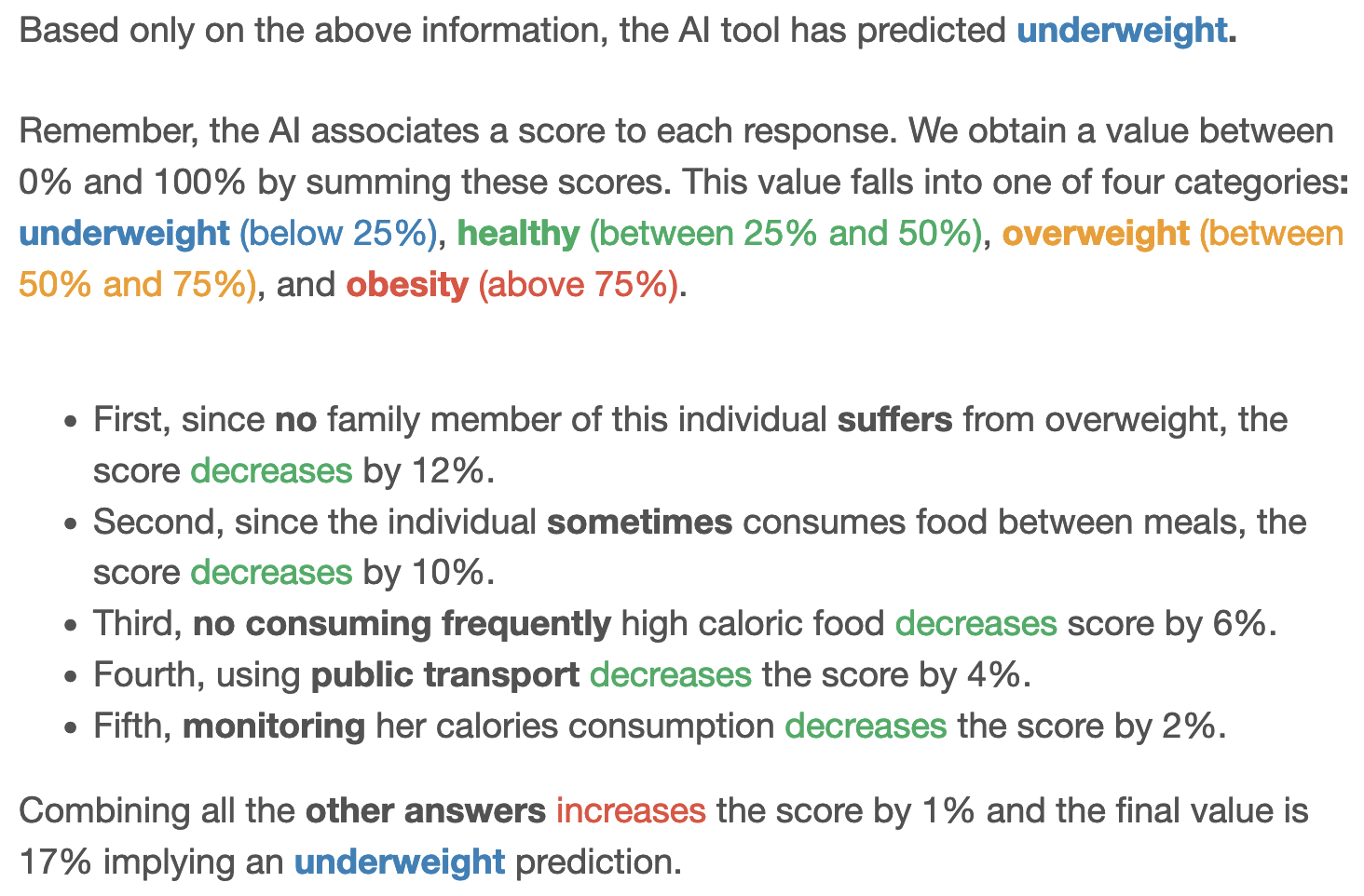} \\ \midrule
        Rule-based & \cincludegraphics[width=5.9cm]{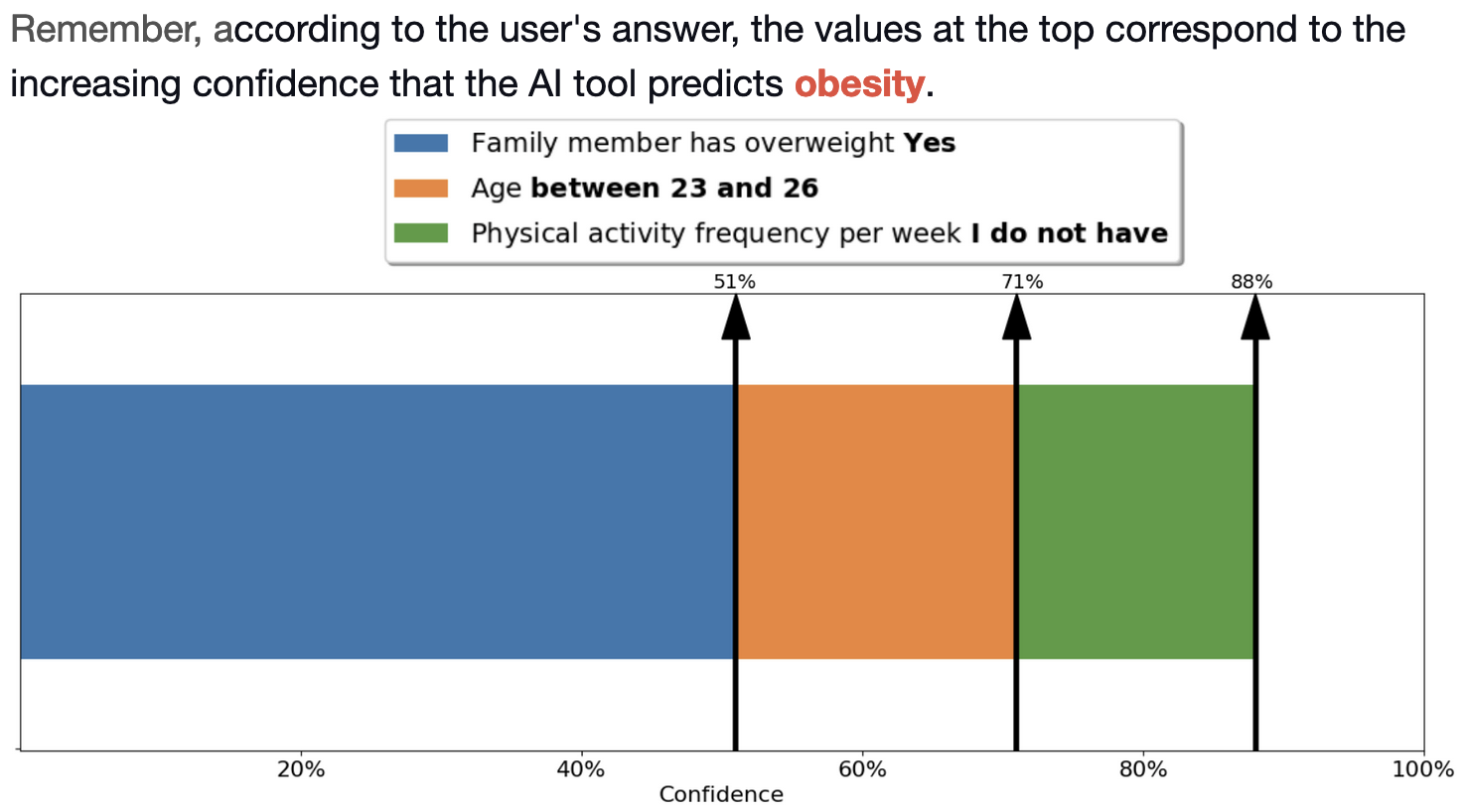} & \includegraphics[width=5.9cm]{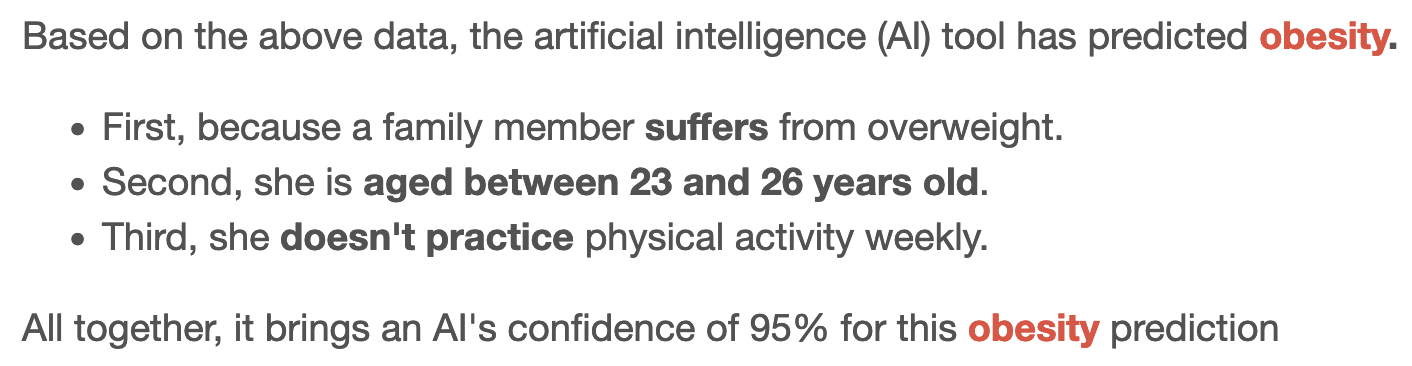} \\ \midrule
        Counterfactual & \cincludegraphics[width=5.9cm]{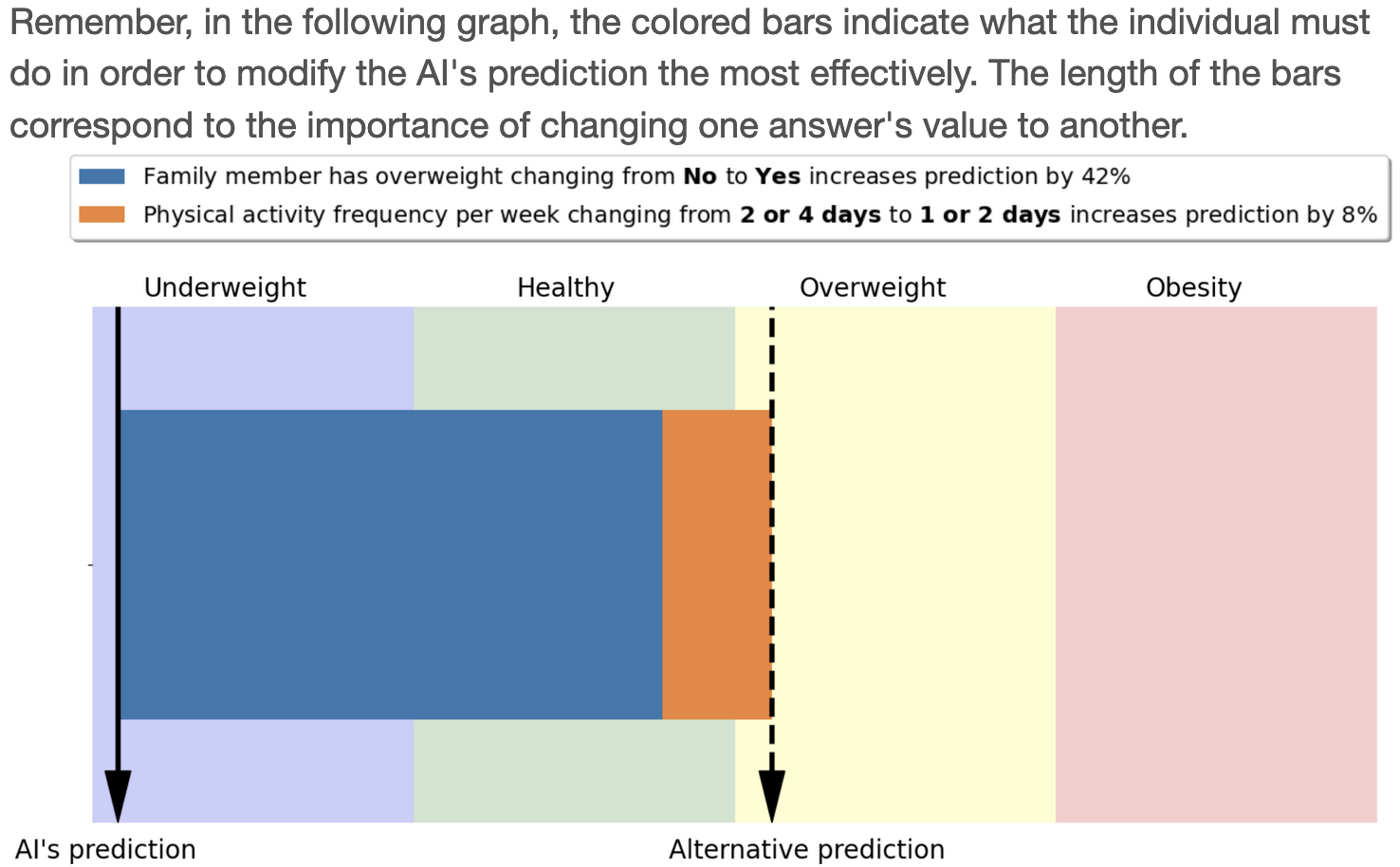} & \includegraphics[width=5.9cm]{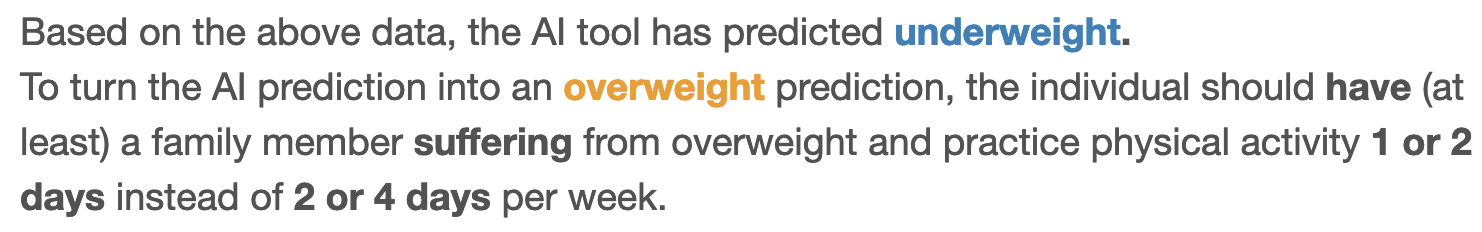} \\ \bottomrule
    \end{tabular}
    \caption{Different explanations for a given individual on the Obesity dataset, as split by explanation technique and representation.}
    \label{fig: visualization}   
\end{figure}
As highlighted in Section~\ref{subsec:evaluating_xai}, feature-attribution, rule-based, and counterfactual explanations reveal different aspects and insights about an AI's prediction process. Therefore, these explanation types are conveyed using different representations, which also depend on the AI agent's input data (e.g., image, text, etc.). When it comes to tabular data, existing XAI toolkits\footnote{AI360, Dalex, H2O, eli5, InterpretML, What-if-Tool, Alibi, Captum.} opt for a graphical representation based on bars for feature-attribution explanations -- as illustrated in Figure~\ref{fig: visualization}. Conversely, for rule-based and counterfactual explanations, the most common representation is natural language (see Figure~\ref{fig: visualization}). In order to control for this visual representation in our experiments, users are confronted with common graphical and textual representations for all the explanation types. 

\textbf{Graphical Representation.}
For each explanation method, we depict the graphical representation through diagrams. The explanation came along within a detailed paragraph reviewed by 20 different people---including 9 computer scientists and 11 laypeople---to verify its comprehensiveness and usefulness. As our AI model predicts four ordinal target outcomes, which range from underweight to obesity and no risk to high risk, we choose a common graphical representation that depicts the spectrum of classes on the x-axis and adds a different background color to the region covered by each of the classes.

\begin{itemize}
    \item As proposed by SHAP~\cite{SHAP} for feature-attribution explanations, the x-axis depicts the contribution of each feature to the predicted class in the form of a directed bar. The length of the bar depends on the magnitude of the attribution, whereas its direction tells towards which side of the spectrum the feature shifts the AI model's prediction (underweight vs. obese, low risk vs. high risk). Furthermore, SHAP computes the attribution score of a limited number of features to reduce the complexity of the explanation. Unlike SHAP, our representation groups features with a marginal attribution score under an artificial feature labeled `Other features'. The aggregated attribution of this label is the sum of the attribution scores of these features. Assuming a ranking of features based on the absolute value of the attribution scores, a feature is considered marginal as soon as the absolute value of its attribution score is less than half the absolute value of the score of the previous feature -- all subsequent features in the ranking are deemed marginal from that point. For instance, in Figure~\ref{fig: visualization}, the features that impact less than 2\% are grouped into the last bar and their cumulative attribution score equals 1\% toward the obesity class.   
    \item For rule-based explanations, we took inspiration from the representation proposed by Molnar~\cite{molnar}. This representation uses stacked bars as well, where each condition of the rule is assigned to a bar having a length proportional to the increase in confidence provided by the condition. As an example, imagine the explanation rule in Figure~\ref{fig: visualization}, stating that ``(a) having family antecedents of obesity, (b) an age between 23 and 26, (c) and practicing no physical activity'' incur an ``obese'' prediction with 90\% confidence. The blue bar tells that condition (a) on its own predicts obesity with 50\% confidence; conditions (a) and (b) increase the confidence to 71\%, and all three conditions increase the confidence to 90\%. 
    \item For counterfactual explanations we use stacked bars. Each feature in the explanation is associated with a bar. For each feature, the explanation incurs a change in the feature value, hence the length of each bar is proportional to the change incurred by the feature in the model's prediction. 
    In other words, the length of the bar tells us to what extent changing the feature's value shifts the black box answer from one predicted class to another -- the counterfactual class. For instance, the counterfactual explanation from Figure~\ref{fig: visualization} depicts that if the patient: ``(a) had family antecedents of obesity, and (b) practiced less often physical activity'' then the AI model would have predicted the patient as being ``overweight''. 
\end{itemize}

\textbf{Text Representation.} For all explanation types we present the explanation using a bulleted list, where each item describes the effect of each feature on the model answer. This effect can be an increase in the confidence of the prediction (for rule-based explanations), how much the feature contributes to the AI model prediction (feature-attribution explanation), or how sensitive is the AI model prediction in regards to the changes in the input features (counterfactual explanation). For feature-attribution explanations, we used colors to highlight the direction of the impact of each feature. Finally, we highlighted the instance's responses (obesity state, charge) by showing the text in bold. 

\section{Method}
\label{sec:method}
While the XAI community has proposed multiple post-hoc explanation methods that fall within the categories of feature attribution, rules, and example-based instances, no user studies have compared the effectiveness of these explanation styles. Hence, our first research question is \textbf{RQ1:} ``Which local explanation technique, \textit{i.e.,} feature-attribution, rule-based, or counterfactuals, provides the best explanations in terms of users' trust and comprehension of the AI model?''. Existing works have shown that explanations improve the users' ability to comprehend a model~\cite{LIME,edit_explanation}. Hence, this question underlies our first hypothesis; (H1) explanations improve the users' trust and understanding of a model. In addition, it has been suggested that explanations based on feature attribution may struggle to consistently assist users in understanding a model~\cite{microsoft_manipulating}. Conversely, decision rules have demonstrated high efficiency in helping users understand the inner mechanism of a model~\cite{Anchor,edit_explanation}. This leads to our second hypothesis; (H2) rule-based explanations improve users' comprehension of a model the most.
Finally, concerning trust, existing works have failed to show significant trust improvement when using feature-attribution~\cite{microsoft_manipulating}, as well as exemplars and rule-based explanation~\cite{van_der_waa}. We therefore follow a more explorative approach to studying the impact of the explanation technique on trust and do not propose a hypothesis on this aspect.

As suggested in~\cite{Niels_fairness} and~\cite{cheng_explanation_style}, the visual representation of an explanation can also impact the users' perception (e.g., trust, understanding, fairness). This leads to our second research question \textbf{RQ2:} ``Does the explanation's visual representation impact the users' trust and understanding?''. As there exists a general tendency to represent feature-attribution explanations graphically and rule-based as well as counterfactual explanations textually, our hypotheses are as follows: for feature-attribution techniques, a graphical representation is preferred (H3), whereas the most preferred explanation for rule-based and counterfactual is text-based (H4).

Therefore, our two independent variables are (i) the explanation style---feature-attribution, rule, and counterfactual, and (ii) the representation---graphical and textual. Conversely, the dependent variables are the users' understanding, satisfaction, and trust in the model.

\subsection{Task}
Our user studies consist of 14 online surveys in which participants are confronted with four prediction tasks. These tasks aim to predict either the risk of recidivism of a defendant given their profile or the risk of obesity of a person given some information about their habits (refer to Figure~\ref{fig:example_users}). To perform those predictions, users count on the recommendations of the AI models described in Section~\ref{subsec:data_and_ai}. 

These surveys were developed through the Qualtrics platform\footnote{\url{https://www.qualtrics.com/}}, with variations in the dataset (Compas and Obesity), explanation techniques (feature-attribution, rule-based, counterfactual) and representation methods (graphical vs. textual). For each dataset, we also created a control group in which no explanations were provided following the AI model's prediction. Furthermore, within the context of the three explanation techniques, two different visualizations were given. 

Figure~\ref{fig: user_study_standard} outlines the process followed by each of these surveys. With a specific dataset as the common ground, the only difference across the seven surveys is the explanation segment. In other words, except for the explanation, every step of the survey is exactly the same. Each survey is composed of three phases: (i) introduction, (ii) task round, and (iii) post-questionnaires as defined in Section~\ref{subsec: task}. The task asked to the users is to select based on the explanation, the features among the list of all possible features, that were used by the AI model to make its recommendation. 

\subsection{Scales \& Metrics (Illustration for One User)}
\label{subsec: metrics_illustration}
In this section, we provide a detailed example of how we employed the scales and metrics introduced in Section~\ref{subsec:metrics} for one user from the rule-based explanation group. This example is designed to provide the reader with a detailed explanation of how we assessed various facets of user behavior and perception. We recall that Figure~\ref{fig: user_study_standard} shows the times at which these parameters are measured. For this illustration, let us refer to this user as ``User J.'' User J participated in predicting the risk of obesity in response to four distinct scenarios, and their responses are reported in Figure~\ref{fig: example_metrics}. 
\begin{figure}[!h]
    \centering
    \addtolength{\leftskip} {-2.5cm}
    \addtolength{\rightskip}{-1cm}
    \includegraphics[width=1.25\textwidth]{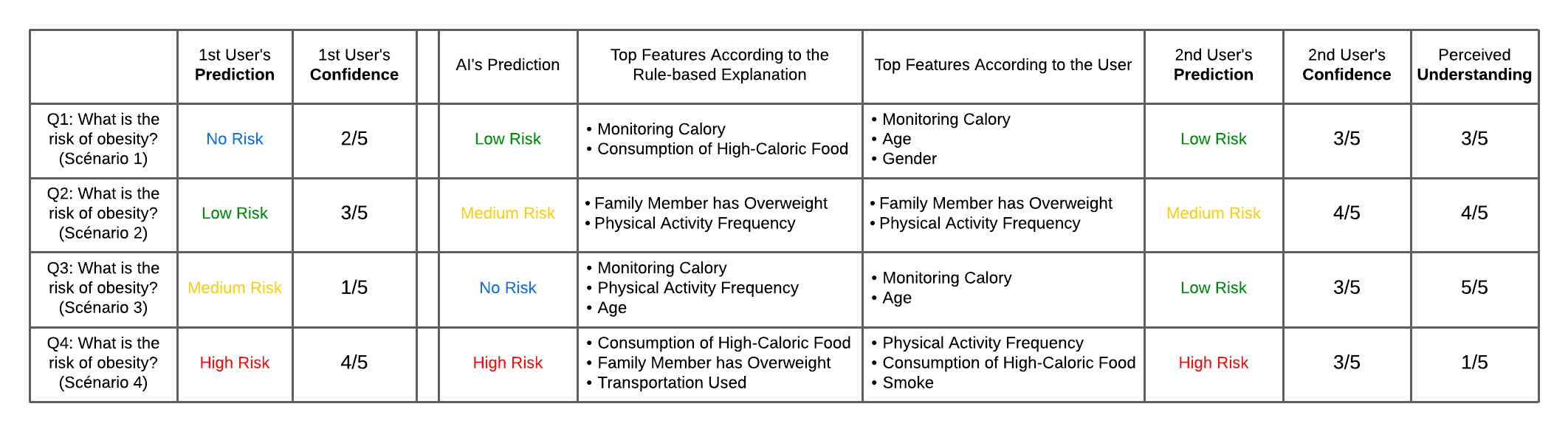}
    \caption[Example of answers from participant ``User J'' from the rule-based explanation group.]{Example of answers from participant ``User J'' from the rule-based explanation group. The values within the columns ``1st User's Confidence'', ``2nd User's Confidence'', and ``Perceived Understanding'' are on a 5-Likert scale.}
    \label{fig: example_metrics}
\end{figure}

\paragraph*{User's Initial Prediction and Confidence.} In Figure~\ref{fig: example_metrics}, User J's initial predictions, scaled from 1 (no risk) to 4 (high risk), are accompanied by their initial confidence levels, measured on a 5-point Likert scale. The Likert scale spans from ``strongly disagree'' to ``strongly agree.'' User J's initial predictions are shown in the ``1st User's Prediction'' column, and their initial confidence is recorded in the ``1st User's Confidence'' column.

\paragraph*{AI Model Predictions and Explanations.} User J's predictions are followed by the AI model's predictions and associated explanations, presented as depicted in Figure~\ref{fig: visualization}. These explanations comprise lists of the most influential features considered by the AI model for each prediction scenario. For example, in Figure~\ref{fig: visualization}, the most important features for the feature attribution are \textit{Family member has overweight}, \textit{Consumption of food between meals}, \textit{Consumption of high caloric food}, \textit{Transportation used}, and \textit{Calories consumption monitoring}. In contrast, for counterfactual, this is only the \textit{Family member has overweight} and \textit{Physical activity frequency} while rule-based also includes the \textit{Age} feature. 

\paragraph*{User's Final Prediction and Confidence.} During the task round, User J was asked to select, from the list of features, which features they considered most important for the AI model's prediction. Subsequently, User J was given the opportunity to reevaluate their prediction in the ``2nd User's Prediction'' column and provide their final confidence in their prediction in the ``2nd User's Confidence'' column.

\paragraph*{User's Perceived Understanding.} User J was also asked to rate their ``Perceived Understanding'' on a 5-point Likert scale to indicate their understanding of how the model made the prediction.

\paragraph*{Metrics Calculation.} The metrics for User J's responses were calculated as follows:
\begin{itemize}
    \item \textbf{$\Delta$-Confidence:} The $\Delta$-Confidence was computed by subtracting the initial confidence from the final confidence for each scenario. User J's $\Delta$-Confidence values are 1, 1, 2, and -1 for the four scenarios. The average $\Delta$-Confidence for User J is thus 3/4.
    \item \textbf{Behavioral Trust (Follow Pred.):} 
    We assessed behavioral trust by tracking instances where the user modified their initial prediction to match the AI model's prediction. It is important to note that we only considered scenarios where the user's initial prediction differed from the AI model's prediction. Thus, User J modified their initial prediction to align with the AI model's prediction in 2 out of 3 such scenarios, resulting in a behavioral trust score of 2/3.
    \item \textbf{Immediate Understanding:} User J's immediate understanding is the average value of their Likert-scale ratings for understanding across all four scenarios. In this case, it is (3 + 4 + 5 + 1) / 4, which equals 13/4.
    \item \textbf{Behavioral Understanding (Precision and Recall.):} To measure User J's precision and recall, we compared the list of features they identified as important to those highlighted in the explanation for each scenario. The precision and recall values for each scenario were calculated as follows:
\end{itemize}
\begin{enumerate}[label= \textbf{Scenario Q\arabic*:}]
    \item 
    \begin{itemize}
        \item Precision = 1/3 (User identified three features, one matched AI explanation),
        \item Recall = 1/2.
    \end{itemize} 
    \item 
    \begin{itemize}
        \item Precision = 1 (User and AI explanation lists are identical),
        \item Recall = 1.
    \end{itemize} 
    \item 
    \begin{itemize}
        \item Precision = 1 (User identified 2 features, both matched AI explanation),
        \item Recall = 2/3.
    \end{itemize}
    \item 
    \begin{itemize}
        \item Precision = 1/3 (User identified 1 feature, which matched AI explanation),
        \item Recall = 1/3.
    \end{itemize}
\end{enumerate}
Please note that these are simplified examples, and in practice, the lists of important features in explanations are typically longer.


\begin{figure}[!h]
    \centering
    \addtolength{\leftskip} {-2.5cm}
    \addtolength{\rightskip}{-1cm}
    \includegraphics[width=1.25\textwidth]{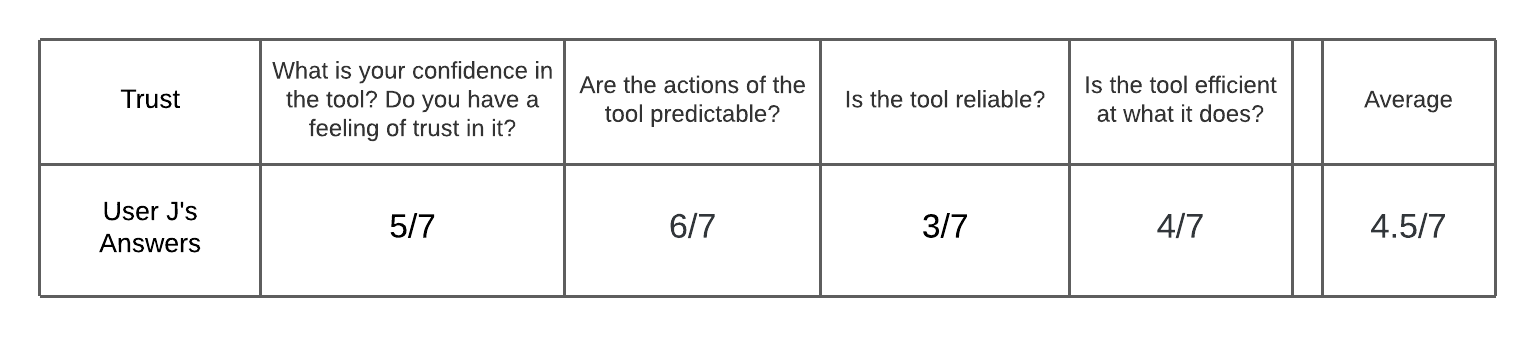}
    \caption[Example of answers from one participant to the Trust survey]{Example of answers from one participant to the Trust survey adapted from Cahour and Forzy~\cite{cahour-forzy}. We measure the users' perceived trust in the AI system on a scale from 1 to 7.}
    \label{fig: example_survey}
\end{figure}

\paragraph*{Post-Questionnaires.} In Figure~\ref{fig: example_survey}, we present an example of a survey measuring User J's perceived trust in the AI system. This survey was adapted from Cahour and Forzy~\cite{cahour-forzy} and employed a Likert scale ranging from 1 to 7. The average of User J's responses to the four survey questions provides a representation of their perceived trust, which, in this case, is 4.5 out of 7. Similar procedures were followed for the understanding and satisfaction questionnaires, each consisting of eight questions rated on a five-point Likert scale.

\subsection{Participants}
We recruited participants through the Prolific Academic platform. First, we restricted participation to crowd-workers with at least a high school degree to guarantee a reasonable response quality. Second, we decided not to limit ourselves to a particular geographical location to promote diversity in our sample. Finally, we ensured that participants could participate only once in our study to avoid situations where a participant is in two groups (e.g., control and feature-attribution). After accepting the task, participants were redirected to the corresponding Qualtrics survey, starting with a short introduction to the algorithm and the dataset. Based on a pilot evaluation with 20 people, we estimated a completion time of 20 minutes for non-control groups and 15 minutes for the control group. We remind the reader that the control group has no explanation to extend the AI prediction, therefore, these participants should be faster at filling out the survey. Participants were paid £9.30 per hour, which translated into a payment of £2.25 for the control-group participants, and £3.10 for the non-control group.

To limit Type II errors and not reject the null hypothesis when it is false, we determined the number of respondents on the basis of a power calculation using G*Power~\cite{gpower}. Given the exploratory nature of our investigation, we used medium-to-large effect sizes ($f^2 = 0.2$), an alpha level of 0.05, and a power of 0.8, in line with established methodological recommendations~\cite{gpower_advise}. Based on our a priori multiple linear regression model with two predictors, the required minimum group size is 107 participants. For the sake of precaution and to maintain consistency, we finally recruited 280 participants -- 140 participants per dataset, or 20 participants per combination of explanation technique and visual representation. 

\begin{table}[ht!]
    \centering 
    $$ 
    \begin{aligned} 
    & \begin{array}{@{}lcccc@{}} 
    \toprule \textbf{Domain} & \multicolumn{2}{c}{\textbf{Healthcare}}  & \multicolumn{2}{c}{\textbf{Law}} \\
       \cmidrule(r){2-3}   \cmidrule(r){4-5}
       \text{Factor} & \boldsymbol{N} & \text{\% sample} & \boldsymbol{N} & \text{\% sample} \\\midrule 
       \textbf{Gender} & & \\
       \text{Female} & 66 & 47.14 & 66 & 47.14  \\
       \text{Male} & 62 & 44.29 & 74 & 52.86  \\
       \text{Prefer not to say} & 1 & 0.71 & 0 & 0.0 \\
       \midrule
       \textbf{\text{Consent revoked}} & 11 & 7.86 & 0 & 0.0 \\
       \midrule
       \textbf{Age} & & \\    
       \text{< 20} & 10 & 7.14 & 11 & 7.86  \\
       \text{20 < 30} & 81 & 57.86  & 88 & 62.86 \\
       \text{30 < 40} & 24 & 17.14  & 27 & 19.29   \\
       \text{40 >} & 14 & 10.0 & 14 & 10.0  \\
       \midrule
       \textbf{Nationality} & & \\
       \text{Africa} & 45 & 32.14 & 37 & 26.43   \\
       \text{Asia} & 2 & 1.43  & 2 & 1.43   \\     
       \text{Australia} & 0 & 0.0 & 1 & 0.71  \\        
       \text{Europe} & 77 & 55.0  & 82 & 58.57  \\
       \text{North America} & 5 & 3.57  & 15 & 10.71  \\
       \text{South America} & 0 & 0.0 & 3 & 2.14  \\
       \midrule
       \textbf{Ethnicity (simplified)} & & \\
       \text{Asian} & 2 & 1.43 & 2 & 1.43  \\
       \text{Black} & 37 & 26.43  & 30 & 21.43  \\
       \text{Mixed} & 10 & 7.14  & 9 & 6.43  \\
       \text{Other} & 3 & 2.14  & 8 & 5.71  \\
        \text{White} & 77 & 55.0 & 91 & 65.0   \\
        \midrule
       \textbf{Highest education} & & \\
       \text{Doctorate degree} & 3 & 2.14 & 1 & 0.71  \\
       \text{Graduate degree} & 27 & 19.29 & 24 & 17.14  \\
       \text{High school diploma} & 47 & 33.57 & 37 & 26.43  \\
       \text{Technical college} & 3 & 2.14  & 14 & 10.0  \\
       \text{Undergraduate degree} & 49 & 35.0 & 64 & 45.71  \\
       \bottomrule
   \end{array}\\
    \end{aligned} $$ 
    \caption{Overview of participants' demographic factors.} 
    \label{tab:user_information}
\end{table}

Table~\ref{tab:user_information} presents the demographic information about our participants. Our study follows a between-subject design, meaning that each participant interacts with one representation and one surrogate model.

Following the introduction of the task, we assessed whether the participants had actually read and comprehended the basic information presented through two questions: `How is Body Mass Index calculated?' for the obesity domain and `Why is recidivism risk calculated?' for the recidivism dataset. We found 10 incorrect answers for the first question and 30 participants who responded erroneously for the second question. This question had the form `The algorithm calculates the risk of obesity (resp. recidivism) for an individual by;'. We asked additional users to participate in our study until we had 20 responses for each group that validated our two understanding questions. This resulted in a final set of 280 participants.

\section{Results}
\label{sec:results}
We present our findings in four sections. We begin by examining the impact of the application domain (\textit{i.e.}, dataset), explanation technique, and representation on users' understanding in Section~\ref{subsec: understanding}. Then, we assess the influence of these factors on users' trust in the AI agent in Section~\ref{subsec:trust}. Additional results concerning satisfaction and completion time are discussed in Section~\ref{subsec:satisfaction}. Subsequently, in Section~\ref{subsec:correlation}, we explore the correlation between behavioral and perceived measurements. We conclude with Section~\ref{subsec:open_q}, where we present an in-depth qualitative study of the users' perception based on the answers to the open questions. For a more comprehensive understanding of our study, including code, participants' comments, survey details, and analysis notebook, these resources are available on GitHub\footnote{\url{https://anonymous.4open.science/r/user_eval-1776}}.

\begin{table*}[ht]
    \centering
    \addtolength{\leftskip} {-1cm}
    \addtolength{\rightskip}{-1cm}
    \begin{tabular}{@{}lcccccccc@{}}
        \toprule & \multicolumn{8}{c}{\textbf{Understanding}}  \\
        \cmidrule(r){2-9}
        & \multicolumn{4}{c}{\textbf{Recidivism}} & \multicolumn{4}{c}{\textbf{Obesity}} \\
        \cmidrule(r){2-5} \cmidrule(r){6-9} 
        & \multicolumn{2}{c}{Self Report} & \multicolumn{2}{c}{Behavioral} & \multicolumn{2}{c}{Self Report} & \multicolumn{2}{c}{Behavioral}\\
        \cmidrule(r){2-3} \cmidrule(r){4-5} \cmidrule(r){6-7} \cmidrule(r){8-9} 
        & Immediate & Survey & Prec. & Rec. & Immediate & Survey & Prec. & Rec. \\
        Explanation Technique & 0.87  & 1.20 & $16.24^{***}$ & 1.58 & $3.75^*$  & 1.35 & $31.42^{***}$ & $6.37^{***}$ \\
        Representation & 0.96 & 0.36 & 0.13 & $3.00^-$ & 0.14 & 0.55 & 0.05 & $2.85^-$\\
        Age & 1.07 & 0.01 & 1.88 & 0.10 & 0.16 & 0.06 & $6.41^*$ & 0.02\\
        Education & 1.63 & 0.93 & 0.94 & 0.43 & 0.50 & 0.34 & 0.25 & 1.31\\
        Gender & 0.54 & 1.07 & 0.35 & 0.30 & 0.14 & 0.03 & 0.18 & 0.36 \\
        Technique:Representation & 0.28 & 0.87 & 1.12 & 0.74 & 0.48 & 0.16 & 0.35 & $4.99^{**}$ \\
        \bottomrule 
        \multicolumn{5}{l}{\begin{small} $ ^{* * *} p<0.001,{ }^{* *} p<0.01,{ }^* p<0.05, { }^- p<0.1$ \end{small}} &
    \end{tabular}
    \caption[F value of the ANOVA Table with understanding measurements grouped by domain and self-reported and behavioral metrics.]{F value of the ANOVA Table with understanding measurements grouped by domain and self-reported and behavioral metrics. `Immediate' corresponds to the perceived comprehension when facing the explanation while `Survey' represents the perceived understanding measured through the post questionnaire. `Prec.' and `Rec.' are respectively the Precision and Recall between the features indicated by the participant and the explanation technique. `Technique:Representation' denotes the interaction between the explanation technique and the visual representation. }
    \label{tab: understanding_regression_coef}
\end{table*}

\subsection{Understanding}
\label{subsec: understanding}
To discern the factors that impact users' understanding of the AI agents, we fit a linear model and conduct an ANOVA analysis for each application domain (Recidivism and Obesity). The linear model incorporated six predictors to estimate four metrics. Those predictors included demographic data (age, gender, education level), along with explanation technique and visual representation. These predictors were categorized to construct the model. The ANOVA f-scores of each predictor and target metric can be found in Table~\ref{tab: understanding_regression_coef}. We show plots only for results that are considered statistically significant according to the ANOVA analysis. The target metrics are elaborated upon in Section~\ref{subsec:metrics}.

\begin{figure}[ht]
    \centering
    \includegraphics[width=0.85\textwidth]{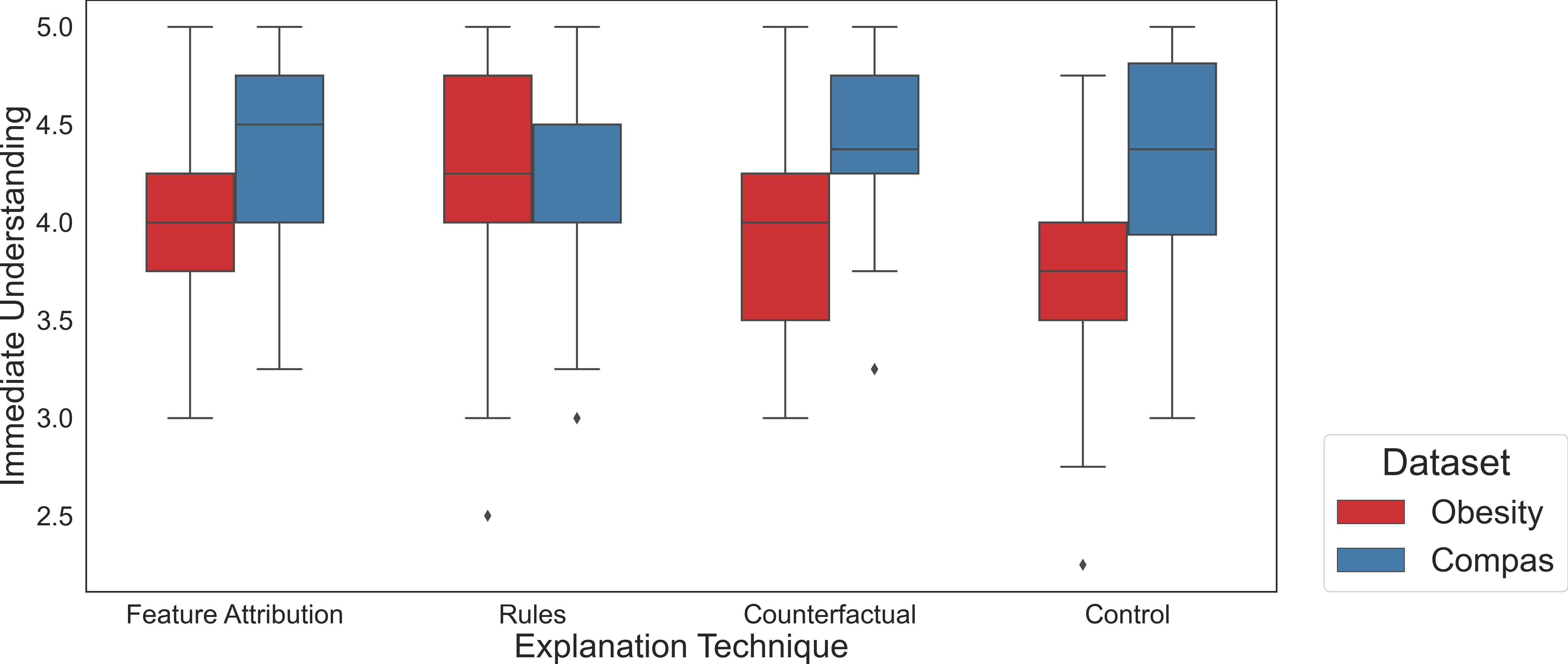}
    \caption{Perceived understanding of the users \textbf{(Immediate Und.)} for both the Obesity and Recidivism datasets based on the explanation technique.}
    \label{fig: perceived}
\end{figure}

We first observe that the users' self-reported understanding of the AI system---based on a post questionnaire \textbf{(Survey)}---does not vary across the different explanation techniques, visual representations, and demographic categories. These observations hold both for the recidivism and obesity datasets.

Conversely, we note that the impact of the chosen explanation technique on the users' perceived understanding when exposed to explanations \textbf{(Immediate Und.)} is statistically significant \textit{(p<0.05)} for the Obesity dataset. Figure~\ref{fig: perceived} depicts the users' perceived understanding of the AI system across the explanation methods for both domains. While statistically significant differences were not found for the recidivism domain, users confronted with rule-based explanations in the obesity domain report a better understanding of the AI model. Notably, on the obesity domain, the third quartile of the Immediate understanding for the rule-based group was around 4.7 (on a scale of 1 to 5), compared to approximately 4.2 for the counterfactual and feature-attribution groups. In contrast, the control group (without explanation) exhibited a lower self-reported understanding of the AI model (around 4). 

\begin{figure}
    \centering
    \includegraphics[width=0.85\textwidth]{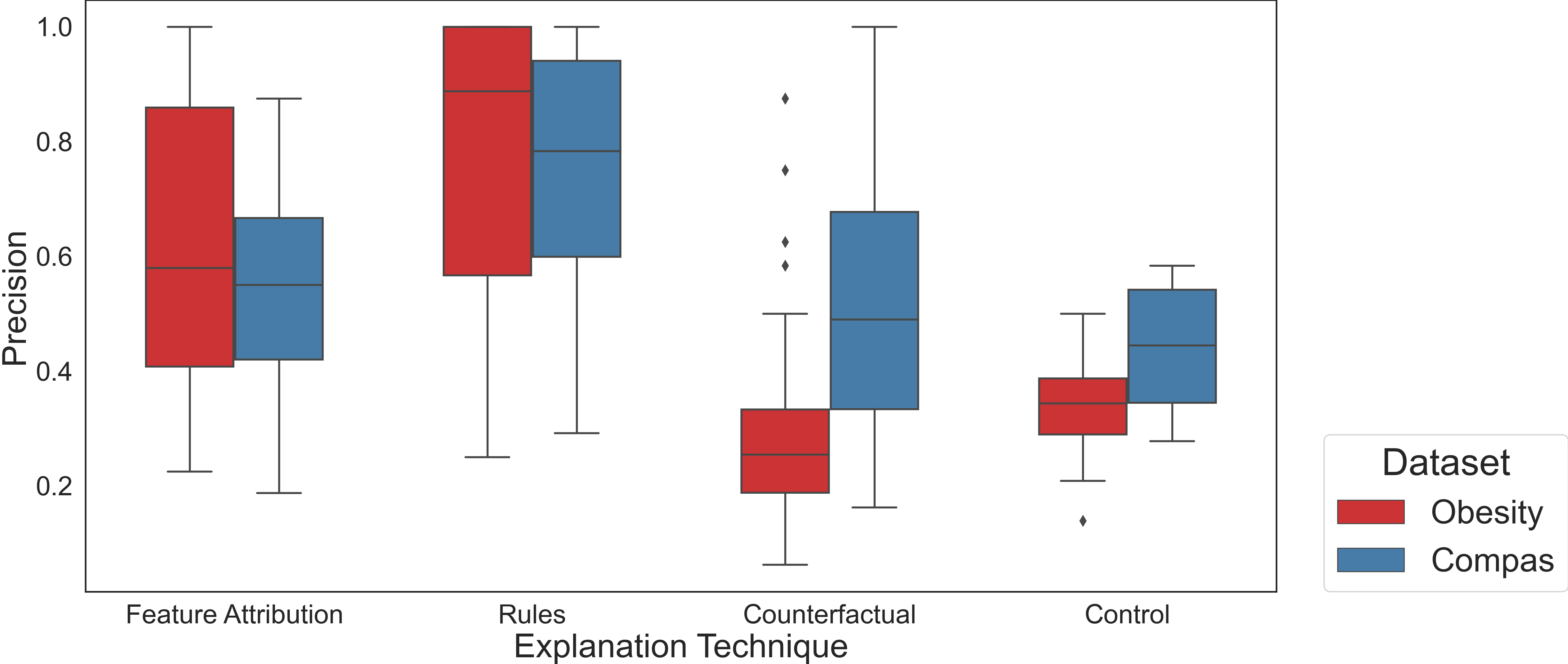}
    \caption[Behavioral precision between the features indicated as important by the users for the AI's prediction and the important features indicated in the explanation.]{Behavioral precision between the features indicated as important by the users for the AI's prediction and the important features indicated in the explanation. Results are shown for the two domains, the tree explanation methods, and the control group.}
    \label{fig:b_precision}
\end{figure}
\begin{figure}
    \centering
    \includegraphics[width=0.85\textwidth]{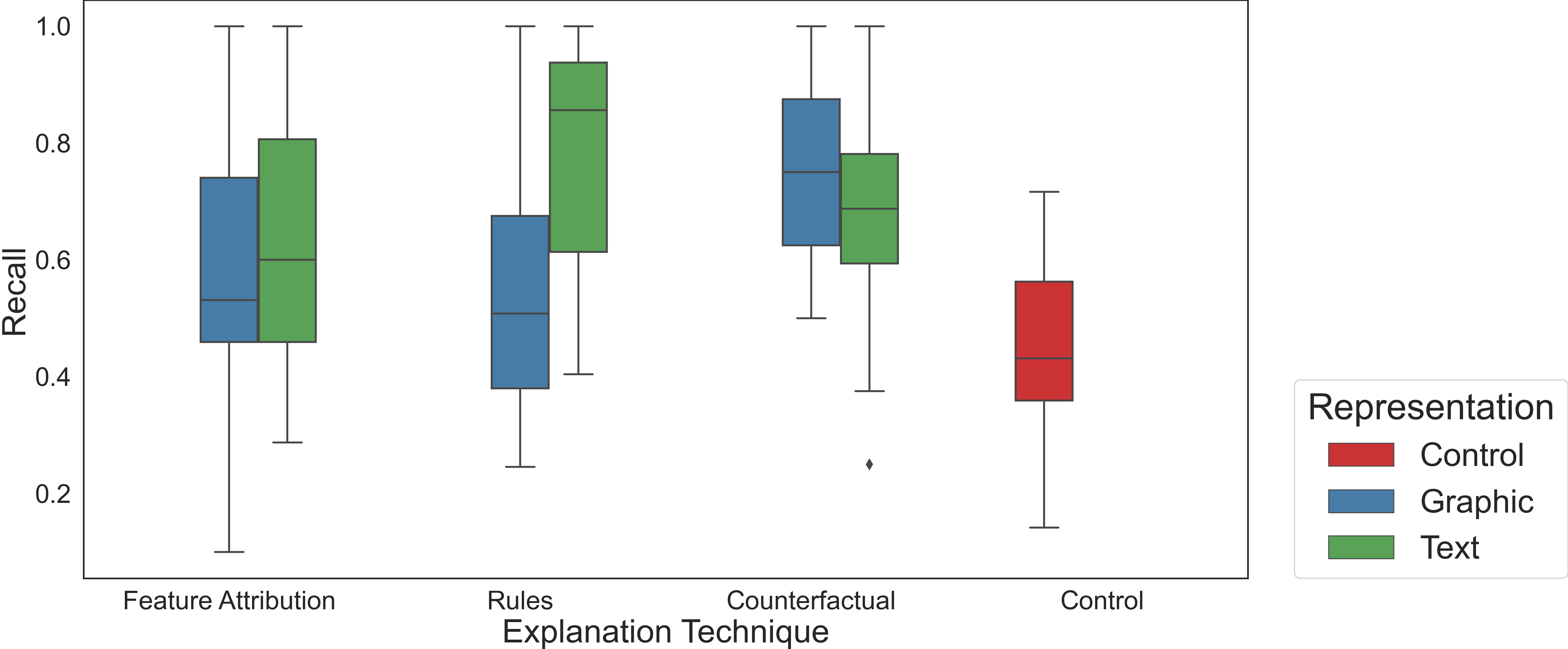}
    \caption{Users' understanding for each explanation technique and representation, computed as the users' recall of the important features according to the explanations in the Obesity dataset.}
    \label{fig:b_recall}
\end{figure}

We then gauge the behavioral understanding through the precision \textbf{(Prec.)} and recall \textbf{(Rec.)}. These scores measure the alignment between the features identified as important by the explanation and features marked as important by the user for prediction. Notably, it might seem straightforward for a researcher in interpretability to identify the features highlighted as significant by the explanation. However, our findings demonstrate that this is not the case for laypersons. The fact that precision and recall values are far from 1 suggests two aspects. Firstly, participants might exhibit a certain level of distrust in the explanation, leading them to choose features they personally consider impactful. Secondly, they could anticipate a higher degree of complexity, which prompts them to indicate additional features as important. 
Table~\ref{tab: understanding_regression_coef} highlights that precision is significantly affected by the explanation method for both domains \textit{(p<0.01)}. Figure~\ref{fig:b_precision} depicts the precision across domains and explanation methods, revealing that rule-based explanations yield the highest precision score in the healthcare domain (median precision of 0.9). On the contrary, counterfactual explanations resulted in poor performances comparable to the control group (precision 0.3). Lastly, feature-attribution explanations fall in between, with median precision scores above 0.5 for both domains. We also note that according to the ANOVA table, the age of the participant significantly impacts the precision in the Obesity dataset.

We also observe in Table~\ref{tab: understanding_regression_coef} that the explanation technique and the visual representation exhibit a statistically significant impact on recall for the healthcare task. Figure~\ref{fig:b_recall} shows the recall of the participants per explanation technique and grouped by representation on the obesity domain. This suggests that for feature-attribution and rule-based explanations, users facing a textual representation are better at selecting important features for the classification than the participants facing a visual presentation. In contrast, users with counterfactual explanations are better at identifying important features when reading a graphical representation than a text representation. We finally note that participants who had an explanation actually achieved a higher recall than those in the control group.

\begin{table}[ht]
    \centering
    \addtolength{\leftskip}{-1cm}
    \addtolength{\rightskip}{-1cm}
    \begin{tabular}{@{}lcccccc@{}}
        \toprule & \multicolumn{6}{c}{\textbf{Trust}}  \\
        \cmidrule(r){2-7}
        & \multicolumn{3}{c}{\textbf{Recidivism}} & \multicolumn{3}{c}{\textbf{Obesity}} \\
        \cmidrule(r){2-4} \cmidrule(r){5-7}  
        & \multicolumn{2}{c}{Self Reported} & \multicolumn{1}{c}{Behavioral} & \multicolumn{2}{c}{Self Reported} & \multicolumn{1}{c}{Behavioral}\\
        \cmidrule(r){2-3} \cmidrule(r){4-4} \cmidrule(r){5-6} \cmidrule(r){7-7} 
        & {\small $\Delta$ Confidence} & Survey & {\small Follow Pred.} & {\small $\Delta$ Confidence} & Survey & {\small Follow Pred.} \\
        Explanation Technique & 1.40 & 0.03 & 0.78 & 0.12 & 0.42  & 0.38 \\
        Representation & 0.04 & 0.32 & 0.00 & 8.22$^{**}$ & 0.55 & 0.12 \\
        Age & 0.46 & 0.18 & 2.76$^-$ & 0.06 & 0.70 & 0.00 \\
        Education & 0.13 & 1.82 & 0.34 & 2.14$^-$ & 0.69 & 0.63 \\
        Gender & 2.16 & 1.35 & 0.31 & 0.12 & 2.32 & 1.11\\
        Technique:Representation & 0.35 & 1.23 & 0.75 & 0.26 & 0.23 & $3.55^*$ \\
        \bottomrule 
        \multicolumn{5}{l}{\begin{small} $ ^{* * *} p<0.001,{ }^{* *} p<0.01,{ }^* p<0.05, { }^- p<0.1$ \end{small}} &
    \end{tabular}
    \caption[F value of the ANOVA Table with trust measurements grouped by domain and by self-reported and behavioral metrics.]{F value of the ANOVA Table with trust measurements grouped by domain and by self-reported and behavioral metrics. The column `$\Delta$ Confidence' is the difference between the self-reported trust before and after facing the AI's prediction. `Survey' corresponds to the perceived trust in the model assessed through the post-questionnaire. `Follow Pred.' is the proportion of times the participants changed their prediction to follow the AI's. `Technique:Representation' refers to the interaction between the explanation technique and the visual representation.}
    \label{tab:trust_regression_coef}
\end{table}

\subsection{Trust}
\label{subsec:trust}
We then measure the impact of the different factors and domains on the users' trust in the AI system. We proceed similarly as in the evaluation of the user's comprehension. That is, we fitted a linear model on our observations with the same set of predictors and the three metrics defined in Section~\ref{subsec:metrics}. We measured the corresponding f-value and reported it in Table~\ref{tab:trust_regression_coef}. 

We first analyze the users' self-reported trust in the AI model's prediction through the post questionnaire \textbf{(Survey)}. Table~\ref{tab:trust_regression_coef} reveals that the perceived trust in the AI system does not show statistically significant differences across the different explanation techniques and visual representations.

\begin{figure}
    \centering
    \includegraphics[width=0.85\textwidth]{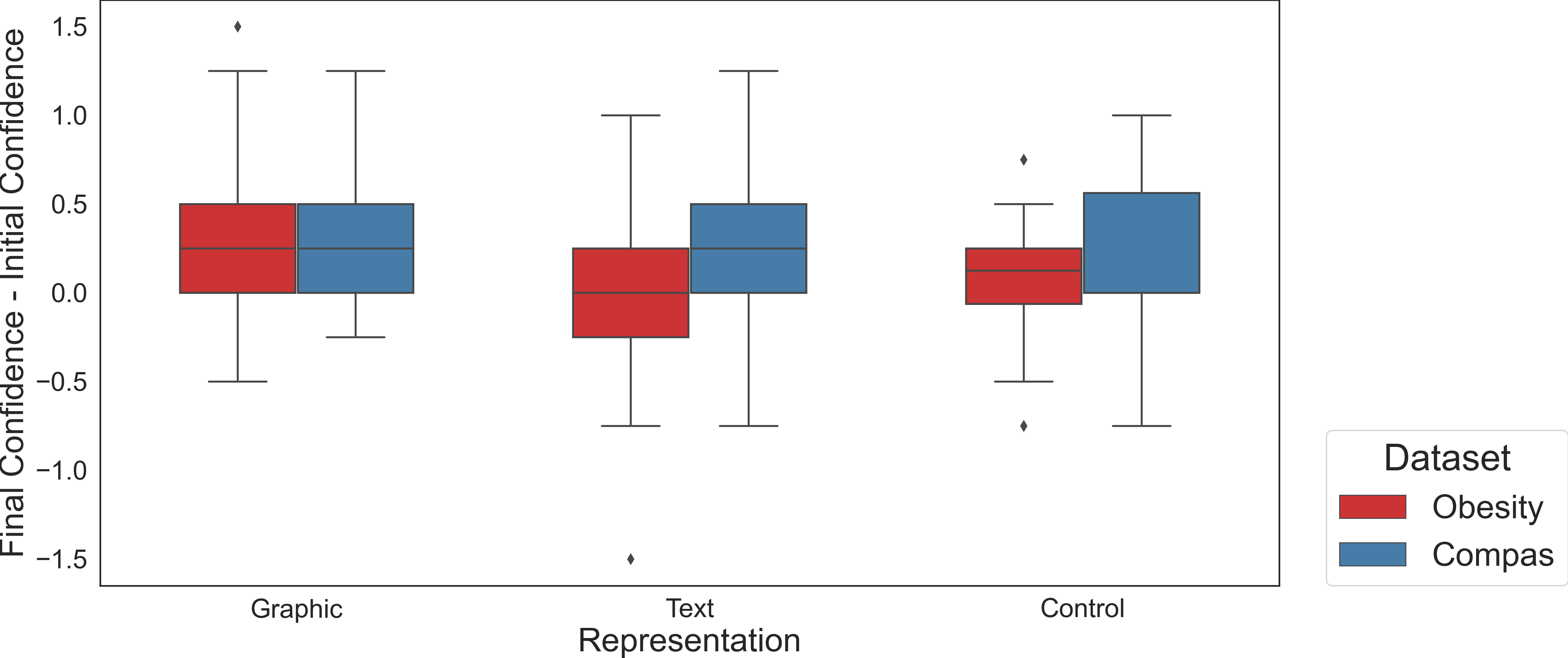}
    \caption[Difference between the self-reported confidence in the users' prediction after and before seeing the AI's prediction and explanation (when provided).]{Difference between the self-reported confidence in the users' prediction after and before seeing the AI's prediction and explanation (when provided). Results are shown for the control group and for each dataset and representation.}
    \label{fig:increased_trust}
\end{figure}

The results in Table~\ref{tab:trust_regression_coef} also suggest that the explanation visual representation has an impact on the difference in self-reported trust before and after seeing the explanation \textbf{($\Delta$ Confidence)}. This impact is statistically significant with a p-value inferior to 0.01 for the Obesity dataset. It is noteworthy that, 56\% of the users' initial predictions aligned with the AI model's prediction for the Recidivism dataset and 39\% for the Obesity dataset. Thus, we limit our evaluation of self-reported trust to scenarios where participants had to adjust their initial predictions. This decision was based on the understanding that assessing behavioral confidence is relevant when participants are prompted to reconsider their own predictions because the AI predicted a different class. Figure~\ref{fig:increased_trust} indicates that for the Obesity dataset, participants exposed to a graphical representation reported an increase in post-explanation trust in their prediction. 

Additionally, we conducted a detailed examination of the participants' perceived confidence splitting participants into two groups. These groups are based on whether their initial prediction matched the AI model's prediction. We conducted this in-depth analysis because participants' confidence in the AI system could be influenced by the matching between their predictions and those of the AI system. This analysis revealed that, in the Obesity dataset, participants with higher educational levels and who initially disagree with the AI model's prediction experienced a decrease in confidence. Conversely, for participants in the Compas dataset, we observed that when the AI model confirmed female participants' prediction, their confidence increased less compared to male participants.

\begin{figure}
    \centering
    \includegraphics[width=0.85\textwidth]{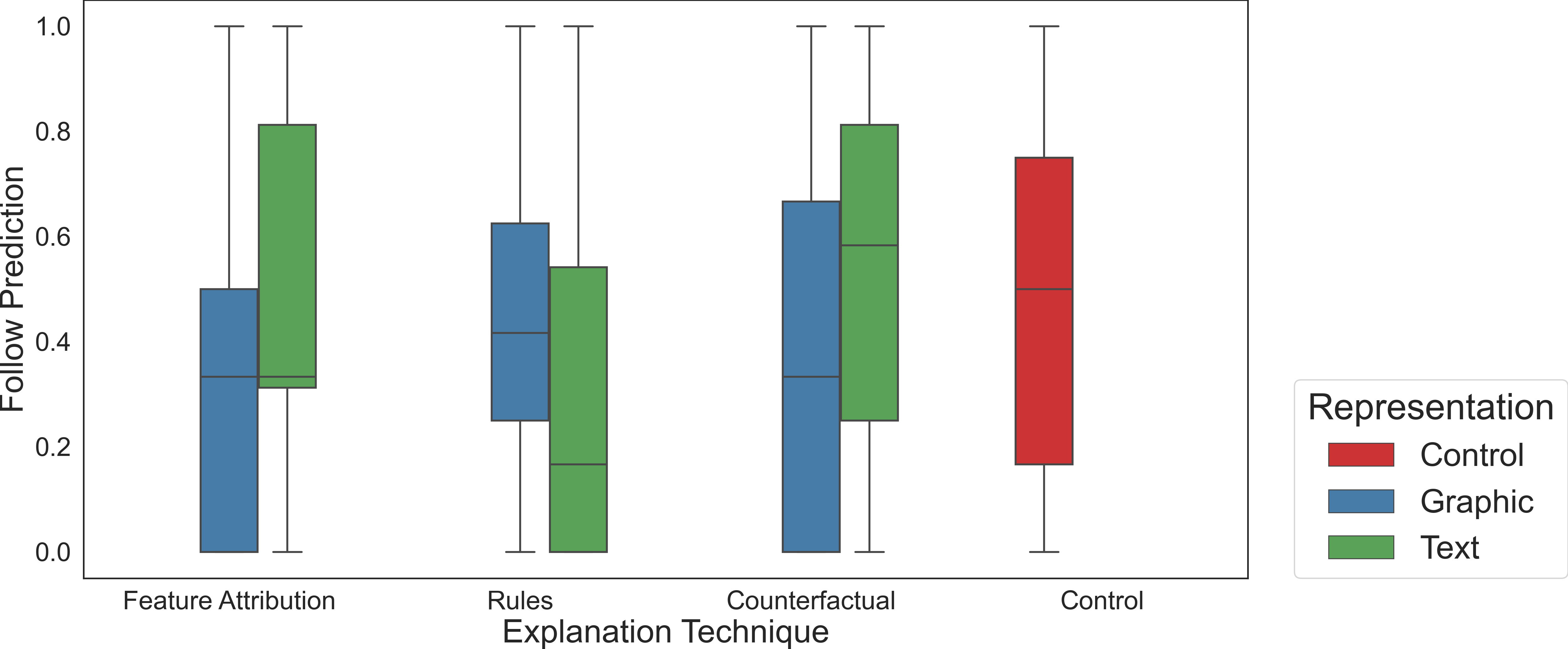}
    \caption[Proportion of cases the participants changed their initial prediction to follow the AI's prediction.]{Proportion of cases the participants changed their initial prediction to follow the AI's prediction. Results are shown for the Obesity dataset on the combination of explanation technique and representation.}
    \label{fig:b_trust}
\end{figure}

Finally, we find a significant interaction (\textit{p<0.05}) between explanation technique and visual representation on participants' behavioral trust, \textit{i.e.}, users who changed their initial prediction to match the AI agent's \textbf{(Follow Pred.)} for the Obesity dataset. The average users' behavioral trust for different explanation methods and representation in the healthcare domain is depicted in Figure~\ref{fig:b_trust}. We observe that users without explanations or with counterfactual explanations are more prone to follow the prediction of the AI system. This suggests that users with feature-attribution and rule-based explanations have lower confidence in the model's prediction.

\begin{table}[!ht]
    \centering
    \begin{tabular}{@{}lcccc@{}}
        \toprule 
        & \multicolumn{2}{c}{\textbf{Recidivism}} & \multicolumn{2}{c}{\textbf{Obesity}} \\
        \cmidrule(r){2-3} \cmidrule(r){4-5}  
        & Time & Survey Sat. & Time & Survey Sat.\\
        Explanation Technique & 2.49$^-$ & 0.39  & 0.78 & 1.60\\
        Representation & 1.04 & 0.00 & 0.88 & 0.02 \\
        Age & 1.76 & 1.08 & 8.97$^{**}$ & 0.09 \\
        Education & 1.31 & 1.75 & 2.07$^-$ & 1.73 \\
        Gender & 2.14 & 0.05 & 0.03 & 1.52 \\
        Technique:Representation & 0.20 & 1.60 & 0.38 & 1.49 \\
        \bottomrule 
        \multicolumn{5}{l}{\begin{small} $ ^{* * *} p<0.001,{ }^{* *} p<0.01,{ }^* p<0.05, { }^- p<0.1$ \end{small}}
    \end{tabular}
    \caption[F value of the ANOVA table with additional measurements.]{F value of the ANOVA table with additional measurements. The `Survey Sat.' column is the self-reported satisfaction measured with a post-questionnaire. The column `Time' represents the time required to complete the task round. `Technique:Representation' is the interaction between the explanation technique and the visual representation.}
    \label{tab:other_regression_coef}
\end{table}

\subsection{Additional Measurements}
\label{subsec:satisfaction}
As stated in Section~\ref{subsec:metrics} we also evaluate perceived satisfaction and completion time. First, we measure the participants' self-reported satisfaction through the Explanation Satisfaction questionnaire \textbf{(Satis.)}. Second, we measure the completion time required by the user to interpret the explanation, that is, the time to indicate the most important features for the prediction \textbf{(Time)}. We fitted a linear model with six factors to predict these two metrics for both domains and report the f-score in Table~\ref{tab:other_regression_coef}. 

This table shows that self-reported satisfaction in the AI system is not impacted by the presence of explanations or by the demographic attributes of the participant. Conversely, we observe that the time to indicate the most important features (completion time) for the obesity dataset is influenced by the age of the participant (\textit{p}<0.01).

\subsection{Perception vs. Behavior}
\label{subsec:correlation}
We compare the self-reported and behavioral measurements of our experiment. We thus report the Pearson correlation between perceived trust (resp. understanding) and behavioral trust (understanding). We compare the two perception metrics with the behavioral measures defined in Section~\ref{subsec:metrics}. We observed a correlation value of 0.43 and 0.49 between the perceived trust in AI when facing an explanation \textbf{($\Delta$ Confidence)} and the proportion of users following the AI's prediction \textbf{(Follow Pred.)} for the recidivism and obesity datasets respectively. This result suggests a moderate positive correlation between these two measurements. Conversely, our results indicate no correlation between the perceived users' understanding \textbf{(Immediate Und.)} or \textbf{(Survey Und.)} and their actual comprehension of the model as measured by the precision and recall scores.

\subsection{Open Questions}
\label{subsec:open_q}
We performed a qualitative analysis of participants' responses to two open-ended questions after our experiment: (1) ``According to the scenarios you have seen and the corresponding explanation, how does the artificial intelligence tool predict?'', and (2) ``What was good in the explanation? What was bad in the explanation?''. Our analysis revealed distinct themes in users' feedback, which were influenced by the explanation style and representation they encountered during the experiment.

For users exposed to counterfactual explanations, their feedback often centered around the scoring or assignment of scores to different categories. They frequently emphasized factors such as physical activity, calorie monitoring, and family history. Additionally, users drew comparisons between individual data and statistical information, as exemplified by comments like, ``based on the individual data compared to statistical information of thousands of other individuals'' (P68). While counterfactual explanations with graphical representation generally provided clarity, occasional misunderstandings arose. Users commented, ``Most of the information was good in the explanation. The bad thing is that previous offenses were not explained in detail.'' (P147) and ``The explanation was good because it listed facts that determine the individual's state of health, but was bad in terms of explaining how something like age can contribute to an individual's state of health.''(P31).

In contrast, participants exposed to rule-based explanations tended to identify key risk factors and correlations. They also recognized the influence of lifestyle choices and standard criteria, with comments such as, ``It focuses on main risk factors and their correlation between with one another'' (P41). These users valued the clarity of rule-based and graphical explanations but expressed a desire for deeper insights. For example, they remarked, ``The explanation was all clear, maybe it could use some more information to be more precise.'' (P06) and ``Good at telling us what it was doing but maybe give more information on which specific characteristics had the most weighting'' (P222).

Linear explanations often encouraged users to reflect on database reliance and comparisons to previous cases, with family history, food consumption, and physical activity emerging as recurring themes (\textit{e.g.,} ``The tool uses the frequency of certain behaviors and circumstances to predict someone’s BMI category using percentages'' (P101)). In the case of linear explanations with graphical representation, users found them practical but sought clarification on variable selection. They expressed, ``The explanation about the AI and how it would review each of the four subject people was good but I didn't understand why the AI kept selecting different criteria to assess each individual.'' (P81) and ``I think the entire explanation is good, but maybe they should give you more information on why some factors have more or less impact on the results'' (P199).

Overall, textual representations were well-praised for their clarity, though some users occasionally pointed out biases. For instance, users stated, ``The explanation showed me clearly how the AI would work. No doubts about that!'' (P22), while also noting, ``I think the tool might be a bit biased and too generalized'' (P24). Similar trends were observed for the second dataset (COMPAS), with participants appreciating clarity and transparency in explanations. However, some expressed concerns about inherent biases in the AI's decision-making process. One user observed, ``The good thing was that the tool scored well in most cases. The bad thing was that it put too much emphasis on previous arrests'' (P194). 

Participants in the control group found their interaction with the AI system useful but underscored the importance of understanding the prediction criteria. Their remarks included, ``The tool predicts randomly'' (P126) and ``The good is that it was easy to read. The bad was that it did not contain specific information about the tool'' (P130). These findings underscore the significance of clear, comprehensive, and unbiased explanations in facilitating effective human-AI interactions across diverse scenarios. We elaborate on these possible biases in the discussion section.

\section{Discussion}
\label{sec:discussion}
In the subsequent sections, we will draw insights from our results. We studied the effects of three explanation techniques (feature attribution, rule-based, and counterfactual) and two representations (graphical, textual) on users' trust and understanding of AI models trained on data from two domains (recidivism and obesity). 
Similarly to prior works~\cite{xai_influences_trust, cheng_explanation_style, what-if}, our results suggest that explanations generally help users to (a) better identify which factors led to an AI's prediction, and (b) increase their perceived comprehension. 
Our findings also reveal that the presentation of explanations plays a crucial role in shaping users' trust in the results. Specifically, graphical representations for explanations tend to be more effective in eliciting user acceptance than textual representations. We next discuss these two aspects in more detail.

\subsection{Impact of Explanation Technique}
\label{subsec:disc_paradigm}
We assessed the effects of three explanation techniques on participants' trust and understanding of an AI model (RQ1). 
First, our findings are in line with existing work and support our first hypothesis (H1), namely that explanations increase both (a) the users' comprehension of the AI model and, (b) the trust in the model's predictions. This is supported by the results in Table~\ref{tab: understanding_regression_coef} that show a significant impact based on the chosen explanation technique.
Furthermore, our study also confirms our second hypothesis that rule-based explanations are the most effective way to explain the inner workings of an AI system. This stands in line with existing results~\cite{edit_explanation, Anchor}. We suspect that users within the group receiving rule-based explanations achieve better comprehension for two reasons: (a) the rules' alignment with common educational reasoning principles, and (b) the clarity in terms of when these rules are applicable, i.e., their simplicity. 
This is supported by the results for both self-reported understanding (Fig.~\ref{fig: perceived}) and precision (Fig.~\ref{fig:b_precision}). 
Interestingly, we observe that the effects of the presence of explanations in AI-assisted tasks are more pronounced for the obesity dataset than for the recidivism dataset. We hypothesize that this is the result of (a) the number of features in the datasets (8 for the recidivism case and 15 for obesity), and (b) participants' prior knowledge of the field. Although participants are unlikely to have firsthand experience with prisoners, they are more likely to harbor preconceptions about the causes of obesity.

On the other hand, our study revealed a relatively low precision and self-reported understanding of the counterfactual explanations. These results were similar to the precision and self-reported understanding observed for the control group. However, it is important to note that we measure the users' precision and recall based on the ground truth of the explanation. As such, these results suggest that users are able to accurately identify the features mentioned in the explanation. Nevertheless, users do not possess information about whether any additional features play a role in the given classification. 
Consequently, the low precision results suggest that they tend to indicate additional features based on their preconceptions, which are not marked as important by the counterfactual explanation. This explains the relatively low precision observed with these explanations and underscores that counterfactual explanations may be perceived as less complete than feature attribution or rule-based. This outcome stands in stark contrast to the high scores obtained for both behavioral trust (as shown in Fig.~\ref{fig:b_trust}) and recall (as illustrated in Fig.~\ref{fig:b_recall}). We attribute these findings to the nature of counterfactual explanations, which have been shown by social sciences to align with humans' ability to comprehend and explain complex events~\cite{miller, wachter}.
Our findings align with this concept, as our participants tended to follow the AI model's predictions, as evidenced by the strong behavioral trust. Additionally, they successfully identified the features mentioned in the explanation, as indicated by the recall scores in the case of counterfactual explanations. 

Our qualitative analysis in Section~\ref{subsec:open_q} regarding the advantages and drawbacks of explanations, underscores that users found explanations to be practical and were able to discern the influential factors in the prediction task. Nonetheless and similarly to existing results obtained from open discussions~\cite{van_der_waa}, some participants expressed concerns about the insufficiency of explanations. Indeed, these explanations do not clarify why specific attributes held importance for the model and their correlation with the assigned score. This qualitative study emphasizes that the current form of explanations is functional but might lack comprehensive depth. Participants also pointed out potential biases within the model that could stem from the model focusing on factors like family history or physical activity. This divergence reveals differences between users' expectations of the model's predictive outcomes and the manner in which these predictions are explained. 

Interestingly, we initially anticipated that participants in the control group, who received no explanations, would report a lower perceived understanding of the model. Surprisingly, when we asked them about the benefits and drawbacks of the explanation, we found that some users did not seem to notice the absence of explanations. This observation brings to light two important phenomena: the concept of placebo explanations, which has been previously discussed in the literature~\cite{placebic_explanations}, and the limited experience of laypersons with AI models. The notion of placebo explanations suggests that users might have attributed their understanding of the model based on the mere presence of explanations, regardless of their actual content. Furthermore, it is possible that these participants did not naturally consider the existence of explanations that have not yet been adapted for explaining ML models.

\subsection{Impact of Representation}
\label{subsec:disc_representation}
The influence of representation on users' perception has been well-established~\cite{Niels_fairness, cheng_explanation_style}, and our results corroborate this phenomenon.
We found that the graphical representation induces a higher perceived trust. Self-reported trust levels were higher for the graphical representation compared to the text representation (Fig.~\ref{fig:increased_trust}). 
We suspect that these results stem from a cognitive bias related to the apparent complexity of a graphical presentation. This complexity may give the impression of a greater underlying effort, thereby increasing users' trust in the system.
It is worth emphasizing that our results do not intend to discourage the use of visual representations. Rather, they underscore the need for improved representation techniques. This is vital to highlight since our experiment studied only one possible visual representation for rule-based and counterfactual explanations. 

Nevertheless, we represent the different explanation techniques based on a common representation. In this context, our findings corroborate our hypothesis 4, which states that textual representation appears to facilitate users' understanding of rule-based methods (Fig.~\ref{fig:b_recall}). Similarly, we observe that users' trust in counterfactual explanations rises with textual representation (Fig.~\ref{fig:b_trust}). Surprisingly, we notice an interesting pattern: with rule-based explanations, a textual representation results in higher comprehension but lower trust, while the reverse is observed with a graphical representation. This phenomenon appears to operate inversely with counterfactual explanations. That is, graphical representation leads to lower confidence but higher understanding, as opposed to the textual representation. 
As such, we find that users tend to trust a model's explanation more when their comprehension of the model is lower. As counterintuitive as this is, this observation stands in line with prior work. For instance, Van der Waa et al.~\cite{van_der_waa} showed that for rule-based explanations, users displayed higher comprehension but lower trust. This contrasts with the results they obtained with example-based explanations, where users showed lower comprehension but higher confidence. 

In line with existing work~\cite{quantifying_interpretability}, our study evaluated the task completion time. Our results suggest a positive correlation between participants' age and completion time (as shown in Table~\ref{tab:other_regression_coef}). As such, older participants generally take longer to complete the task but achieve a higher recall score. This suggests that younger participants may tend to tackle the task more impulsively, resulting in quicker completion times but lower recall scores.

\subsection{Limitations \& Future Work}
\label{subsec:limitations}
We identified several limitations related to the studied application domain and our participant sample.
We intentionally presented participants with decision scenarios typically faced by domain experts. This increased reliance of participants on our explanations; this reliance is presumably less pronounced when domain experts use AI recommendation systems.
Therefore, these results are not directly transferable to domain experts or computer scientists~\cite{delaunay_hcxai, survey_hcxai, ribeira_user_centered} as they may react differently and prefer different kinds of explanations and visual representations. 
We focused instead on crowd-sourcing participants to guarantee diversity and reach a general audience. Further studies could seek to evaluate the effect of different explanation techniques and representations on different user groups.

Prior research has employed questionnaires to assess how explanation techniques impact users' comprehension~\cite{van_der_waa} and how different explanation representations can influence users' trust~\cite{employ_trust_scale_1_do_you_trust_me}. However, the results from our three post-questionnaires (trust, understanding, and satisfaction) did not yield any differences across various explanation techniques and representations. This outcome could be due to the fact that users only engaged with the model a limited number of times and encountered instances that were classified differently. It is conceivable that this limited interaction might have contributed to the absence of statistical significance in our findings, as previously suggested by Van der Waa et al.~\cite{van_der_waa}. To gain a more comprehensive perspective on the model's performance, future evaluation could consider either a larger set of instances or a focus on instances with similar classification outcomes.

We evaluated participant understanding through the identification of the most important features in the decision process. 
Other validation tasks could provide additional insights into participant understanding. For instance, tasks involving using the explanation to reproduce the AI's model behavior on different examples or answering what-if scenarios~\cite{bibal}. Such tasks would have the potential to measure a deeper understanding of the model. However, this would come at the cost of extra participant effort. In our specific experimental context, we observed limited improvements in users' comprehension with counterfactual explanations. We anticipate that what-if scenarios could be particularly beneficial to enhance understanding in the case of counterfactual explanations because these tasks align more closely with the objectives of counterfactual explanations. 
Finally, our study was conducted on AI models trained on tabular data. While the studied explanation techniques also apply to other data types such as text and images, the explanations on these data types may resort to visual representations not covered in our study.

\section{Conclusion}
\label{sec:conclusion}
The majority of XAI research has focused on the technical aspects of creating accurate explanation methods. In contrast, this study explores the human aspects of presenting explanations for AI agents to users. Specifically, we conducted a user study that aimed to examine the impact of explanation techniques and visual representations on users' trust and comprehension when faced with AI-based recommendations.
Our study covered three types of explanations, namely feature-attribution, rule-based, and counterfactual, presented either graphically or as textual statements. We evaluated these explanation strategies in two domains, namely the prediction of recidivism and the prediction of risk of obesity.
Our results indicate that rule-based explanations with textual representation are most effective in terms of precision and self-reported understanding. We also observed a difference in the impact of explanations on users' trust across different domains. Counterfactual explanations presented as text elicited higher levels of trust, while the opposite was observed for graphical feature attribution and textual rule-based explanations. Importantly, our results exhibit some variations across the evaluated domains. This underscores the potential and necessity for future investigations to consider user characteristics, data types, and domains' influence on results.


\clearemptydoublepage

}{
	
}

\clearemptydoublepage
\backmatter
\chapter*{Conclusion and Perspectives}
\addcontentsline{toc}{chapter}{Conclusion and Perspectives}
\chaptermark{Conclusion and Perspectives}
\label{chap: conclusion}
This final chapter marks the end of my Ph.D. journey, a moment both long-awaited and filled with apprehension. Embarking on a three-year-long project dedicated to a single research project might initially appear as an eternity for a research novice. However, in retrospect, it becomes evident that the search for more transparency in machine learning is far from reaching its end. In a time when explanations and transparency in machine learning are significant concerns across various fields, numerous techniques have arisen. Yet fundamental questions persist: under which circumstances should each explainability method be employed, and how do they impact the final users?

Therefore, In this concluding chapter, I begin by extracting from my experiences gained during this journey. I synthesize my contributions and glean common insights from the previous chapters of this Ph.D. thesis. Following this, I delve into the valuable lessons learned during this academic experience. From there, I explore some open challenges that require further investigation, while identifying exciting opportunities for future research. Finally, I conclude this thesis.

\section*{Thesis Objectives and Research Journey}
\label{sec: conclusion_overview}
\addcontentsline{toc}{section}{Thesis Overview and Objectives}
With the growing demand for transparency and explainability in the field of AI, my primary goal throughout this thesis has been to explore the most effective methods to make AI agents fully understandable. This first objective led me to investigate the inner workings of explanation techniques and, more specifically, how the data used to generate these explanations influences their quality.

The journey began with a focus on enhancing explanation techniques, which resulted in my first publication and Chapter~\ref{chap: anchors}. In this research endeavor, I analyzed the impact of discretization techniques on the quality of rule-based explanations. Additionally, I studied the use of pertinent negatives in rule-based explanations. This study reveals which words, when added, have the most substantial impact on the model's prediction probability.

As my doctoral journey progressed, it became evident that new methods were being introduced without a comprehensive understanding of their optimal application scenarios. Therefore, I switched my focus toward the question of when these explanation approaches should be applied. Choosing appropriate explanations became a pivotal aspect of my research. This inquiry led to the development of APE, a framework designed to discern whether a linear explanation is suitable for approximating a classification decision boundary for a given instance.

As I delved into APE, it guided me toward investigating the nearest decision boundary. This exploration naturally led me into the domain of producing counterfactual explanations. Thus, I developed Growing Fields as well as Growing Net and Growing Language, two novel counterfactual explanation techniques tailored for textual data. This phase of my research was motivated by the increasing quality and complexity of machine learning models in text-related tasks. It led me to reflect on whether adding complexity to methods intended to provide transparency was indeed a worthwhile initiative. This critical question is the foundation of the investigation presented in Chapter~\ref{chap: emnlp}. In this chapter, I conducted a comparative study between transparent and black-box counterfactual explanation methods and questioned the need for intricate approaches.

During this academic journey, I also considered the end users of explanation techniques. While my initial focus had been on improving the quality of explanations, I recognized that the true impact of any research on eXplainable AI depends, in the end, on whether these improvements are beneficial to the ultimate users. Thus, I collaborated with Dr. Niels van Berkel on the construction of the user study detailed in Chapter~\ref{chap: chi}. This study aimed to explore the impact of three distinct explanation techniques and two different representations on users' trust and understanding. The findings of this study have given us valuable insights into the various effects of different explanation techniques and their representations. This includes a study of the most effective representation for each explanation technique (e.g., textual for example and rule-based, graphical for feature attribution). Also, our study confirms the improved understanding observed when users were presented with rule-based explanations.

Notably, while I was conducting my research at Aalborg University, the lack of user studies measuring the impact of explanation techniques became evident. Conducting such a study presented its own set of challenges. One of those challenges is the irreversible nature of user interactions, which means that mistakes cannot be rectified after deployment. As a result, careful planning and execution were essential. 

\section*{Key Insights and Lessons}
\label{section: lesson}
\addcontentsline{toc}{section}{Key Insights and Lessons}
Someone told me before the beginning of my Ph.D. that the true essence of the journey lies not merely in the manuscript itself but in the invaluable lessons acquired by the Ph.D. student. Thus, I conclude this manuscript by sharing some lessons I have learned. Firstly, I will explore the ``Explanation El Dorado'' and the quest for a one-size-fits-all solution to transparency in machine learning. Second, I will explore the profound impact of explanations on users, particularly the demand for explanations and the evolving trajectory that explanations could take.

\subsubsection*{The Allure of the ``Explanation El Dorado''}
\addcontentsline{toc}{subsubsection}{The Allure of the ``Explanation El Dorado''}
The phenomenon I refer to as the ``Explanation El Dorado'' represents the collective pursuit within the research community to discover an elusive and all-encompassing solution for explainability in machine learning. This quest has become a veritable treasure hunting and focuses on finding a method that can universally render opaque machine learning models transparent to end users. In this era, explanations have emerged as the best way to bridge the gap between humans and machines, offering the promise of understanding the complex inner workings of algorithms. Machine learning models, once opaque and inscrutable, are now expected to elucidate their decision-making processes. Thus, researchers have tirelessly searched for a unique and innovative approach that could uncover the intricate mechanisms of the ML models. 

This quest, however, is not without its complexities and challenges. The vast landscape of explanation methods prompts questions about the most suitable approach for specific applications. This issue becomes apparent, as researchers have aimed to develop explanations that are faithful but have yet underexplored the aspect of combining different paradigms for instance. Furthermore, generating explanations necessitates careful consideration of the trade-offs between accuracy and interpretability. In this regard, we exemplified in Chapter~\ref{chap: emnlp}, that more and more complex methods are developed to explain black box models while simpler methods may be sufficient in certain conditions. As the ``Explanation El Dorado'' continues to attract, researchers have searched for a one-size-fits-all explanation solution, although such a method may be impossible to catch. It has become increasingly evident that explanations must be tailored to the specifics of the data and the intended users. For instance, in Chapter~\ref{chap: ape}, we demonstrated the necessity of adapting the shape of explanations to the decision boundary for reliable results. In a similar vibe, in Chapter~\ref{chap: chi}, we revealed how the form and representation of explanations can influence the interactions between end users and AI systems. 

Upon reflection, it is clear that the quest for explanations holds the potential to revolutionize how we trust and interact with machine learning systems. The allure of the "Explanation El Dorado" continues to guide our endeavors, urging us to push the boundaries of what we can achieve. It underscores the belief that transparency is not a finite destination but an ongoing journey, one that leads us toward a brighter future of collaboration between humans and machines. To attain this, it is imperative to acknowledge the constraints that presently affect explanation techniques, including data types, model architectures, and the diverse requirements of end users.

\subsubsection*{Users Request for Explanations}
\addcontentsline{toc}{subsubsection}{Users Request for Explanations}
My personal experience with explanations as well as the feedback I received through user studies have shown a growing trend: users are no longer satisfied with explanations that merely explain the surface. Indeed, current explanation methods do not provide reasons behind why a particular feature is significant for a model in generating a prediction. Perhaps, it is time to shift the paradigm of explanations towards elucidating the rationale behind the quality of a prediction. For instance, while counterfactual explanations are effective in revealing what needs to be altered to modify a prediction, there is a growing need to understand the underlying reasons behind specific changes. This includes modifying attributes like an individual's gender, and examining the resulting impact on the model's predictions, such as the likelihood of releasing a prisoner. Is this phenomenon a consequence of the training data, the inherent biases within the model, or other intricate factors that drive this outcome?

These questions drive us beyond the domain of traditional explanation methods into a deeper exploration of causality and the mechanisms that influence the decisions made by complex machine learning models~\cite{miller_dead}. As we delve deeper into the transparency mechanisms, we must not only unveil the ``what'' of predictions but also the ``why'', entering an era where interpretability encompasses not just transparency but also a profound understanding of how those AI systems ended up the way they are. By doing so, we equip ourselves with the knowledge required to handle the ethical, societal, and practical implications of machine learning in an increasingly interconnected world. This transformation represents a pivotal shift, where the quest for insight becomes as vital as the quest for prediction accuracy, forging a path towards responsible and trustworthy AI systems.

\section*{Open Challenges and Opportunities for Future Research}
\addcontentsline{toc}{section}{Open Challenges and Opportunities for Future Research}
The contributions of this thesis lead toward several directions for future research. Beyond the perspectives presented in the concluding remarks of each individual chapter, this thesis incorporates upcoming research efforts focused on the interpretability approaches and users' perspectives. While these opportunities may not encompass every worthwhile research avenue, they contain the pivotal paths we regard as essential for advancing the domain of explainability in AI.

\subsection*{Explanations Adapted to Data}
\addcontentsline{toc}{subsection}{Explanations Adapted to Data}
Throughout this thesis, I have identified two research directions that hold promise for future exploration. Firstly, it is essential to discern the conditions under which a given explanation technique is adapted to its context. This builds upon the research trajectory initiated by APE (cf Chapter~\ref{chap: ape}). Secondly, the research community must investigate the effectiveness of introducing additional layers of complexity to methods designed to explain the inner workings of already complex models. This involves a more in-depth investigation of the research introduced in Chapter~\ref{chap: emnlp}.

Traditionally, the pursuit of better explanation techniques has led to the search for a universal solution applicable to all situations. However, the complexity of objectives, user preferences, model architectures, and data types has rendered this approach insufficient. Consider, for example, the diversity of objectives users may have: some may prioritize the maximum fidelity to the underlying model, while others prefer simplicity or are just curious. These objectives are as varied as the individuals themselves. Therefore, a more nuanced approach involves a systematic examination and comparison of these methods, taking into account their impact on different aspects of the explanation process. As demonstrated in Chapter~\ref{chap: ape}, a proper characterization of the decision boundary can guide the selection of adapted explanation methods. This not only provides insights into the suitability of linear surrogates for approximating specific decision boundaries but also serves as a stepping stone for similar investigations with various explanation techniques. Different contexts may require tailored explanations, as exemplified by the contrast between explaining the decisions of a model classifying the toxicity of a text on a social network versus a model predicting the date of a hurricane. While the former may necessitate a simpler, more intuitive explanation, the latter may require a comprehensive and technically precise elucidation of the model's reasoning. By exploring these nuances, we can uncover the true adaptability of explanation methods. 

Another promising research direction is the quantification of the benefits derived from employing complex methods to explain the inner workings of black-box models. Although there is a common belief that increasing complexity of the ML models naturally results in better accuracy, Rudin~\cite{Rudin} advocates for a rigorous investigation of this hypothesis. Therefore, the extension of this hypothesis to the domain of the trade-off between the complexity of explanations and their accuracy must be thoroughly examined. Consider, for instance, the use of intricate neural network architecture to explain a random forest model with few decision trees. Does the introduction of such sophisticated mechanisms truly enhance the clarity and utility of the explanations, or does it introduce an unnecessary layer of complexity that hinders comprehension?

To address this challenge, future research should aim to develop comprehensive frameworks for evaluating the impact of complexity on the effectiveness of explanation methods. In Chapter~\ref{chap: emnlp}, our focus was on traditional metrics, however, there arises a necessity to shift towards user-centric assessments, considering factors like cognitive load, user satisfaction, and decision-making accuracy. By rigorously quantifying the advantages and disadvantages of complexity, we can establish a foundation for informed design choices in the development of explanation techniques for black-box models.

\subsection*{Explanations Tailored to Users}
\addcontentsline{toc}{subsection}{Explanations Tailored to Users}
It has been largely accepted that explanations should be tailored to factors such as the domain, the AI group~\cite{one_ai_doesnt}, and users role~\cite{ribeira_user_centered}. However, I propose considering broader aspects of the users such as their trust in AI, and their purpose when employing the AI systems~\cite{liao_wortman}. While an expert may seek an explanation to understand why a model has failed, a company deploying a system may prioritize providing users with explanations for why the system makes certain predictions. This adaptability can have a profound impact on user interactions.

Throughout this thesis, we have demonstrated that surrogate explanations approximate complex models by training simpler models over interpretable spaces. Among these simpler models, we identified three kinds of surrogate methods: (a) feature-attribution, (b) example-based, and (c) rule-based explanations. Each surrogate approximates the complex model differently, and we have shown in Chapter~\ref{chap: chi} that the choice of the surrogate impacts how users interpret the explanation. Surprisingly, despite the growing interest in explainability, no prior work has compared the impact of these surrogates on specific user roles (e.g., domain expert, developer). Due to a lack of surrogate explanations comparison, XAI users are presently unable to indicate why they might choose one type of surrogate rather than another. However, the choice of the surrogate and its representation can significantly affect users (e.g., trust, understanding)~\cite{Niels_fairness, van_der_waa}. We hence argue that preferring one type of surrogate over another should be driven by criteria and situations rather than for practical reasons.

As a future research direction, I propose to assess the impact of surrogate techniques across different user roles. Building upon our work~\cite{delaunay_hcxai}, I outline various user roles to guide researchers and practitioners in investigating the impact of selecting a surrogate and representation depending on user roles.

Most existing research has focused on three types of roles~\cite{mapping_user_profiles, ribeira_user_centered}: (a) developers that create or assess AI systems; (b) domain experts, persons with knowledge or authority in a particular area; and (c) lay users, individuals to whom the AI decision is applied (e.g., bank client). Yet, we argue that users and usage scenarios are more complex than those three well-defined categories. Instead, users of AI systems are multi-dimensional (e.g., roles, goals, trust in AI), and various scenarios affect the suitability of different explanation methods (e.g., data types, explanation representation). We thus propose six additional aspects to consider when selecting explanations tailored to users:
\begin{itemize}
    \item \textbf{Motivation for Explanation:} Understanding why a user seeks an explanation, whether out of curiosity, to improve performance or to build trust in the system, is a key criterion for selecting the appropriate explanation model.
    \item \textbf{Trust in AI Systems:} Users' trust in AI systems can vary widely, as some programmers may place excessive trust in the systems they code while others may not have blind faith in it, influencing their reliance on explanations.
    \item \textbf{Use Case:} The context in which users interact with AI systems or explanation methods, whether in a professional setting, educational environment, or everyday life, plays a significant role in choosing the right explanation model.
    \item \textbf{Prior Experience:} Users' prior experience with the domain or explanation techniques can impact their interaction with explanations, as demonstrated in~\cite{cultural_mistrust}, where participants with mistrust in healthcare tended to trust AI systems more compared to doctors. 
    \item \textbf{Data Types:} The challenges posed by representing various data types, such as sound or time series, is one of the reasons why few explanation methods exist for these data types~\cite{survey_pisa}. As such, the data type influences the choice of surrogate explanation. 
    \item \textbf{Visual Representation:} Selecting one explanation representation over another (e.g., graphical rather than textual) is crucial, as demonstrated in this thesis, as it impacts how users perceive AI systems.
\end{itemize}

Evaluating how each dimension of the user roles and usage scenarios impacts the effectiveness of surrogate explanations would allow associating surrogate methods tailored to users. Comparing the impact of different surrogate categories over the different aspects of users would benefit the ML sub-community of XAI by allowing them to manage and carefully select the appropriate proxy. As a future work, I envision investigating the impact of the three distinct explanation techniques in collaboration with computer scientists from different research laboratories, specialists either in HCI or ML, and domain experts in relevant domains (e.g., healthcare). These axes can be both continuous (e.g., trust in AI) and categorical (e.g., data types).

\subsection*{Envisioning the Future of Explainable AI}
\addcontentsline{toc}{subsection}{Envisioning the Future of Explainable AI}
As we progress into the landscape of Explainable AI, it is crucial to pause and envision the path forward. The quest for better explanations should align with the diverse needs and motivations that drive users to seek them. In this higher-level exploration, we contemplate several key aspects that we believe should shape the future of explanations in XAI:

\subsubsection*{Application-Adapted Explanations}
\addcontentsline{toc}{subsubsection}{Application-Adapted Explanations}
In the pursuit of more effective explanations in XAI, a statement emerges: the notion of a single, one-size-fits-all solution is insufficient. Instead, we must explore the concept of generating application-specific explanations. Consider, for example, that explaining a medical diagnosis should be approached quite differently from clarifying a legal decision. This realization highlights the importance of tailoring explanations to suit the specific context of their application. Likewise, the choice of the explanation architecture should be influenced by factors such as the specific instance being explained, the data types involved, and the characteristics of the black-box architectures. Furthermore, user objectives play a pivotal role in shaping the future of XAI. Explanations can serve a multitude of purposes, ranging from satisfying curiosity to fulfilling legal obligations, each with its own unique requirements~(e.g., legal obligations vary~\cite{bibal}). To accomplish this, a thorough analysis of when and for whom a particular explanation technique is adapted is essential. This shift will enable us to transform the elusive dream of a one-size-fits-all solution into a practical reality.

\subsubsection*{Embracing Causality in Explanations}
\addcontentsline{toc}{subsubsection}{Embracing Causality in Explanations}
In the realm of explanations, current XAI techniques excel at identifying influential features in a model's decisions, but we must uncover the ``why'' behind these decisions. Embracing causality in explanations means going beyond surface-level information and providing insights into the causal relationships between input features and model outcomes. For instance, in the context of a hurricane prediction algorithm, understanding why an increase in temperature or a decrease in rainfall impacts hurricane risk is more crucial than just knowing that it does. This shift from ``what'' to ``why'' not only enhances our understanding of AI systems' decision-making processes but also significantly benefits the domain of application. It reduces the impression of being a passive spectator of Artificial Intelligence and paves the way for Augmented Intelligence.

\subsubsection*{Leveraging Language Models for Explanations}
\addcontentsline{toc}{subsubsection}{Leveraging Language Models for Explanations}
One other aspect that should be integrated into existing approaches for more transparency is the use of large language models. The rise of powerful language models opens up exciting possibilities for generating explanations~\cite{talk2model}. However, as we have seen through the second part of this thesis, the explanation can be represented through diverse shapes. All these representations share one important limitation, they are static and non-interactive. Once a graph or a textual explanation is generated to depict the importance of each feature for the final prediction, users cannot seek further clarification or ask follow-up questions. As such, users may perceive themself as spectators not listening and outside of the process. To fill this gap, leveraging models like ChatGPT can lead to the creation of more natural, human-friendly explanations~\cite{joel_chatbot}. By integrating language models into XAI, we can bridge the chasm between technical model insights and human comprehension, making explanations more accessible, informative, and adaptable to users' needs. 

\subsubsection*{Navigating the El Dorado of Explanations}
\addcontentsline{toc}{subsubsection}{Navigating the El Dorado of Explanations}
The journey for better explanations in XAI can sometimes resemble the quest for the mythical El Dorado, a relentless pursuit that presents both formidable challenges and rich rewards. In this context, we recognize the importance of meaningful collaborations with businesses and organizations that rely on machine learning and XAI within their operations. By extending this invitation, we seek to engage with a diverse array of stakeholders, each with their unique goals and objectives when employing ML and XAI technologies. This collaborative effort is aimed at tailoring explanations to align with the requirements of users and organizations. These objectives cover a wide range of considerations, from customizing visualization formats to ensuring the consistent provision of explanations. We understand that the needs of various entities can vary significantly, from corporations implementing predictive analytics to government agencies employing AI systems for critical decision-making processes. By actively involving these key players in the process, we can ensure that the journey towards effective explanations is both purposeful and fruitful. This collective effort toward a deeper understanding of the nuanced needs of the community will facilitate the development of more adaptable and user-centered explanation solutions.


\section*{Conclusion}
\addcontentsline{toc}{section}{Conclusion}
This thesis has addressed the fundamental challenge of communication between humans and machine learning models. In an era where machine learning is ingrained into our everyday lives, this communication is not only beneficial in that it results in better models but also becomes a necessity for legal and moral reasons. To this end, we proposed methods to improve the fidelity of explanations for AI models. Then, we observe that there is a tendency to develop increasingly complex explanation methods. However, this complexity does not translate into increasing transparency and may in some conditions even reduce it. Therefore, as I conclude this thesis, it becomes evident that generating the best explanations cannot be divorced from measuring the impact of these explanations on humans. After all, humans are the ultimate users, and their experience should be the central concern of our research. 

Conversely, solely assessing how humans perceive explanations is not sufficient. We recognize that current explanation techniques are still far from perfect, and there is much work to be done. Therefore, the combination of improving explanation quality and measuring its effectiveness with final users to guide further improvements represents the pivotal aspect towards responsible and effective machine learning interpretability. Within the complex landscape of human-AI interaction, we are reminded that our goal is not just to build transparent models but also to create trustworthy and impactful AI systems that align with ethical principles. This purpose will continue to guide researchers and practitioners toward a future where AI is not just a tool but a responsible and valuable collaborator in our lives.

\clearemptydoublepage
\phantomsection 
\addcontentsline{toc}{chapter}{Bibliography}

\printbibliography[heading=primary,keyword=primary]
\newpage
\printbibliography[]

\clearemptydoublepage
\ifenvsetTF{COMPILE_ALL}{
	\appendix
	
	\input{./Appendix/chi}
}{
	\appendix
}

\listoffigures
\listoftables

\clearemptydoublepage



\cleartoevenpage[\thispagestyle{empty}]
\markboth{}{}
\newgeometry{inner=30mm,outer=20mm,top=40mm,bottom=20mm}

\backcoverheader

\selectfontbackcover{ 

\titleFR{Explicabilité des modèles d'apprentissage automatique : De l'adaptabilité des données à la perception de l'utilisateur}

\keywordsFR{Explicabilité ; Interprétabilité ; Interaction Homme-Machine}

\abstractFR{
Cette thèse se concentre sur la génération d'explications locales pour les modèles de machine learning déjà déployés, en recherchant les conditions optimales pour des explications pertinentes. 
L'objectif principal est de développer des méthodes produisant des explications 
à la fois fidèles au modèle sous-jacent et compréhensibles par les utilisateurs qui les reçoivent.
La thèse est divisée en deux parties. Dans la première, on améliore une méthode d'explication basée sur des règles. On introduit ensuite une approche pour évaluer l'adéquation des explications linéaires pour approximer un modèle à expliquer. Enfin, cette partie présente une expérimentation comparative entre deux familles de méthodes d'explication contrefactuelles, dans le but d'analyser les avantages de l'une par rapport à l'autre. La deuxième partie se concentre sur des expériences utilisateurs évaluant l'impact de trois méthodes d'explication et de deux représentations différentes. Ces expériences mesurent la perception en termes de compréhension et de confiance des utilisateurs en fonction des explications et de leurs représentations. L'ensemble de ces travaux contribue à une meilleure compréhension de la génération d'explications, 
avec des implications potentielles pour l'amélioration de la transparence, de la confiance et de l'utilisabilité des systèmes d'IA déployés.
}

\titleEN{Explainability for Machine Learning Models: From Data Adaptability to User Perception}

\keywordsEN{Explainability, Interpretability, Human-Computer Interaction}

\abstractEN{
This thesis explores the generation of local explanations for already deployed machine learning models, aiming to identify optimal conditions for producing meaningful explanations. 
The primary goal is to develop methods for generating explanations 
faithful to the underlying model and comprehensible to the users.
The thesis is divided into two parts. The first enhances a widely used rule-based explanation method. It then introduces a novel approach for evaluating the suitability of linear explanations to approximate a model. Additionally, it conducts a comparative experiment between two families of counterfactual explanation methods to analyze the advantages of one over the other. 
The second part focuses on user experiments to assess the impact of three explanation methods and two distinct representations. These experiments measure how users perceive their interaction with the model in terms of understanding and trust, depending on the explanations and representations. 
This research contributes to a better explanation generation, with potential implications for enhancing the transparency, trustworthiness, and usability of deployed AI systems.
}

}


\end{document}